\pdfoutput=1
\documentclass[journal]{IEEEtran}
\usepackage{amsmath, amssymb, amsthm}
\usepackage{bm}

\usepackage{graphicx}
\usepackage[caption=false,font=footnotesize]{subfig}
\usepackage{booktabs}
\usepackage{longtable}
\usepackage{array}
\usepackage{multirow}
\newcommand{\blockrule}{\specialrule{\lightrulewidth}{1.6pt}{1.2pt}\specialrule{\lightrulewidth}{0pt}{1.6pt}}
\newcommand{\titlerule}{\specialrule{\cmidrulewidth}{0.8pt}{1.6pt}}
\newcommand{\startrule}{\specialrule{0.2pt}{0.6pt}{1.4pt}}
\usepackage{rotating}
\usepackage{stfloats}

\usepackage{algorithm}
\usepackage[noend]{algpseudocode}
\algrenewcommand\algorithmicindent{1.0em}

\usepackage{tikz}
\usetikzlibrary{positioning, arrows.meta, shapes, decorations.pathreplacing, calc, fit, backgrounds}

\usepackage{cite}

\usepackage{xcolor}
\definecolor{abblue}{HTML}{2A78D6}\definecolor{aborange}{HTML}{EB6834}
\newcommand{\abbetter}[1]{\textcolor{abblue}{#1}}\newcommand{\abworse}[1]{\textcolor{aborange}{#1}}
\newcommand{\abref}{\textcolor{black!45}{ref.}}
\usepackage{url}
\usepackage{placeins}

\makeatletter
\def\input@path{{./}{./sec/}{./table/}{./fig/}}
\makeatother
\graphicspath{{./}{./fig/images/}}

\newcommand{\E}{\mathbb{E}}
\newcommand{\KL}{D_{\mathrm{KL}}}
\newcommand{\Ent}{\mathrm{H}}
\newcommand{\I}{\mathrm{I}}

\newtheorem{theorem}{Theorem}

\newtheorem{proposition}[theorem]{Proposition}
\newtheorem{corollary}[theorem]{Corollary}
\theoremstyle{definition}

\theoremstyle{remark}
\newtheorem{remark}[theorem]{Remark}

\title{TEMPER: Reducing Attribute Leakage\\ from Face Recognition Embeddings}

\author{\"Unsal \"Ozt\"urk, Vedrana Krivoku{\'c}a Hahn, Sushil Bhattacharjee,~and~S\'ebastien~Marcel%
\thanks{This work has been submitted to the IEEE for possible publication. Copyright may be transferred without notice, after which this version may no longer be accessible.}%
\thanks{\"U. \"Ozt\"urk, V. Krivoku{\'c}a Hahn, S. Bhattacharjee, and S. Marcel are with the Idiap Research Institute, Martigny, Switzerland. S. Marcel is also with UNIL, Lausanne, Switzerland.}%
\thanks{Corresponding author: \"U. \"Ozt\"urk (e-mail: unsal.ozturk@idiap.ch).}%
}

\theoremstyle{plain}
\newcommand{\restatedname}{Proposition}
\newtheorem*{restated}{\restatedname}
\newenvironment{restate}[1]{\renewcommand{\restatedname}{Proposition~\ref{#1} (main paper)}\begin{restated}}{\end{restated}}
\begin{document}
\bstctlcite{IEEEexample:BSTcontrol}

\maketitle
\let\PaperAddContentsLine\addcontentsline
\renewcommand{\addcontentsline}[3]{}

\begin{abstract}
Face recognition embeddings encode soft-biometric attributes such as gender and ethnicity, which can be inferred from templates. This paper proposes \textbf{TEMP}late \textbf{E}ntropy \textbf{R}egularisation (TEMPER), a post-hoc method that removes an attribute. TEMPER learns a linear map of the embedding in a split-latent variational autoencoder. The map is initialised at the principal component analysis (PCA) solution of the embeddings and trained to maximise the predictive entropy of an auxiliary attribute classifier. A relational term keeps the similarities of mated pairs close to the embedding similarities. We evaluate the removal of gender and a four-group ethnicity label from ArcFace embeddings against ten prior post-hoc methods and two label-free controls. Every method is trained on VGGFace2 and tested on IJB-C and RFW at false match rates down to $10^{-4}$. Leakage is the balanced accuracy of the strongest of four attacker families trained on transformed VGGFace2 templates. TEMPER retains the highest IJB-C true match rate at a false match rate of $10^{-3}$ at every leakage threshold for both attributes. TEMPER retains a true match rate of 82.4\% at a gender leakage of at most 52.5\%, close to chance, against 76.7\% for the best prior method. In the ablation, this improvement comes from the PCA initialisation and the relational term. Code will be released on acceptance.
\end{abstract}


\section{Introduction}
\label{sec:intro}

\IEEEPARstart{F}{ace} recognition networks map a face image to an embedding that is discriminative for identity, and a recognition system stores one embedding of every enrolled subject as a template. The embedding also encodes soft-biometric attributes such as gender, age and ethnicity, although the network was never trained to predict them~\cite{terhoerst2021soft,Terhorst2020BeyondID}. Whoever obtains the stored templates, by theft or sharing, can therefore infer undisclosed attributes~\cite{OsorioRoig2022Attack}. Ethnic origin is a special category of personal data under the General Data Protection Regulation~\cite{GDPR2016}. The goal of attribute removal is a stored template that supports verification and reveals as little as possible about such attributes.

A deployed recognition network is rarely retrained, since retraining changes every embedding and requires every subject to be re-enrolled. We therefore consider post-hoc removal, in which the network stays unchanged and a transform learned on its embeddings produces the transformed template for storage and comparison (Fig.~\ref{fig:concept}). Leakage is the accuracy of a classifier that an attacker trains on labelled transformed templates of some subjects to predict the attribute of other subjects. Gender and ethnicity are properties of the person, so an attacker recovers the attribute of a new template by matching against a labelled template of the same subject. Removal therefore protects only subjects unseen by the attacker and trades verification accuracy against leakage. A method is assessed by the verification accuracy retained at a given leakage, at the low false match rates (FMRs) of operational systems.

\begin{figure}[!t]
\centering
\scalebox{0.9}{%
\begin{tikzpicture}[
    every node/.style={font=\footnotesize},
    proc/.style={draw, rounded corners=2pt, minimum height=0.5cm,
                 minimum width=0.7cm, align=center, fill=#1!8,
                 line width=0.45pt, inner sep=2pt},
    proc/.default=blue,
    img/.style={draw, minimum width=0.6cm, minimum height=0.72cm,
                inner sep=0pt, line width=0.4pt, fill=gray!6},
    emb/.style={draw, rounded corners=1.5pt, minimum height=0.45cm,
                minimum width=0.75cm, line width=0.4pt, inner sep=1.5pt},
    arr/.style={-{Stealth[length=3pt,width=2.4pt]}, line width=0.4pt},
    lbl/.style={text=black!60, inner sep=1pt},
    restab/.style={draw, rounded corners=2pt, inner sep=3pt,
                   line width=0.4pt, align=left},
    rowlbl/.style={font=\footnotesize\bfseries, rotate=90, anchor=center},
]

\newcommand{\personpic}[2]{%
    \begin{scope}[shift={(#1.center)}, scale=0.85]
    \fill[#2]
        (-0.06, 0.04)
        .. controls (-0.12,-0.01) and (-0.25,-0.08) .. (-0.27,-0.16)
        -- (-0.27,-0.38) -- (0.27,-0.38) -- (0.27,-0.16)
        .. controls (0.25,-0.08) and (0.12,-0.01) .. (0.06, 0.04)
        -- cycle;
    \fill[#2] (-0.055, 0.02) rectangle (0.055, 0.10);
    \fill[#2] (0, 0.21) ellipse (0.115cm and 0.14cm);
    \end{scope}%
}

\pgfmathsetmacro{\imgx}{0.85}
\pgfmathsetmacro{\embx}{4.35}   %
\pgfmathsetmacro{\tabx}{5.35}

\node[rowlbl, text=black!70] at (0.15, 2.55) {No removal};

\node[img] (b_f1) at (\imgx, 3.05) {};
\personpic{b_f1}{black!50}
\node[lbl, above=0pt of b_f1] {enrol};
\node[proc=gray, right=0.18cm of b_f1] (b_fr1) {FR};
\node[emb, fill=gray!12] (b_e1) at (\embx, 3.05) {$\bm{x}^{\mathrm{e}}$};
\draw[arr] (b_f1.east) -- (b_fr1.west);
\draw[arr] (b_fr1.east) -- (b_e1.west);

\node[img] (b_f2) at (\imgx, 2.05) {};
\personpic{b_f2}{black!30}
\node[lbl, below=0pt of b_f2] {probe};
\node[proc=gray, right=0.18cm of b_f2] (b_fr2) {FR};
\node[emb, fill=gray!12] (b_e2) at (\embx, 2.05) {$\bm{x}^{\mathrm{p}}$};
\draw[arr] (b_f2.east) -- (b_fr2.west);
\draw[arr] (b_fr2.east) -- (b_e2.west);

\draw[decorate, decoration={brace, amplitude=3pt, raise=2pt},
      line width=0.4pt, black!40]
    (b_e1.north east) -- (b_e2.south east);
\coordinate (b_brace) at ($(b_e1.east)!0.5!(b_e2.east) + (0.28,0)$);

\node[restab, draw=black!45, fill=red!3, anchor=west] (b_tab) at (\tabx, 2.55) {%
\renewcommand{\arraystretch}{1.15}%
\begin{tabular}{@{}l@{\hspace{1.2em}}r@{\hspace{0.4em}}c@{}}
\textit{TMR}            & 97.6\% & \phantom{$\bm{\times}$} \\
\textit{Gender leakage} & 96.0\% & \textcolor{red!55!black}{$\bm{\times}$}
\end{tabular}};
\draw[arr, black!50] (b_brace) -- (b_tab.west);

\draw[black!18, line width=0.5pt, densely dashed] (-0.1, 1.3) -- (9.2, 1.3);

\node[rowlbl, text=blue!70!black] at (0.15, 0) {With removal};

\node[img] (v_f1) at (\imgx, 0.5) {};
\personpic{v_f1}{black!50}
\node[lbl, above=0pt of v_f1] {enrol};
\node[proc=gray, right=0.18cm of v_f1] (v_fr1) {FR};
\node[proc=blue, right=0.15cm of v_fr1] (v_rm1) {TEMPER};
\node[emb, fill=teal!12] (v_r1) at (\embx, 0.5) {$\bm{r}^{\mathrm{e}}$};
\draw[arr] (v_f1.east) -- (v_fr1.west);
\draw[arr] (v_fr1.east) -- (v_rm1.west);
\draw[arr] (v_rm1.east) -- (v_r1.west);

\node[img] (v_f2) at (\imgx, -0.5) {};
\personpic{v_f2}{black!30}
\node[lbl, below=0pt of v_f2] {probe};
\node[proc=gray, right=0.18cm of v_f2] (v_fr2) {FR};
\node[proc=blue, right=0.15cm of v_fr2] (v_rm2) {TEMPER};
\node[emb, fill=teal!12] (v_r2) at (\embx, -0.5) {$\bm{r}^{\mathrm{p}}$};
\draw[arr] (v_f2.east) -- (v_fr2.west);
\draw[arr] (v_fr2.east) -- (v_rm2.west);
\draw[arr] (v_rm2.east) -- (v_r2.west);

\draw[decorate, decoration={brace, amplitude=3pt, raise=2pt},
      line width=0.4pt, blue!45]
    (v_r1.north east) -- (v_r2.south east);
\coordinate (v_brace) at ($(v_r1.east)!0.5!(v_r2.east) + (0.28,0)$);

\node[restab, draw=blue!50, fill=green!3, anchor=west] (v_tab) at (\tabx, 0) {%
\renewcommand{\arraystretch}{1.15}%
\begin{tabular}{@{}l@{\hspace{1.2em}}r@{\hspace{0.4em}}c@{}}
\textit{TMR}            & 82.4\% & \phantom{$\bm{\times}$} \\
\textit{Gender leakage} & 52.2\% & \phantom{$\bm{\times}$}
\end{tabular}};
\draw[arr, blue!50] (v_brace) -- (v_tab.west);

\node[lbl, anchor=north east] at ($(v_tab.south east) + (-0.33, -0.04)$) {chance 50\%};

\end{tikzpicture}}
\caption{Post-hoc attribute removal. TEMPER replaces each embedding $\bm{x}$ by the transformed template $\bm{r}$ for storage and comparison. TMR at FMR~$10^{-3}$ and leakage are measured on IJB-C.}
\label{fig:concept}
\vspace{-8pt}
\end{figure}

Prior post-hoc methods discard the coordinates or principal components that predict the attribute~\cite{Terhorst2019IVE,Melzi2023MultiIVEPE}, or erase the directions used by a linear classifier~\cite{Ravfogel2020INLP,Belrose2023LEACE}. Split autoencoders match the moments of an identity code across demographic groups~\cite{Bortolato2020PFRNet,Rot2024ASPECD}, and block permutation realigns the blocks of the embedding at comparison time~\cite{Terhorst2020PEMIU}. Other encoders are trained against an attribute detector~\cite{Morales2021SensitiveNets} or under information bounds~\cite{Razeghi2024Funnel,Razeghi2026DVPF}. The studies used different attackers, datasets and operating points, so the verification accuracy at a given leakage is known only within each study. The tradeoff near chance has been reported only with the equal error rate, the LFW accuracy or verification at an FMR of $10^{-1}$ or $10^{-2}$.

We propose \textbf{TEMP}late \textbf{E}ntropy \textbf{R}egularisation (TEMPER), a post-hoc method that transforms embeddings with a linear dimensionality reduction learned to remove the attribute. The transform is trained within a variational autoencoder to maximise the predictive entropy of an auxiliary attribute classifier~\cite{Roy2019MaxEnt}. Training against a classifier can make the templates of different people hard to tell apart. The attribute can also stay predictable for a newly trained classifier while the training classifier is at chance~\cite{Elazar2018Adversarial}. The transform is initialised at the leading principal components of the embeddings, so that training starts from a transformed template that already supports verification. A relational term keeps the similarity of two transformed templates of the same person close to that of their embeddings, which penalises the solution with one transformed template for all identities. The classifier is also trained on transformed templates stored from earlier steps. Our contributions are as follows:

\begin{itemize}
    \item We propose TEMPER, a linear post-hoc transform trained against an auxiliary classifier. TEMPER retains more IJB-C verification accuracy than every other method when the attribute is predicted close to chance. In the ablation, this gain comes from the PCA initialisation and from the relational term.
    \item We evaluate all methods under one protocol, with FMRs down to $10^{-4}$ and leakage as the worst case over four attacker families trained on VGGFace2. The classifier that a method is trained against can be at chance while an attacker trained afterwards still predicts the attribute.
    \item We compare TEMPER with ten prior post-hoc methods on gender and ethnicity removal. We additionally report the leakage and the verification accuracy of label-free dimensionality reductions, such as PCA, as controls, which only one prior study reports~\cite{Melzi2023MultiIVEPE}.
    \item We analyse the tradeoff between fairness and verification accuracy for every method. Attribute removal brings the false match rates of the demographic groups closer, at the cost of a higher false non-match rate.
\end{itemize}

\section{Related Work}
\label{sec:related}

Meden et al.~\cite{Meden2021Survey} survey privacy-enhancing face biometrics. End-to-end methods such as DebFace~\cite{Gong2020DebFace} and Face-CPFNet~\cite{Chen2025FaceCPFNet} retrain the network, and PrivacyNet~\cite{Mirjalili2020PrivacyNet} perturbs the image. Neither approach applies to stored templates. Table~\ref{tab:methods} lists the post-hoc methods that we compare.

\begin{table}[!t]
\centering
\scriptsize
\caption{Compared methods ($^{\ddagger}$ trained with identity labels).}
\label{tab:methods}
\setlength{\tabcolsep}{2pt}
\begin{tabular}{@{}l p{0.28\columnwidth} p{0.31\columnwidth} l@{}}
\toprule
\textbf{Method} & \textbf{Transform} & \textbf{Privacy objective} & \textbf{Param.} \\
\midrule
\multicolumn{4}{c}{\textit{Prior methods}} \\
\midrule
IVE~\cite{Terhorst2019IVE} & \raggedright coordinate subset & \raggedright forest importance ranking & \# removed \\
Multi-IVE~\cite{Melzi2023MultiIVEPE} & \raggedright principal components & \raggedright forest importance ranking & \# removed \\
INLP~\cite{Ravfogel2020INLP} & \raggedright linear projection & \raggedright classifier nullspace & \# iterations \\
PFRNet~\cite{Bortolato2020PFRNet} & \raggedright identity code of a split autoencoder & \raggedright moment matching across groups & $\lambda_{\mathrm{dis}}$ \\
SensitiveNets$^{\ddagger}$~\cite{Morales2021SensitiveNets} & \raggedright stack of linear layers & \raggedright triplet loss, detector driven to a fixed output & \# layers \\
DVPF~\cite{Razeghi2024Funnel,Razeghi2026DVPF} & \raggedright encoder--decoder, sampled output & \raggedright variational privacy funnel & $\alpha$ \\
PE-MIU~\cite{Terhorst2020PEMIU} & \raggedright shuffled blocks, aligned comparison & \raggedright none (no training or labels) & $s$ \\
PASS$^{\ddagger}$~\cite{Dhar2021PASS} & \raggedright linear layer with PReLU & \raggedright discriminator ensemble, retrained in turn & $\lambda_{\mathrm{deb}}$ \\
FaRM~\cite{BasuRoyChowdhury2022FaRM} & \raggedright four-layer network & \raggedright per-class rate--distortion maximisation & \# epochs \\
\midrule
\multicolumn{4}{c}{\textit{Linear erasure}} \\
\midrule
\shortstack[l]{LEACE~\cite{Belrose2023LEACE}\\PCA + LEACE} & \raggedright linear projection, or after PCA truncation & \raggedright closed-form linear erasure & --, $k$ \\
\midrule
\multicolumn{4}{c}{\textit{Label-free controls (reference)}} \\
\midrule
PCA truncation & \raggedright $k$ leading principal components & \raggedright none & $k$ \\
Random dimensions & \raggedright $k$ random dimensions & \raggedright none & $k$ \\
\midrule
TEMPER$^{\ddagger}$ (ours) & \raggedright linear $k$-dim.\ template branch, PCA-initialised & \raggedright maximum entropy of an auxiliary classifier & $k$, $\lambda$ \\
\bottomrule
\end{tabular}
\vspace{-8pt}
\end{table}

\noindent\textbf{Selection and permutation of coordinates.} IVE~\cite{Terhorst2019IVE} trains a decision-tree ensemble to predict the attribute and removes the most important embedding coordinates, a fixed number at a time. Multi-IVE~\cite{Melzi2023MultiIVEPE} applies IVE to principal components for several attributes at once. PE-MIU~\cite{Terhorst2020PEMIU} stores the template as shuffled blocks of coordinates and realigns the blocks at comparison time, and similarity-sensitive noise~\cite{Terhorst2019Noise} perturbs the template. Neither method uses attribute labels. Compressive privacy projects data onto a subspace chosen to keep utility and lose private information~\cite{Kung2017Compressive}. The intrinsic dimensionality of a SphereFace embedding is estimated at 16~\cite{Gong2019Intrinsic}, and a mapping to 16 dimensions reduces the true match rate (TMR) at FMR~$10^{-3}$ on IJB-C from 71.3\% to 59.8\%.

\noindent\textbf{Linear erasure.} INLP~\cite{Ravfogel2020INLP} iteratively trains a linear classifier for the attribute and projects the embeddings onto the nullspace of its weights. LEACE~\cite{Belrose2023LEACE} erases the attribute in closed form, with the smallest change of the embeddings, so that linear classifiers cannot recover the attribute. Fair principal component analysis (PCA) seeks the best low-rank approximation of the data from which a protected attribute cannot be inferred~\cite{Olfat2019FairPCA,Kleindessner2023FairPCA}. These methods protect against linear classifiers only, and a kernelised erasure that defeats one nonlinear adversary does not protect against others~\cite{Ravfogel2022Kernel}.

\noindent\textbf{Adversarial training.} Invariant representations are commonly learned by training an encoder against a classifier that predicts the variable to be removed~\cite{Edwards2016Censoring}, often through a gradient-reversal layer~\cite{Ganin2016DANN}. LAFTR~\cite{Madras2018LAFTR} and Xie et al.~\cite{Xie2017Controllable} learn invariant representations in this way. The variational fair autoencoder~\cite{Louizos2016VFAE} and Moyer et al.~\cite{Moyer2018Invariant} replace the adversary by a maximum mean discrepancy penalty and by a variational bound on the mutual information. Roy \& Boddeti~\cite{Roy2019MaxEnt} let the encoder maximise the entropy of the adversary's prediction instead of minimising its likelihood. SensitiveNets~\cite{Morales2021SensitiveNets} learns a stack of linear layers on fixed embeddings, with a triplet loss that preserves identity and a regulariser that pushes an attribute detector towards a fixed output. PASS~\cite{Dhar2021PASS} trains a transform of fixed face descriptors against an ensemble of attribute discriminators, retrained one at a time so that the other members keep penalising previously used directions. FaRM~\cite{BasuRoyChowdhury2022FaRM}, proposed for text representations, does not use an adversary and decorrelates the representations within each attribute class by maximising a rate--distortion function.

\noindent\textbf{Autoencoders and information objectives.} PFRNet~\cite{Bortolato2020PFRNet} encodes the embedding into an identity-related and an attribute-related code, reconstructs it from both, and matches the moments of the identity-related code across demographic groups. ASPECD~\cite{Rot2024ASPECD} generalises PFRNet to several categorical attributes of arbitrary cardinality. The Variational Autoencoder (VAE)~\cite{Kingma2014VAE} trades reconstruction against a Kullback--Leibler (KL) regulariser towards a prior, and split-latent models reserve latent dimensions for sensitive attributes~\cite{Creager2019Fair}. The privacy funnel~\cite{Makhdoumi2014Funnel} trades the information that the output keeps about the data against the information it reveals about a sensitive variable. Razeghi et al.~\cite{Razeghi2024Funnel,Razeghi2026DVPF} apply it to face recognition with DVPF, a stochastic encoder learned on fixed face embeddings. Such terms may be estimated with neural estimators such as MINE~\cite{Belghazi2018MINE} and CLUB~\cite{Cheng2020CLUB}.

\noindent\textbf{Evaluating removal.} An attribute that an adversary cannot predict during training is not necessarily absent from the representation. Classifiers trained after the fact recovered demographic attributes from adversarially trained text encoders~\cite{Elazar2018Adversarial}. Osorio-Roig et al.~\cite{OsorioRoig2022Attack} recovered gender from protected face templates by matching them against a small labelled database protected with the same method. In the post-hoc methods above, verification is summarised by the equal error rate or the LFW accuracy~\cite{Bortolato2020PFRNet,Terhorst2020PEMIU,Melzi2023MultiIVEPE,Morales2021SensitiveNets}. Verification is reported at FMRs of $10^{-2}$ and $10^{-3}$ for ASPECD~\cite{Rot2024ASPECD}, at $10^{-1}$ for DVPF~\cite{Razeghi2026DVPF} and down to $10^{-5}$ for PASS~\cite{Dhar2021PASS}. Leakage is measured against one classifier in some studies~\cite{Bortolato2020PFRNet,Dhar2021PASS,Razeghi2026DVPF} and against the best of three or four in others~\cite{Terhorst2020PEMIU,Melzi2023MultiIVEPE,Rot2024ASPECD}. Only Multi-IVE compares its method with a label-free reduction of the dimension, a random order of elimination, which removed as much in the embedding space.

\noindent\textbf{Relation to TEMPER.} The split autoencoder of TEMPER follows PFRNet in a variational form. The transformed template has fewer dimensions than the embedding, as in IVE and Multi-IVE, through a linear map initialised at the PCA solution. TEMPER trains this map against an attribute classifier, as in adversarial training and DVPF, and the penalty is the predictive entropy of the classifier. PFRNet matches moments and linear erasure removes the directions of a linear classifier, whereas the auxiliary classifier of TEMPER is nonlinear. DVPF decodes a sampled latent, whereas the template of TEMPER is deterministic. To preserve identity, SensitiveNets uses a triplet loss and PASS uses an identity classifier, whereas TEMPER keeps the similarity of two transformed templates of the same person close to the similarity of their embeddings. PASS trains against an ensemble of discriminators, whereas TEMPER trains one classifier, also on transformed templates stored from earlier steps.

\section{Method}
\label{sec:method}

\subsection{Overview}
\label{sec:method-overview}

TEMPER computes the transformed template from the embedding with an affine map to $k$ dimensions followed by $\ell_2$ normalisation. We train the map within a variational autoencoder that, as in PFRNet~\cite{Bortolato2020PFRNet}, encodes each embedding into two latents and reconstructs the embedding from both (Fig.~\ref{fig:method}). We call the first the template latent, whose normalised posterior mean is the transformed template, and the second the class latent, whose prior depends on the attribute.

The affine map is initialised at the PCA solution of the embeddings (Section~\ref{sec:method-arch}). The map is trained to maximise the predictive entropy of an auxiliary classifier on the transformed template, weighted by the privacy weight $\lambda\ge0$ (Section~\ref{sec:method-obj}). A relational term on mated pairs penalises the trivial solution of this objective, which is one transformed template for all embeddings (Section~\ref{sec:method-rel}). We relate the entropy term to the leakage that an attacker measures (Section~\ref{sec:method-interp}). We give the proofs and further remarks in Appendix~\ref{app:proofs}.

\subsection{Problem Setting and Threat Model}
\label{sec:method-threat}

Let $X$ denote the random variable of $\ell_2$-normalised face embeddings produced by a frozen face recognition network, and let $\bm{x}$ denote a realisation. The embeddings are points of the unit sphere $\mathbb{S}^{d-1}$ of $\mathbb{R}^{d}$. Let $C\in\mathcal{C}=\{1,\ldots,|\mathcal{C}|\}$ denote the sensitive attribute and $Y$ the identity, with realisations $c$ and $y$, and let $\mathcal{D}=\{(\bm{x}_i,c_i,y_i)\}_{i=1}^{N}$ be a labelled training set.

A post-hoc removal method learns a transform $\bm{r}:\mathbb{S}^{d-1}\to\mathbb{S}^{k-1}$ with template dimension $k$ once on $\mathcal{D}$. The transformed template $R\triangleq\bm{r}(X)$ then replaces the embedding in storage and in comparison, and verification compares two transformed templates by their cosine similarity. Entropies $\Ent$ and mutual informations $\I$ are in nats. For a distribution $q$ on $\mathcal{C}$, $\Ent(q)=-\sum_{c\in\mathcal{C}}q(c)\log q(c)$, and $\Ent(C\mid R)$ denotes $\E[\Ent(p(\cdot\mid R))]$ with $p(\cdot\mid R)$ the conditional distribution of $C$ given $R$.

The attacker holds labelled transformed templates of a set of identities, trains any classifier on these templates and predicts the attribute of identities outside this set. We define leakage as the prediction success of the strongest such attacker. The attacker is restricted to identities outside this set, since $C$ is a function of $Y$ when attribute labels are assigned per subject.

The same dependence bounds what a removal method can achieve. If $Y$ is almost surely a function of $R$, then $C$ is almost surely a function of $R$ and $\Ent(C\mid R)=0$. Independence of $R$ and $C$ would instead give $\Ent(C\mid R)=\Ent(C)$, which is positive for every attribute with two or more values of positive probability. A transformed template from which $Y$ is almost surely recoverable cannot be independent of $C$. Removing the attribute therefore costs verification accuracy.

\subsection{Transform and Initialisation}
\label{sec:method-arch}
\label{sec:method-init}%

\begin{figure*}[t]
\centering
\resizebox{0.92\textwidth}{!}{\definecolor{tempercolor}{HTML}{1C5CAB}
\definecolor{dmuted}{HTML}{52514E}
\definecolor{trnline}{HTML}{8A8984}
\definecolor{trnbg}{HTML}{F2F1ED}
\definecolor{losscol}{HTML}{6A3D9A}
\definecolor{hl}{HTML}{C62828}
\begin{tikzpicture}[
    font=\footnotesize,
    >={Stealth[length=4pt,width=3.2pt]},
    box/.style={draw, rounded corners=2pt, minimum height=0.6cm, minimum width=0.8cm, inner sep=2.5pt, align=center},
    inf/.style={box, draw=tempercolor, fill=tempercolor!12, line width=0.9pt},
    rel/.style={box, draw=tempercolor, fill=tempercolor!25, line width=1.4pt},
    frozen/.style={box, draw=dmuted, fill=white, line width=0.5pt},
    trn/.style={box, draw=trnline, fill=white, dashed, line width=0.6pt},
    lat/.style={trn, circle, minimum size=0.64cm, inner sep=0pt},
    var/.style={inner sep=1.5pt, align=center},
    note/.style={font=\scriptsize, text=dmuted, inner sep=1pt, align=center},
    infarr/.style={->, line width=1.1pt, draw=tempercolor},
    arr/.style={->, line width=0.5pt, draw=dmuted},
    tag/.style={draw=losscol!60, fill=losscol!7, text=losscol!85!black, rounded corners=3pt,
                inner sep=2pt, line width=0.5pt, align=center},
    tagline/.style={draw=losscol!60, line width=0.6pt},
    dot/.style={circle, fill=dmuted, inner sep=0pt, minimum size=2.6pt},
]
\fill[tempercolor!7, rounded corners=3pt] (0,0.85) rectangle (17.9,-0.5);
\filldraw[fill=trnbg, draw=trnline, dashed, rounded corners=3pt, line width=0.6pt] (0,-0.66) rectangle (17.9,-3.42);
\node[anchor=east, text=tempercolor, align=right, inner sep=0pt] at (17.75,0.18)
    {\textbf{inference}};
\node[anchor=south east, text=dmuted, align=right, inner sep=0pt] at (17.75,-3.3)
    {\textbf{training only}};

\node[frozen] (fr) at (0.95,0) {frozen FR\\[-1pt]network};
\node[var] (x) at (2.35,0) {$\bm{x}$};
\node[inf] (W) at (4.2,0) {$\bm{W}\bm{x}+\bm{b}$\\[-1pt]{\scriptsize PCA init.}};
\node[note, text=tempercolor, anchor=south] (Wnote) at ([yshift=1pt]W.north) {mean, \textcolor{hl}{$k$} outputs};
\node[var] (mu) at (6.4,0) {$\bm{\mu}_\theta$};
\node[rel] (r) at (10.4,0) {$\bm{r}_\theta=\bm{\mu}_\theta/\|\bm{\mu}_\theta\|_2$};
\node[note, text=tempercolor, anchor=south] at ([yshift=1pt]r.north) {transformed template};
\node[var, anchor=west] (cmp) at (12.45,0) {cosine comparison};
\draw[infarr] (fr) -- (x);
\draw[infarr] (x) -- (W);
\draw[infarr] (W) -- (mu);
\draw[infarr] (mu) -- (r);
\draw[infarr] (r) -- (cmp);

\node[trn] (A) at (4.2,-1.4) {$\bm{A}\bm{x}+\bm{a}$};
\node[lat] (zr) at (6.4,-1.4) {$\bm{z}_r$};
\node[dot] (bx1) at (2.35,-1.4) {};
\draw[arr] (x.south) -- (bx1) -- (A.west);
\draw[arr] (A) -- node[above=-1pt]{$\bm{\sigma}_\theta$} (zr);
\draw[arr] (mu) -- (zr);
\node[tag, anchor=west] (klr) at (7.1,-0.98) {$\mathcal{L}_{\mathrm{KL}}^{r}$ to $\mathcal{N}(\bm{0},\bm{I}_k)$};
\node[draw=dmuted, dashed, line width=0.7pt, rounded corners=3pt, inner xsep=5pt, inner ysep=4pt, fit=(Wnote)(W)(A)] (encbox) {};
\node[note, anchor=north, text=dmuted] at ([yshift=-1pt]encbox.south) {template branch};
\draw[tagline] (zr.north east) -- (klr.west);

\node[trn] (g) at (4.2,-2.6) {$g_\vartheta$\\[-1pt]{\scriptsize class encoder}};
\node[lat] (zc) at (6.4,-2.6) {$\bm{z}_c$};
\draw[arr] (bx1) |- (g.west);
\draw[arr] (g) -- node[above=-1pt]{$\bm{\mu}_\vartheta,\bm{\sigma}_\vartheta$} (zc);
\node[tag, anchor=west] (klc) at (7.1,-3.02) {$\mathcal{L}_{\mathrm{KL}}^{c}$ to $\mathcal{N}(\bm{e}_c,\bm{I}_{d_c})$};
\draw[tagline] (zc.south east) -- (klc.west);

\node[trn] (f) at (10.4,-2.6) {$f_\psi$\\[-1pt]{\scriptsize decoder}};
\draw[arr] (zr) -- node[note, above, sloped]{rest of $\bm{x}$ (identity)} (f);
\draw[arr] (zc) -- node[note, above]{attribute} (f);
\node[var] (xh) at (12.0,-2.6) {$\hat{\bm{x}}$};
\draw[arr] (f) -- node[above=-1pt]{$/\|{\cdot}\|_2$} (xh);
\node[tag] (lrec) at (13.25,-2.6) {$\mathcal{L}_{\mathrm{rec}}$};
\node[var] (xt) at (14.45,-2.6) {$\bm{x}$};
\draw[tagline] (xh) -- (lrec);
\draw[tagline] (lrec) -- (xt);
\node[tag, anchor=west] (lrel) at (14.9,-2.15) {$\gamma\,\mathcal{L}_{\mathrm{rel}}$ {\scriptsize(mated pairs)}};
\draw[tagline] (r.south) -- ++(0,-0.28) -| (17.7,-2.15) -- (lrel.east);

\node[trn] (S) at (10.4,-1.4) {$\mathcal{S}_{\mathcal{A}}$};
\node[trn] (q) at (11.7,-1.4) {$q_\phi$};
\node[note, anchor=south] at ([yshift=1pt]q.north) {auxiliary classifier, memory $\mathcal{M}$};
\node[var] (qh) at (13.3,-1.4) {$q_\phi(\cdot\mid\tilde{\bm{r}})$};
\draw[arr] (r) -- (S);
\draw[arr] (S) -- (q);
\draw[arr] (q) -- (qh);
\node[tag, anchor=west] (ladv) at (14.55,-1.1) {$\mathcal{L}_{\mathrm{adv}}$ {\scriptsize(trains $q_\phi$)}};
\node[tag, anchor=west] (lent) at (14.55,-1.7) {$\textcolor{hl}{\lambda}\,\mathcal{L}_{\mathrm{ent}}$ {\scriptsize(trains $\bm{W},\bm{b}$)}};
\node[var, anchor=west] (c1) at ([xshift=0.3cm]ladv.east) {$c$};
\draw[tagline] (qh.east) -- (ladv.west);
\draw[tagline] (qh.east) -- (lent.west);
\draw[tagline] (ladv) -- (c1);
\end{tikzpicture}}
\caption{Architecture of TEMPER. Only the blue path from $\bm{x}$ to $\bm{r}_\theta(\bm{x})$ is kept at inference, and $k$ and $\lambda$ (red) are the two free parameters.}
\label{fig:method}
\vspace{-8pt}
\end{figure*}

\noindent\textbf{Template branch.} The template branch defines the approximate posterior $q_\theta(\bm{z}_r\mid\bm{x})=\mathcal{N}\bigl(\bm{\mu}_\theta(\bm{x}),\operatorname{diag}\bm{\sigma}_\theta^{2}(\bm{x})\bigr)$ of the template latent $\bm{z}_r\in\mathbb{R}^{k}$, with the affine mean and the standard deviation
\begin{equation}
\begin{aligned}
\bm{\mu}_\theta(\bm{x})&=\bm{W}\bm{x}+\bm{b},\\
\bm{\sigma}_\theta(\bm{x})&=\exp\Bigl(\tfrac{1}{2}\operatorname{clip}\bigl(\bm{A}\bm{x}+\bm{a},\ell_{\min},\ell_{\max}\bigr)\Bigr),
\end{aligned}
\label{eq:enc}
\end{equation}
where $\bm{W},\bm{A}\in\mathbb{R}^{k\times d}$, $\bm{b},\bm{a}\in\mathbb{R}^{k}$, $\theta\triangleq(\bm{W},\bm{b},\bm{A},\bm{a})$. The function $\operatorname{clip}(u,\ell_{\min},\ell_{\max})=\min(\max(u,\ell_{\min}),\ell_{\max})$ with $\ell_{\min}<\ell_{\max}$ bounds the log-variance, and $\operatorname{clip}$ and $\exp$ act elementwise. The transformed template $\bm{r}_\theta(\bm{x})$ is the mean $\bm{\mu}_\theta(\bm{x})$ normalised to unit $\ell_2$ norm, and we write $R=\bm{r}_\theta(X)$ and assume $\bm{\mu}_\theta(X)\neq\bm{0}$ almost surely. The prior of the template latent is $p(\bm{z}_r)=\mathcal{N}(\bm{0},\bm{I}_k)$, and the template latent is sampled in training as $\bm{z}_r=\bm{\mu}_\theta(\bm{x})+\bm{\sigma}_\theta(\bm{x})\odot\bm{\epsilon}$ with $\bm{\epsilon}\sim\mathcal{N}(\bm{0},\bm{I}_k)$, where $\odot$ denotes the elementwise product. This sample feeds only the reconstruction of the embedding and is not part of the transformed template.

\noindent\textbf{PCA initialisation.} A transform trained against a classifier can initially lose the separation of identities, and we therefore initialise the affine map at a transformed template that already discriminates identities. Let $\bm{m},\bm{s}\in\mathbb{R}^{d}$ be the per-feature mean and standard deviation, with $\bm{s}>\bm{0}$, of $n$ randomly drawn training embeddings, and let $\tilde{\bm{X}}\in\mathbb{R}^{n\times d}$ collect their standardised versions $(\bm{x}-\bm{m})\oslash\bm{s}$, where $\oslash$ denotes elementwise division. Let $\bm{V}_k\in\mathbb{R}^{d\times k}$ hold the right singular vectors for the $k$ largest singular values of $\tilde{\bm{X}}$. We initialise
\begin{equation}
\bm{W}_0=\bm{V}_k^{\top}\operatorname{diag}(\bm{s})^{-1},\qquad\bm{b}_0=-\bm{W}_0\bm{m},
\label{eq:pcainit}
\end{equation}
so that, with $\theta_0=(\bm{W}_0,\bm{b}_0,\bm{A},\bm{a})$, $\bm{\mu}_{\theta_0}(\bm{x})=\bm{V}_k^{\top}\bigl((\bm{x}-\bm{m})\oslash\bm{s}\bigr)$ are the top-$k$ principal-component scores of the standardised embedding. The initial transformed template is therefore the normalised truncation to the top-$k$ principal components, which is computed without attribute labels.

\subsection{Privacy Loss}
\label{sec:method-obj}

The privacy loss trains the template branch against an auxiliary attribute classifier on class-balanced minibatches. These minibatches $\mathcal{B}$ hold $PK$ images, $K$ images of each of $P$ identities, drawn so that every value of $C$ contributes $P/|\mathcal{C}|$ identities, with $|\mathcal{C}|$ a divisor of $P$. The class prior is uniform in every minibatch, and the uniform prediction, which maximises the entropy of a classifier, is also the prediction from the class prior. Let $p_{\mathcal{B}}$ denote the distribution of $(X,C,Y)$ with a uniform class prior and the class-conditional distributions of the population. The class-balanced sampler attains the uniform class prior of $p_{\mathcal{B}}$ and approximates the class-conditional distributions.

\noindent\textbf{Classification loss.} The auxiliary classifier has two heads, a multilayer perceptron $q_\phi^{(1)}$ and a linear classifier $q_\phi^{(2)}$, with parameters $\phi$, and $q_\phi$ denotes the pair. Each head maps a standardised transformed template to a distribution on $\mathcal{C}$. Let $\bm{r}_i$ denote the transformed template of entry $i$ of a minibatch, with $\bm{r}_i=\bm{r}_\theta(\bm{x}_i)$ for the current minibatch. For an index set $\mathcal{A}$, an entry $i\in\mathcal{A}$ and a coordinate $j$, the standardisation is
\begin{equation}
\tilde{r}_{ij}=\bigl[\mathcal{S}_{\mathcal{A}}(\bm{r}_i)\bigr]_j=\frac{r_{ij}-\bar{r}_j}{\hat{\tau}_j+\varepsilon},
\label{eq:std}
\end{equation}
where $r_{ij}=[\bm{r}_i]_j$, $\bar{r}_j$ and $\hat{\tau}_j$ are the sample mean and the sample standard deviation with Bessel's correction of coordinate $j$ over $\mathcal{A}$, and $\varepsilon=10^{-6}$. We standardise the inputs of the classifier because the template branch could otherwise raise the predictive entropy by concentrating all transformed templates instead of removing the attribute. A positive scaling of $\bm{\mu}_\theta$ leaves the normalised template unchanged, whereas a large common offset of $\bm{\mu}_\theta$ concentrates the transformed templates in a small spherical cap. A stronger concentration replaces every $\bm{r}_i$ in a minibatch, to first order, by $\bm{\omega}\odot\bm{r}_i+\bm{t}$ with $\bm{\omega}\in\mathbb{R}^{k}$ of positive entries and $\bm{t}\in\mathbb{R}^{k}$. The standardised templates of~\eqref{eq:std} are unchanged, up to $\varepsilon$, under this replacement, and a stronger concentration therefore leaves the predictive entropy of the classifier unchanged.

A transform trained against a classifier can also move the attribute into directions that the classifier no longer uses~\cite{Elazar2018Adversarial}. We therefore store the transformed templates and the labels of the $M$ most recent minibatches in a \emph{memory} $\mathcal{M}$, which follows the history buffers of adversarial training~\cite{Shrivastava2017SimGAN}. The classifier is trained on the current minibatch together with one stored minibatch $\mathcal{B}_s$, drawn uniformly from the memory at every step of the classifier. A stored entry consists of $\bm{r}_i=\bm{r}_{\theta_s}(\bm{x}_i)$ and $c_i$, with $\theta_s$ the parameters at the step of storage. Both heads of $q_\phi$ are trained with $\theta$ fixed by minimising the classification loss
\begin{equation}
\mathcal{L}_{\mathrm{adv}}(\phi)=-\frac{1}{|\mathcal{B}'|}\sum_{i\in\mathcal{B}'}\sum_{h=1}^{2}\log q_\phi^{(h)}\bigl(c_i\mid\tilde{\bm{r}}_i\bigr),
\label{eq:adv}
\end{equation}
where $\mathcal{B}'=\mathcal{B}\sqcup\mathcal{B}_s$ is the disjoint union of the two minibatches and the standardisation~\eqref{eq:std} is taken with $\mathcal{A}=\mathcal{B}'$.

\noindent\textbf{Entropy term.} With $\phi$ fixed, the template branch raises the predictive entropy of both heads on the current minibatch by minimising the entropy term
\begin{equation}
\mathcal{L}_{\mathrm{ent}}(\theta)=-\frac{1}{2|\mathcal{B}|}\sum_{i\in\mathcal{B}}\sum_{h=1}^{2}\Ent\bigl(q_\phi^{(h)}(\cdot\mid\tilde{\bm{r}}_i)\bigr),
\label{eq:ent}
\end{equation}
where the standardisation~\eqref{eq:std} uses $\mathcal{A}=\mathcal{B}$, and gradients flow through the minibatch statistics of $\mathcal{S}_{\mathcal{B}}$ into $\theta$. The template branch minimises $\lambda\mathcal{L}_{\mathrm{ent}}$ within the total objective described later, and we call this term the privacy loss.

\subsection{Relational Term}
\label{sec:method-rel}

The privacy loss alone has a trivial minimiser, one transformed template for all embeddings, which reveals neither the attribute nor the identity. Let $\mathcal{P}_{\mathcal{B}}=\{(i,i')\in\mathcal{B}^2: i<i',\ y_i=y_{i'}\}$ be the mated pairs of the minibatch, of which there are $P\binom{K}{2}$. The relational term,
\begin{equation}
\mathcal{L}_{\mathrm{rel}}=\frac{1}{|\mathcal{P}_{\mathcal{B}}|}\sum_{(i,i')\in\mathcal{P}_{\mathcal{B}}}\bigl(\bm{r}_\theta(\bm{x}_i)^{\top}\bm{r}_\theta(\bm{x}_{i'})-\bm{x}_i^{\top}\bm{x}_{i'}\bigr)^{2}
\label{eq:rel}
\end{equation}
keeps the cosine similarity of two transformed templates of one subject close to the similarity of their embeddings, as in relational knowledge distillation~\cite{Park2019RKD}. The relational term equals the mean of $(1-\bm{x}_i^{\top}\bm{x}_{i'})^2$ over the mated pairs at the solution with one transformed template for all embeddings. This mean is positive unless the two embeddings of every mated pair are equal, and the relational term therefore penalises this solution. The term does not use attribute labels and compares only templates of the same identity.

\subsection{Variational Autoencoder}
\label{sec:method-vae}

We train the template branch within a variational autoencoder, whose decoder reconstructs the embedding from the template latent and a class latent for the attribute.

\noindent\textbf{Class branch and decoder.} A class encoder $g_\vartheta$ with parameters $\vartheta$ maps $\bm{x}$ to the mean $\bm{\mu}_\vartheta(\bm{x})$ and the log-variance $\log\bm{\sigma}_\vartheta^{2}(\bm{x})$ of the approximate posterior $q_\vartheta(\bm{z}_c\mid\bm{x})=\mathcal{N}\bigl(\bm{\mu}_\vartheta(\bm{x}),\operatorname{diag}\bm{\sigma}_\vartheta^{2}(\bm{x})\bigr)$ of the class latent $\bm{z}_c\in\mathbb{R}^{d_c}$. The class latent is sampled in training as $\bm{z}_c=\bm{\mu}_\vartheta(\bm{x})+\bm{\sigma}_\vartheta(\bm{x})\odot\bm{\epsilon}'$ with $\bm{\epsilon}'\sim\mathcal{N}(\bm{0},\bm{I}_{d_c})$ independent of $\bm{\epsilon}$. The prior of the class latent is class-conditional, $p(\bm{z}_c\mid c)=\mathcal{N}(\bm{e}_c,\bm{I}_{d_c})$, with a learnable mean $\bm{e}_c\in\mathbb{R}^{d_c}$ for each value $c$ of the attribute, initialised from $\mathcal{N}(\bm{0},\bm{I}_{d_c})$. This prior pulls the posteriors of each class towards the prior mean of that class. The objective does not separate the prior means of different classes explicitly. A decoder $f_\psi:\mathbb{R}^{k+d_c}\to\mathbb{R}^{d}$ with parameters $\psi$ reconstructs the embedding from the concatenation $[\bm{z}_r;\bm{z}_c]$ of both latents. The reconstruction $\hat{\bm{x}}$ is the decoder output normalised to unit length, provided the output is nonzero.

\noindent\textbf{Reconstruction and KL terms.} The reconstruction term
\begin{equation}
\mathcal{L}_{\mathrm{rec}}=\frac{1}{|\mathcal{B}|}\sum_{i\in\mathcal{B}}\bigl(1-\bm{x}_i^{\top}\hat{\bm{x}}_i\bigr),
\label{eq:rec}
\end{equation}
is the cosine distance, since $\|\bm{x}_i\|_2=\|\hat{\bm{x}}_i\|_2=1$, and makes the two latents retain the information of the embedding. The KL terms pull the two posteriors towards their priors, which limits the information in each latent, and are the minibatch averages of $\KL\bigl(q_\theta(\cdot\mid\bm{x}_i)\,\|\,\mathcal{N}(\bm{0},\bm{I}_k)\bigr)$ and $\KL\bigl(q_\vartheta(\cdot\mid\bm{x}_i)\,\|\,\mathcal{N}(\bm{e}_{c_i},\bm{I}_{d_c})\bigr)$,
\begin{align}
\mathcal{L}_{\mathrm{KL}}^{r}&=\frac{1}{2|\mathcal{B}|}\sum_{i\in\mathcal{B}}\sum_{j=1}^{k}\bigl(\mu_{ij}^{2}+\sigma_{ij}^{2}-\log\sigma_{ij}^{2}-1\bigr),\label{eq:klr}\\
\mathcal{L}_{\mathrm{KL}}^{c}&=\frac{1}{2|\mathcal{B}|}\sum_{i\in\mathcal{B}}\sum_{j=1}^{d_c}\bigl((\mu_{\vartheta,ij}-e_{c_i,j})^{2}+\sigma_{\vartheta,ij}^{2}\notag\\
&\qquad\qquad\qquad\qquad-\log\sigma_{\vartheta,ij}^{2}-1\bigr),\label{eq:klc}
\end{align}
where $\mu_{ij}$, $\sigma_{ij}$, $\mu_{\vartheta,ij}$ and $\sigma_{\vartheta,ij}$ are the $j$-th coordinates of $\bm{\mu}_\theta(\bm{x}_i)$, $\bm{\sigma}_\theta(\bm{x}_i)$, $\bm{\mu}_\vartheta(\bm{x}_i)$ and $\bm{\sigma}_\vartheta(\bm{x}_i)$.

\noindent\textbf{Total objective.} The template branch, the class encoder, the decoder and the prior means minimise
\begin{equation}
\mathcal{L}=\mathcal{L}_{\mathrm{rec}}+\gamma\,\mathcal{L}_{\mathrm{rel}}+\frac{\beta_r}{k}\,\mathcal{L}_{\mathrm{KL}}^{r}+\frac{\beta_c}{d_c}\,\mathcal{L}_{\mathrm{KL}}^{c}+\lambda\,\mathcal{L}_{\mathrm{ent}},
\label{eq:total}
\end{equation}
for a fixed $\phi$, with the weights $\gamma,\beta_r,\beta_c\ge0$ and each KL term divided by the dimension of its latent.

\subsection{Relation to Leakage}
\label{sec:method-interp}

The entropy term is a surrogate for the mutual information $\I(C;R)$ between the attribute and the transformed template. $C$ is uniform under the distribution $p_{\mathcal{B}}$ of Section~\ref{sec:method-obj}, and $\Ent_{p_{\mathcal{B}}}(C)=\log|\mathcal{C}|$. Let $\tilde{R}=\mathcal{S}(R)$ denote the per-feature standardisation of $R$ with the mean and the standard deviation of $R$ under $p_{\mathcal{B}}$, of which $\mathcal{S}_{\mathcal{B}}$ is the minibatch estimate. We now present two propositions (proofs in Appendix~\ref{app:proofs}), the first of which states the lower bound that the log loss of any attacker gives on $\I(C;R)$. The second states the condition under which the expected entropy term equals $\I(C;R)$ up to a constant.

\begin{proposition}[Lower bound on the mutual information]\label{prop:lower}
Let $\mathcal{T}$ be a fixed affine bijection of $\mathbb{R}^{k}$. For any joint distribution of $(C,R)$ and any classifier $q$ that maps $\mathcal{T}(\bm{r})$ to a distribution on $\mathcal{C}$, $\E[-\log q(C\mid\mathcal{T}(R))]\ge\Ent(C\mid R)$ and therefore $\I(C;R)\ge\Ent(C)-\E[-\log q(C\mid\mathcal{T}(R))]$ under that distribution. The gap is $\E\bigl[\KL\bigl(p(\cdot\mid\mathcal{T}(R))\,\|\,q(\cdot\mid\mathcal{T}(R))\bigr)\bigr]$, with $p(\cdot\mid\bm{u})$ the conditional distribution of $C$ given $\mathcal{T}(R)=\bm{u}$.
\end{proposition}

Every standardisation with positive standard deviations is such a bijection, including $\mathcal{S}$ and the standardisation of an attacker. $\Ent(C)$ minus the expected log loss of an attacker trained independently of the test set, both under the distribution of the test data, is therefore a lower bound on $\I(C;R)$. An attacker above chance thereby shows that the attribute is encoded in the transformed template, whereas an attacker at chance is not evidence of removal. We therefore report the strongest of several attackers.

\begin{proposition}[Optimum of the entropy term]\label{prop:ent}
Let every coordinate of $R$ have positive variance under $p_{\mathcal{B}}$. (i)~For every classifier $q$ and every distribution of $\tilde{R}$, $\E[\Ent(q(\cdot\mid\tilde{R}))]\le\log|\mathcal{C}|$, with equality if and only if $q(\cdot\mid\tilde{R})$ is uniform almost surely. (ii)~Under $p_{\mathcal{B}}$, let $q(c\mid\tilde{\bm{r}})=p_{\mathcal{B}}(C=c\mid\tilde{R}=\tilde{\bm{r}})$ be the posterior of $C$, which minimises the expected log loss. Then $-\E[\Ent(q(\cdot\mid\tilde{R}))]=\I(C;R)-\log|\mathcal{C}|\ge-\log|\mathcal{C}|$, with equality if and only if $C$ and $R$ are independent, that is, if and only if the class-conditional distributions of $R$ are equal.
\end{proposition}

By Proposition~\ref{prop:ent}, the expectation of the entropy term under $p_{\mathcal{B}}$ equals $\I(C;R)-\log|\mathcal{C}|$ with the posterior of~(ii) in each head and $\mathcal{S}$ in place of $\mathcal{S}_{\mathcal{B}}$. Minimising the entropy term then minimises the information about the attribute in the transformed template. The auxiliary classifier only approximates this posterior during training, since Algorithm~\ref{alg:training} updates the classifier and the template branch in turn. We therefore treat $\mathcal{L}_{\mathrm{ent}}$ as a surrogate and base every leakage claim on attackers trained against the fixed method.

\label{sec:method-train}

\begin{algorithm}[t]
\caption{TEMPER training.}
\label{alg:training}
\footnotesize
\algtext*{EndFor}
\begin{algorithmic}[1]
\Require $\mathcal{D}$, $k$, $\lambda$ and the hyperparameters of Section~\ref{sec:setup-impl}
\State $(\bm{W},\bm{b})\gets(\bm{W}_0,\bm{b}_0)$, $\bm{e}_c\sim\mathcal{N}(\bm{0},\bm{I}_{d_c})$, $\mathcal{M}\gets\emptyset$ \Comment{\eqref{eq:pcainit}}
\For{each of the $E$ epochs}
  \For{each class-balanced minibatch $\mathcal{B}$}
    \For{each of the $n_{\mathrm{adv}}$ classifier steps}
      \State draw a stored minibatch $\mathcal{B}_s$ uniformly from $\mathcal{M}$, none if $\mathcal{M}=\emptyset$
      \State $\mathcal{B}'\gets\mathcal{B}\sqcup\mathcal{B}_s$, $\tilde{\bm{r}}_i\gets\mathcal{S}_{\mathcal{B}'}(\bm{r}_i)$ for $i\in\mathcal{B}'$ \Comment{\eqref{eq:std}}
      \State $\phi\gets\textsc{Adam}(\phi,\nabla_\phi\mathcal{L}_{\mathrm{adv}})$ \Comment{\eqref{eq:adv}}
    \EndFor
    \State $\mathcal{M}\gets\textsc{Append}(\mathcal{M},\{(\bm{r}_\theta(\bm{x}_i),c_i)\}_{i\in\mathcal{B}})$ \Comment{keep the last $M$}
    \State $\bm{z}_{r,i}\sim q_\theta(\cdot\mid\bm{x}_i)$, $\bm{z}_{c,i}\sim q_\vartheta(\cdot\mid\bm{x}_i)$, $\hat{\bm{x}}_i$ from the decoder, $i\in\mathcal{B}$
    \State $\tilde{\bm{r}}_i\gets\mathcal{S}_{\mathcal{B}}(\bm{r}_\theta(\bm{x}_i))$ for $i\in\mathcal{B}$ \Comment{\eqref{eq:std}}
    \State $\Theta\gets\textsc{Adam}(\Theta,\nabla_{\Theta}\mathcal{L})$ for $\Theta=(\theta,\vartheta,\psi,\{\bm{e}_c\})$ \Comment{\eqref{eq:total}}
  \EndFor
\EndFor
\State \Return $\bm{x}\mapsto(\bm{W}\bm{x}+\bm{b})/\|\bm{W}\bm{x}+\bm{b}\|_2$
\end{algorithmic}
\end{algorithm}

\section{Experimental Setup}
\label{sec:experiments}

\subsection{Datasets and Demographic Labels}
\label{sec:setup-data}

All experiments use a frozen IResNet50 trained with ArcFace~\cite{Deng2019ArcFace} on MS1MV3~\cite{Deng2019LFR}. A reduced grid of TEMPER and PCA truncation is repeated on an IResNet100 trained on Glint360k (Appendix~\ref{app:backbone}). The embeddings are $\ell_2$-normalised before every method except DVPF, which takes the unnormalised embedding as in its public code. We train every method on the VGGFace2~\cite{cao2018vggface2} training split and evaluate on IJB-C~\cite{Whitelam2017IJBC}, RFW~\cite{Wang2019RFW} and LFW~\cite{Huang2007LFW}. Four IJB-C and 552 RFW subjects share a name with a training identity and are excluded from the leakage evaluation.

Each model removes one attribute, gender (female, male) or ethnicity. By \emph{ethnicity} we mean the 4 race groups of RFW, African, Asian, Caucasian and Indian. Appendix~\ref{app:protocol-data} gives further details about the datasets.

\subsection{Verification Protocol}
\label{sec:setup-utility}

We follow the standard 1:1 template protocol on IJB-C. Each method is applied to every face crop and its mirror before the crops are pooled into templates, as in the InsightFace evaluation. PE-MIU acts only on the pooled templates. We report the accuracy of the standard ten-fold pair protocols on RFW and LFW, averaged over the 4 groups for RFW.

We report the IJB-C TMR at FMRs of $10^{-4}$, $10^{-3}$, $10^{-2}$ and $10^{-1}$ and draw our conclusions at $10^{-3}$ and below, the operating range of verification systems. FMR $10^{-1}$ is included because DVPF was evaluated only at this FMR~\cite{Razeghi2026DVPF}, and Fig.~\ref{fig:threeway} adds FMR $10^{-5}$. The RFW threshold at FMR $10^{-3}$ is set by 12 non-mated pairs, so we do not report a lower FMR on RFW. The demographic differentials are the false non-match rate (FNMR) at a global FMR of $10^{-3}$ and the relative FMR of each group. The relative FMR of a group is its FMR over the geometric mean of the group FMRs. The Gini index of ISO/IEC 19795-10~\cite{iso19795-10,garbe} summarises the group FMRs.

\subsection{Leakage Protocol}
\label{sec:setup-leakage}

Leakage is measured by attackers trained after the method is fixed, on its transformed templates of the labelled VGGFace2 training images. They are tested on subjects outside their training data, on IJB-C for gender and on RFW for ethnicity. We test on the 10 face crops with the largest face area of each IJB-C subject, which limits the weight of subjects with many video frames, and on every image of each RFW subject.

We use 4 attacker families, of which the multilayer perceptron (MLP) attacker has 4 hidden layers of 1024 units and the logistic regression (LR) attacker one softmax layer. Both minimise a class-weighted cross-entropy for a fixed number of epochs, without model selection, on inputs standardised with the training statistics. The $k$-nearest-neighbour ($k$NN) attacker votes among the 1, 11 or 51 nearest training templates under centred cosine similarity, with raw votes or votes divided by the class priors. The support vector classifier (SVC) attacker of the DVPF evaluation~\cite{Razeghi2026DVPF} has a radial basis function kernel and class-balanced weights.

We report balanced accuracy, the mean per-class recall, whose chance level is 50\% for gender and 25\% for ethnicity. The leakage of a setting is its \emph{worst-case leakage}, the highest balanced accuracy on the test set over these attackers and the $k$NN variants. We take the worst case since the attackers detect linear, nonlinear and local dependence, and since an attacker at chance is not evidence of removal (Proposition~\ref{prop:lower}).

We call leakage near chance when it is at most 52.5\% for gender or 27.5\% for ethnicity, slightly above the chance levels of 50\% and 25\%, since a maximum of noisy estimates is biased upwards. These thresholds apply to the attackers above, trained on VGGFace2. Furthermore, we test whether these attackers understate the leakage. We train attackers on the test sets, since IJB-C and RFW differ from VGGFace2, and select an MLP attacker on held-out identities, since the attackers are tested on new identities. We also apply the attackers to the pooled IJB-C templates, which are the templates compared in verification.

\noindent\textbf{Settings, operating points and frontiers.} We refer to one trained model of a method, or one parameter value of an untrained method, as a \emph{setting}. The \emph{operating point} of a setting at an FMR is the pair of its worst-case leakage and its IJB-C TMR at this FMR, and a leakage below chance is counted as chance. We say that a setting dominates another setting at an FMR when its TMR is at least as high and its leakage is not higher, and one of the two is strictly better. The frontier of a method is the set of its non-dominated settings, selected on the test sets for every method alike, which favours methods with more settings.

\subsection{Baselines and Label-Free Controls}
\label{sec:setup-methods}

We run the baselines as described in the original works and, where available, from the public code. The learned baselines use the training data of TEMPER, and those without an original schedule train for 10 epochs. We report the details of every baseline, further baselines and the results of every setting in Appendices~\ref{app:baselines} and~\ref{app:results}.

\noindent\textbf{PFRNet.} We reimplement PFRNet~\cite{Bortolato2020PFRNet}, which has no public code, with the original architecture and losses and moment orders 1 and 2. The moment terms are summed over all pairs of groups for ethnicity, as in ASPECD~\cite{Rot2024ASPECD}. We sweep the weight $\lambda_{\mathrm{dis}}\in\{0,0.001,0.003,\dots,3,10\}$ of these terms.

\noindent\textbf{DVPF.} We train the discriminative variant of DVPF~\cite{Razeghi2024Funnel,Razeghi2026DVPF} with a 128-dimensional latent and weight $\alpha\in\{0.1,1,2,3,5,10,100\}$. DVPF (paper) follows the paper and diverges at $\alpha=100$, and DVPF (code) follows the public code. The original schedule of $\alpha$ is compressed into 10 epochs. The transformed template is decoded from one sampled latent, whereas the original work measures the leakage on the latent~\cite{Razeghi2026DVPF}.

\noindent\textbf{SensitiveNets.} We reimplement SensitiveNets~\cite{Morales2019SensitiveNetsArXiv,Morales2021SensitiveNets}, which has no public code, with one to five linear layers. Eq.~(5) of the paper prints the triplet term without the hinge, and we use the hinged form of the works cited by the paper. The detector is a softmax layer in the text of the paper and has 1024 ReLU units before the softmax in its Fig.~1. We run both and report the second as SensitiveNets (Fig.~1 detector).

\noindent\textbf{INLP.} We run INLP~\cite{Ravfogel2020INLP} with a linear softmax classifier for 1, 2, 5, 10 or 20 iterations.

\noindent\textbf{IVE and Multi-IVE.} IVE~\cite{Terhorst2019IVE} removes 250 to 500 of the standardised coordinates in 10 settings, ranked by a random forest for their importance for the attribute. Multi-IVE~\cite{Melzi2023MultiIVEPE} removes 100 to 508 principal components of the standardised embeddings. As in the original work, the five leading components are either locked against removal or not, and the Multi-IVE frontier covers both variants. The forests are fitted on the mean embedding of each identity, a deviation that makes training on 3.1 million images feasible.

\noindent\textbf{PE-MIU.} We run PE-MIU~\cite{Terhorst2020PEMIU} from the public code with block size $s\in\{4,8,16,32,64\}$. PE-MIU permutes the blocks of each template at random, so the coordinates do not keep a fixed position. We therefore add attackers on the set of blocks and an attacker that recovers the block order from a model of the unprotected embeddings. The attackers on the set of blocks include the adaptive attack that the original work proposes against PE-MIU.

\noindent\textbf{PASS.} We run PASS~\cite{Dhar2021PASS} from the public code with the schedule of the paper and the debiasing weight $\lambda_{\mathrm{deb}}\in\{0,1,3,10,30,100\}$.

\noindent\textbf{FaRM.} We run FaRM~\cite{BasuRoyChowdhury2022FaRM} from the public code without a target task, with 15 settings between 0.003 and 30 training epochs.

\noindent\textbf{Linear erasure.} LEACE~\cite{Belrose2023LEACE} is applied to the 512-dimensional embedding and, as PCA + LEACE, to the $k\in\{22,42,62,128\}$ leading principal components of the standardised embeddings.

\noindent\textbf{Label-free controls.} The label-free controls do not use attribute labels. PCA truncation keeps the $k$ leading principal components of the standardised training embeddings, and random dimensions keeps $k$ randomly chosen standardised coordinates, with $k\in\{4,6,8,10,12,16,22,42,62,128\}$.

\subsection{Implementation Details}
\label{sec:setup-impl}

\begin{figure*}[!t]
\centering
\includegraphics[width=0.9\textwidth]{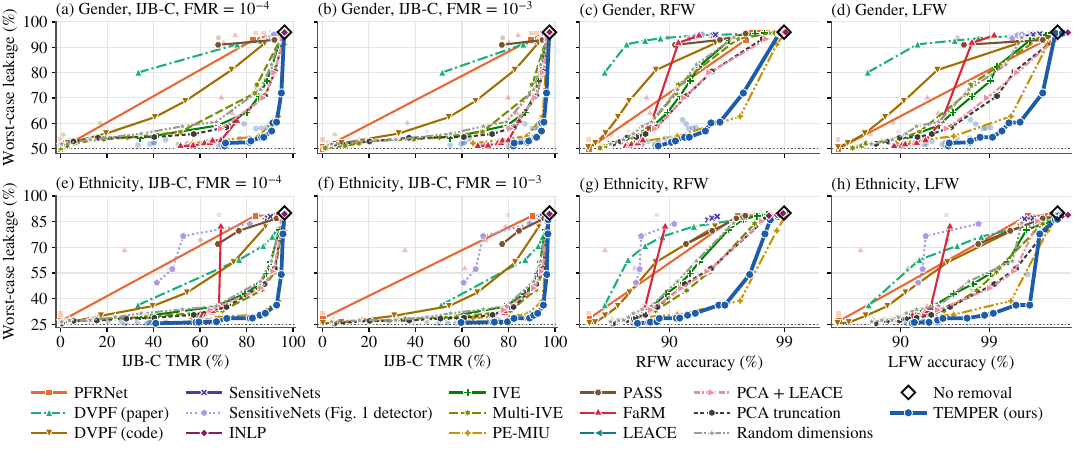}
\caption{Frontiers of every method for gender (a--d) and ethnicity (e--h). Each setting is one point, placed by worst-case leakage and by IJB-C TMR at FMR~$10^{-4}$ or $10^{-3}$, or by RFW or LFW accuracy. Dominated settings are faint, and chance is dotted.}
\label{fig:frontier}
\vspace{-8pt}
\end{figure*}

TEMPER uses one set of hyperparameters for every $k$, $\lambda$ and attribute. The class latent has $d_c=32$ dimensions, the class encoder has 3 hidden PReLU layers of 512 units, and the decoder has one hidden layer of 512 PReLU units. The MLP head of the auxiliary classifier has 3 hidden layers of 1024 LeakyReLU units with dropout 0.2, and the memory holds $M=256$ minibatches. The PCA initialisation uses 500,000 random training images. Minibatches hold $P=64$ identities with $K=4$ images each. The auxiliary classifier takes $n_{\mathrm{adv}}=3$ Adam steps at learning rate $10^{-3}$ per minibatch, and the other components take one step at $10^{-4}$. We use $\beta_r=0.1$, $\beta_c=1$, $\gamma=100$, $(\ell_{\min},\ell_{\max})=(-4,2)$ and $E=10$ epochs. Appendix~\ref{app:protocol-hyper} lists the architecture, the optimisation and the swept settings of every compared method.

We train TEMPER for $k\in\{12,22,42,62,96,128,256\}$ and $\lambda\in\{3,10,30,100\}$ for gender and $\lambda\in\{1,3,10,30\}$ for ethnicity, with $\lambda=300$ at $k=42$ for gender and $\lambda\in\{300,1000,3000\}$ at $k\in\{128,256\}$ for both attributes. Every model uses seed 42, and four settings at low leakage are repeated with seeds 1 and 2: $k=22$, $\lambda=3$ for both attributes, $k=42$, $\lambda=30$ for gender and $k=42$, $\lambda=3$ for ethnicity.

\section{Results and Discussion}
\label{sec:results}

This section presents our results and findings. We start with the measurement of leakage, since the auxiliary classifier of a method can be at chance while a separately trained attacker still predicts the attribute (Section~\ref{sec:res-attacker}). We then identify the settings whose transformed templates concentrate around one direction, because leakage is meaningful only together with the TMR at FMR~$10^{-3}$ of such settings (Section~\ref{sec:res-collapse}). Next, we ask how much of the removal by any method is due only to the lower template dimension (Section~\ref{sec:res-dimension}). We then compare the methods at the same leakage (Section~\ref{sec:res-comparison}) and ablate the components of TEMPER (Section~\ref{sec:res-ablation}). Finally, we test whether attribute removal makes verification fairer across the demographic groups (Section~\ref{sec:res-cross}).

\subsection{Attackers, Linear Erasure and Label-Free Controls}
\label{sec:res-attacker}
\label{sec:res-erasure}

Before we compare the methods, we establish when a lower leakage is evidence that a method removes the attribute. We first compare separately trained attackers with the auxiliary classifier of TEMPER, then test linear erasure against nonlinear attackers, and then compare every method with label-free reductions of the dimension. See Appendices~\ref{app:protocol} to~\ref{app:ssv} for further details on the experiments discussed in this section.

During training, the template branch of TEMPER is optimised against the auxiliary classifier, which can bring this classifier to chance without removing the attribute. To test this, we train a variant of TEMPER in which the privacy loss acts on the sample $\bm{z}_r$ of the template latent, whereas the transformed template is the normalised mean of this latent. We refer to this variant, which has a nonlinear template branch to 480 dimensions, as the sample-side variant. The auxiliary classifier of the sample-side variant predicts gender with at most 51.8\% balanced accuracy at the privacy weights $\lambda\in\{10,10^{2},10^{3},10^{4}\}$, whereas the MLP attacker predicts gender from the transformed templates of the same models with at least 62.2\%. A classifier at chance is therefore not evidence of removal when the classifier was trained together with the method. We report only attackers trained separately from the methods.

A separately trained attacker can also understate the leakage when the transformed templates concentrate around a common direction, as for the sample-side variant at $\lambda=1000$. A deep MLP trained on these templates without standardised inputs and with an unweighted loss assigns every IJB-C crop to the majority class. We therefore train every attacker with standardised inputs and class-weighted losses, and we judge every method by the attacker that predicts the attribute best.

Attackers trained on data closer to the test data could reveal more leakage. We therefore add attackers trained on the test sets or selected on held-out identities, as well as attackers applied to pooled templates. The added attackers change the worst-case leakage by at most 3.5\% for settings whose leakage is near chance and whose TMR at FMR~$10^{-3}$ is at least 50\%. We therefore report the highest balanced accuracy over the 4 attacker families trained on VGGFace2 as the worst-case leakage.

The choice of attacker matters most for linear erasure, since INLP and LEACE make the attribute unpredictable for linear classifiers only. We therefore compare a linear attacker, LR, with a nonlinear attacker, the MLP, with no removal and after INLP, LEACE and PCA + LEACE. We find that the LR attacker predicts gender with 48.0\% balanced accuracy after LEACE, against 65.1\% with no removal. The MLP attacker predicts gender with 96.0\% after LEACE, against 95.8\% with no removal. The same attacker predicts gender from PCA + LEACE as accurately as from PCA truncation to the same number of dimensions, within 1\%. Linear erasure therefore brings only the linear attacker to chance, and PCA truncation does not benefit further from linear erasure.

We also compare every method with the 2 label-free controls, PCA truncation and random dimensions. The label-free controls dominate every setting of DVPF, IVE, Multi-IVE and PASS between 54.6\% and 91.0\% gender leakage at all 4 FMRs, except 3 IVE settings (Fig.~\ref{fig:frontier}). INLP and both variants of SensitiveNets have no setting below 92\% gender leakage, and PFRNet retains less than 0.1\% TMR at FMR~$10^{-3}$ below this leakage. PCA truncation and random dimensions retain at most 5.1\% TMR at FMR~$10^{-3}$ for gender and at most 37.3\% for ethnicity at settings whose leakage is near chance (Table~\ref{tab:matched}). PCA truncation and random dimensions are therefore the reference for a removal method. We credit a removal method with a gain from the attribute labels only if it retains more TMR at FMR~$10^{-3}$ than these controls at the same leakage.

\begin{table*}[!t]
\centering
\scriptsize
\caption{TMR at a given leakage threshold. Highest IJB-C TMR (\%) at FMR~$10^{-3}$ ($10^{-4}$) among the settings with leakage at or below the column threshold (-- where no evaluated setting of the method has such a leakage), and RFW TMR (\%) at FMR~$10^{-3}$ of the setting selected at 60\% (gender) or 40\% (ethnicity). Bold marks the best value at $10^{-3}$ among the compared methods, and $^{\ast}$ marks PE-MIU at its lowest leakage, 28.6\%.}
\label{tab:matched}
\setlength{\tabcolsep}{3pt}
\begin{tabular}{@{}lccccc@{\hspace{8pt}}ccccc@{}}
\toprule
 & \multicolumn{5}{c}{Gender: worst-case leakage $\le$} & \multicolumn{5}{c}{Ethnicity: worst-case leakage $\le$} \\
\cmidrule(lr){2-6}\cmidrule(l){7-11}
Method & 52.5\% & 55\% & 60\% & 70\% & RFW & 27.5\% & 30\% & 40\% & 50\% & RFW \\
\midrule
No removal & \multicolumn{4}{c}{97.6 (96.3) at 96.0\% leakage} & 96.8 & \multicolumn{4}{c}{97.6 (96.3) at 90.0\% leakage} & 96.8 \\
\midrule
\multicolumn{11}{@{}l}{\textit{Label-free controls (reference)}} \\
PCA truncation & 5.1 (0.7) & 60.8 (40.3) & 82.3 (71.3) & 82.3 (71.3) & 24.3 & 37.3 (15.6) & 60.9 (40.4) & 82.2 (71.3) & 91.9 (87.8) & 26.3 \\
Random dimensions & 4.0 (0.5) & 31.6 (12.7) & 69.1 (52.4) & 79.0 (67.0) & 15.5 & 31.5 (12.7) & 45.0 (23.5) & 79.0 (67.0) & 79.0 (67.0) & 23.2 \\
\midrule
LEACE & -- & -- & -- & -- & -- & -- & -- & -- & -- & -- \\
INLP & -- & -- & -- & -- & -- & -- & -- & -- & -- & -- \\
PFRNet & 0.0 (0.0) & 0.0 (0.0) & 0.0 (0.0) & 0.0 (0.0) & 0.0 & 0.0 (0.0) & 0.1 (0.0) & 0.1 (0.0) & 0.1 (0.0) & 0.0 \\
SensitiveNets & -- & -- & -- & -- & -- & -- & -- & -- & -- & -- \\
PASS & -- & -- & -- & -- & -- & -- & -- & -- & -- & -- \\
SensitiveNets (Fig.~1 detector) & -- & -- & -- & -- & -- & -- & -- & -- & 60.7 (41.4) & -- \\
DVPF (paper) & -- & -- & -- & -- & -- & -- & -- & 51.0 (33.3) & 51.0 (33.3) & 1.5 \\
DVPF (code) & 8.2 (3.0) & 8.2 (3.0) & 32.7 (19.5) & 65.8 (53.6) & 0.9 & 8.1 (2.9) & 30.9 (17.3) & 54.4 (39.6) & 66.6 (54.3) & 3.9 \\
PCA + LEACE & -- & -- & 80.9 (69.2) & 80.9 (69.2) & 21.2 & -- & -- & 78.1 (64.7) & 91.2 (86.5) & 21.6 \\
IVE & -- & 54.8 (34.2) & 79.3 (67.7) & 87.2 (80.1) & 27.6 & -- & 54.2 (34.2) & 78.5 (66.6) & 90.5 (85.6) & 20.9 \\
Multi-IVE & 15.7 (6.0) & 50.4 (28.9) & 68.8 (51.5) & 68.8 (51.5) & 17.4 & 20.8 (5.5) & 57.7 (36.7) & 80.8 (69.6) & 91.9 (86.7) & 28.0 \\
PE-MIU & 64.7 (56.8) & 87.9 (85.2) & 87.9 (85.2) & 94.6 (93.6) & 67.9 & 64.7 (56.8)$^{\ast}$ & 64.7 (56.8) & 94.6 (93.6) & 94.6 (93.6) & \textbf{92.5} \\
FaRM & 76.7 (64.2) & 80.0 (69.0) & 80.0 (69.0) & 84.6 (75.8) & 19.5 & -- & 72.9 (58.5) & 79.9 (68.1) & 79.9 (68.1) & 18.9 \\
\midrule
TEMPER (ours) & \textbf{82.4} (71.0) & \textbf{92.1} (87.4) & \textbf{94.2} (90.8) & \textbf{95.2} (92.9) & \textbf{75.4} & \textbf{79.6} (67.0) & \textbf{89.1} (82.5) & \textbf{95.0} (92.9) & \textbf{95.0} (92.9) & 86.5 \\
\bottomrule
\end{tabular}
\vspace{-8pt}
\end{table*}

\subsection{Representation Collapse}
\label{sec:res-collapse}

\begin{figure*}[!t]
\centering
\includegraphics[width=0.9\textwidth]{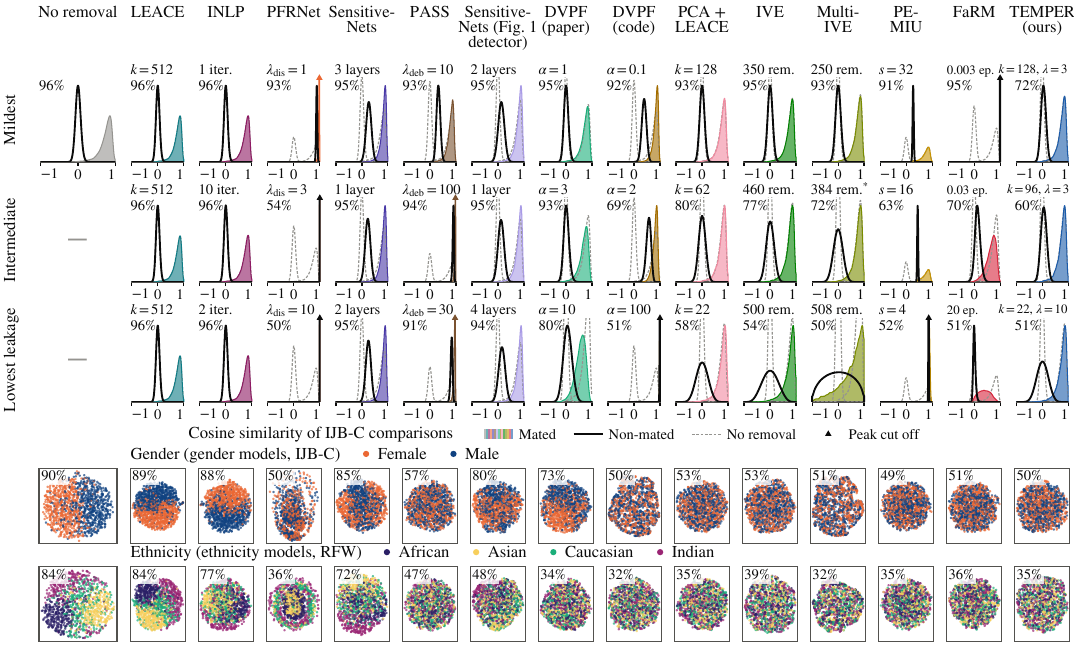}
\caption{Distributions of IJB-C mated and non-mated comparison scores for gender at three settings of each method, and UMAP~\cite{McInnes2018UMAP} maps of IJB-C crops (gender) and RFW images (ethnicity) at the lowest-leakage setting. Each map gives the balanced accuracy of the vote among the 11 nearest templates of other identities, corrected for the class priors, and $^{*}$ marks Multi-IVE with the five leading components locked.}
\label{fig:scores}
\vspace{-8pt}
\end{figure*}

A low leakage is useful in practice only if the transformed templates still discriminate identities. In this subsection, we report the attribute-removal settings whose transformed templates concentrate around one direction and discuss what this concentration does to 1:1 verification and to the measured leakage. We report the score statistics of the compared methods in Appendix~\ref{app:results}.

\looseness=-1 To identify such settings, we examine the distributions of mated and non-mated comparison scores at 3 settings of every method (Fig.~\ref{fig:scores}). We observe that the two distributions move towards each other at some settings and that both become concentrated near a cosine of 1. The two distributions stay apart for PFRNet at $\lambda_{\mathrm{dis}}=1$ and PASS at $\lambda_{\mathrm{deb}}\in\{30,100\}$, but coincide for PFRNet at $\lambda_{\mathrm{dis}}=10$ and DVPF (code) at $\alpha=100$. For instance, the mean mated and non-mated scores of DVPF (code) at $\alpha=100$ are both 0.995, whereas the two means differ by 0.741 with no removal. We say that the transformed templates of a setting are under \textit{representation collapse} when the two distributions coincide near a cosine of 1.

We observe that a method can reduce the measured leakage by degrading the transformed templates of every identity. PFRNet at $\lambda_{\mathrm{dis}}=10$ and DVPF (code) at $\alpha=100$, which are under representation collapse, have a gender leakage near chance and retain less than 0.1\% TMR at FMR~$10^{-3}$. DVPF (code) at $\alpha=10$, where the two means still differ by 0.055, has a gender leakage near chance with 8.2\% TMR at this FMR. The worst-case leakage of a setting is therefore meaningful only together with the TMR at FMR~$10^{-3}$ of the same setting.

As for TEMPER, the mated and non-mated comparison scores stay apart at every setting of Fig.~\ref{fig:scores}. At $k=22$ and $\lambda=10$, where the gender leakage of TEMPER is near chance, the two means differ by 0.743, as much as with no removal. Furthermore, the distribution of non-mated comparison scores of TEMPER is broader than with no removal, since a template with fewer dimensions spreads the cosine between unrelated templates. The mated scores remain as high as with no removal thanks to the relational term. We also report UMAP maps of the transformed templates at the lowest-leakage setting of every method (Fig.~\ref{fig:scores}). The classes occupy separate regions of the maps with no removal, LEACE and INLP, whereas TEMPER and the other methods whose lowest-leakage setting is near chance mix the classes in the maps. TEMPER therefore attains a leakage near chance without representation collapse.

\subsection{Effect of Dimensionality Reduction}
\label{sec:res-dimension}

We now discuss the effect of dimensionality reduction on the measured leakage, since the reference for a removal method is a label-free control that only reduces the dimension. We ask how much a lower template dimension reduces the leakage without attribute labels, and whether a selection of the dimensions with the labels reduces the leakage further. Dimensionality reduction is also a factor in the removal by TEMPER, the effect of which we quantify in the ablation.

We sweep the template dimension $k$ of PCA truncation and random dimensions from 4 to 128. The TMR at FMR~$10^{-3}$ of PCA truncation for gender increases from 82.3\% at $k=22$ to 97.0\% at $k=128$, against 97.6\% with no removal. The gender leakage of PCA truncation increases from 58.4\% to 93.4\% over the same range, against 96.0\% with no removal. The ethnicity leakage increases from 35.1\% to 86.5\% over this range, against 90.0\% with no removal. We observe that PCA truncation retains more TMR at FMR~$10^{-3}$ than random dimensions at every $k$, by at most 6\%, and that the two controls differ by at most 5\% in leakage. The two control curves almost coincide in Fig.~\ref{fig:frontier}. This suggests that the TMR at FMR~$10^{-3}$ and the leakage depend on the number of kept dimensions more than on a choice of these dimensions made without attribute labels.

To test whether a selection with attribute labels does better than the label-free controls, we compare PCA truncation with Multi-IVE, which selects the removed principal components with the labels. With 384 of the 512 components removed and no component locked, Multi-IVE leaks 75.2\% for gender with 90.2\% TMR at FMR~$10^{-3}$, against a leakage of 93.4\% for PCA truncation to the same 128 dimensions. PCA truncation at $k=42$ leaks 70.7\% with 92.0\% TMR at this FMR. The label-based selection therefore removes more of the attribute than PCA truncation at the same dimension, but a label-free truncation to fewer dimensions attains a lower leakage with more TMR at FMR~$10^{-3}$.

Pure dimensionality reduction therefore already gives an identity-preserving template with reduced attribute leakage, which is consistent with~\cite{Gong2019Intrinsic}. Dimensionality reduction alone does not give a leakage near chance at more than 37.3\% TMR at FMR~$10^{-3}$ (Table~\ref{tab:matched}). See Appendix~\ref{app:results} for the full results of the label-free controls and Multi-IVE.

\subsection{Comparison at Equal Leakage}
\label{sec:res-comparison}

We compare TEMPER with the prior methods and the controls at the same leakage. Table~\ref{tab:matched} lists, for each leakage threshold, the highest TMR at FMR~$10^{-3}$ of each method among the settings whose worst-case leakage is at or below the threshold. Comprehensive results for every method are given in Appendices~\ref{app:baselines} to~\ref{app:leakage}.

FaRM and PE-MIU are the prior methods with the highest TMR at FMR~$10^{-3}$ near chance for gender, with 76.7\% and 64.7\%. The prior methods retain at most 20.8\% TMR at FMR~$10^{-3}$ near chance for ethnicity. LEACE, INLP, both variants of SensitiveNets and PASS have no setting at or below 70\% gender leakage. TEMPER retains the highest TMR at FMR~$10^{-3}$ in every IJB-C column of Table~\ref{tab:matched}.

TEMPER with $22\le k\le128$ and $\lambda\le30$ dominates every other setting below 93\% gender and 83\% ethnicity leakage at all 4 FMRs, from $10^{-4}$ to $10^{-1}$ (Fig.~\ref{fig:frontier}). The exceptions are PE-MIU with $s=16$ and $s=32$, FaRM after 30 epochs, and 4 settings that leak less than every TEMPER setting with less than 6\% TMR at FMR~$10^{-3}$. FaRM after 30 epochs leaks 51.0\% for gender with 66.8\% TMR at FMR~$10^{-3}$, below every setting of TEMPER. TEMPER therefore retains the highest TMR at every other leakage threshold in this range as well, apart from these exceptions.

PE-MIU does not remove information from the template and reduces the leakage by obfuscation, and therefore retains more TMR than TEMPER at FMR~$10^{-4}$ beyond 62.8\% gender and 38.7\% ethnicity leakage (Fig.~\ref{fig:threeway}). This advantage is offset by an attacker that recovers the block order for $s\ge32$ from a model of the unprotected embeddings. The worst-case leakage of PE-MIU is then within 11\% of the leakage with no removal.

\begin{figure}[!t]
\centering
\includegraphics[width=0.75\columnwidth]{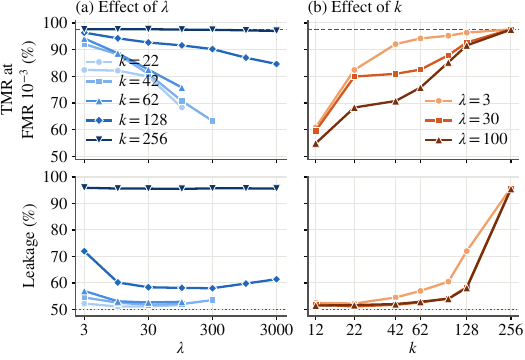}
\caption{IJB-C TMR at FMR~$10^{-3}$ and gender leakage of TEMPER against the privacy weight $\lambda$ (a) and the template dimension $k$ (b). Dotted lines mark chance, and dashed lines mark the TMR at FMR~$10^{-3}$ without removal.}
\label{fig:params}
\vspace{-8pt}
\end{figure}

\begin{figure*}[!t]
\centering
\includegraphics[width=0.8\textwidth]{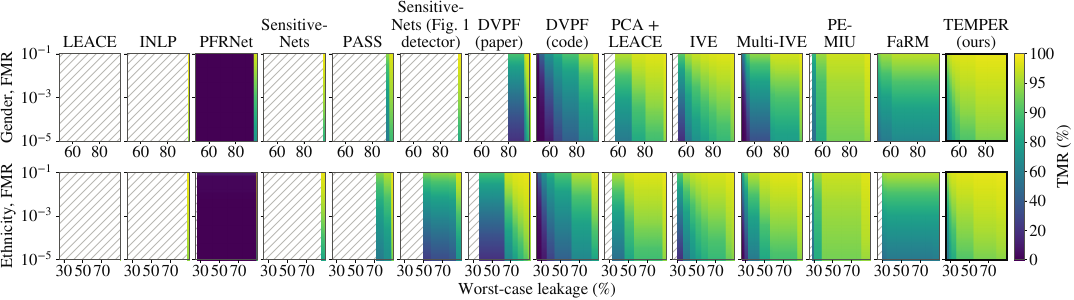}
\caption{Highest IJB-C TMR of each method up to a given worst-case leakage, at FMR $10^{-5}$ to $10^{-1}$ (hatched where no setting reaches it).}
\label{fig:threeway}
\vspace{-8pt}
\end{figure*}

\label{sec:res-stability}
To test whether the ranking in Table~\ref{tab:matched} depends on the test subjects, we build 95\% confidence intervals by resampling the 3{,}531 IJB-C subjects 1000 times. We find that TEMPER exceeds the second-highest TMR at FMR~$10^{-3}$ by at least 4.2\% at the gender thresholds up to 60\% and the ethnicity thresholds up to 30\%, by 0.6\% at the gender threshold of 70\%, and by 0.4\% at the ethnicity thresholds of 40\% and 50\%. Every difference is positive at the 95\% level. The TMR at FMR~$10^{-3}$ of the 4 repeated settings of TEMPER varies by at most 0.4\% over 3 training seeds, and the worst-case leakage by at most 0.8\%. The near-chance settings of TEMPER stay near chance for every seed. On the single-image pairs of LFW, TEMPER near chance attains 97.2\% accuracy for gender and 96.2\% for ethnicity, the highest of the compared methods, against 99.8\% with no removal. The ranking of Table~\ref{tab:matched} therefore holds under resampled test subjects and other training seeds.

To test TEMPER against attackers trained on data closer to the test data, we split each test set into identity-disjoint halves and train attackers on one half to attack the other. TEMPER at $k=42$ and $\lambda=30$ leaks 53.6\% for gender under these attackers, above the near-chance threshold, whereas TEMPER at $k=22$ and $\lambda=3$ stays near chance. Every tested ethnicity setting of TEMPER leaks above the near-chance threshold, and the 2 settings with the lowest leakage leak 30.0\% and 28.7\%. The ethnicity leakage of the TEMPER setting selected at 40\% and 50\% increases from 36.2\% to 53.4\%, which places TEMPER above PE-MIU, FaRM and Multi-IVE in leakage at both thresholds. Attackers trained on the test data are therefore stronger against TEMPER than the attackers trained on VGGFace2.

We conclude that TEMPER achieves near-chance leakage for both attributes with the highest TMR at FMR~$10^{-3}$ of the compared methods, 82.4\% for gender and 79.6\% for ethnicity, against 97.6\% without removal (Table~\ref{tab:matched}). The near-chance ethnicity leakage holds only for the attackers trained on VGGFace2.

\subsection{Ablation}
\label{sec:res-ablation}

\begin{table}[!t]
\centering
\scriptsize
\caption{Ablation of TEMPER for gender at $k=42$ (TMR at FMR~$10^{-3}$ and leakage in \%). Att.\ is the best attacker of each row, and $\Delta$ is the change against the row marked ref.}
\label{tab:ablation}
\setlength{\tabcolsep}{2pt}\renewcommand{\arraystretch}{0.92}
\begin{tabular*}{\linewidth}{@{}l@{\hspace{5pt}}l@{\extracolsep{\fill}}rrrrlrr@{}}
\toprule
 & & \multicolumn{2}{c}{TMR (\%)} & \multicolumn{5}{c}{Leakage (\%)} \\
\cmidrule(lr){3-4}\cmidrule(l){5-9}
Model & Configuration & $10^{-3}$ & $\Delta$ & Worst & $\Delta$ & Att. & MLP & LR \\
\blockrule
\multicolumn{9}{@{}l}{\textit{Removed training terms ($\lambda{=}30$)}} \\
\titlerule
\multirow{6}{*}{TEMPER} & none & 81.0 & \abref & 51.7 & \abref & MLP & 51.7 & 50.1 \\
 & VAE & 81.1 & \abbetter{$+$0.1} & 52.6 & \abworse{$+$0.9} & LR & 51.3 & 52.6 \\
 & VAE, memory & 84.0 & \abbetter{$+$3.0} & 52.4 & \abworse{$+$0.7} & MLP & 52.4 & 51.7 \\
 & VAE, relational & 49.6 & \abworse{$-$31.4} & 53.2 & \abworse{$+$1.5} & LR & 51.1 & 53.2 \\
 & VAE, relational, memory & 70.9 & \abworse{$-$10.1} & 53.6 & \abworse{$+$1.9} & LR & 52.0 & 53.6 \\
 & relational, memory (base) & 77.4 & \abworse{$-$3.6} & 51.2 & \abbetter{$-$0.5} & MLP & 51.2 & 50.2 \\
\blockrule
\multicolumn{9}{@{}l}{\textit{Encoder, initialisation and privacy loss ($\lambda{=}30$)}} \\
\titlerule
TEMPER & random initialisation & 81.9 & \abbetter{$+$0.9} & 65.8 & \abworse{$+$14.1} & MLP & 65.8 & 49.5 \\
\startrule
\multirow{3}{*}{Base} & random initialisation & 1.7 & \abworse{$-$79.3} & 71.8 & \abworse{$+$20.1} & MLP & 71.8 & 52.8 \\
 & nonlinear encoder ($k{=}480$) & 0.4 & \abworse{$-$80.6} & 64.3 & \abworse{$+$12.6} & $k$NN & 50.0 & 50.8 \\
 & \enspace loss on sampled latent & 66.4 & \abworse{$-$14.6} & 82.3 & \abworse{$+$30.6} & MLP & 82.3 & 61.2 \\
\startrule
Control & PCA truncation & 92.0 & \abbetter{$+$11.0} & 70.7 & \abworse{$+$19.0} & SVC & 68.2 & 53.9 \\
\blockrule
\multicolumn{9}{@{}l}{\textit{Other privacy weights}} \\
\titlerule
\multirow{7}{*}{TEMPER} & $\lambda{=}3$ & 92.1 & \abbetter{$+$11.1} & 54.5 & \abworse{$+$2.8} & MLP & 54.5 & 50.6 \\
 & $\lambda{=}10$ & 88.2 & \abbetter{$+$7.2} & 52.5 & \abworse{$+$0.8} & MLP & 52.5 & 50.2 \\
 & $\lambda{=}100$ & 70.8 & \abworse{$-$10.2} & 52.0 & \abworse{$+$0.3} & LR & 51.5 & 52.0 \\
 & random initialisation, $\lambda{=}3$ & 91.3 & \abbetter{$+$10.3} & 54.6 & \abworse{$+$2.9} & MLP & 54.6 & 49.8 \\
 & random initialisation, $\lambda{=}10$ & 88.7 & \abbetter{$+$7.7} & 58.8 & \abworse{$+$7.1} & MLP & 58.8 & 50.8 \\
 & no VAE, $\lambda{=}3$ & 92.0 & \abbetter{$+$11.0} & 54.5 & \abworse{$+$2.8} & MLP & 54.5 & 50.8 \\
 & no VAE, no memory, $\lambda{=}3$ & 91.7 & \abbetter{$+$10.7} & 54.1 & \abworse{$+$2.4} & MLP & 54.1 & 50.9 \\
\startrule
\multirow{6}{*}{Base} & $\lambda{=}0$ & 93.2 & \abbetter{$+$12.2} & 66.6 & \abworse{$+$14.9} & SVC & 65.1 & 53.9 \\
 & $\lambda{=}300$ & 72.0 & \abworse{$-$9.0} & 51.8 & \abworse{$+$0.1} & LR & 51.7 & 51.8 \\
 & random initialisation, $\lambda{=}0$ & 93.2 & \abbetter{$+$12.2} & 66.3 & \abworse{$+$14.6} & SVC & 64.5 & 53.3 \\
 & random initialisation, $\lambda{=}3$ & 50.2 & \abworse{$-$30.8} & 74.0 & \abworse{$+$22.3} & MLP & 74.0 & 50.3 \\
 & nonlinear encoder, $\lambda{=}3$ & 20.1 & \abworse{$-$60.9} & 61.8 & \abworse{$+$10.1} & MLP & 61.8 & 51.2 \\
 & \enspace latent loss, $\lambda{=}1000$ & 0.0 & \abworse{$-$81.0} & 63.4 & \abworse{$+$11.7} & MLP & 63.4 & 55.6 \\
\bottomrule
\end{tabular*}
\vspace{-8pt}
\end{table}

\begin{table*}[!t]
\centering
\fontsize{5.7}{6.5}\selectfont
\caption{Demographic differentials on IJB-C at global FMR~$10^{-3}$, at the 60\% (gender) and 40\% (ethnicity) settings of Table~\ref{tab:matched}. $^{\dagger}$ marks a method without such a setting of at least 1\% TMR at FMR~$10^{-3}$, shown at the lowest leakage with this TMR. The change against no removal is given in brackets. Bold marks the lowest FNMR and, below 50\% FNMR, the lowest Gini index and the relative FMRs closest to 1, excluding $^{\dagger}$ columns.}
\label{tab:fairness}
\setlength{\tabcolsep}{1.0pt}\newcommand{\fd}[1]{\kern0.4pt{\fontsize{4.8}{6}\selectfont(#1)}}
\setlength{\arrayrulewidth}{0.25pt}\setlength{\aboverulesep}{0pt}\setlength{\belowrulesep}{0pt}%
\newcommand{\ft}{\rule{0pt}{6.3pt}}\newcommand{\fb}{\rule[-3pt]{0pt}{0pt}}\newcommand{\fh}{\rule[-3pt]{0pt}{15.5pt}}%
\newcommand{\fdouble}{\specialrule{\lightrulewidth}{0pt}{0pt}&&&&&&&&&&&&&&&&\\[-5.3pt]\specialrule{\lightrulewidth}{0pt}{0pt}}%
\begin{tabular}{@{}cl*{15}{|c}@{}}
\toprule
 & \fh & \shortstack{No\\removal} & PFRNet & \shortstack{DVPF\\(paper)} & \shortstack{DVPF\\(code)} & SensitiveNets & \shortstack{SensitiveNets\\(Fig.~1)} & INLP & IVE & Multi-IVE & PE-MIU & PASS & FaRM & LEACE & \shortstack{PCA +\\LEACE} & \shortstack{TEMPER\\(ours)} \\
\midrule
\multirow{6}{*}{\rotatebox{90}{Gender}} & Leakage (\%)\ft{} & 96.0 & 92.9$^{\dagger}$\fd{\abbetter{$-$3.1}} & 80.1$^{\dagger}$\fd{\abbetter{$-$15.9}} & 56.0\fd{\abbetter{$-$40.0}} & 94.7$^{\dagger}$\fd{\abbetter{$-$1.3}} & 93.6$^{\dagger}$\fd{\abbetter{$-$2.4}} & 95.8$^{\dagger}$\fd{\abbetter{$-$0.2}} & 59.2\fd{\abbetter{$-$36.8}} & 56.6\fd{\abbetter{$-$39.4}} & 54.9\fd{\abbetter{$-$41.1}} & 91.0$^{\dagger}$\fd{\abbetter{$-$5.0}} & 53.3\fd{\abbetter{$-$42.7}} & 96.0$^{\dagger}$\fd{\abbetter{$\pm$0.0}} & 57.9\fd{\abbetter{$-$38.1}} & 57.0\fd{\abbetter{$-$39.0}} \\
 & FNMR (\%) & 2.3 & 10.8\fd{\abworse{$+$8.5}} & 49.7\fd{\abworse{$+$47.4}} & 68.2\fd{\abworse{$+$65.9}} & 4.8\fd{\abworse{$+$2.5}} & 5.9\fd{\abworse{$+$3.6}} & 2.5\fd{\abworse{$+$0.2}} & 21.3\fd{\abworse{$+$19.0}} & 32.3\fd{\abworse{$+$30.0}} & 12.3\fd{\abworse{$+$10.0}} & 23.5\fd{\abworse{$+$21.2}} & 20.6\fd{\abworse{$+$18.3}} & 2.3\fd{\abbetter{$\pm$0.0}} & 19.8\fd{\abworse{$+$17.5}} & \textbf{5.9}\fd{\abworse{$+$3.6}} \\
 & Rel.\ FMR, female & 1.35 & 1.56\fd{\abworse{$+$0.21}} & 0.60\fd{\abworse{$-$0.75}} & 1.17\fd{\abbetter{$-$0.18}} & 1.64\fd{\abworse{$+$0.29}} & 1.48\fd{\abworse{$+$0.13}} & 1.37\fd{\abworse{$+$0.02}} & 1.07\fd{\abbetter{$-$0.28}} & 1.16\fd{\abbetter{$-$0.19}} & \textbf{0.98}\fd{\abbetter{$-$0.37}} & 1.58\fd{\abworse{$+$0.23}} & 0.95\fd{\abbetter{$-$0.40}} & 1.36\fd{\abworse{$+$0.01}} & 1.03\fd{\abbetter{$-$0.32}} & 1.04\fd{\abbetter{$-$0.31}} \\
 & Rel.\ FMR, male & 0.74 & 0.64\fd{\abworse{$-$0.10}} & 1.67\fd{\abworse{$+$0.93}} & 0.86\fd{\abbetter{$+$0.12}} & 0.61\fd{\abworse{$-$0.13}} & 0.67\fd{\abworse{$-$0.07}} & 0.73\fd{\abworse{$-$0.01}} & 0.94\fd{\abbetter{$+$0.20}} & 0.86\fd{\abbetter{$+$0.12}} & \textbf{1.02}\fd{\abbetter{$+$0.28}} & 0.63\fd{\abworse{$-$0.11}} & 1.05\fd{\abbetter{$+$0.31}} & 0.74\fd{\abbetter{$\pm$0.00}} & 0.97\fd{\abbetter{$+$0.23}} & 0.96\fd{\abbetter{$+$0.22}} \\
 & Gini, FMR & 0.29 & 0.42\fd{\abworse{$+$0.13}} & 0.47\fd{\abworse{$+$0.18}} & 0.15\fd{\abbetter{$-$0.14}} & 0.46\fd{\abworse{$+$0.17}} & 0.37\fd{\abworse{$+$0.08}} & 0.30\fd{\abworse{$+$0.01}} & 0.07\fd{\abbetter{$-$0.22}} & 0.15\fd{\abbetter{$-$0.14}} & \textbf{0.02}\fd{\abbetter{$-$0.27}} & 0.43\fd{\abworse{$+$0.14}} & 0.05\fd{\abbetter{$-$0.24}} & 0.29\fd{\abbetter{$\pm$0.00}} & 0.03\fd{\abbetter{$-$0.26}} & 0.04\fd{\abbetter{$-$0.25}} \\
 & Gini, FNMR\fb{} & 0.31 & 0.23\fd{\abbetter{$-$0.08}} & 0.07\fd{\abbetter{$-$0.24}} & 0.03\fd{\abbetter{$-$0.28}} & 0.31\fd{\abbetter{$\pm$0.00}} & 0.30\fd{\abbetter{$-$0.01}} & 0.28\fd{\abbetter{$-$0.03}} & 0.14\fd{\abbetter{$-$0.17}} & \textbf{0.10}\fd{\abbetter{$-$0.21}} & 0.22\fd{\abbetter{$-$0.09}} & 0.13\fd{\abbetter{$-$0.18}} & 0.14\fd{\abbetter{$-$0.17}} & 0.31\fd{\abbetter{$\pm$0.00}} & 0.13\fd{\abbetter{$-$0.18}} & 0.27\fd{\abbetter{$-$0.04}} \\
\fdouble
\multirow{8}{*}{\rotatebox{90}{Ethnicity}} & Leakage (\%)\ft{} & 90.0 & 88.5$^{\dagger}$\fd{\abbetter{$-$1.5}} & 36.3\fd{\abbetter{$-$53.7}} & 35.7\fd{\abbetter{$-$54.3}} & 85.7$^{\dagger}$\fd{\abbetter{$-$4.3}} & 49.3$^{\dagger}$\fd{\abbetter{$-$40.7}} & 88.4$^{\dagger}$\fd{\abbetter{$-$1.6}} & 35.4\fd{\abbetter{$-$54.6}} & 33.0\fd{\abbetter{$-$57.0}} & 38.7\fd{\abbetter{$-$51.3}} & 71.9$^{\dagger}$\fd{\abbetter{$-$18.1}} & 34.5\fd{\abbetter{$-$55.5}} & 89.5$^{\dagger}$\fd{\abbetter{$-$0.5}} & 33.4\fd{\abbetter{$-$56.6}} & 36.2\fd{\abbetter{$-$53.8}} \\
 & FNMR (\%) & 2.3 & 9.5\fd{\abworse{$+$7.2}} & 50.2\fd{\abworse{$+$47.9}} & 46.5\fd{\abworse{$+$44.2}} & 7.1\fd{\abworse{$+$4.8}} & 39.1\fd{\abworse{$+$36.8}} & 2.8\fd{\abworse{$+$0.5}} & 22.3\fd{\abworse{$+$20.0}} & 19.8\fd{\abworse{$+$17.5}} & 5.4\fd{\abworse{$+$3.1}} & 23.3\fd{\abworse{$+$21.0}} & 20.8\fd{\abworse{$+$18.5}} & 2.3\fd{\abbetter{$\pm$0.0}} & 22.5\fd{\abworse{$+$20.2}} & \textbf{5.1}\fd{\abworse{$+$2.8}} \\
 & Rel.\ FMR, African & 1.42 & 1.62\fd{\abworse{$+$0.20}} & 1.16\fd{\abbetter{$-$0.26}} & 2.05\fd{\abworse{$+$0.63}} & 1.57\fd{\abworse{$+$0.15}} & 1.68\fd{\abworse{$+$0.26}} & 1.46\fd{\abworse{$+$0.04}} & 1.02\fd{\abbetter{$-$0.40}} & 1.23\fd{\abbetter{$-$0.19}} & 1.13\fd{\abbetter{$-$0.29}} & 1.56\fd{\abworse{$+$0.14}} & 1.11\fd{\abbetter{$-$0.31}} & 1.49\fd{\abworse{$+$0.07}} & \textbf{1.02}\fd{\abbetter{$-$0.40}} & 1.17\fd{\abbetter{$-$0.25}} \\
 & Rel.\ FMR, Asian & 1.69 & 1.56\fd{\abbetter{$-$0.13}} & 0.94\fd{\abbetter{$-$0.75}} & 1.14\fd{\abbetter{$-$0.55}} & 1.56\fd{\abbetter{$-$0.13}} & 1.40\fd{\abbetter{$-$0.29}} & 1.57\fd{\abbetter{$-$0.12}} & 1.09\fd{\abbetter{$-$0.60}} & 1.02\fd{\abbetter{$-$0.67}} & 1.19\fd{\abbetter{$-$0.50}} & 1.40\fd{\abbetter{$-$0.29}} & 0.96\fd{\abbetter{$-$0.73}} & 1.70\fd{\abworse{$+$0.01}} & \textbf{1.09}\fd{\abbetter{$-$0.60}} & 1.19\fd{\abbetter{$-$0.50}} \\
 & Rel.\ FMR, Caucasian & 0.33 & 0.39\fd{\abbetter{$+$0.06}} & 1.24\fd{\abbetter{$+$0.91}} & 0.58\fd{\abbetter{$+$0.25}} & 0.37\fd{\abbetter{$+$0.04}} & 0.36\fd{\abbetter{$+$0.03}} & 0.36\fd{\abbetter{$+$0.03}} & 0.91\fd{\abbetter{$+$0.58}} & 0.91\fd{\abbetter{$+$0.58}} & 0.78\fd{\abbetter{$+$0.45}} & 0.33\fd{\abbetter{$\pm$0.00}} & 0.94\fd{\abbetter{$+$0.61}} & 0.34\fd{\abbetter{$+$0.01}} & \textbf{0.97}\fd{\abbetter{$+$0.64}} & 0.76\fd{\abbetter{$+$0.43}} \\
 & Rel.\ FMR, Indian & 1.27 & 1.02\fd{\abbetter{$-$0.25}} & 0.74\fd{\abbetter{$-$0.53}} & 0.73\fd{\abbetter{$-$0.54}} & 1.11\fd{\abbetter{$-$0.16}} & 1.18\fd{\abbetter{$-$0.09}} & 1.21\fd{\abbetter{$-$0.06}} & 1.00\fd{\abbetter{$-$0.27}} & 0.87\fd{\abbetter{$-$0.40}} & 0.96\fd{\abbetter{$-$0.31}} & 1.39\fd{\abworse{$+$0.12}} & 1.00\fd{\abbetter{$-$0.27}} & 1.15\fd{\abbetter{$-$0.12}} & \textbf{0.93}\fd{\abbetter{$-$0.34}} & 0.95\fd{\abbetter{$-$0.32}} \\
 & Gini, FMR & 0.30 & 0.31\fd{\abworse{$+$0.01}} & 0.14\fd{\abbetter{$-$0.16}} & 0.36\fd{\abworse{$+$0.06}} & 0.30\fd{\abbetter{$\pm$0.00}} & 0.30\fd{\abbetter{$\pm$0.00}} & 0.28\fd{\abbetter{$-$0.02}} & 0.05\fd{\abbetter{$-$0.25}} & 0.10\fd{\abbetter{$-$0.20}} & 0.12\fd{\abbetter{$-$0.18}} & 0.26\fd{\abbetter{$-$0.04}} & 0.05\fd{\abbetter{$-$0.25}} & 0.31\fd{\abworse{$+$0.01}} & \textbf{0.04}\fd{\abbetter{$-$0.26}} & 0.12\fd{\abbetter{$-$0.18}} \\
 & Gini, FNMR\fb{} & 0.22 & 0.09\fd{\abbetter{$-$0.13}} & 0.09\fd{\abbetter{$-$0.13}} & \textbf{0.04}\fd{\abbetter{$-$0.18}} & 0.12\fd{\abbetter{$-$0.10}} & 0.05\fd{\abbetter{$-$0.17}} & 0.20\fd{\abbetter{$-$0.02}} & 0.06\fd{\abbetter{$-$0.16}} & 0.05\fd{\abbetter{$-$0.17}} & 0.14\fd{\abbetter{$-$0.08}} & 0.05\fd{\abbetter{$-$0.17}} & 0.06\fd{\abbetter{$-$0.16}} & 0.23\fd{\abworse{$+$0.01}} & 0.08\fd{\abbetter{$-$0.14}} & 0.12\fd{\abbetter{$-$0.10}} \\
\bottomrule
\end{tabular}
\vspace{-8pt}
\end{table*}

To explain the performance of TEMPER and the gain from each component, we ablate the components for gender at $k=42$ (Table~\ref{tab:ablation}). We refer to the PCA-initialised linear encoder with the VAE and the privacy loss on the transformed template, without the relational term and the memory, as the base model.

Our attackers predict the attribute from the transformed template, and the privacy loss should therefore act on the transformed template. To test this, we train a model with the privacy loss on the sampled latent, which also has a nonlinear encoder with $k=480$. The leakage then stays above 60\% at every privacy weight (Table~\ref{tab:ablation}). The TMR at FMR~$10^{-3}$ decreases from 66.4\% at $\lambda=30$ to 0.0\% at $\lambda=1000$, while the auxiliary classifier is at chance. A privacy loss on the sampled latent therefore does not remove the attribute from the transformed template of this model.

To see whether a nonlinear encoder removes more of the attribute than the linear encoder, we replace the linear encoder by an MLP with $k=480$. The nonlinear encoder with the privacy loss on the standardised transformed template retains 20.1\% TMR at FMR~$10^{-3}$ with 61.8\% leakage at $\lambda=3$ (Table~\ref{tab:ablation}). The same encoder retains 0.4\% TMR at FMR~$10^{-3}$ with 64.3\% leakage at $\lambda=30$. The nonlinear encoder reduces the leakage only together with the TMR at FMR~$10^{-3}$, whereas the base model retains 77.4\% TMR at this FMR with 51.2\% leakage at $\lambda=30$. We observe that dimensionality reduction with a linear encoder plays a significant role in attribute removal.

To test whether the encoder needs the PCA initialisation, we initialise the linear encoder at random, in the base model and in TEMPER. The randomly initialised base model retains 1.7\% TMR at FMR~$10^{-3}$ and $\lambda=30$, against 77.4\% from the PCA initialisation (Table~\ref{tab:ablation}). The leakage of the randomly initialised base model increases to 71.8\% at $\lambda=30$, above the 66.3\% without the privacy loss. TEMPER from a random initialisation retains the TMR at FMR~$10^{-3}$ of TEMPER, but the gender leakage increases from 54.6\% at $\lambda=3$ to 65.8\% at $\lambda=30$. TEMPER from the PCA initialisation leaks 51.7\% at $\lambda=30$. The PCA initialisation is therefore necessary with and without the relational term and the memory.

We compare TEMPER with PCA truncation at the same $k$ to isolate the gain of the training with attribute labels over the initialisation. PCA truncation at $k=22$ reduces the gender leakage from 96.0\% to 58.4\%. TEMPER at the same $k$ and $\lambda=3$ reduces the leakage further, to 52.2\%. The truncation therefore accounts for more than 80\% of the reduction of the leakage by TEMPER at $k=22$, for both attributes and every privacy weight. The training against the auxiliary classifier removes the remaining leakage. The share of the training grows with $k$, from 14\% of the reduction at $k=22$ to 39\% at $k=42$, at $\lambda=3$ for gender. On the IResNet100 trained on Glint360k, TEMPER at $k=42$ and $\lambda=3$ leaks 53.7\% for gender, against 65.6\% for PCA truncation at $k=42$. TEMPER therefore attains a given leakage at a larger $k$ than the controls and retains more TMR at FMR~$10^{-3}$ at equal leakage (Fig.~\ref{fig:frontier}).

We add the relational term to the models without the VAE and the memory to test whether the term prevents the loss of TMR at FMR~$10^{-3}$. The relational term raises the gender TMR at FMR~$10^{-3}$ and $\lambda=30$ from 70.9\% to 84.0\% and reduces the leakage from 53.6\% to 52.4\% (Table~\ref{tab:ablation}). For ethnicity at $\lambda=3$, the relational term raises the TMR at FMR~$10^{-3}$ from 41.0\% to 83.9\%. The relational term gives the largest gain in TMR at FMR~$10^{-3}$ of the VAE, the relational term and the memory, for both attributes.

The VAE and the memory have smaller effects. The VAE reduces the gender leakage at $\lambda=30$ from 52.6\% to 51.7\% without a cost in TMR at FMR~$10^{-3}$. The VAE also keeps this setting near chance for all 3 training seeds, against 1 of 3 seeds without the VAE. The memory, which trains the auxiliary classifier on the transformed templates of earlier steps, does not raise the TMR at equal leakage. The memory changes the gender operating points by less than 3\% with the relational term, and raises the TMR at FMR~$10^{-3}$ from 83.9\% to 87.6\% and the leakage from 27.2\% to 28.8\% for ethnicity at $\lambda=3$. We keep the memory since it was fixed with the other training weights before the sweeps.

We vary $k$ and $\lambda$ to measure how much each parameter moves the operating point of TEMPER (Fig.~\ref{fig:params}). A larger privacy weight $\lambda$ reduces the TMR at FMR~$10^{-3}$ more than the gender leakage up to $k=62$. As for $k$, the leakage increases with $k$ from $k=96$ on, and a larger $\lambda$ reduces the leakage at $k=128$ from 72.0\% at $\lambda=3$ to 58.0\% at $\lambda=300$. TEMPER at $k=256$ leaks at least 95.5\% for gender at every privacy weight, against 96.0\% with no removal. We find that the privacy loss becomes more effective at a reduced number of dimensions.

We conclude that the reduction of the dimension by the PCA-initialised linear encoder accounts for more than 80\% of the removal by TEMPER at $k=22$, and the privacy loss on the transformed template removes the remaining leakage. The relational term prevents the loss of TMR at FMR~$10^{-3}$ under this loss, whereas the VAE and the memory have smaller effects.

We report the full ablation for both attributes in Appendix~\ref{app:ablation}, the comparison with PCA truncation at every $k$ and the training seeds in Appendix~\ref{app:results}, and the IResNet100 in Appendix~\ref{app:backbone}.

\subsection{Bias and Fairness Analysis}
\label{sec:res-cross}

\label{sec:res-fairness}
Fairness metrics measure whether a verification system treats every group alike, that is, whether the error rates are homogeneous across groups. We therefore compare the error rates of the groups on IJB-C at a common threshold, for the settings selected at the 60\% gender and 40\% ethnicity thresholds of Table~\ref{tab:matched} with at least 1\% TMR at FMR~$10^{-3}$ (Table~\ref{tab:fairness}). Every method with such a setting brings the relative FMRs of the groups closer to 1 than with no removal and reduces the Gini index of the group FMRs, for gender and, except DVPF (code), for ethnicity. The other prior methods have no such setting. TEMPER reduces the Gini index of the gender FMRs from 0.29 to 0.04 at the lowest FNMR of these methods. The same index for ethnicity decreases from 0.30 to 0.12. TEMPER reduces the FMR of every group and raises the FNMR of every group, and the FNMR of women stays above the FNMR of men, at 7.8\% against 4.5\%. The groups are therefore treated more alike at a common threshold, at a higher FNMR. We report the full results, and the same analysis on RFW at the equal error rate for all the methods, in Appendix~\ref{app:fairness}.

\section{Conclusion}
\label{sec:conclusion}

We presented TEMPER, a post-hoc linear transform initialised at the PCA solution and trained against an auxiliary attribute classifier with a relational term on mated pairs. We compared TEMPER with ten prior post-hoc methods and two label-free controls on gender and ethnicity removal, trained on VGGFace2 and evaluated on IJB-C, RFW and LFW, with leakage as the worst case over four attacker families.

We found that pure dimensionality reduction already gives an identity-preserving template with reduced attribute leakage, and PCA truncation retains more TMR at FMR~$10^{-3}$ than at least 7 of the 10 prior methods at every gender leakage threshold above near chance. We also found that a method can reduce the measured leakage by representation collapse, under which the transformed templates no longer discriminate identities. The leakage of a setting is therefore meaningful only together with its TMR at FMR~$10^{-3}$. Attribute removal brings the false match rates of the demographic groups closer and lowers their Gini index, at a higher false non-match rate. TEMPER retains the highest IJB-C TMR at FMR~$10^{-3}$ at every leakage threshold for both attributes, and the improvement near chance is attributed to the PCA initialisation and the relational term.

TEMPER therefore stands out among the compared methods, since it combines this performance with mated and non-mated comparison scores that stay apart at every setting of Fig.~\ref{fig:scores}. Attackers trained on identity-disjoint halves of the test sets raise the ethnicity leakage of every tested TEMPER setting above the near-chance threshold. Future work includes multi-attribute removal.

\section*{Acknowledgments}

This work has received funding from the European Union's Horizon Europe programme under Grant Agreement No.~101189650 (CERTAIN), and the Swiss State Secretariat for Education, Research and Innovation (SERI).

\bibliographystyle{IEEEtran}
\bibliography{refs}

\begin{thebibliography}{10}
\providecommand{\url}[1]{#1}
\csname url@samestyle\endcsname
\providecommand{\newblock}{\relax}
\providecommand{\bibinfo}[2]{#2}
\providecommand{\BIBentrySTDinterwordspacing}{\spaceskip=0pt\relax}
\providecommand{\BIBentryALTinterwordstretchfactor}{4}
\providecommand{\BIBentryALTinterwordspacing}{\spaceskip=\fontdimen2\font plus
\BIBentryALTinterwordstretchfactor\fontdimen3\font minus
  \fontdimen4\font\relax}
\providecommand{\BIBforeignlanguage}[2]{{%
\expandafter\ifx\csname l@#1\endcsname\relax
\typeout{** WARNING: IEEEtran.bst: No hyphenation pattern has been}%
\typeout{** loaded for the language `#1'. Using the pattern for}%
\typeout{** the default language instead.}%
\else
\language=\csname l@#1\endcsname
\fi
#2}}
\providecommand{\BIBdecl}{\relax}
\BIBdecl

\bibitem{terhoerst2021soft}
P.~Terh{\"o}rst, D.~F{\"a}hrmann, N.~Damer, F.~Kirchbuchner, and A.~Kuijper,
  ``On soft-biometric information stored in biometric face embeddings,''
  \emph{IEEE Trans. Biometrics Behav. Identity Sci.}, vol.~3, no.~4, pp.
  519--534, 2021.

\bibitem{Terhorst2020BeyondID}
------, ``Beyond identity: What information is stored in biometric face
  templates?'' in \emph{Proc. IEEE Int. Joint Conf. Biometrics (IJCB)}, 2020,
  pp. 1--10.

\bibitem{OsorioRoig2022Attack}
D.~Osorio-Roig, C.~Rathgeb, P.~Drozdowski, P.~Terh{\"o}rst, V.~{\v{S}}truc, and
  C.~Busch, ``An attack on facial soft-biometric privacy enhancement,''
  \emph{IEEE Trans. Biometrics Behav. Identity Sci.}, vol.~4, no.~2, pp.
  263--275, 2022.

\bibitem{GDPR2016}
{European Parliament and Council of the European Union}, ``Regulation ({EU})
  2016/679 ({General Data Protection Regulation}),'' {Official Journal of the
  European Union}, L 119, pp. 1--88, 2016.

\bibitem{Terhorst2019IVE}
P.~Terh{\"o}rst, N.~Damer, F.~Kirchbuchner, and A.~Kuijper, ``Suppressing
  gender and age in face templates using incremental variable elimination,'' in
  \emph{Proc. Int. Conf. Biometrics (ICB)}, 2019, pp. 1--8.

\bibitem{Melzi2023MultiIVEPE}
P.~Melzi \emph{et~al.}, ``{Multi-IVE}: Privacy enhancement of multiple
  soft-biometrics in face embeddings,'' in \emph{Proc. IEEE/CVF Winter Conf.
  Appl. Comput. Vis. Workshops (WACVW)}, 2023, pp. 323--331.

\bibitem{Ravfogel2020INLP}
S.~Ravfogel, Y.~Elazar, H.~Gonen, M.~Twiton, and Y.~Goldberg, ``Null it out:
  Guarding protected attributes by iterative nullspace projection,'' in
  \emph{Proc. Annu. Meeting Assoc. Comput. Linguistics (ACL)}, 2020, pp.
  7237--7256.

\bibitem{Belrose2023LEACE}
N.~Belrose, D.~Schneider-Joseph, S.~Ravfogel, R.~Cotterell, E.~Raff, and
  S.~Biderman, ``{LEACE}: Perfect linear concept erasure in closed form,'' in
  \emph{Proc. Adv. Neural Inf. Process. Syst. (NeurIPS)}, vol.~36, 2023, pp.
  66\,044--66\,063.

\bibitem{Bortolato2020PFRNet}
B.~Bortolato \emph{et~al.}, ``Learning privacy-enhancing face representations
  through feature disentanglement,'' in \emph{Proc. IEEE Int. Conf. Autom. Face
  Gesture Recognit. (FG)}, 2020, pp. 495--502.

\bibitem{Rot2024ASPECD}
P.~Rot, P.~Terh{\"o}rst, P.~Peer, and V.~{\v{S}}truc, ``{ASPECD}: Adaptable
  soft-biometric privacy-enhancement using centroid decoding for face
  verification,'' in \emph{Proc. IEEE Int. Conf. Autom. Face Gesture Recognit.
  (FG)}, 2024, pp. 1--11.

\bibitem{Terhorst2020PEMIU}
P.~Terh{\"o}rst \emph{et~al.}, ``{PE-MIU}: A training-free privacy-enhancing
  face recognition approach based on minimum information units,'' \emph{IEEE
  Access}, vol.~8, pp. 93\,635--93\,647, 2020.

\bibitem{Morales2021SensitiveNets}
A.~Morales, J.~Fierrez, R.~Vera-Rodriguez, and R.~Tolosana, ``{SensitiveNets}:
  Learning agnostic representations with application to face images,''
  \emph{IEEE Trans. Pattern Anal. Mach. Intell.}, vol.~43, no.~6, pp.
  2158--2164, 2021.

\bibitem{Razeghi2024Funnel}
B.~Razeghi, P.~Rahimi, and S.~Marcel, ``Deep variational privacy funnel:
  General modeling with applications in face recognition,'' in \emph{Proc. IEEE
  Int. Conf. Acoust., Speech Signal Process. (ICASSP)}, 2024, pp. 4920--4924.

\bibitem{Razeghi2026DVPF}
------, ``Deep privacy funnel model: From a discriminative to a generative
  approach with an application to face recognition,'' \emph{IEEE Trans. Inf.
  Forensics Security}, vol.~21, pp. 5181--5192, 2026.

\bibitem{Roy2019MaxEnt}
P.~C. Roy and V.~N. Boddeti, ``Mitigating information leakage in image
  representations: A maximum entropy approach,'' in \emph{Proc. IEEE/CVF Conf.
  Comput. Vis. Pattern Recognit. (CVPR)}, 2019, pp. 2581--2589.

\bibitem{Elazar2018Adversarial}
Y.~Elazar and Y.~Goldberg, ``Adversarial removal of demographic attributes from
  text data,'' in \emph{Proc. Conf. Empirical Methods Natural Lang. Process.
  (EMNLP)}, 2018, pp. 11--21.

\bibitem{Meden2021Survey}
B.~Meden \emph{et~al.}, ``Privacy-enhancing face biometrics: A comprehensive
  survey,'' \emph{IEEE Trans. Inf. Forensics Security}, vol.~16, pp.
  4147--4183, 2021.

\bibitem{Gong2020DebFace}
S.~Gong, X.~Liu, and A.~K. Jain, ``Jointly de-biasing face recognition and
  demographic attribute estimation,'' in \emph{Proc. Eur. Conf. Comput. Vis.
  (ECCV)}, 2020, pp. 330--347.

\bibitem{Chen2025FaceCPFNet}
Z.~Chen, Z.~Yao, B.~Jin, J.~Ning, and M.~Lin, ``{Face-CPFNet}: Leveraging
  disentangled representations for dual-level soft-biometric
  privacy-enhancement,'' \emph{IEEE Trans. Dependable Secure Comput.}, vol.~22,
  no.~6, pp. 7060--7076, 2025.

\bibitem{Mirjalili2020PrivacyNet}
V.~Mirjalili, S.~Raschka, and A.~Ross, ``{PrivacyNet}: Semi-adversarial
  networks for multi-attribute face privacy,'' \emph{IEEE Trans. Image
  Process.}, vol.~29, pp. 9400--9412, 2020.

\bibitem{Dhar2021PASS}
P.~Dhar, J.~Gleason, A.~Roy, C.~D. Castillo, and R.~Chellappa, ``{PASS}:
  Protected attribute suppression system for mitigating bias in face
  recognition,'' in \emph{Proc. IEEE/CVF Int. Conf. Comput. Vis. (ICCV)}, 2021,
  pp. 15\,067--15\,076.

\bibitem{BasuRoyChowdhury2022FaRM}
S.~Basu Roy~Chowdhury and S.~Chaturvedi, ``Learning fair representations via
  rate-distortion maximization,'' \emph{Trans. Assoc. Comput. Linguistics},
  vol.~10, pp. 1159--1174, 2022.

\bibitem{Terhorst2019Noise}
P.~Terh{\"o}rst, N.~Damer, F.~Kirchbuchner, and A.~Kuijper, ``Unsupervised
  privacy-enhancement of face representations using similarity-sensitive noise
  transformations,'' \emph{Appl. Intell.}, vol.~49, no.~8, pp. 3043--3060,
  2019.

\bibitem{Kung2017Compressive}
S.-Y. Kung, ``Compressive privacy: From information/estimation theory to
  machine learning [lecture notes],'' \emph{IEEE Signal Process. Mag.},
  vol.~34, no.~1, pp. 94--112, 2017.

\bibitem{Gong2019Intrinsic}
S.~Gong, V.~N. Boddeti, and A.~K. Jain, ``On the intrinsic dimensionality of
  image representations,'' in \emph{Proc. IEEE/CVF Conf. Comput. Vis. Pattern
  Recognit. (CVPR)}, 2019, pp. 3987--3996.

\bibitem{Olfat2019FairPCA}
M.~Olfat and A.~Aswani, ``Convex formulations for fair principal component
  analysis,'' in \emph{Proc. AAAI Conf. Artif. Intell.}, vol.~33, no.~1, 2019,
  pp. 663--670.

\bibitem{Kleindessner2023FairPCA}
M.~Kleindessner, M.~Donini, C.~Russell, and M.~B. Zafar, ``Efficient fair {PCA}
  for fair representation learning,'' in \emph{Proc. Int. Conf. Artif. Intell.
  Statist. (AISTATS)}, vol. 206, 2023, pp. 5250--5270.

\bibitem{Ravfogel2022Kernel}
S.~Ravfogel, F.~Vargas, Y.~Goldberg, and R.~Cotterell, ``Adversarial concept
  erasure in kernel space,'' in \emph{Proc. Conf. Empirical Methods Natural
  Lang. Process. (EMNLP)}, 2022, pp. 6034--6055.

\bibitem{Edwards2016Censoring}
H.~Edwards and A.~Storkey, ``Censoring representations with an adversary,'' in
  \emph{Proc. Int. Conf. Learn. Represent. (ICLR)}, 2016.

\bibitem{Ganin2016DANN}
Y.~Ganin \emph{et~al.}, ``Domain-adversarial training of neural networks,''
  \emph{J. Mach. Learn. Res.}, vol.~17, no.~59, pp. 1--35, 2016.

\bibitem{Madras2018LAFTR}
D.~Madras, E.~Creager, T.~Pitassi, and R.~Zemel, ``Learning adversarially fair
  and transferable representations,'' in \emph{Proc. Int. Conf. Mach. Learn.
  (ICML)}, vol.~80, 2018, pp. 3384--3393.

\bibitem{Xie2017Controllable}
Q.~Xie, Z.~Dai, Y.~Du, E.~Hovy, and G.~Neubig, ``Controllable invariance
  through adversarial feature learning,'' in \emph{Proc. Adv. Neural Inf.
  Process. Syst. (NIPS)}, vol.~30, 2017.

\bibitem{Louizos2016VFAE}
C.~Louizos, K.~Swersky, Y.~Li, M.~Welling, and R.~Zemel, ``The variational fair
  autoencoder,'' in \emph{Proc. Int. Conf. Learn. Represent. (ICLR)}, 2016.

\bibitem{Moyer2018Invariant}
D.~Moyer, S.~Gao, R.~Brekelmans, A.~Galstyan, and G.~Ver~Steeg, ``Invariant
  representations without adversarial training,'' in \emph{Proc. Adv. Neural
  Inf. Process. Syst. (NeurIPS)}, vol.~31, 2018.

\bibitem{Kingma2014VAE}
D.~P. Kingma and M.~Welling, ``Auto-encoding variational {B}ayes,'' in
  \emph{Proc. Int. Conf. Learn. Represent. (ICLR)}, 2014.

\bibitem{Creager2019Fair}
E.~Creager \emph{et~al.}, ``Flexibly fair representation learning by
  disentanglement,'' in \emph{Proc. Int. Conf. Mach. Learn. (ICML)}, vol.~97,
  2019, pp. 1436--1445.

\bibitem{Makhdoumi2014Funnel}
A.~Makhdoumi, S.~Salamatian, N.~Fawaz, and M.~M{\'e}dard, ``From the
  information bottleneck to the privacy funnel,'' in \emph{Proc. IEEE Inf.
  Theory Workshop (ITW)}, 2014, pp. 501--505.

\bibitem{Belghazi2018MINE}
M.~I. Belghazi \emph{et~al.}, ``Mutual information neural estimation,'' in
  \emph{Proc. Int. Conf. Mach. Learn. (ICML)}, vol.~80, 2018, pp. 531--540.

\bibitem{Cheng2020CLUB}
P.~Cheng, W.~Hao, S.~Dai, J.~Liu, Z.~Gan, and L.~Carin, ``{CLUB}: A contrastive
  log-ratio upper bound of mutual information,'' in \emph{Proc. Int. Conf.
  Mach. Learn. (ICML)}, vol. 119, 2020, pp. 1779--1788.

\bibitem{Shrivastava2017SimGAN}
A.~Shrivastava, T.~Pfister, O.~Tuzel, J.~Susskind, W.~Wang, and R.~Webb,
  ``Learning from simulated and unsupervised images through adversarial
  training,'' in \emph{Proc. IEEE Conf. Comput. Vis. Pattern Recognit. (CVPR)},
  2017, pp. 2242--2251.

\bibitem{Park2019RKD}
W.~Park, D.~Kim, Y.~Lu, and M.~Cho, ``Relational knowledge distillation,'' in
  \emph{Proc. IEEE/CVF Conf. Comput. Vis. Pattern Recognit. (CVPR)}, 2019, pp.
  3967--3976.

\bibitem{Deng2019ArcFace}
J.~Deng, J.~Guo, N.~Xue, and S.~Zafeiriou, ``{ArcFace}: Additive angular margin
  loss for deep face recognition,'' in \emph{Proc. IEEE/CVF Conf. Comput. Vis.
  Pattern Recognit. (CVPR)}, 2019, pp. 4685--4694.

\bibitem{Deng2019LFR}
J.~Deng, J.~Guo, D.~Zhang, Y.~Deng, X.~Lu, and S.~Shi, ``Lightweight face
  recognition challenge,'' in \emph{Proc. IEEE/CVF Int. Conf. Comput. Vis.
  Workshops (ICCVW)}, 2019, pp. 2638--2646.

\bibitem{cao2018vggface2}
Q.~Cao, L.~Shen, W.~Xie, O.~M. Parkhi, and A.~Zisserman, ``{VGGFace2}: A
  dataset for recognising faces across pose and age,'' in \emph{Proc. IEEE Int.
  Conf. Autom. Face Gesture Recognit. (FG)}, 2018, pp. 67--74.

\bibitem{Whitelam2017IJBC}
B.~Maze \emph{et~al.}, ``{IARPA Janus Benchmark-C}: Face dataset and
  protocol,'' in \emph{Proc. Int. Conf. Biometrics (ICB)}, 2018, pp. 158--165.

\bibitem{Wang2019RFW}
M.~Wang, W.~Deng, J.~Hu, X.~Tao, and Y.~Huang, ``{Racial Faces in the Wild}:
  Reducing racial bias by information maximization adaptation network,'' in
  \emph{Proc. IEEE/CVF Int. Conf. Comput. Vis. (ICCV)}, 2019, pp. 692--702.

\bibitem{Huang2007LFW}
G.~B. Huang, M.~Ramesh, T.~Berg, and E.~Learned-Miller, ``Labeled faces in the
  wild: A database for studying face recognition in unconstrained
  environments,'' University of Massachusetts, Amherst, Tech. Rep. 07-49, Oct.
  2007.

\bibitem{iso19795-10}
{ISO/IEC}, ``Information technology---{B}iometric performance testing and
  reporting---{P}art~10: {Q}uantifying biometric system performance variation
  across demographic groups,'' {ISO/IEC} 19795-10:2024, International
  Organization for Standardization, Geneva, Switzerland, 2024.

\bibitem{garbe}
J.~J. Howard, E.~J. Laird, R.~E. Rubin, Y.~B. Sirotin, J.~L. Tipton, and A.~R.
  Vemury, ``Evaluating proposed fairness models for face recognition
  algorithms,'' in \emph{Proc. Int. Conf. Pattern Recognit. (ICPR) Workshops},
  2023, pp. 431--447.

\bibitem{Morales2019SensitiveNetsArXiv}
A.~Morales, J.~Fierrez, R.~Vera-Rodriguez, and R.~Tolosana, ``{SensitiveNets}:
  Learning agnostic representations with application to face images,''
  arXiv:1902.00334v3, 2019.

\bibitem{McInnes2018UMAP}
L.~McInnes, J.~Healy, and J.~Melville, ``{UMAP}: Uniform manifold approximation
  and projection for dimension reduction,'' arXiv:1802.03426, 2018.

\bibitem{CoverThomas2006}
T.~M. Cover and J.~A. Thomas, \emph{Elements of Information Theory},
  2nd~ed.\hskip 1em plus 0.5em minus 0.4em\relax Hoboken, NJ, USA:
  Wiley-Interscience, 2006.

\bibitem{Morales2020Patent}
A.~{Morales Moreno}, J.~{Ortega Garc{\'i}a}, J.~{Fierrez Aguilar}, and R.~{Vera
  Rodr{\'i}guez}, ``M{\'e}todo para la eliminaci{\'o}n del sesgo en sistemas de
  reconocimiento biom{\'e}trico,'' Spanish Patent ES~2\,769\,093~A1, 2020.

\bibitem{Pena2021EmotionBlind}
A.~Pe{\~n}a, J.~Fierrez, A.~Morales, and A.~Lapedriza, ``Learning
  emotional-blinded face representations,'' in \emph{Proc. Int. Conf. Pattern
  Recognit. (ICPR)}, 2021, pp. 3566--3573.

\bibitem{Pena2020Thesis}
A.~{Pe{\~n}a Almansa}, ``Aprendizaje de representaciones faciales
  emocionalmente ciegas con redes neuronales profundas,'' Master's thesis,
  Univ. Aut{\'o}noma de Madrid, 2020.

\bibitem{Zaheer2017DeepSets}
M.~Zaheer, S.~Kottur, S.~Ravanbakhsh, B.~P{\'o}czos, R.~Salakhutdinov, and
  A.~J. Smola, ``Deep sets,'' in \emph{Proc. Adv. Neural Inf. Process. Syst.
  (NIPS)}, vol.~30, 2017.

\bibitem{Lee2019SetTransformer}
J.~Lee, Y.~Lee, J.~Kim, A.~R. Kosiorek, S.~Choi, and Y.~W. Teh, ``Set
  transformer: A framework for attention-based permutation-invariant neural
  networks,'' in \emph{Proc. Int. Conf. Mach. Learn. (ICML)}, vol.~97, 2019,
  pp. 3744--3753.

\end{thebibliography}

\clearpage
\onecolumn
\let\addcontentsline\PaperAddContentsLine
\setcounter{table}{0}
\setcounter{figure}{0}
\setcounter{equation}{0}
\setcounter{theorem}{0}
\renewcommand{\thetable}{S\arabic{table}}
\renewcommand{\thefigure}{S\arabic{figure}}
\renewcommand{\theequation}{S\arabic{equation}}
\renewcommand{\thetheorem}{S\arabic{theorem}}
\makeatletter
\expandafter\renewcommand\expandafter\section\expandafter{\expandafter\@fb@secFB\section}%
\newcommand\@fb@secFB{\FloatBarrier
  \gdef\@fb@afterHHook{\@fb@topbarrier \gdef\@fb@afterHHook{}}}
\g@addto@macro\@afterheading{\@fb@afterHHook}
\gdef\@fb@afterHHook{}
\makeatother
\begin{center}
{\Huge Supplementary Material for\\
TEMPER: Reducing Attribute Leakage\\ from Face Recognition Embeddings\par}
\end{center}
\vspace{\baselineskip}
This supplementary material holds the proofs, the protocol, the implementation of every compared method and the full results behind the main paper. The appendices follow the order of the main paper, and the tables and figures of the supplement carry the prefix S. The appendices hold the following material.
\begin{itemize}
\item Appendix~\ref{app:proofs}: the proofs of the propositions of the main paper, the bound on the information in the class latent (Proposition~\ref{prop:s-zc}) and further standard remarks.
\item Appendix~\ref{app:protocol}: the data, the labels, the verification protocol, the attackers and the worst-case convention, and the hyperparameters of every compared method, the label-free controls and the attackers (Table~\ref{tab:s-hyper}).
\item Appendix~\ref{app:baselines}: the implementation of each baseline, the choices left open by the original papers, our deviations from these papers, and our DVPF numbers next to those of the DVPF paper (Table~\ref{tab:s-dvpf-reported}).
\item Appendix~\ref{app:results}: the score statistics of every setting, the frontiers at all 4 operating points, the LFW and RFW accuracy, the leakage on each test set, the full per-setting results, the comparison at equal $k$, the seed variation, the bootstrap intervals and the ROC curves.
\item Appendix~\ref{app:leakage}: the log loss of the attackers and the leakage under pooled templates, under attackers trained in the target domain and under an MLP attacker selected on held-out identities.
\item Appendix~\ref{app:ssv}: the sample-side variant of TEMPER, whose privacy loss acts on the sampled latent, with the diagnostics of the auxiliary classifier (Sections~\ref{sec:res-attacker} and~\ref{sec:res-ablation}).
\item Appendix~\ref{app:ablation}: the full ablation of TEMPER and the measurements of the class latent.
\item Appendix~\ref{app:fairness}: the per-group error rates behind the demographic differentials of the main paper, and the differentials on IJB-C and RFW at three operating points.
\item Appendix~\ref{app:backbone}: a reduced grid of TEMPER and PCA truncation on an IResNet100 trained on Glint360k.
\end{itemize}
Symbols not defined in the appendices have the meaning given in Sections~\ref{sec:method} and~\ref{sec:experiments} of the main paper.

\newpage
\setcounter{tocdepth}{2}
\begingroup\small\tableofcontents\endgroup
\clearpage

\appendices
\section{Proofs and Standard Results}
\label{app:proofs}

We prove Propositions~\ref{prop:lower} and~\ref{prop:ent} of the main paper and a result on a privacy loss on the sampled latent (Proposition~\ref{prop:s-mask}), which the sample-side variant of Section~\ref{sec:res-attacker} realises.
We also prove a bound on the class latent (Proposition~\ref{prop:s-zc}) and the corollary on the decoder input (Corollary~\ref{cor:s-decoder}).
Two further standard remarks are collected in Appendix~\ref{app:proofs-facts}.
All results are standard and are stated here for the quantities of TEMPER.
The training objective is a surrogate, since the entropy term computed with the trained auxiliary classifier differs in general from a mutual information.

\subsection{Setting and Notation}
\label{app:proofs-setting}

Let $(X,C)$ be distributed according to a distribution $p$ on $\mathbb{S}^{d-1}\times\mathcal{C}$, where $X$ is the $\ell_2$-normalised embedding and $C$ is the sensitive attribute with label set $\mathcal{C}=\{1,\dots,|\mathcal{C}|\}$.
Statements without a named distribution are made under the distribution $p$ of the population behind the training set.
Proposition~\ref{prop:ent} and Proposition~\ref{prop:s-zc} are stated under the distribution $p_{\mathcal{B}}$ with a uniform class prior and the class-conditional distributions of the population.
The class-balanced sampler approximates $p_{\mathcal{B}}$ on the training set (Section~\ref{sec:method-obj} and Remark~\ref{rem:s-surrogate}).
Every statement on leakage is made under the joint distribution of $(C,R)$ in the population from which the test set is drawn, which we call the distribution of the test data.
The transformed template is $R=\bm{r}_\theta(X)$ with $\bm{r}_\theta(\bm{x})=\bm{\mu}_\theta(\bm{x})/\|\bm{\mu}_\theta(\bm{x})\|_2$, and we assume $\bm{\mu}_\theta(X)\neq\bm{0}$ almost surely.
We write $\mathcal{S}$ for the per-feature standardisation $\mathcal{S}(\bm{r})=\operatorname{diag}(\bm{\tau})^{-1}(\bm{r}-\bm{m})$, with the mean $\bm{m}$ and the standard deviations $\bm{\tau}>\bm{0}$ of $R$ under $p_{\mathcal{B}}$, and $\tilde{R}=\mathcal{S}(R)$.
A classifier $q$ is a measurable map from $\mathbb{R}^{k}$ to the distributions on $\mathcal{C}$, and $p(\cdot\mid\tilde{\bm{r}})$ is a regular conditional distribution of $C$ given $\tilde{R}=\tilde{\bm{r}}$.
Since $C$ is discrete, the conditional entropy is $\Ent(C\mid R)=\E[\Ent(p(\cdot\mid R))]$, with $p(\cdot\mid R)$ the conditional distribution of $C$ given $R$.
Entropies and mutual informations are in nats, and $\Ent(q)=-\sum_{c\in\mathcal{C}}q(c)\log q(c)$ is the entropy of a distribution $q$ on $\mathcal{C}$.

\subsection{Proof of Proposition~\ref{prop:lower}}
\label{app:proof-lower}

\begin{restate}{prop:lower}
Let $\mathcal{T}$ be a fixed affine bijection of $\mathbb{R}^{k}$. For any joint distribution of $(C,R)$ and any classifier $q$ that maps $\mathcal{T}(\bm{r})$ to a distribution on $\mathcal{C}$,
\begin{equation}
\label{eq:s-ce}
\E\bigl[-\log q(C\mid\mathcal{T}(R))\bigr]=\Ent(C\mid R)+\E\Bigl[\KL\bigl(p(\cdot\mid\mathcal{T}(R))\,\big\|\,q(\cdot\mid\mathcal{T}(R))\bigr)\Bigr]\;\ge\;\Ent(C\mid R),
\end{equation}
with $p(\cdot\mid\bm{u})$ the conditional distribution of $C$ given $\mathcal{T}(R)=\bm{u}$, and hence $\I(C;R)\ge\Ent(C)-\E[-\log q(C\mid\mathcal{T}(R))]$.
\end{restate}

\begin{proof}[Proof of Proposition~\ref{prop:lower}]
Let $U=\mathcal{T}(R)$, and let $p(\cdot\mid\bm{u})$ be the conditional distribution of $C$ given $U=\bm{u}$, which exists because $C$ takes finitely many values.
If $q(c\mid\bm{u})=0$ for some $c$ with $p(c\mid\bm{u})>0$ on a set of positive probability, both sides of the identity in~\eqref{eq:s-ce} are infinite and the inequality holds trivially.
Otherwise, for almost every $\bm{u}$ under the distribution of $U$,
\begin{equation}
\begin{aligned}
\E\bigl[-\log q(C\mid U)\,\big|\,U=\bm{u}\bigr]
&=-\sum_{c\in\mathcal{C}}p(c\mid\bm{u})\log p(c\mid\bm{u})
+\sum_{c\in\mathcal{C}}p(c\mid\bm{u})\log\frac{p(c\mid\bm{u})}{q(c\mid\bm{u})}\\
&=\Ent\bigl(p(\cdot\mid\bm{u})\bigr)+\KL\bigl(p(\cdot\mid\bm{u})\,\big\|\,q(\cdot\mid\bm{u})\bigr).
\end{aligned}
\end{equation}
Taking the expectation over $U$ gives $\E[-\log q(C\mid U)]=\Ent(C\mid U)+\E[\KL(p(\cdot\mid U)\,\|\,q(\cdot\mid U))]$, and the second term is nonnegative by Gibbs' inequality.

The random variables $R$ and $U$ generate the same $\sigma$-algebra, since, by hypothesis, $\mathcal{T}$ is an affine bijection of $\mathbb{R}^{k}$ with an affine inverse.
The conditional distributions of $C$ given $R$ and given $U$ therefore coincide almost surely, so that $\Ent(C\mid U)=\Ent(C\mid R)$ and $\I(C;U)=\I(C;R)$, which gives~\eqref{eq:s-ce}.
With $\I(C;R)=\Ent(C)-\Ent(C\mid R)$, the inequality in~\eqref{eq:s-ce} gives $\I(C;R)\ge\Ent(C)-\E[-\log q(C\mid U)]$.
The standardisation $\mathcal{S}$ and the standardisation of the attackers with training statistics are such bijections whenever their standard deviations are positive.
\end{proof}

\begin{remark}[Lower bound from an attacker]
\label{rem:s-certify}
The statement holds for any joint distribution, in particular for the distribution of the test data and for $p_{\mathcal{B}}$.
The held-out log loss $L_q$ of an attacker trained on independent data estimates the left-hand side of~\eqref{eq:s-ce} under the distribution of the held-out data.
The difference $\Ent(C)-L_q$ then estimates a lower bound on $\I(C;R)$.
The bound is one-sided, and an attacker with a log loss below $\Ent(C)$ is evidence that the attribute is predictable from the transformed template.
An attacker whose log loss is at least $\Ent(C)$ is not evidence of independence, and a trained classifier, the auxiliary classifier included, does not give an upper bound on $\I(C;R)$.
Let $L_0$ be the held-out log loss of the prediction from the class prior of the test set, which estimates $\Ent(C)$.
The normalised log-loss reduction $G_{\log}=(L_0-L_q)/L_0$ of Appendix~\ref{app:leakage-logloss} then estimates $(\Ent(C)-\E[-\log q(C\mid\mathcal{T}(R))])/\Ent(C)$.
By~\eqref{eq:s-ce}, this quantity is at most $\I(C;R)/\Ent(C)$, with equality for the posterior of $C$.
The posteriors are reweighted to the prior of the test set before the computation of $L_q$, since our attackers are trained with a class-weighted loss.
\end{remark}

\begin{remark}[Balanced accuracy and Fano's inequality]
\label{rem:s-fano}
Let $p^{\mathrm{bal}}$ be the distribution with the class-conditional distributions of the test data and a uniform class prior.
The 0--1 error of a classifier under $p^{\mathrm{bal}}$ equals one minus the balanced accuracy $b$ of the classifier under the distribution of the test data.
The Bayes error under $p^{\mathrm{bal}}$ therefore satisfies $P_e^\star\le1-b$, and the balanced accuracy on the test set is an estimate of $b$.
The right-hand side of Fano's inequality~\cite{CoverThomas2006}, $\Ent_{p^{\mathrm{bal}}}(C\mid R)\le h_2(P_e^\star)+P_e^\star\log(|\mathcal{C}|-1)$ with $h_2$ the binary entropy, is nondecreasing in $P_e^\star$ on $[0,1-1/|\mathcal{C}|]$.
Whenever $b\ge1/|\mathcal{C}|$, the two inequalities and $\Ent_{p^{\mathrm{bal}}}(C)=\log|\mathcal{C}|$ give
\begin{equation}
\I_{p^{\mathrm{bal}}}(C;R)\;\ge\;\log|\mathcal{C}|-h_2(1-b)-(1-b)\log(|\mathcal{C}|-1).
\end{equation}
A balanced accuracy above chance implies a positive lower bound on the information, in the same way as the log loss.
A balanced accuracy at chance implies only the trivial bound zero.
\end{remark}

\subsection{Proof of Proposition~\ref{prop:ent}}
\label{app:proof-ent}

\begin{restate}{prop:ent}
Let every coordinate of $R$ have positive variance under $p_{\mathcal{B}}$. (i)~For every classifier $q$ and every distribution of $\tilde{R}$, $\E[\Ent(q(\cdot\mid\tilde{R}))]\le\log|\mathcal{C}|$, with equality if and only if $q(\cdot\mid\tilde{R})$ is uniform almost surely. (ii)~Under $p_{\mathcal{B}}$, let $q(c\mid\tilde{\bm{r}})=p_{\mathcal{B}}(C=c\mid\tilde{R}=\tilde{\bm{r}})$ be the posterior of $C$, which minimises the expected log loss. Then $-\E[\Ent(q(\cdot\mid\tilde{R}))]=\I(C;R)-\log|\mathcal{C}|\ge-\log|\mathcal{C}|$, with equality if and only if $C$ and $R$ are independent, that is, if and only if the class-conditional distributions of $R$ are equal.
\end{restate}

\begin{proof}[Proof of Proposition~\ref{prop:ent}]
Under $p_{\mathcal{B}}$ the class prior is uniform by definition.
Therefore, $\Ent_{p_{\mathcal{B}}}(C)=\log|\mathcal{C}|<\infty$.

(i) Let $u$ denote the uniform distribution on $\mathcal{C}$, so that every distribution $q$ on $\mathcal{C}$ satisfies
\begin{equation}
\log|\mathcal{C}|-\Ent(q)=\sum_{c\in\mathcal{C}}q(c)\log\frac{q(c)}{1/|\mathcal{C}|}=\KL(q\,\|\,u)\ge0,
\end{equation}
with equality if and only if $q=u$, by Gibbs' inequality.
Applying this to $q(\cdot\mid\tilde{R})$ and taking expectations gives $\E[\Ent(q(\cdot\mid\tilde{R}))]\le\log|\mathcal{C}|$, with equality if and only if $\KL(q(\cdot\mid\tilde{R})\,\|\,u)=0$ almost surely, that is, if and only if $q(\cdot\mid\tilde{R})$ is uniform almost surely.

(ii) If $q(\cdot\mid\tilde{\bm{r}})=p_{\mathcal{B}}(\cdot\mid\tilde{\bm{r}})$, then by the definition of conditional entropy and the invariance shown in the proof of Proposition~\ref{prop:lower}, which applies to $\mathcal{S}$ under the positive-variance condition,
\begin{equation}
\E\bigl[\Ent\bigl(p_{\mathcal{B}}(\cdot\mid\tilde{R})\bigr)\bigr]=\Ent(C\mid\tilde{R})=\Ent(C\mid R)=\Ent(C)-\I(C;R)=\log|\mathcal{C}|-\I(C;R).
\end{equation}
Since $\I(C;R)\ge0$, with equality if and only if $C$ and $R$ are independent, $-\E[\Ent(q(\cdot\mid\tilde{R}))]=\I(C;R)-\log|\mathcal{C}|$ attains its minimum $-\log|\mathcal{C}|$ if and only if $C$ and $R$ are independent.
Because every class has positive probability under $p_{\mathcal{B}}$, independence holds if and only if the conditional distribution of $R$ given $C=c$ is the same for every $c$.
\end{proof}

\begin{remark}[Limits of the entropy term]
\label{rem:s-surrogate}
(a) Part (ii) of Proposition~\ref{prop:ent} is stated for the posterior classifier.
The equality need not hold for any other $q$, since a constant uniform $q$ attains the maximum in (i) for every value of $\I(C;R)$.
The attackers attain more than chance on the test data after the training of TEMPER.
For example, the LR attacker achieves a balanced accuracy of 53.8\% for the gender model of TEMPER with $k=96$ and $\lambda=30$ on IJB-C (Appendix~\ref{app:results-cross}).
This accuracy exceeds the null percentile of the setting (Appendix~\ref{app:protocol-attackers}), which is evidence against independence under the distribution of the test data.
Even independence under the distribution of the sampler would not imply independence under the distribution of the test data (part~(c)).
(b) The auxiliary classifier $q_\phi$ is fixed in the step of the template branch, whereas the posterior classifier depends on $\theta$.
The gradient of the predictive entropy $\E[\Ent(q_\phi(\cdot\mid\tilde{R}))]$ with respect to $\theta$ therefore differs in general from $\nabla_\theta\Ent(C\mid R)$, even if $q_\phi=p_{\mathcal{B}}(\cdot\mid\tilde{\bm{r}})$ before the step.
The alternating updates approximate a bilevel problem.
The auxiliary classifier $q_\phi$ is fitted by cross-entropy in the inner problem, and the total loss is minimised at the inner solution in the outer problem.
The auxiliary classifier is also trained on the stored templates of the memory, which were computed with earlier parameters of the template branch.
The solution of the inner problem is in general also not the posterior classifier of the current transformed template.
(c) The class-balanced sampler approximates $p_{\mathcal{B}}$ on the training set.
An epoch ends when every image of every identity has entered a minibatch.
A class whose identities run out earlier receives identities drawn again from the same class.
The images within a class are therefore weighted roughly per image for the class that finishes last and partly per identity for the other classes.
The class prior has no effect on the equality of the class-conditional distributions, whereas the within-class weighting changes these distributions.
Independence under the distribution of the sampler therefore does not imply independence under $p_{\mathcal{B}}$ or under the distribution of the test data.
(d) The transformed templates are standardised in training with the minibatch statistics of $\mathcal{S}_{\mathcal{B}}$ in place of the statistics of $\mathcal{S}$ under $p_{\mathcal{B}}$.
The expectation of $\lambda\,\mathcal{L}_{\mathrm{ent}}$ under $p_{\mathcal{B}}$ is $\lambda\,\I_{p_{\mathcal{B}}}(C;R)-\lambda\log|\mathcal{C}|$ with the posterior classifier in both heads and $\mathcal{S}$ in place of $\mathcal{S}_{\mathcal{B}}$.
This identity need not hold for any other classifier.
We therefore call $\mathcal{L}_{\mathrm{ent}}$ a predictive-entropy surrogate and measure leakage only with attackers trained on the transformed templates of the fixed method.
\end{remark}

\subsection{Proof of Proposition~\ref{prop:s-mask}}
\label{app:proof-mask}

\begin{proposition}[Privacy loss on the sampled latent]
\label{prop:s-mask}
Let $\bm{\mu}_r$ and $\bm{\sigma}_r$ be the mean and the standard deviation of a Gaussian template branch with $d_r$ outputs and a continuous mean ($\bm{\mu}_\theta$, $\bm{\sigma}_\theta$ and $d_r=k$ for TEMPER).
Let $(X,C)$ have any distribution on $\mathbb{S}^{d-1}\times\mathcal{C}$, and suppose that $\bm{\sigma}_r(\bm{x})=\sigma_0\bm{1}$ for every $\bm{x}$, so that $Z_r=\bm{\mu}_r(X)+\sigma_0\bm{\epsilon}$ with $\bm{\epsilon}\sim\mathcal{N}(\bm{0},\bm{I})$ independent of $(X,C)$.
Then
\begin{equation}
\label{eq:s-mask}
\I(C;Z_r)\;\le\;\I\bigl(\bm{\mu}_r(X);Z_r\bigr)\;\le\;\frac{1}{2}\sum_{j=1}^{d_r}\log\Bigl(1+\frac{\operatorname{Var}[\mu_{r,j}(X)]}{\sigma_0^2}\Bigr),
\end{equation}
whereas the information in the mean satisfies $\I(C;\zeta\bm{\mu}_r(X)+\bm{v})=\I(C;\bm{\mu}_r(X))$ for every $\zeta\neq0$ and $\bm{v}\in\mathbb{R}^{d_r}$.
In particular, replacing $\bm{\mu}_r$ by $\zeta\bm{\mu}_r$ with $0<\zeta<1$ multiplies every variance in the bound by $\zeta^2$ and leaves the transformed template unchanged.
\end{proposition}

\begin{proof}[Proof of Proposition~\ref{prop:s-mask}]
Write $\bar{Z}_r=\bm{\mu}_r(X)\in\mathbb{R}^{d_r}$ for the conditional mean of $Z_r$ given $X$.
The encoder of the template branch is affine for TEMPER.
For the nonlinear variants of Appendix~\ref{app:ablation}, the encoder is a composition of affine maps and PReLU activations.
The encoder is therefore continuous in both cases.
Since $X$ lies on the compact sphere $\mathbb{S}^{d-1}$, $\bar{Z}_r$ is bounded and every $\operatorname{Var}[\bar{Z}_{r,j}]$ is finite.

\emph{First inequality.} Since $\bm{\epsilon}$ is independent of $(X,C)$ and $Z_r=\bar{Z}_r+\sigma_0\bm{\epsilon}$, the conditional distribution of $Z_r$ given $(C,X)$ is $\mathcal{N}(\bar{Z}_r,\sigma_0^2\bm{I})$, which depends on $(C,X)$ only through $\bar{Z}_r$.
The sequence $C\to\bar{Z}_r\to Z_r$ is therefore a Markov chain, and the data processing inequality~\cite{CoverThomas2006} gives $\I(C;Z_r)\le\I(\bar{Z}_r;Z_r)$.

\emph{Second inequality.} Given $\bar{Z}_r$, $Z_r$ is Gaussian with covariance $\sigma_0^2\bm{I}$.
Therefore, $h(Z_r\mid\bar{Z}_r)=\tfrac{d_r}{2}\log(2\pi e\sigma_0^2)$, where $h$ denotes differential entropy.
The vector $Z_r$ has a density, and its covariance is $\bm{\Sigma}=\operatorname{Cov}(\bar{Z}_r)+\sigma_0^2\bm{I}$, which is positive definite.
Among all densities with covariance $\bm{\Sigma}$, the Gaussian has the largest differential entropy, and Hadamard's inequality bounds the determinant of a positive definite matrix by the product of its diagonal entries.
Since $h(Z_r)\ge h(Z_r\mid\bar{Z}_r)>-\infty$, every quantity below is finite, and the two facts give
\begin{equation}
h(Z_r)\le\tfrac{1}{2}\log\bigl((2\pi e)^{d_r}\det\bm{\Sigma}\bigr)\le\tfrac{1}{2}\sum_{j=1}^{d_r}\log\bigl(2\pi e\,(\operatorname{Var}[\bar{Z}_{r,j}]+\sigma_0^2)\bigr),
\end{equation}
and
\begin{equation}
\I(\bar{Z}_r;Z_r)=h(Z_r)-h(Z_r\mid\bar{Z}_r)\le\frac{1}{2}\sum_{j=1}^{d_r}\log\Bigl(1+\frac{\operatorname{Var}[\bar{Z}_{r,j}]}{\sigma_0^2}\Bigr).
\end{equation}

\emph{Invariance.} For $\zeta\neq0$ and $\bm{v}\in\mathbb{R}^{d_r}$, the map $\bm{z}\mapsto\zeta\bm{z}+\bm{v}$ is a bijection of $\mathbb{R}^{d_r}$ with the measurable inverse $\bm{z}\mapsto(\bm{z}-\bm{v})/\zeta$.
Both $C\to\bar{Z}_r\to\zeta\bar{Z}_r+\bm{v}$ and $C\to\zeta\bar{Z}_r+\bm{v}\to\bar{Z}_r$ are Markov chains.
The data processing inequality applied in both directions therefore gives $\I(C;\zeta\bar{Z}_r+\bm{v})=\I(C;\bar{Z}_r)$.

\emph{Rescaling.} Replace $\bm{\mu}_r$ by $\zeta\bm{\mu}_r$ with $0<\zeta<1$, keeping $\bm{\sigma}_r\equiv\sigma_0\bm{1}$.
Then $\operatorname{Var}[\zeta\bar{Z}_{r,j}]=\zeta^2\operatorname{Var}[\bar{Z}_{r,j}]$.
The bound therefore becomes $\frac{1}{2}\sum_j\log(1+\zeta^2\operatorname{Var}[\bar{Z}_{r,j}]/\sigma_0^2)$, which tends to zero as $\zeta\to0$.
For every $\bm{x}$ with $\bm{\mu}_r(\bm{x})\neq\bm{0}$, $\zeta\bm{\mu}_r(\bm{x})/\|\zeta\bm{\mu}_r(\bm{x})\|_2=\bm{\mu}_r(\bm{x})/\|\bm{\mu}_r(\bm{x})\|_2$.
The transformed template and $\I(C;R)$ therefore stay the same.
\end{proof}

\begin{remark}[Reading Proposition~\ref{prop:s-mask}]
\label{rem:s-mask}
The transformed template $R=\bar{Z}_r/\|\bar{Z}_r\|_2$ is a function of $\bar{Z}_r$.
Therefore, $\I(C;R)\le\I(C;\bar{Z}_r)$.
The transformed template is not a function of $Z_r$, and $\I(C;Z_r)$ and $\I(C;R)$ are not ordered in general.
The information in the transformed template is therefore bounded by the information in the mean.
The bound on the attribute information in the sampled latent can therefore be reduced by a rescaling of the mean without a change of the transformed template.
The assumption $\bm{\sigma}_r\equiv\sigma_0\bm{1}$ is an idealisation of the clipped log-variance of the implementation.
The log-variance is clipped to $[-4,2]$, and the clipping has zero gradient below $-4$.
An input whose log-variance is below $-4$ therefore does not contribute a gradient to the log-variance head.
The share of log-variances of the template branch clipped to $-4$ is 92.2\% at $\lambda=100$ in the gender models of the sample-side variant (Appendix~\ref{app:ssv}).
This share increases to 99.99\% at $\lambda=1000$.
Every log-variance of the template branch is clipped to $-4$ in the ethnicity model at $\lambda=1000$ (Table~\ref{tab:s-diag}).
\end{remark}

\begin{remark}[Common offset, informal]
\label{rem:s-offset}
The trained mean of the sample-side variant also has a large common offset, $\bm{\mu}_r(\bm{x})=\bm{v}+\zeta\bm{\delta}(\bm{x})$ with $\|\bm{v}\|_2\gg\zeta\|\bm{\delta}(\bm{x})\|_2$ (Table~\ref{tab:s-diag}).
With $\hat{\bm{v}}=\bm{v}/\|\bm{v}\|_2$, a first-order expansion in $\zeta$ gives
\begin{equation}
\bm{r}=\hat{\bm{v}}+\frac{\zeta}{\|\bm{v}\|_2}\bigl(\bm{I}-\hat{\bm{v}}\hat{\bm{v}}^\top\bigr)\bm{\delta}(\bm{x})+\mathcal{O}(\zeta^2).
\end{equation}
The transformed templates therefore concentrate in a small spherical cap around $\hat{\bm{v}}$.
This is the replacement of Remark~\ref{rem:s-scale} with $\bm{\omega}=(\zeta/\|\bm{v}\|_2)\bm{1}$ and $\bm{t}=\hat{\bm{v}}$, applied to $\bm{u}_i=(\bm{I}-\hat{\bm{v}}\hat{\bm{v}}^\top)\bm{\delta}(\bm{x}_i)$ in place of $\bm{r}_i$.
A per-feature standardisation removes the constant $\hat{\bm{v}}$ and the factor $\zeta/\|\bm{v}\|_2$, and the standardised transformed template is, to first order, an affine image of $\bm{\delta}(\bm{x})$.
The attribute is then, to first order, as predictable from the standardised transformed template as from the component of $\bm{\delta}(\bm{x})$ orthogonal to $\hat{\bm{v}}$.
The MLP attacker with standardised inputs predicts the attribute from transformed templates on which the weak MLP of Appendix~\ref{app:protocol-weak} attains chance.
The non-mated comparison scores of these templates are close to one (Table~\ref{tab:s-scores}).
\end{remark}

\subsection{Capacity of the Class Latent}
\label{app:proofs-zc}

The decoder of TEMPER receives the class latent $\bm{z}_c$ as well as the training-only sample $\bm{z}_r$ of the template branch.
Identity information can therefore be encoded in $\bm{z}_c$ as well as in $\bm{z}_r$.
Proposition~\ref{prop:s-zc} gives an upper bound on the information about $X$ in the class latent for the prior covariance $\bm{I}$ of the implementation.
The proof follows from 2 standard facts about the KL divergence.

\begin{proposition}[Capacity of the class latent]
\label{prop:s-zc}
Let $q_\vartheta(\cdot\mid\bm{x})=\mathcal{N}\bigl(\bm{\mu}_\vartheta(\bm{x}),\operatorname{diag}\bm{\sigma}_\vartheta^{2}(\bm{x})\bigr)$ and $Z_c\sim q_\vartheta(\cdot\mid X)$, with $\bm{\mu}_\vartheta$ and $\log\bm{\sigma}_\vartheta^{2}$ continuous and $(X,C)$ distributed according to $p_{\mathcal{B}}$.
Then
\begin{equation}
\label{eq:s-zc}
\I(X;Z_c)\le\E\Bigl[\KL\bigl(q_\vartheta(\cdot\mid X)\,\big\|\,\mathcal{N}(\bm{e}_C,\bm{I})\bigr)\Bigr]+\Ent(C)
=\E\Bigl[\KL\bigl(q_\vartheta(\cdot\mid X)\,\big\|\,\mathcal{N}(\bm{e}_C,\bm{I})\bigr)\Bigr]+\log|\mathcal{C}|.
\end{equation}
\end{proposition}

\begin{proof}
Let $\bar{q}$ denote the marginal density of $Z_c$.
For every density $\rho$ on $\mathbb{R}^{d_c}$, the expected KL divergence decomposes as
\begin{equation}
\E\bigl[\KL\bigl(q_\vartheta(\cdot\mid X)\,\|\,\rho\bigr)\bigr]=\I(X;Z_c)+\KL(\bar{q}\,\|\,\rho)\ge\I(X;Z_c),
\end{equation}
whenever the left-hand side is finite.
For the mixture $\rho=\sum_{c\in\mathcal{C}}\pi_c\,\mathcal{N}(\bm{e}_c,\bm{I})$ with $\pi_c=p_{\mathcal{B}}(C=c)=1/|\mathcal{C}|$, the KL divergence inside the expectation is a continuous function of $\bm{x}$ through $\bm{\mu}_\vartheta$ and $\log\bm{\sigma}_\vartheta^{2}$.
The left-hand side is finite, since the sphere $\mathbb{S}^{d-1}$ is compact.
Every $c$ and $\bm{z}$ satisfy $\rho(\bm{z})\ge\pi_c\,\mathcal{N}(\bm{z};\bm{e}_c,\bm{I})$, so that $-\log\rho(\bm{z})\le-\log\pi_c-\log\mathcal{N}(\bm{z};\bm{e}_c,\bm{I})$.
For an embedding $\bm{x}$ with label $c$, it follows that
\begin{equation}
\KL\bigl(q_\vartheta(\cdot\mid\bm{x})\,\|\,\rho\bigr)\le\KL\bigl(q_\vartheta(\cdot\mid\bm{x})\,\|\,\mathcal{N}(\bm{e}_c,\bm{I})\bigr)-\log\pi_c.
\end{equation}
Taking the expectation over $(X,C)\sim p_{\mathcal{B}}$, and using $\E[-\log\pi_C]=\Ent(C)=\log|\mathcal{C}|$, gives the claim.
\end{proof}

\begin{remark}[Size of the bound]
\label{rem:s-zc}
The term $\mathcal{L}^{c}_{\mathrm{KL}}$ in~\eqref{eq:klc} is the minibatch estimate of the expectation in~\eqref{eq:s-zc}.
The term sums the full KL divergence over the $d_c$ coordinates and averages over the minibatch, and the weight $1/d_c$ enters only the total loss.
The value of this term in the last training epoch of TEMPER at $k=42$ is 0.03~nats for gender and at most 0.11~nats for ethnicity (Table~\ref{tab:s-zc}).
The bound is then $\I(X;Z_c)\lesssim\log2+0.03\approx0.72$~nats for gender and $\I(X;Z_c)\lesssim\log4+0.11\approx1.50$~nats for ethnicity.
The information about $X$ in the class latent therefore exceeds the entropy of the attribute label by at most the value of $\mathcal{L}^{c}_{\mathrm{KL}}$.
The information about $X$ in the decoder input exceeds the information in $\bm{z}_r$ by at most this bound (Corollary~\ref{cor:s-decoder}).
The bound applies to the sampled $Z_c$ under $p_{\mathcal{B}}$, and the values above are estimates on the training set.
The prediction of the attribute and of identity from the class mean on test data is measured in Appendix~\ref{app:ablation-zc}.
\end{remark}

\begin{corollary}[Information in the decoder input]
\label{cor:s-decoder}
Under $p_{\mathcal{B}}$, let $Z_r$ and $Z_c$ be the sampled latents, conditionally independent given $X$, with $\I(X;Z_r)$ and $\I(X;Z_c)$ finite.
Then $\I(X;Z_r,Z_c)-\I(X;Z_r)\le\I(X;Z_c)$, with the bound of Proposition~\ref{prop:s-zc} on the right-hand side.
\end{corollary}

\begin{proof}
By the chain rule of mutual information, $\I(X;Z_r,Z_c)-\I(X;Z_r)=\I(X;Z_c\mid Z_r)$.
The chain rule applied to $\I(Z_c;X,Z_r)$ in both orders, with $\I(Z_r;Z_c\mid X)=0$, gives $\I(X;Z_c\mid Z_r)=\I(X;Z_c)-\I(Z_r;Z_c)\le\I(X;Z_c)$.
\end{proof}

\subsection{Standard Remarks}
\label{app:proofs-facts}

\begin{remark}[Invariance of the minibatch standardisation]
\label{rem:s-scale}
Let every $\bm{r}_i$ in a minibatch be replaced by $\bm{\omega}\odot\bm{r}_i+\bm{t}$, where $\bm{\omega}$ has positive entries and $\odot$ denotes the elementwise product.
The minibatch mean then becomes $\bm{\omega}\odot\bar{\bm{r}}+\bm{t}$ and the standard deviation of coordinate $j$ becomes $\omega_j\hat{\tau}_j$.
Therefore,
\begin{equation}
\frac{\omega_jr_{ij}+t_j-\omega_j\bar{r}_j-t_j}{\omega_j\hat{\tau}_j+\varepsilon}=\frac{r_{ij}-\bar{r}_j}{\hat{\tau}_j+\varepsilon/\omega_j},
\end{equation}
which is the standardisation~\eqref{eq:std} of the original templates with $\varepsilon$ replaced by $\varepsilon/\omega_j$.
The same holds for the standardisation of the attackers with training-set statistics.
The predictive entropy of a fixed classifier on the standardised transformed templates is therefore unchanged by such a rescaling or offset.
The condition for this invariance is that every $\hat{\tau}_j$ and every $\omega_j\hat{\tau}_j$ are large compared with $\varepsilon$.
The bound of Proposition~\ref{prop:s-mask} on the attribute information in the sampled latent can instead be reduced by a rescaling.
\end{remark}

\begin{remark}[Von Mises--Fisher reading of the cosine loss]
\label{rem:s-vmf}
A von Mises--Fisher density on $\mathbb{S}^{d-1}$ with concentration $\kappa>0$ is $p(\bm{x}\mid\hat{\bm{x}})=\nu_d(\kappa)\exp(\kappa\hat{\bm{x}}^\top\bm{x})$, with the normalising constant $\nu_d(\kappa)$.
The negative log-density $-\log p(\bm{x}\mid\hat{\bm{x}})=\kappa(1-\hat{\bm{x}}^\top\bm{x})-\kappa-\log\nu_d(\kappa)$ is $1-\bm{x}^\top\hat{\bm{x}}$ up to an additive constant and the factor $\kappa$.
The total loss~\eqref{eq:total} is not a multiple of the negative evidence lower bound, since the two KL terms carry the different weights $\beta_r/k$ and $\beta_c/d_c$ of Section~\ref{sec:setup-impl}, and the relational term and the entropy term enter~\eqref{eq:total}.
\end{remark}

\section{Evaluation Protocol}
\label{app:protocol}

We give the details of the protocol of Section~\ref{sec:experiments} that do not fit in the main paper.

\subsection{Datasets and Demographic Labels}
\label{app:protocol-data}

Every method takes as input the 512-dimensional embedding that an IResNet50 trained with ArcFace on MS1MV3 computes from the standard $112\times112$ aligned crop.
The methods with fitted parameters are trained on the labelled images of the VGGFace2 training split of 8{,}631 identities.
Leakage is measured on subjects that do not occur in training, on IJB-C for gender and on RFW for ethnicity.
We exclude the 4 IJB-C and 552 RFW subjects that share a name with a VGGFace2 training identity from the leakage sets.

The IJB-C leakage set has the distinct crops with the largest face area of each remaining subject, up to 10 per subject.
The IJB-C leakage set comprises 33{,}584 crops of 3{,}527 subjects.
The RFW leakage set comprises all 38{,}676 images of the 10{,}877 remaining subjects.

The attribute is a function of identity, since labels are assigned per subject.
Table~\ref{tab:s-datasets} lists the datasets with their labelled images and comparisons, and Table~\ref{tab:s-labels} lists the label sources and the mapping to 4 ethnicity groups.
The gender labels come from the metadata of VGGFace2 and IJB-C and from the identifiers of the RFW identities.
The subjects of VGGFace2 and IJB-C are public figures listed by name, and the ethnicity labels of both datasets are our own annotation of these subjects.
RFW has no group for the IJB-C subjects that we annotated as Hispanic/Latin American or Middle Eastern, and we assign these subjects to the Caucasian group.

\begin{table}[!t]
\centering
\caption{Datasets and verification protocols. Class shares are percentages of the images labelled for each attribute, and IJB-C images are distinct face crops. The counts include the subjects excluded from the leakage sets, and LFW has no attribute label.}
\label{tab:s-datasets}
\footnotesize
\setlength{\tabcolsep}{4pt}
\begin{tabular}{@{}lrrrrrrrrrrr@{}}
\toprule
& & \multicolumn{2}{c}{Labelled images} & \multicolumn{2}{c}{Comparisons} & \multicolumn{2}{c}{Gender (\%)} & \multicolumn{4}{c}{Ethnicity (\%)} \\
\cmidrule(lr){3-4} \cmidrule(lr){5-6} \cmidrule(lr){7-8} \cmidrule(l){9-12}
Set & Identities & Gender & Ethnicity & Mated & Non-mated & Female & Male & African & Asian & Caucasian & Indian \\
\midrule
VGGFace2 train & 8,631 & 3,142,284 & 3,107,508 & -- & -- & 41.4 & 58.6 & 8.3 & 6.3 & 77.3 & 8.1 \\
IJB-C & 3,531 & 140,694 & 140,604 & 19,557 & 15,638,932 & 37.4 & 62.6 & 10.2 & 9.6 & 68.3 & 11.8 \\
RFW & 11,429 & 40,546 & 40,607 & 12,000 & 12,000 & 24.5 & 75.5 & 25.6 & 23.9 & 25.1 & 25.4 \\
LFW (View 2) & 4,281 & -- & -- & 3,000 & 3,000 & -- & -- & -- & -- & -- & -- \\
\bottomrule
\end{tabular}
\end{table}

\begin{table}[!t]
\centering
\caption{Demographic labels and their mapping to four ethnicity groups. Labels are assigned per subject, and images whose label lies outside the listed classes are excluded for that attribute.
}
\label{tab:s-labels}
\footnotesize
\setlength{\tabcolsep}{4pt}
\begin{tabular}{@{}p{0.11\textwidth}p{0.25\textwidth}p{0.12\textwidth}p{0.42\textwidth}@{}}
\toprule
Dataset & Source (subject level) & Gender & Ethnicity, source categories $\rightarrow$ group used here \\
\midrule
VGGFace2 & Gender from the VGGFace2 identity metadata, ethnicity from our annotation & male, female & African, Asian, Caucasian, Indian \\
\addlinespace
IJB-C & Gender from the IJB-C metadata, ethnicity from our annotation & male, female & African $\rightarrow$ African; East Asian and Southeast Asian $\rightarrow$ Asian; South Asian $\rightarrow$ Indian; Caucasian, Hispanic/Latin American and Middle Eastern $\rightarrow$ Caucasian \\
\addlinespace
RFW & Race groups of RFW~\cite{Wang2019RFW}, gender from the identifiers of the identities & male, female & African, Asian, Caucasian, Indian (RFW's own race labels) \\
\bottomrule
\end{tabular}
\end{table}

\subsection{Verification Protocol}
\label{app:protocol-verification}

Each method is applied to every IJB-C face crop and to the horizontal mirror of the crop, and the templates are then formed with the standard InsightFace protocol.
The transformed crop and the transformed mirror are $\ell_2$-normalised and summed, and the sum is normalised to unit length.
Crops are averaged within each media item with their detector scores as weights, the media items of a template are summed, and the template is normalised to unit length.
PE-MIU and the similarity-sensitive noise transformations are applied only to the pooled templates (Appendices~\ref{app:baselines-pemiu} and~\ref{app:baselines-other}).
We call a setting \emph{usable} when its IJB-C TMR at FMR~$10^{-3}$ is at least 50\%.
Templates are compared by cosine similarity on the 1:1 protocol, which has 19{,}557 mated and 15{,}638{,}932 non-mated comparisons.

Each RFW group has 3{,}000 mated and 3{,}000 non-mated pairs.
LFW has the 3{,}000 mated and 3{,}000 non-mated pairs of View~2.
Each method is applied to the single crop of an RFW or LFW image, without the mirror image.
Every pair of IJB-C, RFW and LFW is used for verification, including the pairs of subjects that share a name with a VGGFace2 training identity.
RFW was drawn from MS-Celeb-1M, of which MS1MV3 is a cleaned subset, so RFW subjects may occur in the training data of the backbone.
This overlap can raise the RFW TMR at FMR~$10^{-3}$, whereas the attackers are trained on VGGFace2 and are independent of the overlap.

The threshold at a target FMR is the $(\lfloor\mathrm{FMR}\cdot N_{\mathrm{nm}}\rfloor+1)$-th highest of the $N_{\mathrm{nm}}$ non-mated comparison scores.
The TMR at the target FMR is the share of mated comparison scores strictly above this threshold.
The RFW threshold at FMR~$10^{-3}$ is therefore set by 12 of the 12{,}000 non-mated pairs, and a threshold at FMR~$10^{-4}$ would be set by one pair.
The RFW TMR at FMR~$10^{-3}$ is noisier than the IJB-C TMR at this FMR, whose threshold is set by about 15{,}600 non-mated comparisons.

The RFW and LFW accuracies are computed with the ten-fold protocols of these datasets, with 300 mated and 300 non-mated pairs per fold.
LFW has 10 folds, and RFW has 10 folds per group.
The threshold with the highest accuracy on the other 9 folds is applied to each fold.
The accuracy is the mean over the folds and, for RFW, over the 4 groups.

\subsection{Leakage Protocol}
\label{app:protocol-attackers}

An attacker is a classifier trained after the method is fixed, on transformed templates of labelled VGGFace2 training images.
The attackers do not share parameters, data order or optimiser with the auxiliary classifier of a method.
Each attacker is tested on the transformed templates of subjects outside the training data, on IJB-C for gender and on RFW for ethnicity (Appendix~\ref{app:protocol-data}).
Each attacker is also scored on RFW for gender and on IJB-C for ethnicity, to measure the transfer of leakage between datasets (Table~\ref{tab:s-cross}).

Table~\ref{tab:s-hyper} lists the settings of the MLP, LR, $k$NN and SVC attackers.
The MLP attacker is a multilayer perceptron with 4 hidden layers of 1024 units and a class-weighted cross-entropy loss.
The inputs of the MLP attacker are standardised per feature with the statistics of the training templates.
The LR attacker is a linear softmax classifier on the same standardised templates, with the same loss.
The MLP and LR attackers are trained on the transformed template of every labelled training image.
We train both attackers once, with seed 0 and without model selection, and keep the model of the last epoch.

The $k$NN attacker votes among the 1, 11 or 51 nearest reference templates by cosine similarity after centring on the training mean.
The reference templates are a random draw of $10^6$ training templates with the class proportions of the training split.
We take the raw votes and, as a second variant, the votes divided by the class prior.

The SVC attacker is the support vector classifier of the DVPF evaluation~\cite{Razeghi2026DVPF}, with a radial basis function kernel and class-balanced weights.
The regularisation parameter of the SVC attacker is 1, and the kernel coefficient is scaled by the input variance, as in the defaults of scikit-learn.
The SVC attacker is fitted on 86{,}000 training templates drawn at random and standardised with the statistics of this sample.
We fit the SVC attacker with ThunderSVM on a GPU, with the stopping tolerance of libsvm and one-vs-one classifiers for ethnicity.

We report the accuracy, the balanced accuracy and the AUC of each attacker, with the AUC as a one-vs-rest macro average for ethnicity.
Balanced accuracy is the mean of the per-class recalls.
The no-information value of the balanced accuracy is 50\% for gender and 25\% for ethnicity, whatever the class prior.
The leakage of a setting is the worst-case balanced accuracy of the attackers on the test set.
A balanced accuracy above chance gives a positive lower bound on the attribute information in the transformed template, whereas a balanced accuracy at chance gives only the trivial bound (Remark~\ref{rem:s-fano}).
The worst case is the highest balanced accuracy over the MLP, LR and SVC attackers and the variants of the $k$NN attacker.
These variants combine the 3 numbers of neighbours with raw or prior-corrected votes.
The maximum is therefore taken over 9 estimates.
The 9 estimates reduce to 8 distinct estimates, since prior correction does not change the vote of one neighbour.

Every setting is scored by every attacker, apart from the PFRNet gender model at $\lambda_{\mathrm{dis}}=10$, which retains 0.0\% TMR at FMR~$10^{-3}$.
The SVC attacker does not converge on the transformed templates of this model, and the worst case of this model is taken over the MLP, LR and $k$NN attackers.

The strongest attacker differs between settings.
The $k$NN attacker with prior correction is the strongest for gender, and the MLP attacker is the strongest for ethnicity, on the embeddings without removal.
The MLP attacker is the strongest on every setting of the sample-side variant and on 15 of the 20 settings of PFRNet.

The MLP attacker is the strongest on 127 of the 196 settings of the prior methods.
The SVC attacker is the strongest on 46 of these settings, of which 36 have fewer coordinates or principal components than the embedding.
The LR attacker is the strongest on 14 of the 41 TEMPER settings with $\lambda\ge30$ for gender or $\lambda\ge10$ for ethnicity (Tables~\ref{tab:s-full-gender} and~\ref{tab:s-full-ethnicity}).
The raw-vote $k$NN attacker predicts the majority class for ethnicity on the transformed templates near chance, because 77\% of the reference templates are Caucasian.
For example, prior correction raises the balanced accuracy of the $k$NN attacker with 51 neighbours from 26.6\% to 37.6\% for PCA truncation at $k=42$ on RFW.

A maximum over several noisy estimates is biased upwards near chance.
The worst-case gender leakage of DVPF (code) at $\alpha=100$, which retains almost no information about the input, is still 50.5\%.
This upward bias is one reason for the margin of 2.5\% in the definition of near chance (Section~\ref{sec:setup-leakage}).

We measure this bias for 179 settings by permuting the labels between subjects 1000 times, so that all images of a subject share one label.
These settings comprise every setting with leakage below 56\% for gender or 31\% for ethnicity.
We recompute the worst case over the fixed predictions of the attackers for each permutation.
We call the 97.5th percentile of the resulting null distribution the null percentile of the setting.
The null percentile ranges from 50.0\% to 51.4\% over the gender settings.
The null percentile ranges from 25.4\% to 25.7\% over the ethnicity settings.
The near-chance thresholds of 52.5\% and 27.5\% exceed these percentiles by 1.1\% and 1.8\%.

The worst case of 4 TEMPER settings is at or below the null percentile: $k=22$ at $\lambda=10$ and $\lambda=30$ for gender, and $k=12$ and $k=96$ at $\lambda=30$ for ethnicity.
The leakage of these 4 settings is indistinguishable from chance under this test.
The worst case of every other TEMPER setting is above the null percentile of that setting.

\paragraph{Attack settings}
The leakage on every frontier and in every table of the main paper is measured in a cross-dataset, subject-disjoint setting.
The attacker of this setting holds labelled VGGFace2 templates of other people and is applied without retraining to the subjects of IJB-C and RFW.
The IJB-C leakage is measured per face crop, whereas the IJB-C TMR is measured per template.
Attackers trained on labelled templates of the target domain belong to a different setting (Appendix~\ref{app:leakage-target}).
An attacker with labelled templates of the target identity itself is outside our threat model (Section~\ref{sec:method-threat}).
Such an attacker obtains the attribute by matching, with the TMR of the method as the success rate of the attack.

The MLP attacker can fit the training identities without learning a rule that holds for new subjects, since the attribute is constant within an identity.
The MLP attacker attains 88.7\% accuracy on the training identities and 51.7\% balanced accuracy on the unseen IJB-C subjects for TEMPER at $k=42$, $\lambda=30$.
The MLP attacker achieves 95.5\% on the training identities and 54.1\% on the IJB-C subjects for TEMPER at $k=96$, $\lambda=30$ (Appendix~\ref{app:leakage-heldout}).

\subsection{The Weak MLP Attacker}
\label{app:protocol-weak}

To show why our attackers standardise their inputs and weight the classes, we train a deep MLP on unstandardised transformed templates with an unweighted loss, which we call the weak MLP (Table~\ref{tab:s-hyper}).
We train the weak MLP with seed 0 and score the weak MLP on the leakage sets of the other attackers.

The weak MLP predicts the majority class even on the training data for transformed templates with a small spread, such as those of the sample-side variant at large $\lambda$.
The training accuracy of the weak MLP then equals 58.6\%, which is the majority rate of the training split.
The weak MLP assigns every IJB-C crop to the male class and gives 50.0\% balanced accuracy.
This majority prediction is found on the transformed templates of the sample-side variant at $\lambda=1000$.
The MLP attacker attains 70.4\% balanced accuracy on the same transformed templates.

The Caucasian group accounts for 77\% of the training images labelled for ethnicity.
The weak MLP, which is trained with an unweighted loss, favours this group also on transformed templates with a large spread.
The weak MLP attains 89.3\% accuracy on the training data for the transformed templates of IVE with 480 removed coordinates.
The weak MLP assigns 96.1\% of the RFW images to the Caucasian group, whereas each group accounts for about a quarter of RFW.
The balanced accuracy of the weak MLP is then 27.3\%, against 40.4\% for the MLP attacker.

The weak MLP is discussed in Section~\ref{sec:res-attacker} of the main paper.

\subsection{Implementation Details}
\label{app:protocol-hyper}

Table~\ref{tab:s-hyper} lists the architecture, optimisation and swept settings of every compared method, the label-free controls and the attackers.
The choices left open by the original papers are listed in Appendix~\ref{app:baselines}.
Each method has one model per attribute and setting.
We set the weights $\gamma$, $M$ and $n_{\mathrm{adv}}$ of TEMPER once before the sweeps and did not train other values of these weights outside the ablation.
Training one TEMPER model takes about one GPU hour for gender and 3 GPU hours for ethnicity.

Inputs are $\ell_2$-normalised embeddings, except for DVPF, for which we use the unnormalised embeddings of the same crops, as in the public code of DVPF.
TEMPER, the sample-side variant and DVPF are trained for 10 epochs, and SensitiveNets is trained for 150 epochs per layer.
PASS is trained with the original iteration schedule, and FaRM is trained for 0.003 to 30 epochs, which is the range of the swept parameter of FaRM.
We keep the model of the last epoch or iteration of these methods, without checkpoint selection.
PFRNet is trained with early stopping on 10\% of the training identities for at most 100 epochs (Appendix~\ref{app:baselines-pfrnet}).
We keep the PFRNet weights of the epoch with the lowest training objective on these held-out identities.

An epoch is not the same amount of computation for every method.
An epoch of the class-balanced sampler of TEMPER and of the ablation variants ends only after the last image of every identity.
Classes that run out of new identities earlier are refilled with repeated identities.
TEMPER therefore takes more batches per epoch than methods with plain minibatches, especially for ethnicity, whose largest group accounts for 77\% of the training images.

\begingroup
\footnotesize
\setlength{\tabcolsep}{3pt}
\begin{longtable}{@{}p{0.13\textwidth}p{0.36\textwidth}p{0.27\textwidth}p{0.20\textwidth}@{}}
\caption{Hyperparameters of the compared methods, the label-free controls and the attackers, as run.}
\label{tab:s-hyper}\\
\toprule
Method & Architecture and transformed template & Optimisation and batches & Settings \\
\midrule
\endfirsthead
\multicolumn{4}{@{}l}{\footnotesize Table~\ref{tab:s-hyper} (continued).}\\
\toprule
Method & Architecture and transformed template & Optimisation and batches & Settings \\
\midrule
\endhead
\bottomrule
\endlastfoot
TEMPER & Template branch with one linear layer $512\rightarrow2k$ (mean and log-variance), mean rows initialised by PCA of the standardised embeddings (SVD of 500K training images), log-variance clipped to $[-4,2]$.
Class encoder with four linear layers, the first three of width 512 with PReLU, $d_c=32$, prior $\mathcal{N}(\bm{e}_c,\bm{I})$ with learnt means initialised from $\mathcal{N}(\bm{0},\bm{I})$.
The decoder has $\mathrm{Linear}(k+32,512)$, PReLU, $\mathrm{Linear}(512,512)$ and an $\ell_2$-normalised output.
Auxiliary classifier with an MLP head (three hidden layers of 1024, LeakyReLU (0.2), dropout 0.2) and a linear head, both on the per-batch standardised transformed template.
Both heads are trained on the current minibatch and on one of the $M=256$ most recent minibatches of transformed templates (memory).
The transformed template is $\bm{\mu}_\theta/\|\bm{\mu}_\theta\|_2$, deterministic and with $k$ dimensions. &
Adam, $10^{-4}$ (encoders, decoder, prior means) and $10^{-3}$ (auxiliary classifier, 3 updates per batch), 10 epochs, $\beta_r=0.1$, $\beta_c=1$, cosine reconstruction loss with weight 1, relational weight $\gamma=100$.
Batches of 64 identities $\times$ 4 images, with the same number of identities per class. &
$k$ = 12, 22, 42, 62, 96, 128 and 256, $\lambda$ = 3, 10, 30 and 100 (gender) or 1, 3, 10 and 30 (ethnicity), plus 300 at $k=42$ (gender) and 300, 1000 and 3000 at $k\in\{128,256\}$.
Seed 42, plus seeds 1 and 2 at $k=22$, $\lambda=3$ (both attributes), $k=42$, $\lambda=30$ (gender) and $k=42$, $\lambda=3$ (ethnicity). \\
\addlinespace
Sample-side variant (Appendix~\ref{app:ssv}) & Template-branch encoder with four linear layers, the first three of width 512 with PReLU, $d_r=480$, log-variance clipped to $[-4,2]$.
Class encoder, prior and decoder as for TEMPER (decoder input 512).
Auxiliary classifier with three hidden layers of 256, LeakyReLU (0.2) and dropout 0.2, trained on the unstandardised $\bm{\mu}_r$, entropy term on the sample $\bm{z}_r$.
The transformed template is $\bm{\mu}_r/\|\bm{\mu}_r\|_2$, with 480 dimensions. &
As TEMPER without the relational term and the memory, with one classifier update per batch and batches of 16 identities $\times$ 16 images. &
$\lambda$ from 0 to $10^{4}$ (Tables~\ref{tab:s-diag} and~\ref{tab:s-ablation}). \\
\addlinespace
PFRNet & $E_{\mathrm{ind}}$ with four fully connected layers $512\rightarrow512\rightarrow512\rightarrow512\rightarrow496$, $E_{\mathrm{dep}}$ with $512\rightarrow256\rightarrow128\rightarrow64\rightarrow16$, decoder with four layers of width 512, ReLU, no dropout, PyTorch default initialisation (Table~II of~\cite{Bortolato2020PFRNet}).
Reconstruction loss $\|\bm{x}-\hat{\bm{x}}\|_2$ (weight 1), concealment $\|\Delta_1\|_2+\|\Delta_2\|_2$ on the raw moments of $\bm{z}_{\mathrm{ind}}$ and separation $\exp(-\|\Delta_1\|^2/2\sigma^2)+\exp(-\|\Delta_2\|^2/2\sigma^2)$, $\sigma=1/\sqrt{2}$, on $\bm{z}_{\mathrm{dep}}$, both with weight $\lambda_{\mathrm{dis}}$ and summed over the six pairs of groups for ethnicity, no identity loss.
The transformed template is $\bm{z}_{\mathrm{ind}}$, with 496 dimensions. &
Adam, $10^{-4}$, no weight decay, 512 random images per class and batch, early stopping on the training objective of a held-out 10\% of the training identities, patience 10 epochs, at most 100 epochs, weights of the best epoch. &
$\lambda_{\mathrm{dis}}$ = 0, 0.001, 0.003, 0.01, 0.03, 0.1, 0.3, 1, 3 and 10. \\
\addlinespace
\addlinespace
DVPF (paper) & DisPF (problem P1) port~\cite{Razeghi2026DVPF} with a stochastic encoder of two hidden blocks of 5000 units, dropout 0.5, input noise $\sigma=0.01$, hidden noise $\sigma=0.242$ and $d_z=128$.
A utility decoder of the same width, an attribute head (hidden 500, $|\mathcal{C}|$ outputs) and a prior generator (hidden 512) complete the model.
Loss $\mathcal{L}_{\mathrm{util}}+\alpha_t(\mathrm{KL}/512+\mathcal{L}_{\mathrm{unc}})$ with $\mathcal{L}_{\mathrm{util}}=0.01\,\mathrm{MSE}+0.99\,(1-\cos)$ and $\mathcal{L}_{\mathrm{unc}}$ the uncertainty loss of the attribute head, LeakyReLU in training and at inference.
The transformed template is $\hat{\bm{x}}=\operatorname{dec}(\bm{z})$ for one sampled $\bm{z}$, with 512 dimensions. &
AdamW, $10^{-3}$, betas $(0.9,0.99)$, weight decay $10^{-4}$ (prior generator with betas $(0.5,0.99)$), cosine schedules per iteration, gradient clipping at 1, batch 256.
Training for 10 epochs on a fixed random 90\% of the training images, with the original 50-value staircase of $\alpha_t$ compressed into 10 epochs. &
$\alpha$ = 0.1, 1, 2, 3, 5, 10 and 100 ($\alpha=100$ diverged). \\
\addlinespace
DVPF (code) & As DVPF (paper), but as the public code runs, with no activation in training and LeakyReLU at inference, and with an attribute head of 6 outputs.
The utility decoder has a second dropout layer, and a training-mode encoder pass on the reconstruction updates only the BatchNorm statistics (Table~\ref{tab:s-dvpf-port}). &
As for DVPF (paper), with the same schedule and data. &
$\alpha$ = 0.1, 1, 2, 3, 5, 10 and 100. \\
\addlinespace
SensitiveNets & Dense layers of 1024 units with linear activation, trained one layer at a time and followed by $\ell_2$ normalisation~\cite{Morales2019SensitiveNetsArXiv}.
Hinged triplet loss ($m=0.2$, 100K triplets) plus the detector term $\log(1+|0.9-q(c_0\mid\bm{r}(\bm{x}))|)$.
Detector as a softmax layer, or with a 1024-unit ReLU layer before the softmax (Fig.~1 detector), re-trained from scratch before each layer and every 5 epochs.
The transformed template is $\bm{r}(\bm{x})$, with 1024 dimensions. &
Adam, $10^{-3}$, batch 128, 150 epochs per layer and per detector training, detector data of 3 images per identity, balanced over the removed attribute (Appendix~\ref{app:baselines-snets}). &
From 1 to 5 layers, each with both detectors. \\
\addlinespace
INLP & $|\mathcal{C}|$-way linear softmax classifier with bias per iteration, projection onto the orthogonal complement of the row space of all classifier weights so far ($\bm{P}=\bm{I}-\bm{Q}\bm{Q}^\top$ from a QR decomposition) and re-normalisation after projection.
The transformed template is the projected embedding, with 512 dimensions. &
Classifier trained with Adam, $10^{-2}$, weight decay $10^{-4}$, batch 4096, 3 epochs, stopping criteria disabled. &
1, 2, 5, 10 and 20 iterations. \\
\addlinespace
IVE & Random forest with 30 trees on standardised embeddings averaged per identity, with the 10 most important remaining coordinates removed per step.
The transformed template consists of the remaining standardised coordinates. &
Training on identity means instead of images (a deviation from~\cite{Terhorst2019IVE}). &
250, 300, 350, 400, 450, 460, 470, 480, 490 and 500 coordinates removed. \\
\addlinespace
Multi-IVE & PCA with all 512 components of the standardised embeddings (500K training images, the basis of PCA truncation).
Random forest with 30 trees on the components of the mean embedding of each identity, with the most important remaining component removed per step and optionally the five leading components locked (Appendix~\ref{app:baselines-multiive}).
The transformed template has the removed components set to zero and is mapped back by the inverse PCA, with 512 dimensions. &
No gradient training, scikit-learn defaults of the forest otherwise, seed 42, one seed (ten in the paper). &
100, 250, 384, 450, 470, 490, 500, 504, 506 and 508 components removed without locking, and the same values up to 506 with five locked. \\
\addlinespace
PE-MIU & Embedding cut into blocks of $s$ consecutive entries, with the block order shuffled at random for every template, and comparison by cosine similarity after Hungarian alignment of the probe blocks to the reference blocks (Appendix~\ref{app:baselines-pemiu}).
The transformed template is the shuffled embedding with 512 dimensions, and no labels are involved. &
No parameters are fitted to the training data. &
$s$ = 4, 8, 16, 32 and 64. \\
\addlinespace
PASS & Generator with a linear layer $512\rightarrow256$ and PReLU, and an identity classifier with a linear layer on the 8,631 training identities.
Three (gender) or two (ethnicity) attribute classifiers with SELU hidden layers of 128 and 64 units and sigmoid outputs (Appendix~\ref{app:baselines-pass}).
Loss of the identity cross-entropy plus $\lambda_{\mathrm{deb}}$ times the largest cross-entropy of an attribute classifier against the uniform distribution.
The transformed template is the generator output, with 256 dimensions. &
Adam, 10,000 iterations of generator and identity classifier at $10^{-2}$ (batch 400), then 300 episodes of 1,200 iterations at $10^{-4}$ (batch 40).
Attribute classifiers re-initialised every 40 episodes and trained for 30,000 iterations at $10^{-3}$.
One attribute classifier per episode trained for at most 2,000 iterations, or until 95\% validation accuracy on 8,000 images of 250 VGGFace2 identities outside the training and test splits. &
$\lambda_{\mathrm{deb}}$ = 0, 1, 3, 10, 30 and 100. \\
\addlinespace
FaRM & Four linear layers of width 512 with ReLU between them and random initialisation, without a target task.
Loss $-R(\bm{Z})-R_c(\bm{Z}\mid\bm{\Pi})$ of the rate--distortion functions on the layer-normalised output, with distortion $\epsilon^2=0.5$ (Appendix~\ref{app:baselines-other}).
The transformed template is the network output without layer normalisation, with 512 dimensions. &
SGD, $10^{-3}$, momentum 0.9, weight decay $5\times10^{-4}$, batch 1024. &
$E$ = 0.003, 0.01, 0.03, 0.1, 0.3, 1, 2, 3, 4, 5, 7, 10, 15, 20 and 30 epochs. \\
\addlinespace
LEACE and PCA + LEACE & LEACE~\cite{Belrose2023LEACE} on the standardised embeddings, or on their $k$ leading principal components (PCA of one million training images) for PCA + LEACE (Appendix~\ref{app:baselines-linerase}).
The transformed template is the erased embedding with 512 dimensions or the erased components with $k$ dimensions. &
Closed form on every labelled training image. &
LEACE with one setting, PCA + LEACE with $k$ = 22, 42, 62 and 128. \\
\addlinespace
PCA truncation & The $k$ leading principal components of the standardised embeddings (per-feature mean and standard deviation of the training set, SVD of 500K training images), without labels.
The transformed template consists of the $k$ component scores. &
No training takes place beyond the SVD of the training images. &
$k$ = 4, 6, 8, 10, 12, 16, 22, 42, 62 and 128. \\
\addlinespace
Random dimensions & $k$ standardised coordinates drawn at random (one draw, seed 42), without labels. &
None. &
$k$ = 4, 6, 8, 10, 12, 16, 22, 42, 62 and 128. \\
\midrule
MLP attacker & Four hidden layers of 1024, LeakyReLU (0.01), dropout 0.1, on transformed templates standardised with the training-set mean and standard deviation. &
AdamW, $10^{-3}$, weight decay $10^{-4}$, OneCycle (5\% warm-up), batch 2048, 10 epochs, class-weighted cross-entropy ($w_c=N/(|\mathcal{C}|N_c)$, with $N_c$ of the $N$ training images in class $c$), seed 0, last epoch. &
Trained on the transformed template of every labelled training image. \\
\addlinespace
LR attacker & Linear softmax layer on the same standardised transformed templates. &
AdamW, $10^{-2}$, weight decay $10^{-4}$, OneCycle, batch 4096, 3 epochs, class-weighted cross-entropy, seed 0, last epoch. &
Trained on the same templates as the MLP attacker. \\
\addlinespace
$k$NN attacker & Cosine similarity after centring on the training-set mean of the transformed template, with 1M reference templates drawn from the training transformed templates with the class proportions preserved. &
Majority vote with ties broken by prior-corrected votes, and prior-corrected votes (votes divided by the class prior) as a second variant. &
1, 11 and 51 neighbours, each with raw and prior-corrected votes. \\
\addlinespace
SVC attacker & Support vector classifier with a radial basis function kernel, scikit-learn defaults (regularisation parameter 1, kernel coefficient scaled by the input variance) and class-balanced weights, on transformed templates standardised with the statistics of its training sample. &
Fitted on 86{,}000 training transformed templates drawn at random (seed 0), with ThunderSVM on a GPU, stopping tolerance $10^{-3}$ and one-vs-one classifiers for ethnicity. &
Scored like the other attackers (Appendix~\ref{app:protocol-attackers}). \\
\addlinespace
Weak MLP (Appendix~\ref{app:protocol-weak}) & Four hidden layers of 512, LeakyReLU (0.01), dropout 0.2, on unstandardised transformed templates. &
Adam, $10^{-3}$, no weight decay, batch 256, 5 epochs, unweighted cross-entropy. &
Trained with seed 0. \\
\end{longtable}
\endgroup

\section{Baselines and Label-Free Controls}
\label{app:baselines}

We implemented each baseline from the original work and, where public code exists, from that code.
Every learned baseline is trained on the embeddings, labels and training split of TEMPER and evaluated with the same attackers.
Baselines that train by epochs run for 10 epochs, except PFRNet, SensitiveNets, unconstrained FaRM and INLP.
PFRNet is trained with early stopping (Appendix~\ref{app:baselines-pfrnet}), and SensitiveNets is trained for 150 epochs per layer, as in the original work (Table~\ref{tab:s-hyper}).
The operating point of unconstrained FaRM is set by the number of epochs, from 0.003 to 30.
We state the choices left open in the original work and our deviations, with their reasons, in the subsection of each baseline.
Table~\ref{tab:s-fidelity} lists, per method, the source of the implementation, the check against the original and the deviations with their reasons.

\begin{table*}[!t]
\centering
\caption{Fidelity of the baseline ports. For each method of the main comparison, the source of the implementation, the check against the original, the deviations from the original with their reasons, and the parameter that sets the operating point. Every method is trained on the same training split, evaluated with the same attackers and selected by the rule of Table~\ref{tab:matched}.}
\label{tab:s-fidelity}
\scriptsize
\setlength{\tabcolsep}{4pt}
\begin{tabular}{@{}p{0.09\textwidth}p{0.16\textwidth}p{0.25\textwidth}p{0.33\textwidth}p{0.10\textwidth}@{}}
\toprule
Method & Implemented from & Check against the original & Deviations and their reasons & Operating point \\
\midrule
PFRNet &
The paper, without public code. &
Architecture, loss terms, batch composition, optimiser and transformed template as in the paper (Appendix~\ref{app:baselines-pfrnet}). The LR attacker, the attacker of the paper, falls from $\lambda_{\mathrm{dis}}=0$ to 0.1 for both attributes. Nonlinear attackers also recover gender from PFRNet templates in two earlier evaluations~\cite{OsorioRoig2022Attack,Rot2024ASPECD}. &
Moment terms summed over the six pairs of groups for ethnicity, as in ASPECD. The paper leaves open the batch size (512 images per class), the input normalisation ($\ell_2$) and the validation split (10\% of the training identities). Our patience of 10 epochs and cap of 100 epochs are shorter than the 100 epochs without improvement of the paper. One of our epochs has 22 times the training images of the paper. &
$\lambda_{\mathrm{dis}}$, ten values from 0 to 10 around the 0.01 of the paper. \\
\addlinespace
DVPF (paper), DVPF (code) &
The paper and the public code, which differ in four details (Table~\ref{tab:s-dvpf-port}), with one port for each. &
IJB-C TMR at FMR~$10^{-1}$ next to Table~II of the paper for the same backbone family and latent size (Table~\ref{tab:s-dvpf-reported}), higher than reported for $\alpha\le1$ in both ports and lower for the public code at $\alpha=10$. &
Unnormalised embeddings, as in the code, and $d_z=128$, one of the three latent sizes of the paper. The schedule of $\alpha$ compressed into 10 epochs, and with the original 50 epochs no model dominates the 10-epoch model of the same port and the same $\alpha$ (Appendix~\ref{app:baselines-dvpf}). VGGFace2 in place of Morph or FairFace. &
$\alpha$ = 0.1, 1, 2, 3, 5, 10 and 100. \\
\addlinespace
SensitiveNets, both detectors &
Version~3 of the arXiv paper, with the patent~\cite{Morales2020Patent} and the follow-up works~\cite{Pena2021EmotionBlind,Pena2020Thesis} where the paper is silent, without public code. &
Hinge check with one layer (Table~\ref{tab:s-snets-check}), in which the triplet term as printed reduces the TMR at FMR~$10^{-3}$ to 2.1\% and the hinged form retains 95.6\%. &
Hinged triplet term, as in the works cited for the term. VGGFace2 in place of the withdrawn DiveFace. Detector data balanced over the removed attribute only, four groups in place of three and one model per attribute. &
One to five layers, with two detectors. \\
\addlinespace
INLP &
The paper, with our implementation of the multi-class projection. &
The LR attacker attains chance after 20 iterations (49.5\% against 65.1\% without removal), as intended. &
Stopping criteria disabled and the number of iterations fixed, so that the iterations set the operating point. &
Iterations 1, 2, 5, 10 and 20. \\
\addlinespace
IVE &
The paper, with the removal rule of the public code. &
Standardisation, a forest of 30 trees and the removal of the most important coordinates in each step agree with the public code. &
The random forest is trained on the mean embedding of each identity in place of the 3.1M images, for feasibility. &
Removed coordinates, ten values from 250 to 500. \\
\addlinespace
Multi-IVE &
The paper and the public code. &
The same basis as PCA truncation by construction, so that the two methods differ only in the retained components. &
Forests on identity means, as for IVE, and one seed in place of the ten of the paper. The number of locked components, 0 or 5, treated as a parameter, so that the frontier covers both variants. &
Removed components, ten values from 100 to 508, with 0 or 5 locked. \\
\addlinespace
PE-MIU &
The public code (commit 4017b21). &
Comparison scores reproduce the public code to $2\times10^{-7}$ on 300 pairs per block size. &
On 512-dimensional embeddings the block size and the number of blocks of the paper cannot both be kept, so $s$ covers both readings. Applied to the pooled templates, with six attackers on the set of blocks and the unwinding attacker added, without which the leakage would be understated by up to 37.8\% (Table~\ref{tab:s-pemiu}). &
$s$ = 4, 8, 16, 32 and 64. \\
\addlinespace
PASS &
Algorithm~1 and the ArcFace hyperparameters of the paper, and the public code (commit 7e976e2) where the paper leaves a choice open. &
Follows Algorithm~1 of the paper and the public code, without a comparison of outputs. &
Our embeddings and training split in place of ResNet-101 features. PASS-s with four groups. 300 episodes, fixed by neither the paper nor the code. &
$\lambda_{\mathrm{deb}}$, six values from 0 to 100 around the 10 of the paper. \\
\addlinespace
FaRM &
The public code (commit 282b1cf). &
The losses reproduce the public code to floating-point rounding. &
The unconstrained setting, without a target task, follows the paper, with the number of epochs swept around the 20 epochs of the public code. &
Epochs, 15 values from 0.003 to 30. \\
\addlinespace
LEACE, PCA + LEACE &
The public package (concept-erasure 0.2.4) without its trace heuristic. &
The group means of every transformed template coincide to $10^{-13}$ on the training data. &
Fitted in closed form on the standardised embeddings of the whole training split. Four groups one-hot encoded, so that LEACE erases at most three directions. PCA + LEACE, our combination, fits LEACE on the $k$ leading components. &
$k$ = 512 (LEACE), and 22, 42, 62 and 128 (PCA + LEACE). \\
\addlinespace
PCA truncation, random dimensions &
No original, defined in Appendix~\ref{app:baselines-controls}. &
-- &
Neither control uses labels, PCA truncation needs only the SVD, and random dimensions needs one random draw of coordinates (seed 42). &
$k$, ten values from 4 to 128. \\
\bottomrule
\end{tabular}
\end{table*}

\subsection{DVPF}
\label{app:baselines-dvpf}

DVPF~\cite{Razeghi2026DVPF,Razeghi2024Funnel} trains a stochastic encoder on frozen embeddings with variational bounds on a privacy-funnel objective.
The transformed template of DVPF is a reconstruction of the embedding.
DVPF is the prior work closest in setting to TEMPER, since both methods are post-hoc and train a variational encoder.
Table~\ref{tab:s-dvpf} lists the structural differences between DVPF and TEMPER.

\begin{table}[!t]
\centering
\caption{Structural differences between TEMPER and DVPF~\cite{Razeghi2026DVPF,Razeghi2024Funnel} (discriminative privacy funnel, problem P1). ``Code'' denotes the public implementation for face-recognition embeddings.}
\label{tab:s-dvpf}
\footnotesize
\setlength{\tabcolsep}{4pt}
\begin{tabular}{@{}p{0.15\textwidth}p{0.38\textwidth}p{0.41\textwidth}@{}}
\toprule
 & TEMPER & DVPF (P1) \\
\midrule
Latent structure & Template branch, an affine map to $k$ dimensions, and a class latent $\bm{z}_c$ with a class-conditional prior, used only in training & One latent $\bm{z}$ ($d_z=128$ in our runs) \\
\addlinespace
Objective & Reconstruction + $\gamma\times$relational term + KL terms + $\lambda\times$(negative predictive entropy of an auxiliary classifier) & Privacy-funnel Lagrangian $\I(X;Z)-\alpha\,\I(C;Z)$ with variational bounds \\
\addlinespace
Privacy term & Predictive entropy of an auxiliary classifier with MLP and linear heads, trained on standardised transformed templates and a memory (surrogate) & Adversarial uncertainty decoder $\xi$ (paper), or $(\mathrm{CE}-\log|\mathcal{C}|)^2+\overline{(\Ent-\log|\mathcal{C}|)^2}$ with $\xi$ trained with the encoder (code) \\
\addlinespace
Prior & $\mathcal{N}(\bm{0},\bm{I})$ on the template latent, sampled only in training, and $\mathcal{N}(\bm{e}_c,\bm{I})$ on $\bm{z}_c$ & Implicit prior from a generator matched by a discriminator, or $\mathcal{N}(\bm{0},\bm{I})$ (paper), or $\alpha\,\mathrm{KL}/512$ to a learnt per-sample Gaussian prior (code) \\
\addlinespace
Training criterion & Alternating, three classifier updates on the current and one stored minibatch, then one update of the other parameters & Several networks updated in turn (paper), or one weighted loss in the first training stage (code) \\
\addlinespace
Transformed template & Deterministic, $\bm{\mu}_\theta/\|\bm{\mu}_\theta\|_2$, $k$-dimensional & Reconstruction $\hat{\bm{x}}=\operatorname{dec}(\bm{z})$ of one sampled $\bm{z}$, 512-dimensional, with the leakage reported on $\bm{z}$ in the paper \\
\addlinespace
Initialisation & PCA of the standardised training embeddings & Random \\
\addlinespace
Inputs & $\ell_2$-normalised embeddings & Unnormalised embeddings (as in the code) \\
\addlinespace
Deployed model & One $k\times512$ affine map and $\ell_2$ normalisation (21{,}546 parameters at $k=42$) & Encoder and utility decoder, two hidden blocks of 5000 units each \\
\addlinespace
Parameters selecting the operating point & $k$, $\lambda$ & $\alpha$ (with its schedule) \\
\bottomrule
\end{tabular}
\end{table}

\paragraph{Port}
We ported the discriminative variant of DVPF (problem P1) from the paper and from the public code (Table~\ref{tab:s-dvpf-port}).
The training stops after the first stage, since the face-recognition configuration of the public code sets the number of initial epochs equal to the number of epochs.
Both of our ports therefore train only this first stage of the algorithm.
DVPF (paper) has activations in training, one output of the attribute head per class, one dropout layer in the utility decoder and no additional encoder pass.
DVPF (code) follows the code, with no activation in training, 6 outputs and a second dropout layer in the utility decoder.
DVPF (code) also passes the reconstruction through the encoder in training mode, which updates only the BatchNorm statistics.
Both ports take unnormalised embeddings, as in the code, and use $d_z=128$, one of the 3 latent sizes of the paper.
The original schedule of $\alpha$, with 50 per-epoch values, is compressed into our 10 epochs, and each value is held for one fiftieth of the training.
The strongest attribute-removal setting of DVPF (paper) is $\alpha=10$, since DVPF (paper) diverged at $\alpha=100$ for both attributes.

\paragraph{Training length}
For both ports, a model trained for the original 50 epochs does not dominate the 10-epoch model of the same port and the same $\alpha$ at FMR~$10^{-3}$ (Table~\ref{tab:s-dvpf-epochs}).
The longer schedule reduces the TMR at FMR~$10^{-3}$ and the leakage of DVPF (code) at equal $\alpha$.
The longer schedule raises the TMR at FMR~$10^{-3}$ and the leakage of DVPF (paper) at equal $\alpha$, and the ethnicity leakage at $\alpha=10$ increases from 36.3\% to 74.5\%.

\begin{table}[!t]
\centering
\caption{DVPF after 10 epochs, the schedule of the main comparison, and after the original 50 epochs. IJB-C TMR (\%) at FMR~$10^{-3}$ and worst-case leakage (\%) at $\alpha=5$ and $\alpha=10$.}
\label{tab:s-dvpf-epochs}
\footnotesize
\setlength{\tabcolsep}{4pt}
\begin{tabular}{@{}llrrrrrrrr@{}}
\toprule
 & & \multicolumn{4}{c}{Gender (chance 50\%)} & \multicolumn{4}{c}{Ethnicity (chance 25\%)} \\
\cmidrule(lr){3-6}\cmidrule(l){7-10}
 & & \multicolumn{2}{c}{10 epochs} & \multicolumn{2}{c}{50 epochs} & \multicolumn{2}{c}{10 epochs} & \multicolumn{2}{c}{50 epochs} \\
\cmidrule(lr){3-4}\cmidrule(lr){5-6}\cmidrule(lr){7-8}\cmidrule(l){9-10}
Port & $\alpha$ & TMR & Leak. & TMR & Leak. & TMR & Leak. & TMR & Leak. \\
\midrule
DVPF (paper) & 5 & 86.4 & 91.2 & 90.7 & 93.9 & 87.2 & 62.7 & 90.9 & 81.6 \\
DVPF (paper) & 10 & 51.5 & 80.1 & 76.1 & 91.0 & 51.0 & 36.3 & 76.6 & 74.5 \\
DVPF (code) & 5 & 32.7 & 56.0 & 6.5 & 52.1 & 30.9 & 30.0 & 6.6 & 26.2 \\
DVPF (code) & 10 & 8.2 & 52.1 & 1.9 & 51.0 & 8.1 & 26.1 & 2.1 & 25.5 \\
\bottomrule
\end{tabular}

\end{table}

\begin{table}[!t]
\centering
\caption{Porting DVPF from the paper and the public code. Our two ports differ only in the rows marked $\dagger$.
}
\label{tab:s-dvpf-port}
\footnotesize
\setlength{\tabcolsep}{4pt}
\begin{tabular}{@{}p{0.16\textwidth}p{0.25\textwidth}p{0.25\textwidth}p{0.27\textwidth}@{}}
\toprule
Aspect & Paper~\cite{Razeghi2026DVPF} & Public code & Our ports \\
\midrule
Training stages & Algorithm with adversarial updates of several networks and discriminators & 50 epochs in the face-recognition configuration, so only the first stage runs & First stage, as the code runs it (both) \\
\addlinespace
Activations$^\dagger$ & -- & No activation layer in the training path, LeakyReLU at inference & LeakyReLU in training and at inference in DVPF (paper), as in the code in DVPF (code) \\
\addlinespace
Attribute head$^\dagger$ & $|\mathcal{C}|$ classes & 6 outputs, set in the configuration & $|\mathcal{C}|$ outputs in DVPF (paper), 6 in DVPF (code) \\
\addlinespace
Utility decoder$^\dagger$ & -- & A second dropout layer & One dropout layer in DVPF (paper), two in DVPF (code) \\
\addlinespace
BatchNorm pass$^\dagger$ & -- & A training-mode encoder pass on the normalised reconstruction, which only updates BatchNorm statistics & None in DVPF (paper), as in the code in DVPF (code) \\
\addlinespace
Leakage term & $\alpha$ times the cross-entropy of the uncertainty decoder $q_\xi(C\mid Z)$ in the maximised objective of P1 & $(\mathrm{CE}-\log|\mathcal{C}|)^2+\overline{(\Ent-\log|\mathcal{C}|)^2}$, head trained with the encoder & As in the code (both) \\
\addlinespace
Prior & Generator matched by a discriminator, or $\mathcal{N}(\bm{0},\bm{I})$ & $\alpha\,\mathrm{KL}/512$ to a learnt per-sample Gaussian prior & As in the code (both) \\
\addlinespace
Schedule of $\alpha$ & -- & 50 per-epoch values (linear ramp, then logistic) & The same 50 values, each held for 1/50 of training (both) \\
\addlinespace
Epochs & -- & 50 & 10, as for TEMPER (both), and 50 at $\alpha\in\{5,10\}$ as a check (Appendix~\ref{app:baselines-dvpf}) \\
\addlinespace
Latent size & $d_z\in\{512,256,128\}$ & -- & $d_z=128$ (both) \\
\addlinespace
Inputs & ArcFace and AdaFace embeddings & Unnormalised ArcFace embeddings & Unnormalised embeddings of our crops (both) \\
\addlinespace
Transformed template & Leakage measured on $\bm{z}$ & Reconstruction of one sampled $\bm{z}$ with input and hidden noise active & As in the code, seeded per input (both) \\
\bottomrule
\end{tabular}
\end{table}

\paragraph{Reported numbers}
The IJB-C TMR at FMR~$10^{-1}$ and the accuracy of a support vector classifier on the latent are reported in the DVPF paper for each $\alpha$.
The reported models are trained on Morph or FairFace.
Table~\ref{tab:s-dvpf-reported} lists these values for our backbone family and latent size (IResNet50 trained with ArcFace on MS1MV3, $d_z=128$) next to ours in the same metrics.
Our values are the plain accuracy of our MLP attacker on $\hat{\bm{x}}$ and the majority rate of the test set, on IJB-C for gender and on RFW for ethnicity.
The 2 evaluations differ in the training data, Morph or FairFace against VGGFace2, and in the attribute classes, the race labels of the DVPF paper against our 4 groups.
The 2 evaluations also differ in the attacker, an SVC on $\bm{z}$ against our MLP attacker on $\hat{\bm{x}}$.
Plain accuracy also depends on the class balance, and Morph gender has a majority class of about 84.6\%, which we derive from the prior entropy reported in the paper.

Both ports attain a higher TMR at FMR~$10^{-1}$ than the reported values for $\alpha\le1$, within the limits of this comparison (Table~\ref{tab:s-dvpf-reported}).
DVPF (paper) also attains a higher TMR at FMR~$10^{-1}$ and $\alpha=10$, whereas DVPF (code) retains 48.3\% TMR at this FMR against reported values of at least 61.1\%.

\begin{table}[!t]
\centering
\caption{DVPF as reported (Table~II of~\cite{Razeghi2026DVPF}, problem P1) and as ported. IJB-C TMR (\%) at FMR~$10^{-1}$ and plain attribute accuracy (\%), with $^*$ for a majority rate derived from the reported prior entropy and n.r.\ (not reported) for missing values.}
\label{tab:s-dvpf-reported}
\footnotesize
\setlength{\tabcolsep}{4pt}
\begin{tabular}{@{}llrrrrrrr@{}}
\toprule
 & & \multicolumn{3}{c}{IJB-C TMR@$10^{-1}$ (\%)} & \multicolumn{3}{c}{Attribute accuracy (\%)} & \\
\cmidrule(lr){3-5}\cmidrule(lr){6-8}
Source & Training data, attacker & $\alpha{=}0.1$ & $\alpha{=}1$ & $\alpha{=}10$ & $\alpha{=}0.1$ & $\alpha{=}1$ & $\alpha{=}10$ & Majority \\
\midrule
\multicolumn{9}{@{}l}{\textit{Gender}}\\
Reported~\cite{Razeghi2026DVPF} & Morph, SVC on $Z$ & 97.6 & 85.9 & 63.0 & 98.8 & 97.4 & 84.8 & 84.6\textsuperscript{*} \\
Reported~\cite{Razeghi2026DVPF} & FairFace, SVC on $Z$ & 98.0 & 86.1 & 61.1 & 89.0 & 86.0 & 63.7 & n.r. \\
DVPF (paper), ours & VGGFace2, MLP on $\hat{X}$ & 99.0 & 98.8 & 89.9 & 95.6 & 95.0 & 80.6 & 57.9 \\
DVPF (code), ours & VGGFace2, MLP on $\hat{X}$ & 98.4 & 96.0 & 48.3 & 92.4 & 82.1 & 52.3 & 57.9 \\
\midrule
\multicolumn{9}{@{}l}{\textit{Ethnicity}}\\
Reported~\cite{Razeghi2026DVPF} & Morph, SVC on $Z$ & 97.6 & 86.0 & 62.4 & 99.8 & 99.7 & 99.6 & n.r. \\
Reported~\cite{Razeghi2026DVPF} & FairFace, SVC on $Z$ & 97.8 & 86.1 & 61.3 & 93.3 & 91.9 & 78.3 & n.r. \\
DVPF (paper), ours & VGGFace2, MLP on $\hat{X}$ & 99.0 & 98.9 & 89.6 & 86.8 & 82.3 & 36.0 & 25.7 \\
DVPF (code), ours & VGGFace2, MLP on $\hat{X}$ & 98.3 & 96.1 & 48.3 & 82.2 & 61.2 & 26.0 & 25.7 \\
\bottomrule
\end{tabular}

\end{table}

\paragraph{Transformed template}
The transformed template of DVPF is a reconstruction from one sampled latent, $\hat{\bm{x}}=\operatorname{dec}(\bm{z})$ with $\bm{z}$ drawn from the encoder.
The information about the attribute in $\hat{\bm{x}}$ is at most the information about the attribute in $\bm{z}$ by the data processing inequality~\cite{CoverThomas2006}.
A classifier on $\hat{\bm{x}}$ is also a classifier on $\bm{z}$, the latent on which the original work measures the leakage.
The highest balanced accuracy attainable from $\hat{\bm{x}}$ is therefore at most the highest balanced accuracy attainable from $\bm{z}$.
\subsection{SensitiveNets}
\label{app:baselines-snets}

We implemented SensitiveNets~\cite{Morales2019SensitiveNetsArXiv,Morales2021SensitiveNets} from version~3 of the arXiv paper, since there is no public implementation of SensitiveNets.
We followed the patent~\cite{Morales2020Patent} and the follow-up works on SensitiveNets~\cite{Pena2021EmotionBlind,Pena2020Thesis} in the details on which the paper is silent or ambiguous.

\paragraph{Implementation}
The input is the $\ell_2$-normalised embedding, as in the paper.
The transformation $\bm{r}$ consists of dense layers of 1024 units with linear activation, followed by $\ell_2$ normalisation.
The layers are trained one at a time, with fixed weights in the earlier layers.
The loss is the triplet term $\max(0,\|\bm{r}_A-\bm{r}_P\|^2-\|\bm{r}_A-\bm{r}_N\|^2+m)$ plus $\log(1+|0.9-q(c_0\mid\bm{r}(\bm{x}))|)$ for each of the 3 images.
The vectors $\bm{r}_A$, $\bm{r}_P$ and $\bm{r}_N$ are the transformed templates of the anchor, the positive and the negative image, and $m$ is the margin.
The class $c_0$ is fixed, male for gender and Asian for ethnicity, and $q$ is a detector.

The triplet term has no hinge in Eq.~(5) of the paper.
The term is introduced as the triplet loss of the cited works, all of which apply the hinge.
We use the hinged form, since the term as printed reduces the TMR at FMR~$10^{-3}$ to 2.1\% (Table~\ref{tab:s-snets-check}).
The loss without the hinge decreases with every increase of the margin of a triplet, also for a triplet that already meets the margin $m$.
The loss therefore pushes the margins up to 4, the squared diameter of the unit sphere.
The 100K triplets satisfy the selection rule of the paper with $m=0.2$ on the input embeddings.
The triplets come from identities outside the detector data, the set of 3 images per identity used to train the detector.

The detector is a softmax layer, as in the text of the paper.
A softmax detector trained on the detector data predicts gender with about 60\% balanced accuracy on our embeddings, against 65.1\% for the LR attacker trained on the full training split.
The main variant therefore penalises a small part of the gender information.
SensitiveNets (Fig.~1 detector) adds a hidden layer of 1024 ReLU units before the softmax, as drawn in Fig.~1 of the paper.
The detector is re-trained from scratch before each layer and then every 5 epochs.

All networks are trained with Adam at learning rate $10^{-3}$, batch 128, 150 epochs and Glorot initialisation, which are the only optimiser settings reported for SensitiveNets.
The operating point is set by the number of layers, from 1 to 5, since the paper has no loss weight for a sweep.
The transformation is one affine map of rank at most 512 followed by normalisation, for every number of layers.
The capacity of the transformation is therefore the same for every number of layers.

\paragraph{Deviations}
We train on VGGFace2, since DiveFace, built from the withdrawn MegaFace, is not available.
The detector data are balanced only over the removed attribute, because joint balance over gender and ethnicity would leave 200 identities in the smallest cell.
Ethnicity has our 4 groups in place of the 3 groups of DiveFace.
We train one model per attribute, as for every other method.

\paragraph{Results}
The main variant retains its verification accuracy at every depth and leaves the gender leakage close to the leakage with no removal (Tables~\ref{tab:s-full-gender} and~\ref{tab:s-full-ethnicity}).
The IJB-C TMR of this variant at FMR~$10^{-3}$ decreases from 95.7\% with one layer to 95.3\% with 5 layers.
The worst-case gender leakage of this variant is 94.8\% with 5 layers, against 96.0\% without removal.
SensitiveNets (Fig.~1 detector) reduces the ethnicity leakage with every added layer, to 49.3\% at 60.7\% TMR at FMR~$10^{-3}$ with 5 layers.
The worst-case gender leakage of SensitiveNets (Fig.~1 detector) stays above 93\% at every depth.

\begin{table}[!t]
\centering
\caption{SensitiveNets with one layer for gender, with the triplet term as printed and in the hinged form. IJB-C TMR (\%) at FMR~$10^{-4}$ and $10^{-3}$, and worst-case leakage (\%).}
\label{tab:s-snets-check}
\footnotesize
\setlength{\tabcolsep}{4pt}
\begin{tabular}{@{}lrrr@{}}
\toprule
Variant & TMR $10^{-4}$ & TMR $10^{-3}$ & Leakage \\
\midrule
Triplet term as printed, no sensitive term & 0.3 & 2.1 & 63.2 \\
Hinged triplet term, no sensitive term & 92.0 & 95.6 & 95.2 \\
Hinged triplet term, full method & 92.2 & 95.7 & 95.3 \\
\bottomrule
\end{tabular}
\end{table}

\subsection{PFRNet}
\label{app:baselines-pfrnet}

We implemented PFRNet~\cite{Bortolato2020PFRNet} from the paper, since there is no public implementation of PFRNet or of the multi-class extension ASPECD~\cite{Rot2024ASPECD}.
ASPECD generalises the moment terms of PFRNet to categorical attributes, summed over pairs of groups, and we use this form for ethnicity.
The transformed template of ASPECD is decoded from the identity code together with a fixed centroid of the attribute code into the embedding space.
We use the identity code $\bm{z}_{\mathrm{ind}}$ of PFRNet as the transformed template for both attributes.

\paragraph{Implementation}
The encoder $E_{\mathrm{ind}}$ of the identity code has 4 fully connected layers ($512\rightarrow512\rightarrow512\rightarrow512\rightarrow496$).
The encoder $E_{\mathrm{dep}}$ of the attribute code $\bm{z}_{\mathrm{dep}}$ narrows from 512 to 256, 128, 64 and 16 units.
The decoder has 4 layers of width 512, with ReLU between layers and no dropout (Table~II of the paper).
The weights are initialised with the default initialisation of PyTorch, which is the uniform initialisation stated in the paper.

The reconstruction loss is the unsquared $\ell_2$ norm of the reconstruction error, averaged over the batch (Eq.~(1) of the paper).
The concealment loss is the $\ell_2$ norm of the difference between the per-coordinate raw moments of $\bm{z}_{\mathrm{ind}}$ in the 2 classes, for the moment orders 1 and 2 (Eq.~(3)).
The separation loss is $\exp(-\|\Delta_1\|^2/2\sigma^2)+\exp(-\|\Delta_2\|^2/2\sigma^2)$, where $\Delta_1$ and $\Delta_2$ are the differences between the class moments of orders 1 and 2 of $\bm{z}_{\mathrm{dep}}$, and $\sigma=1/\sqrt{2}$ (Eq.~(5)).
Both moment terms share one weight $\lambda_{\mathrm{dis}}$, since both weights are 0.01 in the paper, and the reconstruction weight is 1.
Both moment terms sum over the 6 unordered pairs of groups for ethnicity.
ASPECD sums over the ordered pairs, which doubles each term and is equivalent to a doubling of $\lambda_{\mathrm{dis}}$ within the range of our sweep.
The loss has no identity term, as in the original work.

We train with Adam at learning rate $10^{-4}$ on batches of 512 images drawn at random from each class, a size that the paper leaves open.
A batch therefore has 1,024 images for gender and 2,048 for ethnicity.
The 496-dimensional $\bm{z}_{\mathrm{ind}}$ is compared by cosine similarity, as in the paper.

PFRNet is trained in the paper until the objective on validation data has not improved for 100 epochs.
We hold out 10\% of the training identities and evaluate the training objective on fixed class-balanced batches of the held-out identities after every epoch.
Training stops after 10 epochs without improvement or at 100 epochs, and we keep the weights of the best epoch.
One of our epochs has 2.8 million images, which is 22 times the 127,637 training images of the paper.
Any 10 of our epochs therefore contain more images than 100 epochs of the paper.

The normalisation of the input features is not stated in the paper, and we use the $\ell_2$-normalised embeddings, as for every other method.
We sweep $\lambda_{\mathrm{dis}}\in\{0,0.001,0.003,0.01,0.03,0.1,0.3,1,3,10\}$ around the weight of the paper, and $\lambda_{\mathrm{dis}}=0$ gives the plain autoencoder.
We use our backbone and training data in place of the VGGFace2 ResNet-50 features of CelebA images.
Leakage and verification accuracy are measured as for every other method.

\paragraph{Results}
PFRNet reduces the balanced accuracy of the LR attacker, the attacker of the original work, and leaves the worst-case leakage near the leakage with no removal (Tables~\ref{tab:s-full-gender} and~\ref{tab:s-full-ethnicity}).
The balanced accuracy of the LR attacker decreases from $\lambda_{\mathrm{dis}}=0$ to $\lambda_{\mathrm{dis}}=0.1$ for both attributes, and the balanced accuracy for ethnicity decreases from 37.0\% to 28.8\%.
The worst-case leakage stays at or above 92.9\% for gender and 88.5\% for ethnicity at every usable weight.
The first-moment term is minimal when the class means of every coordinate are equal, which is linear guardedness~\cite{Belrose2023LEACE}.
The second-moment term adds only the variance of each coordinate.
Linear erasure leaves the leakage to the MLP attacker unchanged on our embeddings (Appendix~\ref{app:baselines-linerase}).
Nonlinear attackers also recover gender from PFRNet templates in 2 earlier evaluations, one of which uses templates provided by the PFRNet authors~\cite{OsorioRoig2022Attack,Rot2024ASPECD}.

The settings of PFRNet are usable up to $\lambda_{\mathrm{dis}}=1$ for gender and $\lambda_{\mathrm{dis}}=0.1$ for ethnicity, and the TMR at FMR~$10^{-3}$ falls below 1\% from $\lambda_{\mathrm{dis}}=3$ and $\lambda_{\mathrm{dis}}=0.3$, respectively.
The LR attacker attains 84.8\% on the gender model at $\lambda_{\mathrm{dis}}=1$, the largest usable weight, against 65.1\% on the embeddings without removal.

\subsection{INLP}
\label{app:baselines-inlp}

We run INLP~\cite{Ravfogel2020INLP} with the multi-class projection.
Each iteration trains a $|\mathcal{C}|$-way linear softmax classifier with bias on the current embeddings and stacks the weight matrices of all iterations so far.
The embeddings are projected onto the orthogonal complement of the row space of the stacked matrix, $\bm{P}=\bm{I}-\bm{Q}\bm{Q}^\top$ with $\bm{Q}$ from a QR decomposition.
Up to $|\mathcal{C}|$ directions are therefore removed per iteration.
The embeddings are $\ell_2$-normalised after each projection.
The operating point is set by the number of iterations, 1, 2, 5, 10 or 20, and we disable the stopping criteria of INLP.
The gender LR attacker attains 49.5\% after 20 iterations, against 65.1\% without removal (Table~\ref{tab:s-full-gender}).
The worst-case leakage stays at 95.8\%, as for linear erasure on our embeddings (Appendix~\ref{app:baselines-linerase}).

\subsection{IVE}
\label{app:baselines-ive}

IVE~\cite{Terhorst2019IVE} standardises the embeddings and trains a random forest of 30 trees to predict the attribute.
IVE removes the 10 coordinates with the highest importance in each round, up to the number of removed coordinates of the setting.
The settings remove between 250 and 500 of the 512 coordinates.
The transformed template consists of the remaining standardised coordinates.
The standardisation, the forest of 30 trees and the removal rule agree with the public code, in which the number of removals per step is a parameter.
To make training on 3.1 million images feasible, we train the forest on the mean embedding of each identity, which deviates from the published IVE.

\subsection{Multi-IVE}
\label{app:baselines-multiive}

Multi-IVE~\cite{Melzi2023MultiIVEPE} applies IVE to the principal components of the embeddings, and we port Multi-IVE from the paper and the public code.
The embeddings are standardised and transformed by a PCA with all 512 components and without whitening.
A random forest of 30 trees predicts the attribute from the remaining components.
The component with the highest impurity-based importance is then removed, which is variant~A of the paper for one attribute.
The transformed template has 512 values, since the removed components are set to zero and the PCA is inverted after the last removal.

We apply the PCA without a centring step, since the standardised training embeddings have zero mean.
The inverse PCA is then an orthogonal map of the component scores, and the cosine similarities of the transformed template equal the cosine similarities of the remaining component scores.
Multi-IVE and PCA truncation with the same number of kept components therefore share one basis and differ only in the choice of these components.
The forests are trained on the mean embedding of each identity, as for IVE, and the PCA is fitted on the same 500K training images as PCA truncation.

Multi-IVE can also lock the leading components against removal, and locking the 5 leading components reduces the verification error in the original work.
We evaluate both variants and treat the number of locked components as a parameter of the method, so that the frontier of Multi-IVE covers both variants.
We sweep the number of removed components over 100, 250, 384, 450, 470, 490, 500, 504, 506 and 508 without locking.
The sweep with 5 locked components covers the same values up to 506.
We use one seed, whereas the results of the original work are averaged over 10 seeds.

\subsection{Linear Erasure}
\label{app:baselines-linerase}

We run LEACE~\cite{Belrose2023LEACE} with the public package (concept-erasure 0.2.4) without its trace heuristic.
LEACE is fitted in closed form on the standardised embeddings of the whole training split.
LEACE erases at most 3 directions for ethnicity, since the 4 groups are encoded as one-hot vectors.
PCA + LEACE retains the $k$ leading principal components of the standardised embeddings, computed from one million training images, and fits LEACE on these components.
Fair PCA~\cite{Kleindessner2023FairPCA} retains the $k$ leading principal components within the subspace of equal group means, with centred one-hot indicators for ethnicity.
We run fair kernel PCA with the public implementation, an RBF kernel and 4000 stratified basis points.
Fair kernel PCA is run only for gender, since the public implementation of fair kernel PCA handles 2 groups.
Every map is fitted on the training split and applied unchanged to the test sets.
The group means of every LEACE and fair PCA transformed template coincide to $10^{-13}$ on the training data.

These methods are compared with PCA truncation at the same $k$ in Table~\ref{tab:s-linerase}.
All of these methods reduce the balanced accuracy of the LR attacker to about chance or below.
The balanced accuracy of the MLP attacker on these methods stays within 1\% of its value on PCA truncation for gender and within 4\% for ethnicity.
The TMR at FMR~$10^{-3}$ differs from the TMR at this FMR of PCA truncation by at most 4.1\%.
Fair PCA and fair kernel PCA~\cite{Olfat2019FairPCA,Kleindessner2023FairPCA} coincide with PCA truncation at the same $k$ within 0.3\% TMR at FMR~$10^{-3}$ and 1.2\% worst-case leakage.
LEACE on the full embedding retains the leakage of the embedding without removal.

\begin{table*}[!t]
\centering
\caption{Linear erasure and PCA truncation at the same template dimension $k$. IJB-C TMR (\%) at FMR~$10^{-3}$ and leakage (\%) of the worst case and of the MLP, LR and $k$NN-51 attackers, on IJB-C for gender and on RFW for ethnicity. The worst case also includes the SVC attacker and the other $k$NN variants.}
\label{tab:s-linerase}
\footnotesize
\setlength{\tabcolsep}{4pt}
\begin{tabular}{@{}rlcccccccccc@{}}
\toprule
 & & \multicolumn{5}{c}{Gender (chance 50\%)} & \multicolumn{5}{c}{Ethnicity (chance 25\%)} \\
\cmidrule(lr){3-7}\cmidrule(l){8-12}
$k$ & Method & TMR & Worst & MLP & LR & $k$NN-51 & TMR & Worst & MLP & LR & $k$NN-51 \\
\midrule
512 & No removal & 97.6 & 96.0 & 95.8 & 65.1 & 95.5 & 97.6 & 90.0 & 90.0 & 41.2 & 83.7 \\
512 & LEACE & 97.6 & 96.0 & 96.0 & 48.0 & 95.4 & 97.6 & 89.5 & 89.5 & 26.0 & 81.1 \\
\midrule
22 & PCA truncation & 82.3 & 58.4 & 55.8 & 52.3 & 54.9 & 82.2 & 35.1 & 32.8 & 25.8 & 25.4 \\
22 & PCA + LEACE & 80.9 & 57.9 & 55.5 & 48.7 & 53.9 & 78.1 & 33.4 & 31.0 & 25.4 & 25.2 \\
22 & Fair PCA & 82.0 & 58.4 & 56.0 & 48.9 & 54.5 & 82.2 & 35.2 & 32.6 & 25.8 & 25.4 \\
\midrule
42 & PCA truncation & 92.0 & 70.7 & 68.2 & 53.9 & 63.7 & 91.9 & 48.5 & 44.4 & 29.1 & 26.6 \\
42 & PCA + LEACE & 91.7 & 70.1 & 67.5 & 47.4 & 62.9 & 91.2 & 46.1 & 41.1 & 25.5 & 26.2 \\
42 & Fair PCA & 92.0 & 70.3 & 68.0 & 47.7 & 63.4 & 91.9 & 48.5 & 43.4 & 25.6 & 26.3 \\
42 & Fair kernel PCA & 92.2 & 70.3 & 68.1 & 51.4 & 63.1 & -- & -- & -- & -- & -- \\
\midrule
62 & PCA truncation & 94.9 & 80.5 & 79.8 & 53.9 & 72.9 & 94.9 & 57.6 & 57.2 & 29.7 & 28.1 \\
62 & PCA + LEACE & 94.9 & 80.0 & 79.6 & 48.9 & 72.1 & 94.6 & 55.9 & 53.5 & 24.9 & 27.3 \\
62 & Fair PCA & 94.9 & 80.6 & 80.0 & 49.2 & 72.3 & 94.9 & 57.6 & 55.8 & 25.3 & 27.5 \\
\midrule
128 & PCA truncation & 97.0 & 93.4 & 93.4 & 54.5 & 89.5 & 97.0 & 86.5 & 86.5 & 33.5 & 51.8 \\
128 & PCA + LEACE & 97.0 & 93.3 & 93.3 & 48.7 & 89.4 & 97.0 & 85.3 & 85.3 & 25.4 & 48.4 \\
128 & Fair PCA & 97.1 & 93.4 & 93.4 & 48.2 & 89.7 & 97.0 & 85.3 & 85.3 & 25.1 & 49.9 \\
\bottomrule
\end{tabular}

\end{table*}

\subsection{PE-MIU}
\label{app:baselines-pemiu}

PE-MIU~\cite{Terhorst2020PEMIU} splits a template into blocks of $s$ consecutive coordinates and stores the blocks in a random order, drawn anew for every template.
A pair of templates is compared by assigning the blocks of one template to the blocks of the other template with the Hungarian algorithm, with Euclidean cost.
The comparison score is the cosine similarity of the aligned vectors.
Our port follows the public code and reproduces the comparison scores of the public code to $2\times10^{-7}$ on 300 pairs per block size.
The verification accuracy of PE-MIU is the same for the gender and the ethnicity models, since PE-MIU needs neither training nor labels.

PE-MIU was proposed for 128-dimensional FaceNet embeddings with $s=16$, that is 8 blocks.
On our 512-dimensional embeddings, a value of $s$ cannot retain both the block size and the number of blocks of the paper.
The values $s\in\{4,8,16,32,64\}$ cover both readings, since $s=16$ retains the block size and $s=64$ retains the number of blocks.
The crops of an IJB-C template are pooled as for every other method, since PE-MIU protects the stored template.
Every enrolment and verification template is then shuffled with an independent permutation and compared by the aligned cosine similarity.
Shuffling single crops before pooling would average blocks from different positions.
Every transformed template of the attack data is shuffled with an independent permutation, which is the setting named S-ATK in the PE-MIU paper.

The assignment matches the blocks of every pair as closely as possible, which raises the non-mated comparison scores.
The mean non-mated IJB-C comparison score increases from 0.003 without removal to 0.152 at $s=64$, 0.443 at $s=16$ and 0.865 at $s=4$ (Table~\ref{tab:s-score-stats-full}).
The IJB-C TMR at FMR~$10^{-3}$ of PE-MIU decreases as these scores increase, to 64.7\% at $s=4$ (Table~\ref{tab:s-pemiu}).

The shared attackers (Appendix~\ref{app:protocol-attackers}) treat the transformed template coordinate by coordinate.
A permutation drawn anew for every template therefore does not leave the shared attackers any fixed coordinates.
We add 6 attackers on the set of blocks and an attacker that recovers the order of the blocks, since PE-MIU keeps every coordinate of the template.
These attackers are trained on 500,000 transformed training templates and tested on the same sets as the shared attackers.

The adaptive attack of the PE-MIU paper (A-ATK) uses one MLP trained on single blocks.
The class probabilities of this MLP are averaged over the blocks of a template.
Every transformed template is aligned to the normalised mean of the training embeddings with the Hungarian step of PE-MIU in 2 alignment attacks.
The MLP and LR attackers are then trained on the aligned transformed templates.
Every transformed template is compared with the class-mean templates by the PE-MIU comparison in a class-mean attack, and LR is trained on these scores.
We also train 2 set classifiers, which take the blocks of a template as an unordered set.
The DeepSets classifier~\cite{Zaheer2017DeepSets} uses mean and max pooling.
Our set transformer, in the manner of~\cite{Lee2019SetTransformer}, is a 2-layer transformer encoder with mean pooling and no positional encoding.
The worst-case leakage of PE-MIU is the highest balanced accuracy over the shared attackers, the 6 attackers on the set of blocks and the unwinding attacker (Table~\ref{tab:s-pemiu}).

The original work attributes the protection to the random block order and states that this order can be reconstructed only with an unmodified embedding of the same identity.
The order can however be inferred from the template and the distribution of the unprotected embeddings, which an attacker with the feature extractor estimates from face images of other people.
The coordinates of an embedding are correlated within and between blocks, and these correlations do not depend on the face.
The content of a block and the content of the other blocks of the same template therefore identify the position of the block.

A classifier with 2 hidden layers predicts the position of a block from its $s$ values.
The Hungarian algorithm then assigns the blocks of a template to distinct positions with the highest total log-probability.
A Gaussian model of the unprotected embedding refines this assignment.
The energy of an order is the Mahalanobis norm of the reordered template under this model.
We reduce this energy by reassigning blocks given the positions of the other blocks, and then by pairwise swaps until a swap no longer reduces the energy.
The classifier is trained on the blocks of 200,000 unprotected training embeddings, and the Gaussian model is fitted on 500,000 unprotected training embeddings.
The MLP and LR attackers are then trained on the reordered training templates, and the MLP attacker trained on unprotected embeddings is applied to the reordered test templates.
The Unwound column of Table~\ref{tab:s-pemiu} is the highest balanced accuracy of these 3 attackers.

The refinement recovers more than 99.9\% of the block positions at $s=64$ and 91.0\% of the positions at $s=32$, measured on 20,000 shuffled embeddings of the VGGFace2 test split with known permutations.
The exact order of 89.4\% of the templates is recovered at $s=32$.
The reordered templates of PE-MIU with $s=64$ leak 95.6\% gender (Table~\ref{tab:s-pemiu}), against 96.0\% without removal.
The unwinding attacker is the strongest attacker at $s\ge32$ for both attributes.
The refinement recovers few block positions at $s\le16$, and the leakage of the reordered templates at these block sizes stays below the balanced accuracy of the set transformer.
The TMR of PE-MIU is unchanged by the attack at every FMR, since the comparison of PE-MIU realigns the blocks of every pair.

The set transformer is the strongest of the attackers on the set of blocks at every setting except $s=64$ for gender, where A-ATK is stronger.
The shared attackers alone attain 41.9\% at $s=32$ for ethnicity, against 79.7\% for the unwinding attacker.
The gender leakage on IJB-C increases with the block size, from 52.1\% at $s=4$ to 95.6\% at $s=64$.

\begin{table*}[!t]
\centering
\caption{PE-MIU under the shared attackers, the attackers that act on the set of blocks and the unwinding attacker. TMR (\%) at FMR~$10^{-3}$ and balanced accuracy (\%), the latter on IJB-C for gender and on RFW for ethnicity. The $k$NN column is the highest over the numbers of neighbours and both vote rules, Unwound is the highest over the 3 attackers on the reordered templates, and Worst is the highest over all attackers, including the SVC attacker.}
\label{tab:s-pemiu}
\scriptsize
\setlength{\tabcolsep}{3pt}
\begin{tabular}{@{}rccccccccccccc@{}}
\toprule
 & \multicolumn{2}{c}{TMR@$10^{-3}$} & \multicolumn{3}{c}{Shared attackers} & \multicolumn{7}{c}{Attackers that exploit the permutation} & \\
\cmidrule(lr){2-3}\cmidrule(lr){4-6}\cmidrule(lr){7-13}
$s$ & IJB-C & RFW & MLP & LR & $k$NN & A-ATK & Align.\ + MLP & Align.\ + LR & Class mean + LR & DeepSets & Set transf. & Unwound & Worst \\
\midrule
\multicolumn{14}{@{}l}{\textit{Gender} (chance 50\%)}\\
4 & 64.7 & 13.8 & 50.6 & 51.4 & 50.3 & 50.5 & 50.2 & 51.4 & 51.0 & 51.6 & 52.1 & 50.9 & 52.1 \\
8 & 87.9 & 67.9 & 52.6 & 51.4 & 50.6 & 51.7 & 52.9 & 52.9 & 51.6 & 53.2 & 54.9 & 52.5 & 54.9 \\
16 & 94.6 & 92.5 & 59.1 & 53.2 & 51.9 & 54.9 & 59.4 & 55.5 & 51.2 & 58.5 & 62.8 & 59.0 & 62.8 \\
32 & 96.8 & 96.5 & 69.6 & 53.0 & 57.2 & 67.0 & 69.5 & 57.1 & 52.6 & 69.3 & 73.9 & 91.1 & 91.1 \\
64 & 97.6 & 96.8 & 85.6 & 52.7 & 84.4 & 87.9 & 84.0 & 58.9 & 54.2 & 82.1 & 86.1 & 95.6 & 95.6 \\
\midrule
\multicolumn{14}{@{}l}{\textit{Ethnicity} (chance 25\%)}\\
4 & 64.7 & 13.8 & 26.1 & 27.4 & 25.4 & 26.3 & 26.1 & 27.1 & 25.3 & 26.9 & 28.6 & 26.7 & 28.6 \\
8 & 87.9 & 67.9 & 27.3 & 26.1 & 25.8 & 26.3 & 27.9 & 28.7 & 25.2 & 28.4 & 30.6 & 27.1 & 30.6 \\
16 & 94.6 & 92.5 & 31.6 & 27.5 & 25.8 & 28.2 & 32.1 & 31.5 & 26.5 & 33.7 & 38.7 & 32.4 & 38.7 \\
32 & 96.8 & 96.5 & 41.9 & 28.2 & 30.1 & 35.0 & 40.0 & 32.7 & 29.5 & 43.4 & 52.5 & 79.7 & 79.7 \\
64 & 97.6 & 96.8 & 63.1 & 30.8 & 60.5 & 46.1 & 55.7 & 35.8 & 31.5 & 61.8 & 69.0 & 88.1 & 88.1 \\
\bottomrule
\end{tabular}

\end{table*}

\subsection{PASS}
\label{app:baselines-pass}

PASS~\cite{Dhar2021PASS} trains a linear layer with PReLU and 256 outputs, together with an identity classifier, against an ensemble of attribute discriminators.
The discriminators are re-initialised periodically and trained one at a time.
We follow Algorithm~1 and the ArcFace hyperparameters of the supplementary material of PASS, with 3 discriminators for gender and 2 for ethnicity.
We follow the public code for the choices left open in the paper.
Our embeddings and training split replace the ResNet-101 features and the training data of the paper.
The ethnicity models are PASS-s with our 4 groups.
We set the number of episodes, which neither the paper nor the code fixes, to 300.
We sweep the debiasing weight $\lambda_{\mathrm{deb}}\in\{0,1,3,10,30,100\}$ around the value of the paper, $\lambda_{\mathrm{deb}}=10$.

PASS reduces the balanced accuracy of the gender LR attacker to about chance from $\lambda_{\mathrm{deb}}=10$.
The worst-case leakage of PASS stays above 90\% for gender and above 70\% for ethnicity at every usable setting.

\subsection{Similarity-Sensitive Noise and FaRM}
\label{app:baselines-other}

Table~\ref{tab:s-other} lists the similarity-sensitive noise transformations together with PASS and FaRM.
The table lists the setting with the lowest leakage and the usable setting with the lowest leakage for each method and attribute.
The table also lists the setting with the highest TMR at FMR~$10^{-3}$ at or below the near-chance threshold and the 2 upper leakage thresholds of Table~\ref{tab:matched} of the main paper.

\paragraph{Similarity-sensitive noise}
The similarity-sensitive noise transformations~\cite{Terhorst2019Noise} mix the unit-length template with random noise.
CSN outputs a combination of the template and a noisy copy, with a cosine similarity of the result to the template equal to a target $O$.
ESN adds a random vector of fixed norm $\eta$ to the template.
Our port of the public code reproduces the coefficients of this code to $2\times10^{-16}$, and both transformations in the code match their description in~\cite{Terhorst2020PEMIU}.
We sweep $O\in\{0.3,0.4,\dots,0.9\}$ and $\eta\in\{0.25,0.5,0.75,1,1.5,2,3\}$, which cover a similar range of cosine similarities to the input template.
Both transformations act on the stored templates after pooling, as for PE-MIU.
The public code scales the CSN transformed template to a random length between 1 and 100, which leaves the cosine comparison scores unchanged.
We output unit-length templates, since our shared attackers do not normalise their inputs, whereas the attackers of the PE-MIU paper normalise every template.

\paragraph{FaRM}
FaRM~\cite{BasuRoyChowdhury2022FaRM} maps the representation through a network and maximises a rate--distortion function of all samples.
FaRM also maximises the rate--distortion function of the samples within each attribute class, which spreads out the samples of each class.
Our port reproduces the losses of the public code to floating-point rounding.
The constrained setting of FaRM, with identity as the target task, retains the leakage of the embeddings without removal up to $\lambda_{\mathrm{rd}}=10^{-4}$ and less than 1\% TMR at FMR~$10^{-3}$ from $\lambda_{\mathrm{rd}}=10^{-3}$, and we therefore use the unconstrained setting.

The unconstrained setting, which does not use identity labels, follows the paper, with a randomly initialised 4-layer network trained with SGD.
We train the unconstrained setting for $E\in\{0.003,0.01,0.03,0.1,0.3,1,2,3,4,5,7,10,15,20,30\}$ epochs, around the 20 epochs of the public code.
Training with $E<1$ stops after the share $E$ of the minibatches of one epoch, for example after 30\% at $E=0.3$.
One epoch of training reduces the gender leakage to 53.2\%, and every setting below one epoch leaves the gender leakage at or above 61.0\% (Table~\ref{tab:s-full-gender}).

The noise transformations reduce the leakage only together with the TMR at FMR~$10^{-3}$, and ESN retains 58.2\% TMR at this FMR with its lowest usable gender leakage of 85.4\%.

\begin{table*}[!t]
\centering
\caption{Similarity-sensitive noise, PASS and FaRM at the selected settings. TMR (\%) at FMR~$10^{-3}$ and leakage (\%) of the worst case and of the MLP, LR and $k$NN-51 attackers (raw and prior-corrected votes).}
\label{tab:s-other}
\scriptsize
\setlength{\tabcolsep}{3pt}
\begin{tabular}{@{}lllrrrlrrr@{}}
\toprule
 & & & \multicolumn{2}{c}{TMR@$10^{-3}$} & \multicolumn{5}{c}{Leakage (\%)} \\
\cmidrule(lr){4-5}\cmidrule(l){6-10}
Method & Setting & Selected as & IJB-C & RFW & Worst & Attacker & MLP & LR & $k$NN-51 \\
\midrule
\multicolumn{10}{@{}l}{\textit{Gender} (chance 50\%)}\\
Similarity-sensitive noise & ESN, $\eta{=}2$ & lowest, TMR $\ge$50\% & 58.2 & 9.7 & 85.4 & MLP & 85.4 & 54.5 & 65.8/72.4 \\
 & CSN, $O{=}0.3$ & lowest & 5.8 & 0.6 & 70.2 & MLP & 70.2 & 53.1 & 52.8/55.2 \\
PASS & $\lambda_{\mathrm{deb}}{=}30$ & lowest & 77.1 & 48.1 & 91.0 & MLP & 91.0 & 52.1 & 67.3/69.1 \\
FaRM & 0.3 epochs & $\le$70\% & 84.6 & 29.6 & 61.0 & MLP & 61.0 & 53.7 & 54.2/54.8 \\
 & 2 epochs & $\le$60\% & 80.0 & 19.5 & 53.3 & MLP & 53.3 & 51.9 & 51.3/51.3 \\
 & 5 epochs & $\le$52.5\% & 76.7 & 15.8 & 52.2 & MLP & 52.2 & 51.4 & 51.1/51.1 \\
 & 30 epochs & lowest & 66.8 & 10.8 & 51.0 & MLP & 51.0 & 50.3 & 50.5/50.7 \\
\midrule
\multicolumn{10}{@{}l}{\textit{Ethnicity} (chance 25\%)}\\
Similarity-sensitive noise & ESN, $\eta{=}2$ & lowest, TMR $\ge$50\% & 58.2 & 9.7 & 68.0 & MLP & 68.0 & 31.7 & 25.1/43.9 \\
 & CSN, $O{=}0.3$ & lowest, $\le$50\% & 5.8 & 0.6 & 48.2 & MLP & 48.2 & 29.4 & 25.0/28.8 \\
PASS & $\lambda_{\mathrm{deb}}{=}30$ & lowest & 77.3 & 60.1 & 71.9 & MLP & 71.9 & 28.3 & 29.6/55.1 \\
FaRM & 0.3 epochs & $\le$40\%, $\le$50\% & 79.9 & 18.9 & 34.5 & MLP & 34.5 & 32.8 & 25.3/30.2 \\
 & 20 epochs & lowest & 72.9 & 17.5 & 29.7 & MLP & 29.7 & 28.9 & 25.1/27.5 \\
\bottomrule
\end{tabular}

\end{table*}

\subsection{Label-Free Controls}
\label{app:baselines-controls}

PCA truncation standardises the embeddings with the per-feature mean and standard deviation of the training split.
PCA truncation computes the principal components from 500K training images and outputs the scores of the $k$ leading components.
Random dimensions outputs $k$ standardised coordinates drawn at random, with one draw (seed 42).
Neither control uses attribute or identity labels in any step.
Both controls are fitted on the training split and applied unchanged to the test sets, with $k\in\{4,6,8,10,12,16,22,42,62,128\}$.
Random dimensions is the label-free counterpart of IVE, with the standardisation and the type of transformed template of IVE and with coordinates drawn at random.
PCA truncation is the label-free counterpart of Multi-IVE in the same sense, since both methods keep components of one PCA basis (Appendix~\ref{app:baselines-multiive}).
The original work of Multi-IVE~\cite{Melzi2023MultiIVEPE} used a random order of elimination, the same control, as a baseline.

\section{Additional Results}
\label{app:results}

We give further results for Section~\ref{sec:results} of the main paper in this appendix.
Leakage is the worst-case balanced accuracy on IJB-C for gender and on RFW for ethnicity (Appendix~\ref{app:protocol-attackers}).
The TMR is measured on IJB-C, and on RFW where stated.
Near chance is defined in Section~\ref{sec:setup-leakage}, and a usable setting in Appendix~\ref{app:protocol-verification}.

\subsection{Score Distributions}
\label{app:results-scores}

\begin{figure}[!t]
\centering
\includegraphics[width=\textwidth]{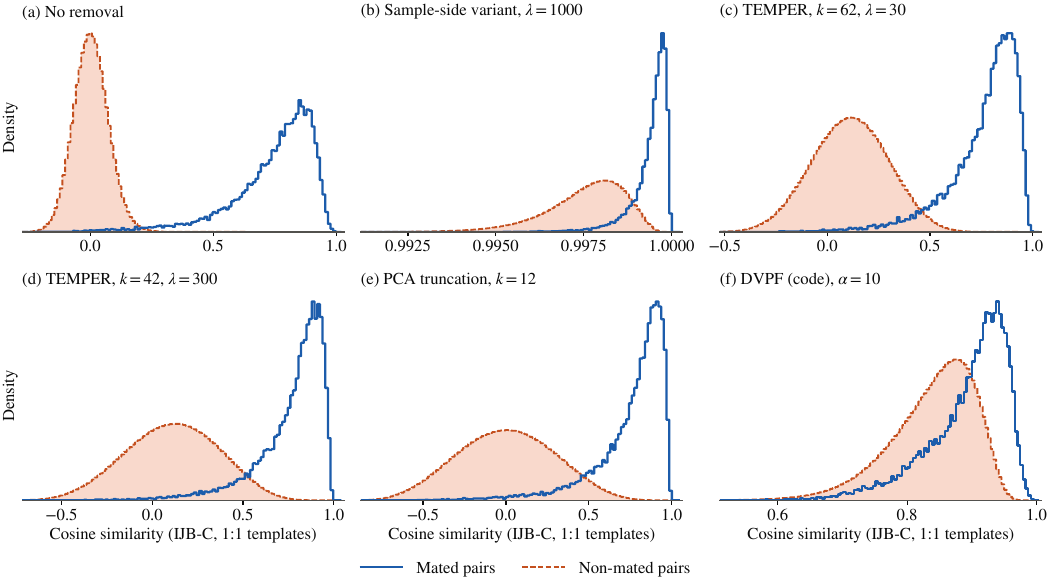}
\caption{Distributions of IJB-C mated and non-mated comparison scores of the gender models, without removal and for the sample-side variant, TEMPER at 2 settings, PCA truncation and DVPF (code).}
\label{fig:s-scores}
\end{figure}

The distributions of IJB-C mated and non-mated comparison scores of the transformed templates are plotted in Fig.~\ref{fig:s-scores}, and the moments of these distributions are listed in Table~\ref{tab:s-scores}.
The mated and non-mated comparison scores of the sample-side variant at $\lambda=1000$ are both close to 1, with a non-mated mean of 0.998 against 0.003 without removal.
The transformed templates of this variant therefore concentrate in a small spherical cap, as in Remark~\ref{rem:s-offset}.
The two distributions of this variant stay apart, with an equal error rate of 10.4\%, whereas the two distributions coincide under the representation collapse of Section~\ref{sec:res-collapse} of the main paper.
The transformed templates of DVPF (code) at $\alpha=10$ also concentrate, with a non-mated mean of 0.839.

The transformed templates of TEMPER are not under representation collapse at $k=62$ and $\lambda=30$, with a non-mated mean of 0.111.
The distributions of non-mated comparison scores of TEMPER and of PCA truncation are broader than without removal, and the templates of these methods have fewer dimensions than the embeddings.
The cosine between a pair of independent uniform directions in $k$ dimensions has a standard deviation of $1/\sqrt{k}$.
The non-mated comparison scores of PCA truncation to $k=12$ have a standard deviation of 0.290, close to $1/\sqrt{12}\approx0.289$.
The standard deviation of the non-mated comparison scores of TEMPER exceeds $1/\sqrt{k}$ at both settings.

\begin{table}[!t]
\centering
\caption{IJB-C score statistics of the transformed templates of the gender models. Mean and standard deviation of the mated and non-mated comparison scores, equal error rate and decidability index $d'$, at the settings of Fig.~\ref{fig:s-scores}.}
\label{tab:s-scores}
\footnotesize
\setlength{\tabcolsep}{4pt}
\begin{tabular}{lrrrrrr}
\toprule
 & \multicolumn{2}{c}{Mated} & \multicolumn{2}{c}{Non-mated} & & \\
\cmidrule(lr){2-3}\cmidrule(lr){4-5}
Method, setting & Mean & s.d. & Mean & s.d. & EER (\%) & $d'$ \\
\midrule
No removal & 0.744 & 0.171 & 0.003 & 0.071 & 1.4 & 5.65 \\
Sample-side variant, $\lambda{=}1000$ & 0.999 & 0.001 & 0.998 & 0.001 & 10.4 & 2.08 \\
TEMPER, $k{=}62$, $\lambda{=}30$ & 0.763 & 0.171 & 0.111 & 0.185 & 4.8 & 3.66 \\
TEMPER, $k{=}42$, $\lambda{=}300$ & 0.768 & 0.189 & 0.110 & 0.254 & 7.7 & 2.94 \\
PCA truncation, $k{=}12$ & 0.756 & 0.214 & 0.002 & 0.290 & 7.7 & 2.96 \\
DVPF (code), $\alpha{=}10$ & 0.894 & 0.062 & 0.839 & 0.066 & 31.1 & 0.86 \\
\bottomrule
\end{tabular}

\end{table}

Table~\ref{tab:s-score-stats-full} lists the same statistics for every setting of the compared methods.
In this table, $E$ is the number of epochs of FaRM, $n$ is the number of iterations of INLP and of layers of SensitiveNets, and $\lambda$ is the weight $\lambda_{\mathrm{dis}}$ of PFRNet and $\lambda_{\mathrm{deb}}$ of PASS.
The mated and non-mated comparison scores of TEMPER stay apart at every setting of both attributes.
The decidability index $d'$ of TEMPER is at least 2.77 over the 35 gender settings and at least 1.88 over the 34 ethnicity settings.
The equal error rate of TEMPER is at most 8.5\% for gender and 13.3\% for ethnicity.

\begingroup\footnotesize\setlength{\tabcolsep}{4pt}
\begin{longtable}{@{}llccccc@{}}
\caption{IJB-C score statistics of every setting of the compared methods. Worst-case leakage (\%), mean and standard deviation of the mated and of the non-mated comparison scores, equal error rate (\%) and decidability index $d'$, for the gender and the ethnicity models.}\label{tab:s-score-stats-full}\\
\toprule
Method & Setting & Leakage & Mated & Non-mated & EER & $d'$ \\
\midrule
\endfirsthead
\toprule
Method & Setting & Leakage & Mated & Non-mated & EER & $d'$ \\
\midrule
\endhead
\bottomrule
\endfoot
\multicolumn{7}{@{}l}{\textit{Gender}}\\
No removal & -- & 96.0 & 0.744 (0.171) & 0.003 (0.071) & 1.4 & 5.65 \\
PCA truncation & $k{=}4$ & 51.9 & 0.731 (0.324) & 0.001 (0.500) & 17.8 & 1.73 \\
PCA truncation & $k{=}6$ & 52.8 & 0.746 (0.266) & 0.002 (0.409) & 13.3 & 2.16 \\
PCA truncation & $k{=}8$ & 54.3 & 0.753 (0.238) & 0.002 (0.354) & 10.6 & 2.49 \\
PCA truncation & $k{=}10$ & 54.7 & 0.754 (0.222) & 0.002 (0.317) & 8.8 & 2.75 \\
PCA truncation & $k{=}12$ & 54.6 & 0.756 (0.214) & 0.002 (0.290) & 7.7 & 2.96 \\
PCA truncation & $k{=}16$ & 55.8 & 0.755 (0.205) & 0.003 (0.252) & 6.2 & 3.28 \\
\end{longtable}
\endgroup

\subsection{Frontiers at All Operating Points}
\label{app:results-frontiers}

\begin{figure*}[!t]
\centering
\includegraphics[width=\textwidth]{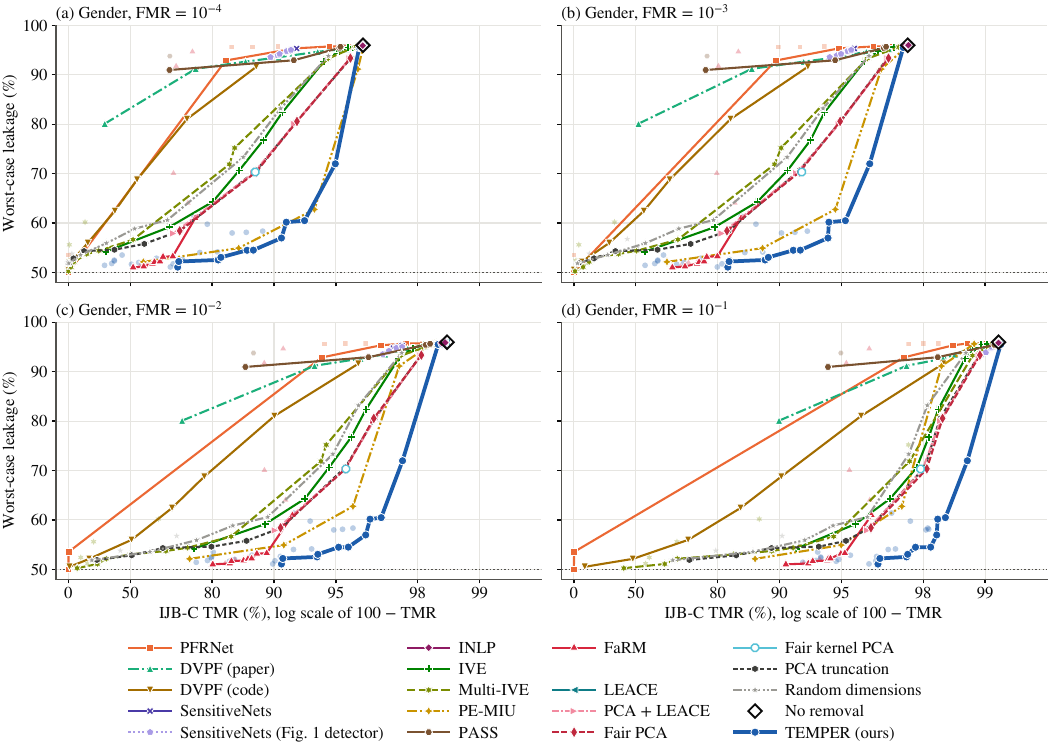}
\caption{Frontiers for gender at the 4 operating points. Worst-case leakage, clipped at chance (50\%, dotted), against IJB-C TMR at the 4 FMRs on an axis logarithmic in $100-\mathrm{TMR}$. Faint markers are dominated settings.}
\label{fig:s-frontier-gender}
\end{figure*}

\begin{figure*}[!t]
\centering
\includegraphics[width=\textwidth]{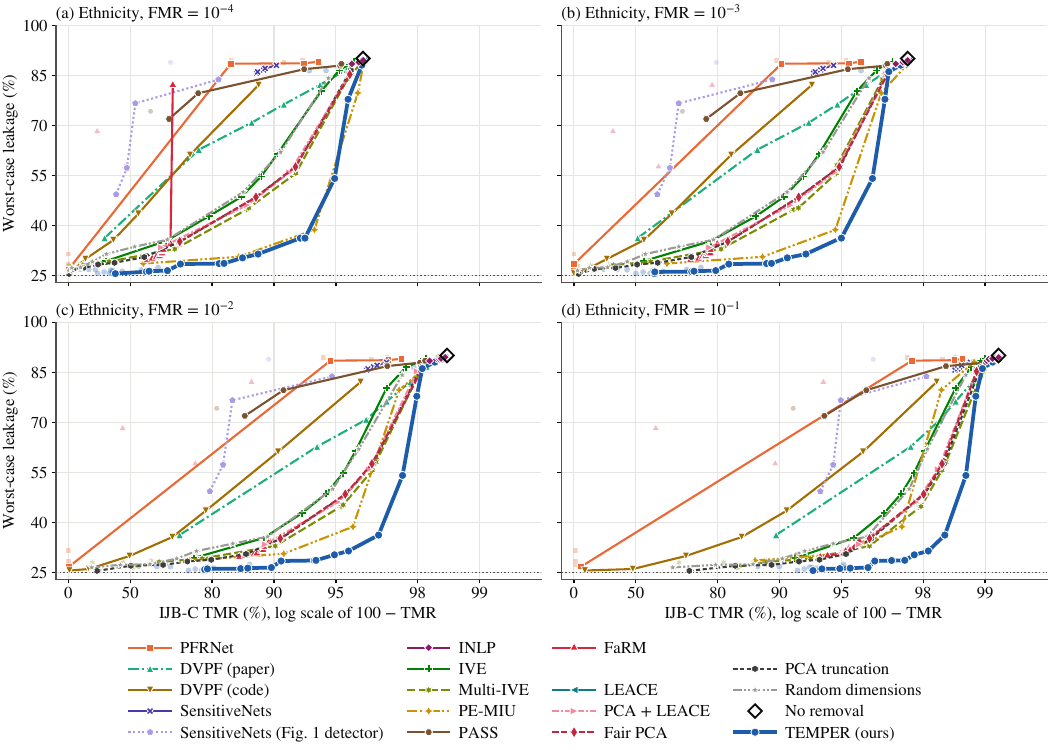}
\caption{Frontiers for ethnicity at the 4 operating points, as in Fig.~\ref{fig:s-frontier-gender} (chance 25\%).}
\label{fig:s-frontier-ethnicity}
\end{figure*}

The IJB-C frontiers of Fig.~\ref{fig:frontier} are repeated in Figs.~\ref{fig:s-frontier-gender} and~\ref{fig:s-frontier-ethnicity} at FMR~$10^{-4}$, $10^{-3}$, $10^{-2}$ and $10^{-1}$, with fair PCA and fair kernel PCA added.
Each panel shows the IJB-C TMR at one FMR on an axis logarithmic in $100-\mathrm{TMR}$, which spreads out the markers of the methods above 90\%.
Each line is the frontier of one method over the settings of that method, selected on the test sets for every method alike.
Every setting on these frontiers is listed in Tables~\ref{tab:s-full-gender} and~\ref{tab:s-full-ethnicity} of Appendix~\ref{app:results-full}, apart from the settings of fair PCA and fair kernel PCA in Table~\ref{tab:s-linerase}.
The markers of INLP lie on or next to the marker of no removal, since INLP changes the TMR at the 4 FMRs by at most 0.5\% and the leakage by at most 1.6\%.

TEMPER retains the highest TMR at every gender threshold of Table~\ref{tab:matched} at FMR~$10^{-3}$, $10^{-2}$ and $10^{-1}$, with smaller TMR gaps between the methods at larger FMR.
At FMR~$10^{-4}$, PE-MIU retains 0.7\% more TMR than TEMPER at the 70\% threshold.
The only operating point of the DVPF paper is FMR~$10^{-1}$.
DVPF (code) at $\alpha=10$ retains 48.3\% TMR at this FMR for gender, against 96.6\% for TEMPER at $k=42$, $\lambda=30$ at a similar leakage.
These 2 settings retain 8.2\% and 81.0\% TMR at FMR~$10^{-3}$.
The same holds for ethnicity, where PE-MIU retains 0.7\% more TMR than TEMPER at FMR~$10^{-4}$ at the 40\% and 50\% thresholds.
For example, PCA truncation at $k=12$ retains 93.6\% TMR at FMR~$10^{-1}$, against 97.2\% for TEMPER at $k=42$, $\lambda=3$ at a lower leakage.

\subsection{Frontiers Under the MLP Attacker Alone and on RFW}
\label{app:results-mlp-rfw}

We recompute the frontiers at the 4 FMRs with the MLP attacker as the only attacker (Fig.~\ref{fig:s-frontier-mlponly}).
The difference from Figs.~\ref{fig:s-frontier-gender} and~\ref{fig:s-frontier-ethnicity} is the contribution of the other attackers to the worst case.
The differences are largest for PE-MIU. The unwinding attacker attains 79.7\% balanced accuracy against 41.9\% for the MLP attacker, for ethnicity at $s=32$ (Appendix~\ref{app:baselines-pemiu}).
The LR attacker is often the strongest attacker on TEMPER at large $\lambda$ (Appendix~\ref{app:protocol-attackers}).
The MLP attacker alone therefore attains less than the worst case on these settings.
The MLP attacker attains about chance on many transformed templates of TEMPER for ethnicity, for example 25.3\% at $k=96$, $\lambda=30$.
The MLP attacker does not fit the training data on these transformed templates, with a training accuracy below the majority rate of the training split.
A balanced accuracy at chance then indicates a failed attacker and is not evidence of removal.
We take the worst case over several attackers partly for this reason.

We recompute the frontiers with RFW TMR at FMR~$10^{-3}$, $10^{-2}$ and $10^{-1}$ in place of IJB-C TMR at these FMRs (Fig.~\ref{fig:s-frontier-rfw}).
Every usable setting that reduces the leakage by more than 4\% retains less TMR at FMR~$10^{-3}$ on RFW than on IJB-C.
TEMPER at $k=42$, $\lambda=30$ retains 26.8\% RFW TMR at FMR~$10^{-3}$ for gender, against 96.8\% without removal.
The RFW TMR of this setting at FMR~$10^{-1}$ is 83.1\%, against 99.7\% without removal.

PE-MIU with $s=8$ retains more RFW TMR at FMR~$10^{-3}$ than any setting of TEMPER below 56\% gender leakage.
PE-MIU with $s=8$ retains 67.9\%, against 61.2\% for TEMPER at $k=42$, $\lambda=3$ at a lower leakage (Table~\ref{tab:s-cross}).
The highest RFW TMR at FMR~$10^{-3}$ of the other compared methods and the controls below 56\% gender leakage is 19.5\%, for FaRM after 2 epochs.
The RFW panels at FMR~$10^{-2}$ and $10^{-1}$ are more reliable than the panel at $10^{-3}$, since the RFW threshold at FMR~$10^{-3}$ is set by 12 non-mated pairs.

\begin{figure*}[!t]
\centering
\includegraphics[width=\textwidth]{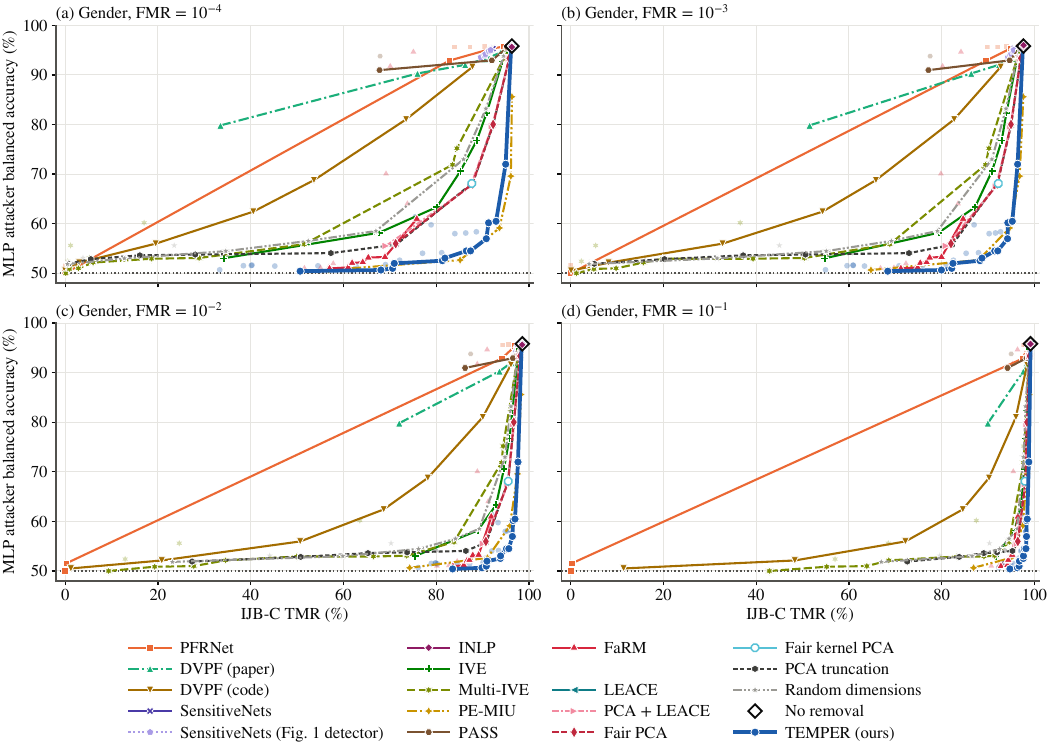}
\caption{Frontiers under the MLP attacker alone, gender (chance 50\%). Balanced accuracy of the MLP attacker against IJB-C TMR at the 4 FMRs on a linear axis, otherwise as in Fig.~\ref{fig:s-frontier-gender}.}
\label{fig:s-frontier-mlponly}
\end{figure*}

\begin{figure*}[!t]
\addtocounter{figure}{-1}
\centering
\includegraphics[width=\textwidth]{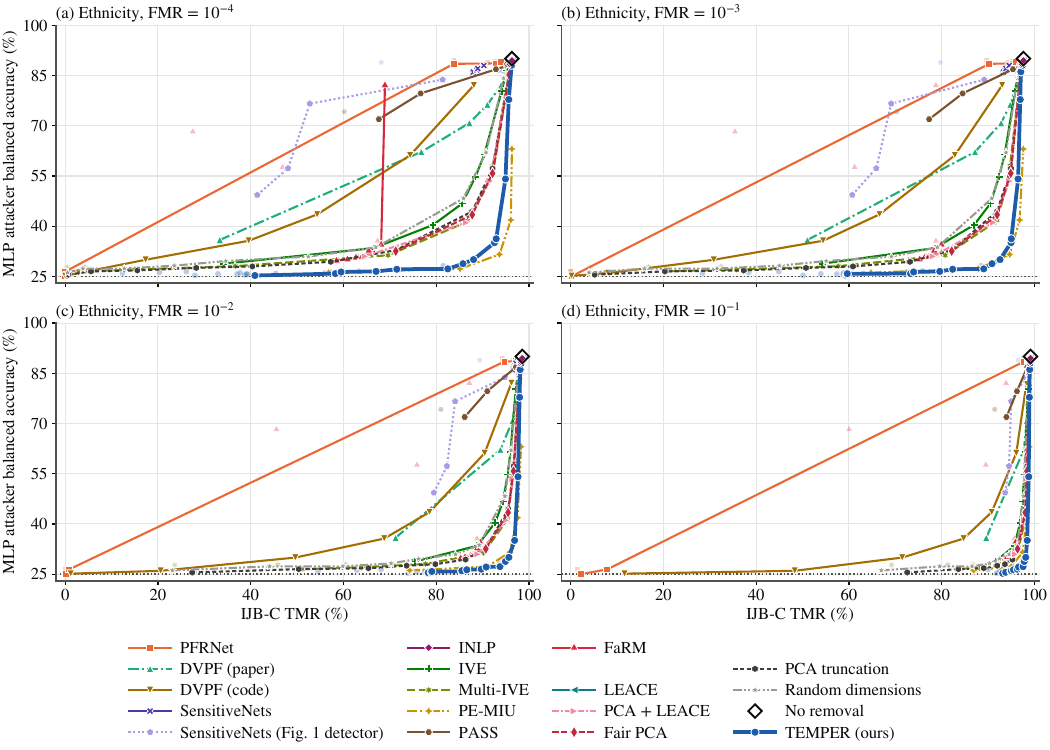}
\caption{(Continued) Frontiers under the MLP attacker alone, ethnicity (chance 25\%).}
\end{figure*}

\begin{figure*}[!t]
\centering
\includegraphics[width=\textwidth]{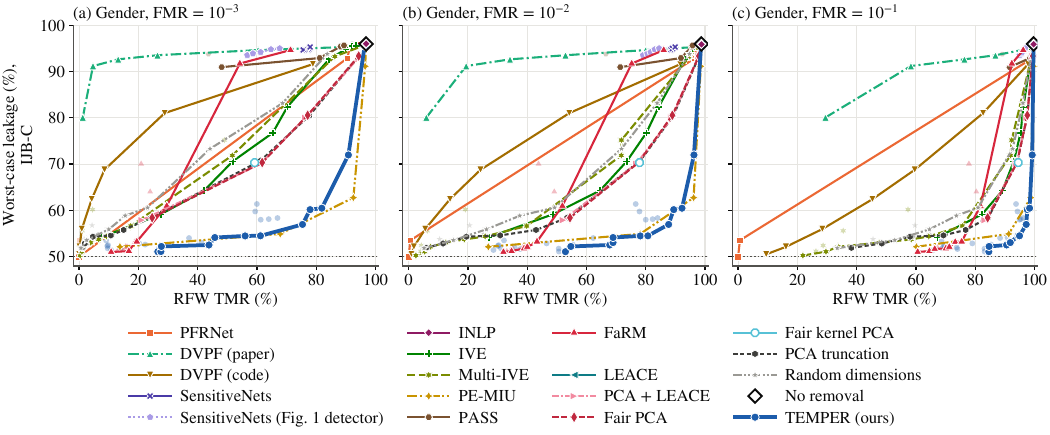}\\[4pt]
\includegraphics[width=\textwidth]{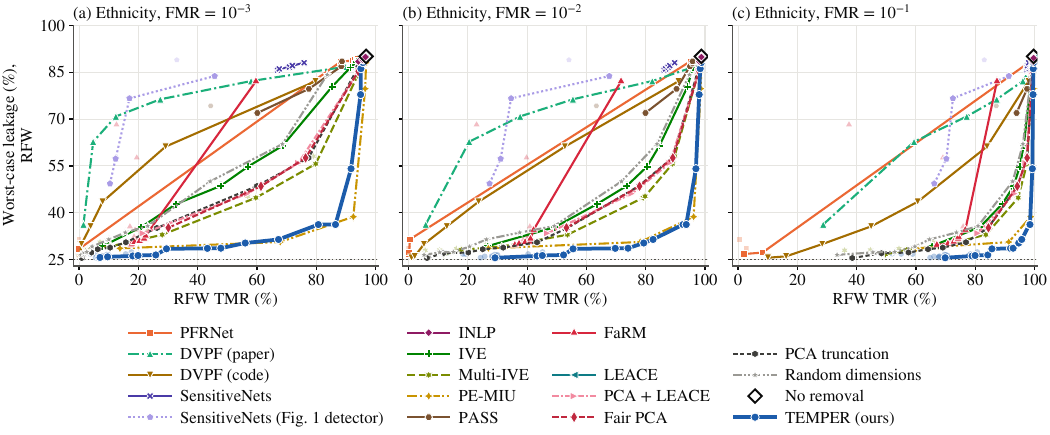}
\caption{Frontiers with RFW TMR at FMR $10^{-3}$, $10^{-2}$ and $10^{-1}$, gender (top) and ethnicity (bottom). Worst-case leakage against RFW TMR at these FMRs, with the frontiers recomputed on this TMR.}
\label{fig:s-frontier-rfw}
\end{figure*}

\subsection{Accuracy on LFW and RFW}
\label{app:results-acc}

The LFW and RFW accuracy in Table~\ref{tab:s-acc} is the accuracy of the settings selected on IJB-C in Table~\ref{tab:matched}, without a further selection on LFW or RFW.
TEMPER attains the highest LFW accuracy in every column, with a lead of 3.8\% over FaRM near chance for gender.
The LFW lead of TEMPER over PE-MIU with $s=16$ is below 0.5\% at 70\% gender and at 40\% and 50\% ethnicity leakage.
PE-MIU with $s=16$ exceeds the RFW accuracy of TEMPER by about 1\% in these 3 columns.
TEMPER also attains the highest RFW accuracy in every other column.
PFRNet attains about 50\% accuracy on LFW and RFW in every column, which is chance for the balanced pairs of both datasets.
PFRNet also retains at most 0.1\% IJB-C TMR at FMR~$10^{-3}$ in every column of Table~\ref{tab:matched}.

The LFW ranking of the methods below TEMPER differs from the IJB-C ranking of these methods.
For example, PE-MIU with $s=4$ retains the third-highest IJB-C TMR at FMR~$10^{-3}$ at 30\% ethnicity leakage.
The LFW accuracy of PE-MIU with $s=4$ is the lowest of this column apart from DVPF and PFRNet.

\begin{table*}[!t]
\centering
\caption{LFW and RFW accuracy at a given leakage threshold. Accuracy (\%) on LFW (RFW) of the setting selected on IJB-C in Table~\ref{tab:matched} for the column (-- if none). Bold marks the highest accuracy per column among the compared methods, and $^{\ast}$ marks PE-MIU at its lowest leakage, 28.6\%.}
\label{tab:s-acc}
\footnotesize
\setlength{\tabcolsep}{3pt}
\begin{tabular}{@{}lcccc@{\hspace{8pt}}cccc@{}}
\toprule
 & \multicolumn{4}{c}{Gender: worst-case leakage $\le$} & \multicolumn{4}{c}{Ethnicity: worst-case leakage $\le$} \\
\cmidrule(lr){2-5}\cmidrule(l){6-9}
Method & 52.5\% & 55\% & 60\% & 70\% & 27.5\% & 30\% & 40\% & 50\% \\
\midrule
No removal & \multicolumn{4}{c}{99.8 (99.0)} & \multicolumn{4}{c}{99.8 (99.0)} \\
\midrule
\multicolumn{9}{@{}l}{\textit{Label-free controls (reference)}} \\
PCA truncation & 81.1 (69.1) & 93.6 (80.6) & 97.3 (87.1) & 97.3 (87.1) & 89.7 (76.5) & 93.4 (80.4) & 97.2 (87.3) & 99.1 (92.7) \\
Random dimensions & 78.7 (67.5) & 87.6 (75.1) & 93.4 (82.8) & 96.3 (85.7) & 87.6 (75.0) & 89.4 (77.4) & 96.3 (85.7) & 96.3 (85.7) \\
\midrule
LEACE & -- & -- & -- & -- & -- & -- & -- & -- \\
INLP & -- & -- & -- & -- & -- & -- & -- & -- \\
PFRNet & 50.0 (50.0) & 50.0 (50.0) & 50.0 (50.0) & 50.0 (50.0) & 50.4 (50.4) & 50.2 (50.6) & 50.2 (50.6) & 50.2 (50.6) \\
SensitiveNets & -- & -- & -- & -- & -- & -- & -- & -- \\
PASS & -- & -- & -- & -- & -- & -- & -- & -- \\
SensitiveNets (Fig.~1 detector) & -- & -- & -- & -- & -- & -- & -- & 93.7 (80.5) \\
DVPF (paper) & -- & -- & -- & -- & -- & -- & 77.2 (63.6) & 77.2 (63.6) \\
DVPF (code) & 60.5 (56.1) & 60.5 (56.1) & 74.8 (64.1) & 90.8 (77.6) & 61.7 (55.9) & 75.3 (63.9) & 85.9 (72.0) & 91.2 (77.8) \\
PCA + LEACE & -- & -- & 97.0 (86.8) & 97.0 (86.8) & -- & -- & 96.3 (85.6) & 99.0 (92.2) \\
IVE & -- & 91.2 (80.1) & 96.3 (86.4) & 97.4 (89.6) & -- & 91.4 (79.5) & 95.6 (85.9) & 98.5 (91.9) \\
Multi-IVE & 75.4 (66.4) & 90.6 (78.8) & 93.9 (83.2) & 93.9 (83.2) & 86.6 (73.4) & 92.8 (80.3) & 96.8 (86.8) & 98.4 (92.8) \\
PE-MIU & 88.8 (77.2) & 97.5 (91.5) & 97.5 (91.5) & 99.4 (\textbf{97.6}) & 88.8 (77.2)$^{\ast}$ & 88.8 (77.2) & 99.4 (\textbf{97.6}) & 99.4 (\textbf{97.6}) \\
FaRM & 93.4 (80.6) & 94.8 (83.0) & 94.8 (83.0) & 96.6 (85.8) & -- & 92.1 (78.8) & 95.4 (83.7) & 95.4 (83.7) \\
\midrule
TEMPER (ours) & \textbf{97.2} (\textbf{87.4}) & \textbf{99.1} (\textbf{93.3}) & \textbf{99.3} (\textbf{95.1}) & \textbf{99.6} (96.4) & \textbf{96.2} (\textbf{87.2}) & \textbf{98.1} (\textbf{91.9}) & \textbf{99.7} (96.7) & \textbf{99.7} (96.7) \\
\bottomrule
\end{tabular}

\end{table*}

\subsection{Leakage on Each Test Set}
\label{app:results-cross}

We list the leakage on each test set in Table~\ref{tab:s-cross}, for the settings selected in Table~\ref{tab:matched} and for other settings discussed in the main paper.
SensitiveNets, PASS and LEACE have no setting in Table~\ref{tab:matched} and are not listed.
Tables~\ref{tab:s-full-gender} and~\ref{tab:s-full-ethnicity} list the worst case of the MLP and LR attackers on both test sets for every setting of every method.
The attackers are trained on VGGFace2, and the transfer of the attackers between datasets is listed in the RFW columns for gender and in the IJB-C columns for ethnicity.
The worst case of the MLP and LR attackers for gender on RFW is within 2.6\% of the worst case on IJB-C for every TEMPER setting (Table~\ref{tab:s-full-gender}).
For example, the gender leakage of TEMPER at $k=42$, $\lambda=3$ is 53.1\% on RFW and 54.5\% on IJB-C.
The worst case over all 4 attackers is 52.2\% on IJB-C and 51.4\% on RFW for TEMPER at $k=22$, $\lambda=3$, which is the setting of the abstract.
The SVC attacker attains this worst case on RFW, and the MLP and LR attackers of the table attain at most 50.7\%.
The LR attacker transfers in the same way, with 53.8\% on IJB-C and 53.3\% on RFW at $k=96$, $\lambda=30$.

The worst case of the MLP and LR attackers on IJB-C is within about 2\% of the worst case on RFW for ethnicity below 50\% leakage.
This worst case is lower on IJB-C than on RFW by 6.9\% to 9.5\% for every setting above 85\% ethnicity leakage in Table~\ref{tab:s-full-ethnicity} apart from PE-MIU at $s=64$, and by 8.7\% without removal.

\begin{table*}[!t]
\centering
\caption{Leakage on each test set at the settings selected in Table~\ref{tab:matched} of the main paper and at other settings discussed in the text. TMR (\%) at FMR~$10^{-3}$ on IJB-C and RFW, leakage (\%) on IJB-C for gender and on RFW for ethnicity with raw and prior-corrected $k$NN votes, and the MLP and LR attackers with the worst case of the two on IJB-C and on RFW.}
\label{tab:s-cross}
\scriptsize
\setlength{\tabcolsep}{2pt}
\begin{tabular}{@{}llrrrlrrrrrrrrr@{}}
\toprule
 & & \multicolumn{2}{c}{TMR@$10^{-3}$} & \multicolumn{5}{c}{Leakage test set} & \multicolumn{3}{c}{IJB-C} & \multicolumn{3}{c}{RFW} \\
\cmidrule(lr){3-4}\cmidrule(lr){5-9}\cmidrule(lr){10-12}\cmidrule(lr){13-15}
Method & Setting & IJB-C & RFW & Worst & Attacker & MLP & LR & $k$NN-51 & Worst & MLP & LR & Worst & MLP & LR \\
\midrule
\multicolumn{15}{@{}l}{\textit{Gender} (chance 50\%)}\\
No removal & -- & 97.6 & 96.8 & 96.0 & $k$NN-51\,pc & 95.8 & 65.1 & 95.5/96.0 & 95.8 & 95.8 & 65.1 & 96.0 & 96.0 & 56.0 \\
PFRNet & $\lambda_{\mathrm{dis}}{=}0.1$ & 97.2 & 95.0 & 95.7 & MLP & 95.7 & 57.1 & 95.0/95.5 & 95.7 & 95.7 & 57.1 & 96.0 & 96.0 & 52.1 \\
PFRNet & $\lambda_{\mathrm{dis}}{=}10$ & 0.0 & 0.0 & 50.0 & MLP & 50.0 & 50.0 & 50.0/50.0 & 50.0 & 50.0 & 50.0 & 50.0 & 50.0 & 50.0 \\
DVPF (code) & $\alpha{=}2$ & 65.8 & 8.6 & 68.8 & MLP & 68.8 & 53.8 & 59.4/62.5 & 68.8 & 68.8 & 53.8 & 67.6 & 67.6 & 49.6 \\
DVPF (code) & $\alpha{=}5$ & 32.7 & 0.9 & 56.0 & MLP & 56.0 & 52.4 & 52.0/52.6 & 56.0 & 56.0 & 52.4 & 55.0 & 55.0 & 49.7 \\
DVPF (code) & $\alpha{=}10$ & 8.2 & 0.2 & 52.1 & MLP & 52.1 & 51.0 & 50.6/50.7 & 52.1 & 52.1 & 51.0 & 51.3 & 51.3 & 50.0 \\
INLP & 20 iterations & 97.6 & 96.1 & 95.8 & MLP & 95.8 & 49.5 & 95.3/95.8 & 95.8 & 95.8 & 49.5 & 96.0 & 96.0 & 51.6 \\
IVE & 500 removed & 54.8 & 7.5 & 54.2 & SVC & 53.0 & 50.3 & 52.6/53.3 & 53.0 & 53.0 & 50.3 & 52.9 & 52.9 & 50.4 \\
IVE & 490 removed & 79.3 & 27.6 & 59.2 & SVC & 58.2 & 50.5 & 56.1/57.4 & 58.2 & 58.2 & 50.5 & 56.3 & 56.3 & 49.8 \\
IVE & 480 removed & 87.2 & 42.2 & 64.3 & SVC & 63.3 & 51.1 & 60.6/61.9 & 63.3 & 63.3 & 51.1 & 61.7 & 61.7 & 49.5 \\
Multi-IVE & 500 removed & 15.7 & 1.5 & 52.1 & MLP & 52.1 & 50.7 & 50.7/51.8 & 52.1 & 52.1 & 50.7 & 52.2 & 52.2 & 50.8 \\
Multi-IVE & 470 removed, 5 locked & 50.4 & 9.2 & 54.4 & SVC & 53.1 & 52.3 & 51.9/52.7 & 53.1 & 53.1 & 52.3 & 52.3 & 52.3 & 49.8 \\
Multi-IVE & 450 removed, 5 locked & 68.8 & 17.4 & 56.6 & SVC & 55.8 & 53.1 & 54.0/54.7 & 55.8 & 55.8 & 53.1 & 55.9 & 55.9 & 50.1 \\
PE-MIU & $s{=}4$ & 64.7 & 13.8 & 52.1 & set transf. & 50.6 & 51.4 & 50.1/50.3 & 51.4 & 50.6 & 51.4 & 50.8 & 50.8 & 50.3 \\
PE-MIU & $s{=}8$ & 87.9 & 67.9 & 54.9 & set transf. & 52.6 & 51.4 & 50.4/50.2 & 52.6 & 52.6 & 51.4 & 52.7 & 52.7 & 50.6 \\
PE-MIU & $s{=}16$ & 94.6 & 92.5 & 62.8 & set transf. & 59.1 & 53.2 & 51.0/51.9 & 59.1 & 59.1 & 53.2 & 57.7 & 57.7 & 52.0 \\
FaRM & 0.3 epochs & 84.6 & 29.6 & 61.0 & MLP & 61.0 & 53.7 & 54.2/54.8 & 61.0 & 61.0 & 53.7 & 59.9 & 59.9 & 53.6 \\
FaRM & 2 epochs & 80.0 & 19.5 & 53.3 & MLP & 53.3 & 51.9 & 51.3/51.3 & 53.3 & 53.3 & 51.9 & 53.5 & 53.5 & 52.1 \\
FaRM & 5 epochs & 76.7 & 15.8 & 52.2 & MLP & 52.2 & 51.4 & 51.1/51.1 & 52.2 & 52.2 & 51.4 & 52.0 & 52.0 & 51.4 \\
PCA + LEACE & $k{=}22$ & 80.9 & 21.2 & 57.9 & SVC & 55.5 & 48.7 & 53.9/54.9 & 55.5 & 55.5 & 48.7 & 55.2 & 55.2 & 50.0 \\
PCA truncation & $k{=}4$ & 5.1 & 0.9 & 51.9 & MLP & 51.9 & 51.6 & 50.2/50.6 & 51.9 & 51.9 & 51.6 & 51.2 & 51.2 & 49.4 \\
PCA truncation & $k{=}12$ & 60.8 & 10.7 & 54.6 & SVC & 54.1 & 51.9 & 51.7/52.2 & 54.1 & 54.1 & 51.9 & 52.5 & 52.5 & 49.6 \\
PCA truncation & $k{=}22$ & 82.3 & 24.3 & 58.4 & SVC & 55.8 & 52.3 & 54.9/55.7 & 55.8 & 55.8 & 52.3 & 55.5 & 55.5 & 48.8 \\
Random dimensions & $k{=}4$ & 4.0 & 0.6 & 51.8 & MLP & 51.8 & 51.1 & 50.4/51.2 & 51.8 & 51.8 & 51.1 & 52.2 & 52.2 & 50.3 \\
Random dimensions & $k{=}8$ & 31.6 & 2.6 & 53.5 & SVC & 53.0 & 51.3 & 51.3/52.3 & 53.0 & 53.0 & 51.3 & 51.9 & 51.9 & 50.1 \\
Random dimensions & $k{=}16$ & 69.1 & 15.5 & 58.8 & SVC & 56.5 & 52.7 & 56.5/57.6 & 56.5 & 56.5 & 52.7 & 55.4 & 55.4 & 50.0 \\
Random dimensions & $k{=}22$ & 79.0 & 23.1 & 60.6 & SVC & 58.5 & 54.0 & 58.3/59.4 & 58.5 & 58.5 & 54.0 & 56.5 & 56.5 & 50.0 \\
TEMPER (ours) & $k{=}22$, $\lambda{=}3$ & 82.4 & 27.8 & 52.2 & SVC & 52.0 & 51.5 & 51.2/51.8 & 52.0 & 52.0 & 51.5 & 50.7 & 50.7 & 47.4 \\
TEMPER (ours) & $k{=}42$, $\lambda{=}3$ & 92.1 & 61.2 & 54.5 & MLP & 54.5 & 50.6 & 52.2/53.1 & 54.5 & 54.5 & 50.6 & 53.1 & 53.1 & 48.7 \\
TEMPER (ours) & $k{=}42$, $\lambda{=}30$ & 81.0 & 26.8 & 51.7 & MLP & 51.7 & 50.1 & 50.8/51.6 & 51.7 & 51.7 & 50.1 & 51.4 & 51.4 & 50.9 \\
TEMPER (ours) & $k{=}62$, $\lambda{=}3$ & 94.2 & 75.4 & 57.0 & MLP & 57.0 & 50.8 & 53.4/54.3 & 57.0 & 57.0 & 50.8 & 55.7 & 55.7 & 50.2 \\
TEMPER (ours) & $k{=}96$, $\lambda{=}3$ & 95.2 & 81.9 & 60.5 & MLP & 60.5 & 51.6 & 54.5/55.7 & 60.5 & 60.5 & 51.6 & 59.6 & 59.6 & 49.4 \\
\midrule
\multicolumn{15}{@{}l}{\textit{Ethnicity} (chance 25\%)}\\
No removal & -- & 97.6 & 96.8 & 90.0 & MLP & 90.0 & 41.2 & 83.7/89.7 & 81.4 & 81.4 & 47.5 & 90.0 & 90.0 & 41.2 \\
PFRNet & $\lambda_{\mathrm{dis}}{=}0.1$ & 90.2 & 88.4 & 88.5 & MLP & 88.5 & 28.8 & 62.7/85.5 & 79.7 & 79.7 & 29.0 & 88.5 & 88.5 & 28.8 \\
PFRNet & $\lambda_{\mathrm{dis}}{=}1$ & 0.0 & 0.0 & 26.8 & SVC & 26.4 & 25.0 & 25.0/25.3 & 26.3 & 26.3 & 25.3 & 26.4 & 26.4 & 25.0 \\
PFRNet & $\lambda_{\mathrm{dis}}{=}3$ & 0.1 & 0.0 & 28.5 & SVC & 25.0 & 26.4 & 25.0/25.0 & 26.5 & 25.0 & 26.5 & 26.4 & 25.0 & 26.4 \\
DVPF (paper) & $\alpha{=}10$ & 51.0 & 1.5 & 36.3 & LR & 35.9 & 36.3 & 26.2/33.0 & 36.9 & 36.2 & 36.9 & 36.3 & 35.9 & 36.3 \\
DVPF (code) & $\alpha{=}2$ & 66.6 & 7.9 & 43.5 & MLP & 43.5 & 28.9 & 25.0/33.2 & 41.8 & 41.8 & 31.7 & 43.5 & 43.5 & 28.9 \\
DVPF (code) & $\alpha{=}3$ & 54.4 & 3.9 & 35.7 & MLP & 35.7 & 28.1 & 25.0/29.6 & 34.9 & 34.9 & 30.3 & 35.7 & 35.7 & 28.1 \\
DVPF (code) & $\alpha{=}5$ & 30.9 & 0.8 & 30.0 & MLP & 30.0 & 27.0 & 25.0/26.6 & 29.3 & 29.3 & 28.0 & 30.0 & 30.0 & 27.0 \\
DVPF (code) & $\alpha{=}10$ & 8.1 & 0.2 & 26.1 & SVC & 26.0 & 25.2 & 25.0/25.5 & 26.3 & 26.3 & 26.1 & 26.0 & 26.0 & 25.2 \\
SensitiveNets (Fig.~1 detector) & 5 layers & 60.7 & 10.4 & 49.3 & MLP & 49.3 & 36.4 & 30.1/41.3 & 47.3 & 47.3 & 37.2 & 49.3 & 49.3 & 36.4 \\
INLP & 20 iterations & 97.3 & 94.3 & 88.4 & MLP & 88.4 & 26.1 & 70.9/85.2 & 79.4 & 79.4 & 26.2 & 88.4 & 88.4 & 26.1 \\
IVE & 500 removed & 54.2 & 7.9 & 29.3 & SVC & 28.8 & 27.6 & 25.1/28.2 & 28.9 & 28.8 & 28.9 & 28.8 & 28.8 & 27.6 \\
IVE & 490 removed & 78.5 & 20.9 & 35.4 & SVC & 33.5 & 27.9 & 25.4/32.3 & 33.8 & 33.8 & 29.7 & 33.5 & 33.5 & 27.9 \\
IVE & 480 removed & 87.0 & 32.7 & 42.8 & SVC & 40.4 & 29.7 & 26.0/38.1 & 40.4 & 40.4 & 32.2 & 40.4 & 40.4 & 29.7 \\
IVE & 470 removed & 90.5 & 47.8 & 48.6 & SVC & 46.8 & 29.8 & 26.8/42.6 & 45.7 & 45.7 & 32.4 & 46.8 & 46.8 & 29.8 \\
Multi-IVE & 506 removed, 5 locked & 20.8 & 1.7 & 26.7 & SVC & 26.5 & 24.9 & 25.0/25.3 & 26.9 & 26.9 & 26.9 & 26.5 & 26.5 & 24.9 \\
Multi-IVE & 500 removed, 5 locked & 57.7 & 10.4 & 29.2 & SVC & 28.1 & 26.2 & 25.1/27.2 & 28.5 & 28.5 & 26.5 & 28.1 & 28.1 & 26.2 \\
Multi-IVE & 490 removed, 5 locked & 80.8 & 28.0 & 33.0 & SVC & 31.3 & 26.3 & 25.3/29.7 & 32.1 & 32.1 & 27.8 & 31.3 & 31.3 & 26.3 \\
Multi-IVE & 470 removed & 91.9 & 56.9 & 45.3 & SVC & 41.7 & 27.3 & 26.2/36.5 & 41.3 & 41.3 & 27.7 & 41.7 & 41.7 & 27.3 \\
PE-MIU & $s{=}4$ & 64.7 & 13.8 & 28.6 & set transf. & 26.1 & 27.4 & 25.0/25.4 & 25.7 & 25.7 & 25.5 & 27.4 & 26.1 & 27.4 \\
PE-MIU & $s{=}16$ & 94.6 & 92.5 & 38.7 & set transf. & 31.6 & 27.5 & 25.0/25.8 & 30.9 & 30.9 & 27.4 & 31.6 & 31.6 & 27.5 \\
FaRM & 0.3 epochs & 79.9 & 18.9 & 34.5 & MLP & 34.5 & 32.8 & 25.3/30.2 & 34.7 & 34.7 & 32.9 & 34.5 & 34.5 & 32.8 \\
FaRM & 20 epochs & 72.9 & 17.5 & 29.7 & MLP & 29.7 & 28.9 & 25.1/27.5 & 29.8 & 29.8 & 29.0 & 29.7 & 29.7 & 28.9 \\
PCA + LEACE & $k{=}22$ & 78.1 & 21.6 & 33.4 & SVC & 31.0 & 25.4 & 25.2/28.7 & 30.3 & 30.3 & 26.3 & 31.0 & 31.0 & 25.4 \\
PCA + LEACE & $k{=}42$ & 91.2 & 56.8 & 46.1 & SVC & 41.1 & 25.5 & 26.2/35.9 & 39.7 & 39.7 & 26.9 & 41.1 & 41.1 & 25.5 \\
PCA truncation & $k{=}8$ & 37.3 & 4.2 & 27.3 & SVC & 26.8 & 25.2 & 25.0/25.8 & 27.2 & 27.0 & 27.2 & 26.8 & 26.8 & 25.2 \\
PCA truncation & $k{=}12$ & 60.9 & 10.5 & 28.8 & SVC & 28.0 & 25.6 & 25.1/26.6 & 28.2 & 28.2 & 26.4 & 28.0 & 28.0 & 25.6 \\
PCA truncation & $k{=}22$ & 82.2 & 26.3 & 35.1 & SVC & 32.8 & 25.8 & 25.4/30.4 & 32.5 & 32.5 & 29.4 & 32.8 & 32.8 & 25.8 \\
PCA truncation & $k{=}42$ & 91.9 & 60.0 & 48.5 & SVC & 44.4 & 29.1 & 26.6/37.6 & 43.0 & 43.0 & 31.0 & 44.4 & 44.4 & 29.1 \\
Random dimensions & $k{=}8$ & 31.5 & 2.6 & 27.5 & SVC & 27.4 & 25.2 & 25.0/26.6 & 27.4 & 27.4 & 26.7 & 27.4 & 27.4 & 25.2 \\
Random dimensions & $k{=}10$ & 45.0 & 4.4 & 29.2 & SVC & 28.2 & 25.7 & 25.0/28.3 & 28.1 & 28.1 & 25.6 & 28.2 & 28.2 & 25.7 \\
Random dimensions & $k{=}22$ & 79.0 & 23.2 & 35.8 & SVC & 33.7 & 25.6 & 25.6/33.6 & 32.8 & 32.8 & 26.1 & 33.7 & 33.7 & 25.6 \\
TEMPER (ours) & $k{=}22$, $\lambda{=}3$ & 79.6 & 25.9 & 26.5 & MLP & 26.5 & 25.5 & 25.1/26.3 & 27.0 & 27.0 & 26.1 & 26.5 & 26.5 & 25.5 \\
TEMPER (ours) & $k{=}42$, $\lambda{=}3$ & 88.4 & 43.3 & 28.6 & SVC & 28.2 & 26.7 & 25.3/27.1 & 28.4 & 28.4 & 25.0 & 28.2 & 28.2 & 26.7 \\
TEMPER (ours) & $k{=}42$, $\lambda{=}10$ & 72.5 & 17.6 & 26.4 & MLP & 26.4 & 25.3 & 25.0/25.7 & 26.7 & 26.7 & 24.5 & 26.4 & 26.4 & 25.3 \\
TEMPER (ours) & $k{=}62$, $\lambda{=}3$ & 89.1 & 47.8 & 28.7 & SVC & 27.3 & 27.0 & 25.3/27.0 & 28.5 & 28.5 & 26.4 & 27.3 & 27.3 & 27.0 \\
TEMPER (ours) & $k{=}128$, $\lambda{=}3$ & 95.0 & 86.5 & 36.2 & MLP & 36.2 & 24.9 & 25.7/34.6 & 37.0 & 37.0 & 26.0 & 36.2 & 36.2 & 24.9 \\
\bottomrule
\end{tabular}

\end{table*}

\clearpage   %
\subsection{Full Results}
\label{app:results-full}

We list every evaluated setting of TEMPER, the label-free controls and the compared methods in Tables~\ref{tab:s-full-gender} and~\ref{tab:s-full-ethnicity}.
The settings selected in Table~\ref{tab:matched} are marked by superscripts.
The verification columns list IJB-C TMR at 4 FMRs and RFW TMR at FMR~$10^{-3}$.
The leakage columns list the worst case and the attacker that attains the worst case, on IJB-C for gender and on RFW for ethnicity.
The abbreviation pc denotes the $k$NN attacker with prior-corrected votes, and set transf.\ and unwinding denote the attackers of PE-MIU on the set of blocks and on the reordered templates (Appendix~\ref{app:baselines-pemiu}).
The leakage columns also list the MLP attacker, the AUC of the MLP attacker, the LR attacker and the $k$NN-51 attacker with raw and prior-corrected votes.
The transfer columns list the worst case of the MLP and LR attackers on IJB-C and on RFW.
The transfer is the value on RFW for gender and on IJB-C for ethnicity.
The SVC attacker and the $k$NN attackers with 1 and 11 neighbours enter only through the worst case.
The selection of Table~\ref{tab:matched} uses the unrounded leakage.
TEMPER at $k=42$, $\lambda=10$ for gender, with 52.52\%, and random dimensions at $k=42$ for ethnicity, with 50.03\%, are therefore not selected at the thresholds of 52.5\% and 50\%.

\newcommand{\SFullCaptionGender}{Full results, gender. TMR (\%) on IJB-C at 4 FMRs and on RFW at FMR~$10^{-3}$, and leakage (\%, chance 50\%) on IJB-C, with the worst case of the MLP and LR attackers on IJB-C and RFW as transfer. Superscripts a--d mark the settings selected in Table~\ref{tab:matched} at worst-case leakage $\le52.5$, 55, 60 and 70\%.}
\newcommand{\SFullCaptionEthnicity}{Full results, ethnicity, with the columns of Table~\ref{tab:s-full-gender} and leakage on RFW (chance 25\%). Superscripts a--d mark the settings selected in Table~\ref{tab:matched} at worst-case leakage $\le27.5$, 30, 40 and 50\%.}

\begingroup\scriptsize\setlength{\tabcolsep}{2.6pt}
\begin{longtable}{@{}l*{5}{r}rlrrrrrr@{}}
\caption{\SFullCaptionGender}\label{tab:s-full-gender}\\
\toprule
 & \multicolumn{5}{c}{Verification: TMR (\%)} & \multicolumn{6}{c}{Leakage, IJB-C (\%)} & \multicolumn{2}{c}{Transfer (\%)} \\
\cmidrule(lr){2-6}\cmidrule(lr){7-12}\cmidrule(lr){13-14}
 & \multicolumn{4}{c}{IJB-C, FMR} & RFW & Worst & & & & & $k$NN-51 & IJB-C & RFW \\
Setting & $10^{-4}$ & $10^{-3}$ & $10^{-2}$ & $10^{-1}$ & $10^{-3}$ & case & Attacker & MLP & AUC & LR & raw/pc & worst & worst \\
\midrule
\endfirsthead
\multicolumn{14}{@{}l}{\footnotesize Table~\ref{tab:s-full-gender} (continued).}\\
\toprule
 & \multicolumn{5}{c}{Verification: TMR (\%)} & \multicolumn{6}{c}{Leakage, IJB-C (\%)} & \multicolumn{2}{c}{Transfer (\%)} \\
\cmidrule(lr){2-6}\cmidrule(lr){7-12}\cmidrule(lr){13-14}
 & \multicolumn{4}{c}{IJB-C, FMR} & RFW & Worst & & & & & $k$NN-51 & IJB-C & RFW \\
Setting & $10^{-4}$ & $10^{-3}$ & $10^{-2}$ & $10^{-1}$ & $10^{-3}$ & case & Attacker & MLP & AUC & LR & raw/pc & worst & worst \\
\midrule
\endhead
\bottomrule
\endlastfoot
\multicolumn{14}{@{}l}{\textit{No removal}}\\
-- & 96.3 & 97.6 & 98.6 & 99.1 & 96.8 & 96.0 & $k$NN-51\,pc & 95.8 & 0.989 & 65.1 & 95.5/96.0 & 95.8 & 96.0 \\
\addlinespace[2pt]
\multicolumn{14}{@{}l}{\textit{PFRNet}}\\
$\lambda_{\mathrm{dis}}{=}0$ & 83.9 & 89.9 & 94.3 & 97.7 & 90.9 & 95.6 & MLP & 95.6 & 0.988 & 59.3 & 94.0/94.1 & 95.6 & 96.1 \\
$\lambda_{\mathrm{dis}}{=}0.001$ & 87.3 & 92.1 & 95.5 & 98.0 & 92.8 & 95.6 & MLP & 95.6 & 0.989 & 57.5 & 93.9/94.2 & 95.6 & 96.1 \\
$\lambda_{\mathrm{dis}}{=}0.003$ & 90.4 & 94.0 & 96.4 & 98.3 & 93.6 & 95.7 & MLP & 95.7 & 0.989 & 59.0 & 94.3/94.6 & 95.7 & 96.0 \\
$\lambda_{\mathrm{dis}}{=}0.01$ & 94.6 & 96.5 & 97.7 & 98.8 & 95.2 & 95.7 & MLP & 95.7 & 0.988 & 58.6 & 94.6/95.1 & 95.7 & 96.0 \\
$\lambda_{\mathrm{dis}}{=}0.03$ & 95.4 & 97.1 & 98.1 & 99.0 & 95.2 & 95.8 & MLP & 95.8 & 0.989 & 56.9 & 95.0/95.5 & 95.8 & 96.0 \\
\end{longtable}
\endgroup

\begingroup\scriptsize\setlength{\tabcolsep}{2.6pt}
\begin{longtable}{@{}l*{5}{r}rlrrrrrr@{}}
\caption{\SFullCaptionEthnicity}\label{tab:s-full-ethnicity}\\
\toprule
 & \multicolumn{5}{c}{Verification: TMR (\%)} & \multicolumn{6}{c}{Leakage, RFW (\%)} & \multicolumn{2}{c}{Transfer (\%)} \\
\cmidrule(lr){2-6}\cmidrule(lr){7-12}\cmidrule(lr){13-14}
 & \multicolumn{4}{c}{IJB-C, FMR} & RFW & Worst & & & & & $k$NN-51 & IJB-C & RFW \\
Setting & $10^{-4}$ & $10^{-3}$ & $10^{-2}$ & $10^{-1}$ & $10^{-3}$ & case & Attacker & MLP & AUC & LR & raw/pc & worst & worst \\
\midrule
\endfirsthead
\multicolumn{14}{@{}l}{\footnotesize Table~\ref{tab:s-full-ethnicity} (continued).}\\
\toprule
 & \multicolumn{5}{c}{Verification: TMR (\%)} & \multicolumn{6}{c}{Leakage, RFW (\%)} & \multicolumn{2}{c}{Transfer (\%)} \\
\cmidrule(lr){2-6}\cmidrule(lr){7-12}\cmidrule(lr){13-14}
 & \multicolumn{4}{c}{IJB-C, FMR} & RFW & Worst & & & & & $k$NN-51 & IJB-C & RFW \\
Setting & $10^{-4}$ & $10^{-3}$ & $10^{-2}$ & $10^{-1}$ & $10^{-3}$ & case & Attacker & MLP & AUC & LR & raw/pc & worst & worst \\
\midrule
\endhead
\bottomrule
\endlastfoot
\multicolumn{14}{@{}l}{\textit{No removal}}\\
-- & 96.3 & 97.6 & 98.6 & 99.1 & 96.8 & 90.0 & MLP & 90.0 & 0.986 & 41.2 & 83.7/89.7 & 81.4 & 90.0 \\
\addlinespace[2pt]
\multicolumn{14}{@{}l}{\textit{PFRNet}}\\
$\lambda_{\mathrm{dis}}{=}0$ & 83.8 & 89.7 & 94.2 & 97.6 & 93.1 & 89.4 & MLP & 89.4 & 0.985 & 37.0 & 72.6/82.7 & 80.5 & 89.4 \\
$\lambda_{\mathrm{dis}}{=}0.001$ & 91.1 & 94.2 & 96.6 & 98.4 & 93.1 & 89.0 & MLP & 89.0 & 0.984 & 29.7 & 72.9/83.9 & 80.0 & 89.0 \\
$\lambda_{\mathrm{dis}}{=}0.003$ & 92.7 & 95.3 & 97.2 & 98.6 & 94.1 & 89.4 & MLP & 89.4 & 0.984 & 29.4 & 74.9/85.6 & 80.2 & 89.4 \\
$\lambda_{\mathrm{dis}}{=}0.01$ & 93.9 & 96.0 & 97.6 & 98.7 & 93.8 & 89.1 & MLP & 89.1 & 0.984 & 30.1 & 75.4/86.2 & 79.8 & 89.1 \\
$\lambda_{\mathrm{dis}}{=}0.03$ & 92.9 & 95.4 & 97.2 & 98.6 & 92.3 & 88.6 & MLP & 88.6 & 0.984 & 29.2 & 74.5/87.2 & 80.1 & 88.6 \\
\end{longtable}
\endgroup

\subsection{Comparison at Equal Template Dimension}
\label{app:results-matchedk}

The frontier of each method consists of the best settings of that method on the test sets, and this selection favours TEMPER, which has the most settings.
We avoid this selection by comparing TEMPER with PCA truncation and random dimensions at the same $k$ (Table~\ref{tab:s-matchedk}).
TEMPER at $\lambda=3$ retains the TMR at FMR~$10^{-3}$ of PCA truncation at a leakage of 54.5\% against 70.7\%, for gender at $k=42$.
TEMPER at $\lambda=1$ retains at least the TMR at FMR~$10^{-3}$ of PCA truncation at a leakage of 31.5\% against 48.5\%, for ethnicity at $k=42$.
The ethnicity leakage of TEMPER with $\lambda\ge10$ stays within 2.4\% of chance up to $k=128$.
The ethnicity leakage of PCA truncation increases from 28.8\% at $k=12$ to 86.5\% at $k=128$.

\begin{table}[!t]
\centering
\caption{TEMPER and the label-free controls at equal template dimension $k$. IJB-C TMR at FMR~$10^{-3}$ / worst-case leakage (\%) at each $k$ (-- for a setting outside the grid of the method).}
\label{tab:s-matchedk}
\scriptsize
\setlength{\tabcolsep}{3pt}
\begin{tabular}{@{}rcccccccc@{}}
\toprule
 & \multicolumn{2}{c}{Label-free controls} & \multicolumn{6}{c}{TEMPER, $\lambda$} \\
\cmidrule(lr){2-3}\cmidrule(lr){4-9}
$k$ & PCA trunc. & Random & $1$ & $3$ & $10$ & $30$ & $100$ & $300$ \\
\midrule
\multicolumn{9}{@{}l}{\textit{Gender} (chance 50\%)}\\
12 & 60.8/54.6 & 55.1/55.9 & -- & 60.9/52.5 & 60.9/52.3 & 59.6/51.8 & 54.9/51.4 & -- \\
22 & 82.3/58.4 & 79.0/60.6 & -- & 82.4/52.2 & 82.2/51.1 & 79.9/51.1 & 68.4/51.6 & -- \\
42 & 92.0/70.7 & 90.8/73.3 & -- & 92.1/54.5 & 88.2/52.5 & 81.0/51.7 & 70.8/52.0 & 63.4/53.5 \\
62 & 94.9/80.5 & 93.9/83.2 & -- & 94.2/57.0 & 88.7/53.0 & 82.5/52.6 & 75.8/52.9 & -- \\
96 & -- & -- & -- & 95.2/60.5 & 91.4/54.5 & 87.9/54.1 & 85.3/54.0 & -- \\
128 & 97.0/93.4 & 96.3/93.8 & -- & 96.4/72.0 & 94.2/60.2 & 92.7/58.3 & 91.6/58.1 & 90.2/58.0 \\
256 & -- & -- & -- & 97.6/95.9 & 97.6/95.6 & 97.6/95.6 & 97.5/95.5 & 97.4/95.7 \\
\midrule
\multicolumn{9}{@{}l}{\textit{Ethnicity} (chance 25\%)}\\
12 & 60.9/28.8 & 55.1/31.4 & 59.6/26.5 & 58.8/26.6 & 53.9/26.0 & 50.1/25.6 & -- & -- \\
22 & 82.2/35.1 & 79.0/35.8 & 82.4/28.5 & 79.6/26.5 & 73.9/26.3 & 57.4/25.7 & -- & -- \\
42 & 91.9/48.5 & 90.8/50.0 & 92.6/31.5 & 88.4/28.6 & 72.5/26.4 & 59.6/26.1 & -- & -- \\
62 & 94.9/57.6 & 93.9/62.0 & 95.0/36.2 & 89.1/28.7 & 71.8/26.8 & 58.8/25.9 & -- & -- \\
96 & -- & -- & 96.5/54.1 & 90.8/30.3 & 73.3/26.2 & 59.4/25.6 & -- & -- \\
128 & 97.0/86.5 & 96.3/84.2 & 96.9/77.8 & 95.0/36.2 & 75.9/27.4 & 63.1/26.3 & -- & 44.9/26.8 \\
256 & -- & -- & 97.6/89.3 & 97.6/89.0 & 97.6/88.9 & 97.4/88.1 & -- & 97.1/86.1 \\
\bottomrule
\end{tabular}

\end{table}

\subsection{Training Seeds}
\label{app:results-seeds}

We repeat 4 settings of TEMPER at low leakage with 2 further training seeds (Table~\ref{tab:s-seeds}).
These settings are $k=22$, $\lambda=3$ for both attributes, $k=42$, $\lambda=30$ for gender and $k=42$, $\lambda=3$ for ethnicity.
We also repeat the 2 settings at $k=42$ without the variational autoencoder.

The IJB-C TMR at FMR~$10^{-3}$ and the worst-case leakage of TEMPER vary by at most 0.4\% and 0.8\%, respectively, over the seeds.
The 3 settings near chance stay near chance for every seed.
The IJB-C TMR at FMR~$10^{-3}$ varies by at most 1.0\% over all repeated settings.
The RFW TMR at FMR~$10^{-3}$ varies by up to 6.6\%, for TEMPER at $k=42$, $\lambda=3$ (ethnicity).

\begin{table}[!t]
\centering
\caption{Variation over training seeds, gender unless stated. IJB-C TMR (\%) at FMR~$10^{-4}$, $10^{-3}$ and $10^{-1}$, RFW TMR (\%) at FMR~$10^{-3}$ and leakage (\%) for seed 42, the run used everywhere else, and for 2 further seeds, with the range over the 3 seeds.}
\label{tab:s-seeds}
\footnotesize
\setlength{\tabcolsep}{4pt}
\begin{tabular}{@{}llrrrrrlrr@{}}
\toprule
 & & \multicolumn{3}{c}{IJB-C TMR, FMR} & RFW & \multicolumn{4}{c}{Leakage} \\
\cmidrule(lr){3-5}\cmidrule(lr){7-10}
Setting & Seed & $10^{-4}$ & $10^{-3}$ & $10^{-1}$ & $10^{-3}$ & Worst & Attacker & MLP & LR \\
\midrule
TEMPER, $k{=}22$, $\lambda{=}3$ & 42 & 71.0 & 82.4 & 96.8 & 27.8 & 52.2 & SVC & 52.0 & 51.5 \\
 & 1 & 71.0 & 82.3 & 96.7 & 29.5 & 52.4 & SVC & 51.6 & 51.5 \\
 & 2 & 71.0 & 82.4 & 96.7 & 26.2 & 52.3 & SVC & 52.0 & 51.4 \\
 & \textit{range} & 0.0 & 0.1 & 0.1 & 3.3 & 0.2 &  & 0.4 & 0.1 \\
\addlinespace[2pt]
TEMPER, $k{=}22$, $\lambda{=}3$ (ethnicity) & 42 & 67.0 & 79.6 & 96.3 & 25.9 & 26.5 & MLP & 26.5 & 25.5 \\
 & 1 & 67.1 & 79.8 & 96.3 & 26.5 & 27.0 & SVC & 26.5 & 25.6 \\
 & 2 & 67.2 & 79.8 & 96.3 & 29.5 & 26.9 & MLP & 26.9 & 25.2 \\
 & \textit{range} & 0.2 & 0.2 & 0.0 & 3.7 & 0.5 &  & 0.4 & 0.3 \\
\addlinespace[2pt]
TEMPER, $k{=}42$, $\lambda{=}30$ & 42 & 69.1 & 81.0 & 96.6 & 26.8 & 51.7 & MLP & 51.7 & 50.1 \\
 & 1 & 69.5 & 80.9 & 96.5 & 23.0 & 51.8 & MLP & 51.8 & 51.8 \\
 & 2 & 69.0 & 80.8 & 96.6 & 24.8 & 52.0 & MLP & 52.0 & 51.8 \\
 & \textit{range} & 0.5 & 0.1 & 0.2 & 3.8 & 0.3 &  & 0.3 & 1.7 \\
\addlinespace[2pt]
TEMPER, $k{=}42$, $\lambda{=}3$ (ethnicity) & 42 & 81.5 & 88.4 & 97.2 & 43.3 & 28.6 & SVC & 28.2 & 26.7 \\
 & 1 & 80.9 & 88.1 & 97.2 & 42.9 & 28.2 & SVC & 27.6 & 26.8 \\
 & 2 & 81.6 & 88.5 & 97.4 & 49.5 & 29.1 & SVC & 28.3 & 26.0 \\
 & \textit{range} & 0.6 & 0.4 & 0.3 & 6.6 & 0.8 &  & 0.7 & 0.7 \\
\addlinespace[2pt]
without the variational autoencoder, $k{=}42$, $\lambda{=}30$ & 42 & 69.2 & 81.1 & 96.6 & 26.8 & 52.6 & LR & 51.3 & 52.6 \\
 & 1 & 68.2 & 80.3 & 96.5 & 27.2 & 52.3 & MLP & 52.3 & 51.4 \\
 & 2 & 69.7 & 81.3 & 96.7 & 25.7 & 52.7 & MLP & 52.7 & 52.3 \\
 & \textit{range} & 1.5 & 1.0 & 0.3 & 1.5 & 0.4 &  & 1.4 & 1.2 \\
\addlinespace[2pt]
without the variational autoencoder, $k{=}42$, $\lambda{=}3$ (ethnicity) & 42 & 80.2 & 87.6 & 97.2 & 46.2 & 28.8 & SVC & 27.7 & 26.7 \\
 & 1 & 80.3 & 87.5 & 97.2 & 47.3 & 28.3 & SVC & 27.4 & 25.2 \\
 & 2 & 80.3 & 87.6 & 97.2 & 47.4 & 28.1 & SVC & 27.7 & 25.6 \\
 & \textit{range} & 0.2 & 0.1 & 0.0 & 1.2 & 0.7 &  & 0.4 & 1.5 \\
\addlinespace[2pt]
\bottomrule
\end{tabular}

\end{table}

\subsection{Identity-Level Bootstrap on IJB-C}
\label{app:results-bootstrap}

We give 95\% percentile intervals for the IJB-C TMR at FMR~$10^{-3}$ and $10^{-4}$ of the settings in Table~\ref{tab:matched} (Table~\ref{tab:s-bootstrap}).
The 3,531 subjects of the 1:1 protocol are resampled with replacement 1000 times.
The threshold at the target FMR is recomputed on the resampled non-mated comparison scores in each draw, and the TMR at this FMR is recomputed on the resampled mated comparison scores.
The same draws are used for every method.
The difference between TEMPER and the other compared method with the highest TMR at FMR~$10^{-3}$ in a column of Table~\ref{tab:matched} is therefore a paired statistic.
The intervals cover only the sampling of the test set, since the training seed and the selected settings are fixed in every draw.

The paired interval of the difference excludes zero in every column at FMR~$10^{-3}$.
The difference is at least 4.2\% in the gender columns up to 60\% and the ethnicity columns up to 30\%. This difference exceeds 6 times the half-width of its interval.
The smallest lead of TEMPER at this FMR is 0.4\%, at 40\% and 50\% ethnicity leakage.
PE-MIU retains 0.7\% more TMR than TEMPER at FMR~$10^{-4}$ in the 3 columns in which $s=16$ is the selected setting of PE-MIU.
The paired interval of this difference also excludes zero.

\begin{table}[!t]
\centering
\scriptsize
\caption{Identity-level bootstrap on IJB-C. TMR (\%) at FMR~$10^{-3}$ and $10^{-4}$ with 95\% intervals for TEMPER and for the other compared method with the highest TMR at FMR~$10^{-3}$ in each column of Table~\ref{tab:matched}, and the paired difference. $^{\ast}$ marks the closest setting above the column threshold.}
\label{tab:s-bootstrap}
\setlength{\tabcolsep}{3pt}
\begin{tabular}{@{}llllll@{}}
\toprule
Threshold & FMR & TEMPER & Other method & TMR & TEMPER $-$ other \\
\midrule
gender $\le52.5$\% & $10^{-3}$ & $k{=}22$, $\lambda{=}3$: 82.4 [81.5, 83.5] & FaRM, 5 epochs & 76.7 [75.7, 77.8] & +5.7 [+4.8, +6.5] \\
gender $\le52.5$\% & $10^{-4}$ & $k{=}22$, $\lambda{=}3$: 71.0 [69.7, 72.4] & FaRM, 5 epochs & 64.2 [62.7, 65.6] & +6.8 [+5.5, +8.1] \\
\addlinespace[1pt]
gender $\le55$\% & $10^{-3}$ & $k{=}42$, $\lambda{=}3$: 92.1 [91.4, 92.8] & PE-MIU, $s{=}8$ & 87.9 [87.1, 88.7] & +4.2 [+3.7, +4.7] \\
gender $\le55$\% & $10^{-4}$ & $k{=}42$, $\lambda{=}3$: 87.4 [86.6, 88.4] & PE-MIU, $s{=}8$ & 85.2 [84.2, 86.1] & +2.2 [+1.6, +2.8] \\
\addlinespace[1pt]
gender $\le60$\% & $10^{-3}$ & $k{=}62$, $\lambda{=}3$: 94.2 [93.6, 94.7] & PE-MIU, $s{=}8$ & 87.9 [87.1, 88.7] & +6.3 [+5.7, +6.8] \\
gender $\le60$\% & $10^{-4}$ & $k{=}62$, $\lambda{=}3$: 90.8 [90.0, 91.6] & PE-MIU, $s{=}8$ & 85.2 [84.2, 86.1] & +5.6 [+5.1, +6.2] \\
\addlinespace[1pt]
gender $\le70$\% & $10^{-3}$ & $k{=}96$, $\lambda{=}3$: 95.2 [94.7, 95.7] & PE-MIU, $s{=}16$ & 94.6 [94.1, 95.2] & +0.6 [+0.4, +0.8] \\
gender $\le70$\% & $10^{-4}$ & $k{=}96$, $\lambda{=}3$: 92.9 [92.2, 93.5] & PE-MIU, $s{=}16$ & 93.6 [93.0, 94.2] & -0.7 [-1.0, -0.4] \\
\addlinespace[1pt]
ethnicity $\le27.5$\% & $10^{-3}$ & $k{=}22$, $\lambda{=}3$: 79.6 [78.6, 80.6] & Multi-IVE, 506 removed, 5 locked & 20.8 [19.7, 21.9] & +58.8 [+57.5, +60.1] \\
ethnicity $\le27.5$\% & $10^{-3}$ & $k{=}22$, $\lambda{=}3$: 79.6 [78.6, 80.6] & PE-MIU, $s{=}4$$^{\ast}$ & 64.7 [63.5, 66.0] & +14.8 [+13.9, +15.8] \\
ethnicity $\le27.5$\% & $10^{-4}$ & $k{=}22$, $\lambda{=}3$: 67.0 [65.6, 68.6] & Multi-IVE, 506 removed, 5 locked & 5.5 [4.9, 6.2] & +61.6 [+60.1, +63.1] \\
ethnicity $\le27.5$\% & $10^{-4}$ & $k{=}22$, $\lambda{=}3$: 67.0 [65.6, 68.6] & PE-MIU, $s{=}4$$^{\ast}$ & 56.8 [55.4, 58.2] & +10.2 [+9.0, +11.4] \\
\addlinespace[1pt]
ethnicity $\le30$\% & $10^{-3}$ & $k{=}62$, $\lambda{=}3$: 89.1 [88.3, 89.8] & FaRM, 20 epochs & 72.9 [71.7, 74.1] & +16.2 [+15.2, +17.1] \\
ethnicity $\le30$\% & $10^{-4}$ & $k{=}62$, $\lambda{=}3$: 82.5 [81.5, 83.5] & FaRM, 20 epochs & 58.5 [57.1, 60.0] & +24.0 [+22.6, +25.3] \\
\addlinespace[1pt]
ethnicity $\le40$\% & $10^{-3}$ & $k{=}128$, $\lambda{=}3$: 95.0 [94.5, 95.5] & PE-MIU, $s{=}16$ & 94.6 [94.1, 95.2] & +0.4 [+0.1, +0.6] \\
ethnicity $\le40$\% & $10^{-4}$ & $k{=}128$, $\lambda{=}3$: 92.9 [92.3, 93.6] & PE-MIU, $s{=}16$ & 93.6 [93.0, 94.2] & -0.7 [-1.0, -0.4] \\
\addlinespace[1pt]
ethnicity $\le50$\% & $10^{-3}$ & $k{=}128$, $\lambda{=}3$: 95.0 [94.5, 95.5] & PE-MIU, $s{=}16$ & 94.6 [94.1, 95.2] & +0.4 [+0.1, +0.6] \\
ethnicity $\le50$\% & $10^{-4}$ & $k{=}128$, $\lambda{=}3$: 92.9 [92.3, 93.6] & PE-MIU, $s{=}16$ & 93.6 [93.0, 94.2] & -0.7 [-1.0, -0.4] \\
\bottomrule
\end{tabular}

\end{table}

\subsection{ROC Curves}
\label{app:results-roc}

\begin{figure}[!t]
\centering
\includegraphics[width=\textwidth]{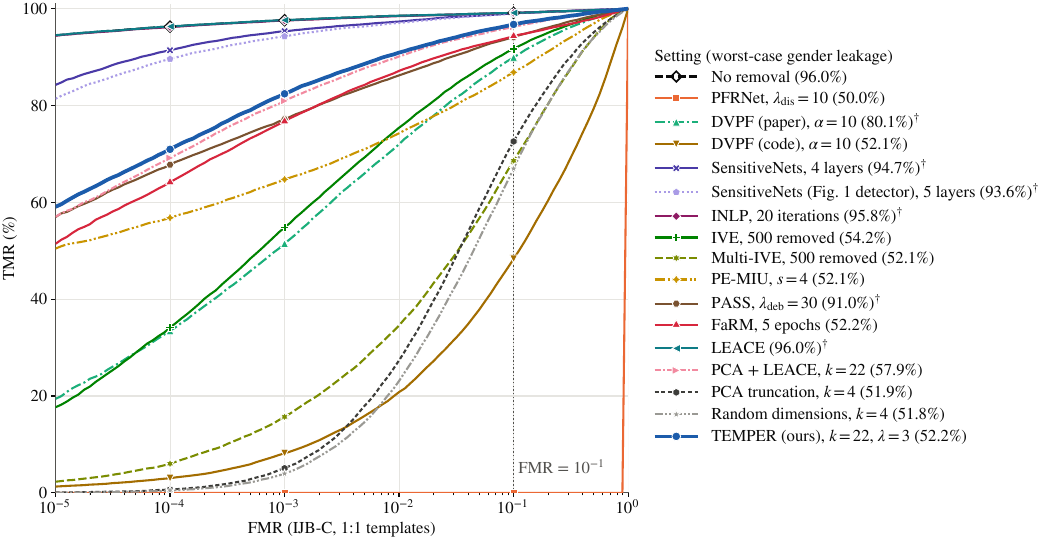}
\caption{IJB-C ROC curves of every compared method and control at one gender setting, and without removal. Each method is drawn at its setting of Table~\ref{tab:matched} at the lowest leakage threshold at which the method has a setting. $^{\dagger}$ marks a method without a setting at any threshold, drawn at its lowest leakage with at least 1\% TMR at FMR~$10^{-3}$. Markers at FMR $10^{-4}$, $10^{-3}$ and $10^{-1}$.}
\label{fig:s-roc-matched}
\end{figure}

To compare the methods over the whole range of FMRs, we draw the IJB-C ROC curve of every compared method and control at one gender setting (Fig.~\ref{fig:s-roc-matched}).
Each method is drawn at its setting of Table~\ref{tab:matched} at the lowest leakage threshold at which the method has a setting.
DVPF (paper), both variants of SensitiveNets, INLP, PASS and LEACE have no setting at any threshold of Table~\ref{tab:matched} and are drawn at their lowest leakage with at least 1\% TMR at FMR~$10^{-3}$, marked~$^{\dagger}$.
TEMPER retains the highest TMR at the 4 FMRs of Table~\ref{tab:s-full-gender} among the 8 methods and controls drawn at a setting near chance, and FaRM and PE-MIU retain the second-highest and the third-highest TMR at these FMRs.
PCA + LEACE, drawn at 57.9\% leakage, retains less TMR than TEMPER at the 4 FMRs.
The curves of INLP and LEACE are within 0.1\% TMR of the curve without removal at these FMRs.

We draw the ROC curves of every evaluated setting of the methods of Table~\ref{tab:matched}, of fair PCA and of fair kernel PCA, coloured by worst-case leakage (Fig.~\ref{fig:s-roc-all}).
Near-chance settings with a usable TMR at FMR~$10^{-3}$ appear in the gender panels of FaRM, PE-MIU and TEMPER and in the ethnicity panel of TEMPER only.
The near-chance settings of the other methods and of the controls retain at most 15.7\% TMR for gender and 37.3\% for ethnicity at FMR~$10^{-3}$.

\begin{figure*}[!t]
\centering
\includegraphics[width=\textwidth,height=0.82\textheight,keepaspectratio]{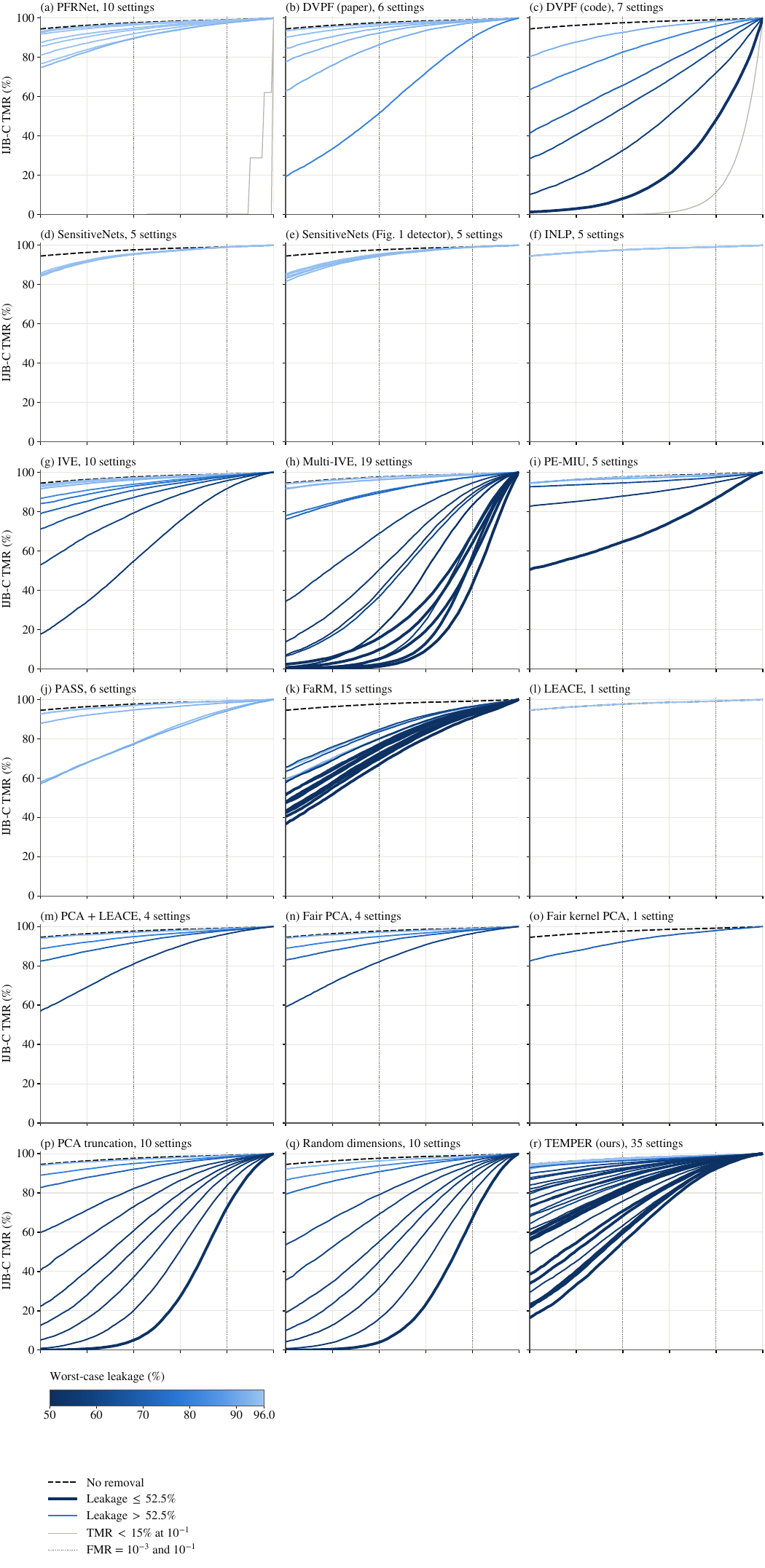}
\caption{IJB-C ROC curves of every evaluated setting, gender models. One panel per method, with each curve coloured by the worst-case leakage of the setting on IJB-C (chance 50\%, 96.0\% without removal).}
\label{fig:s-roc-all}
\end{figure*}

\begin{figure*}[!t]
\addtocounter{figure}{-1}
\centering
\includegraphics[width=\textwidth,height=0.82\textheight,keepaspectratio]{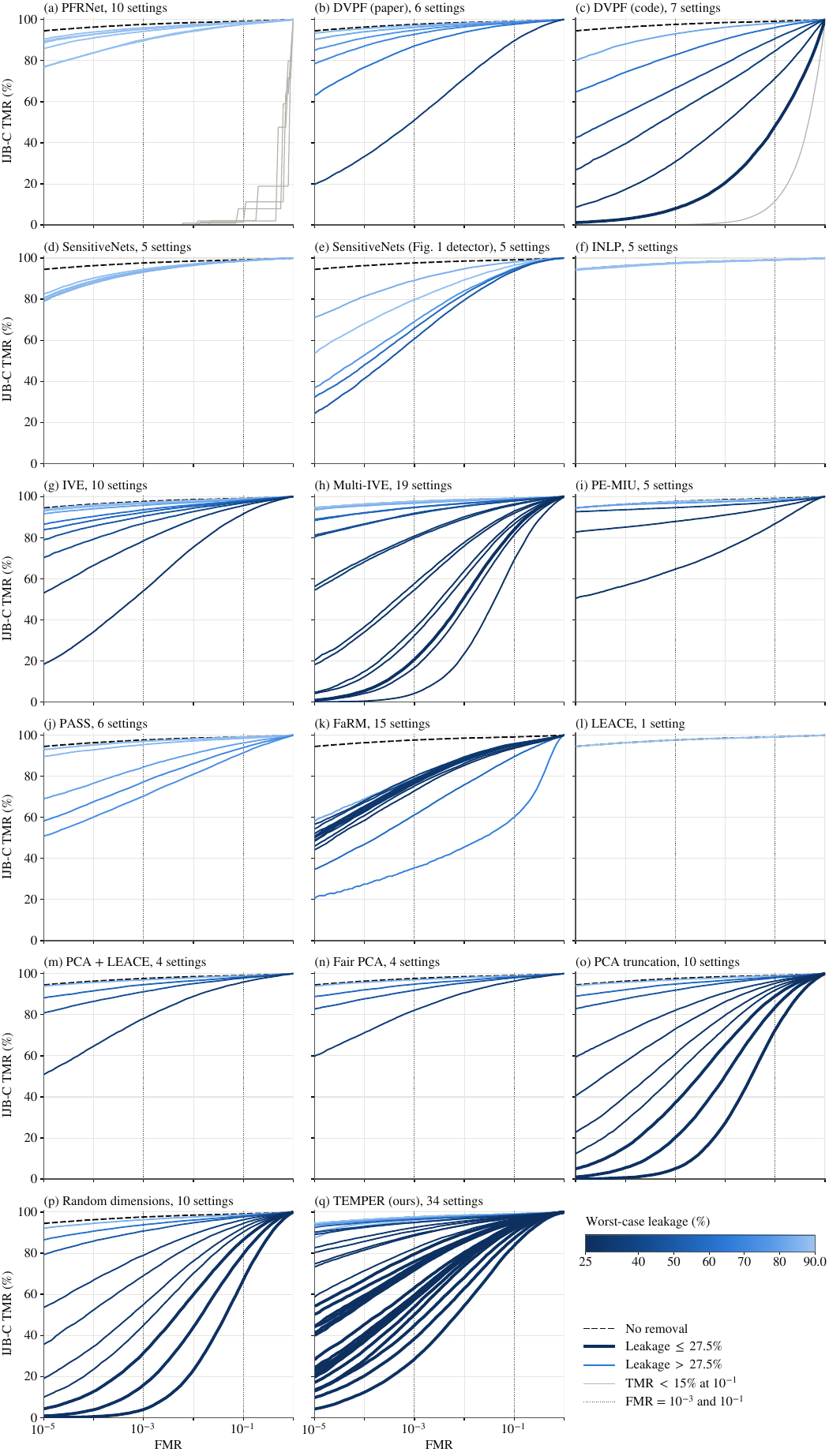}
\caption{(Continued) IJB-C ROC curves of every evaluated setting, ethnicity models, coloured by worst-case leakage on RFW (chance 25\%, 90.0\% without removal).}
\end{figure*}

\section{Further Leakage Measures}
\label{app:leakage}

The leakage measure of the main paper, worst-case balanced accuracy, is a classification measure and does not depend on the class prior.
We add an information measure based on the log loss of the attackers.
We also report attackers trained in the target domain, attackers on pooled templates and an MLP attacker selected on held-out identities, none of which enters the worst case.
The worst-case leakage of usable settings near chance changes by at most 3.5\% under these attackers (Tables~\ref{tab:s-target}, \ref{tab:s-pooled} and~\ref{tab:s-heldout}).

\subsection{Log-Loss Leakage}
\label{app:leakage-logloss}

Let $L_0=\Ent(\hat{p}(C))$ be the entropy of the empirical class prior of a test set, which is the log loss of the best constant prediction.
Let $L_q$ be the held-out log loss of an attacker $q$ on the same test set.
The normalised log-loss reduction is
\begin{equation}
\label{eq:s-glog}
G_{\log}=\frac{L_0-L_q}{L_0},
\end{equation}
whose population counterpart, with $\Ent(C)$ in place of $L_0$ and the expected log loss in place of $L_q$, is at most $\I(C;R)/\Ent(C)$ by Proposition~\ref{prop:lower} (Remark~\ref{rem:s-certify}).
The bound is attained by the posterior of $C$.
$G_{\log}$ therefore estimates a lower bound on the share of the entropy of the attribute that is encoded in the transformed template.
A positive value is evidence that the attribute is predictable from the transformed template.
A value near zero means only that this attacker is not better than the class prior.
A negative value means only that the attacker is worse than the class prior.
The normalisation by $L_0$ makes $G_{\log}$ comparable between test sets with different class priors, and we report $G_{\log}$ on each test set.
The posteriors of the attackers correspond to a uniform class prior, since our attackers are trained with a class-weighted loss.
We reweight these posteriors to the prior $\hat{p}$ of the test set before computing $L_q$, with $q'(c\mid\tilde{\bm{r}})\propto q(c\mid\tilde{\bm{r}})\,\hat{p}(c)$.
The attackers are those of Appendix~\ref{app:protocol-attackers}, trained on VGGFace2 and applied to every test set without retraining.
We compare the training-time surrogate with the attackers through the held-out log loss of the auxiliary classifier of TEMPER and of the sample-side variant (Table~\ref{tab:s-logloss}).
The auxiliary classifier is applied to the same input as in training, which is the normalised mean for TEMPER and the unnormalised posterior mean for the sample-side variant.
The per-batch standardisation of the auxiliary classifier is taken over the whole test set, and the posteriors of the classifier are reweighted to the test-set prior as for the attackers.

The MLP attacker attains a $G_{\log}$ of at least 0.56 without removal, for gender on IJB-C and for ethnicity on RFW (Table~\ref{tab:s-logloss}).
The LR attacker gives less than 0.1 on the same test sets.
An attacker attains a $G_{\log}$ above 0.01 on the sample-side variant, with the largest value of 0.10 for ethnicity on RFW.
The log loss of the auxiliary classifier of the sample-side variant is within 0.005 nats of the prior entropy.
The auxiliary classifier of the sample-side variant is therefore at chance in log loss as well, while the attackers predict the attribute (Appendix~\ref{app:ssv}).

Both attackers attain a $G_{\log}$ of at most 0.01 on every other transformed template of the table.
The MLP attacker attains a negative $G_{\log}$ on every setting of TEMPER and PCA truncation in the table.
The MLP attacker fits the training identities, and its log loss on unseen subjects exceeds the log loss of the class prior (Table~\ref{tab:s-logloss}).
The log loss increases with the confidence of these errors, whereas the balanced accuracy does not depend on the confidence of a prediction.

The log loss of the auxiliary classifier of TEMPER is between the log losses of the LR and MLP attackers and 0.02 to 0.25 nats above the prior entropy.
Neither the training-time surrogate nor the attackers attain a log loss more than 0.01 nats below the prior entropy on the transformed templates of TEMPER, on IJB-C or RFW.
Balanced accuracy is therefore the measure of the main paper, since a negative $G_{\log}$ is uninformative about the attribute.
A maximum of balanced accuracies is also less sensitive to the calibration of an overfitted attacker than the log loss.

\begin{table}[!t]
\centering
\caption{Log-loss leakage per test set. Prior entropy $L_0$, held-out log loss $L_q$ of the auxiliary classifier (Aux.) and of the attackers, both in nats, and normalised log-loss reduction $G_{\log}$~\eqref{eq:s-glog}. The auxiliary classifier is evaluated for TEMPER and the sample-side variant.}
\label{tab:s-logloss}
\footnotesize
\setlength{\tabcolsep}{4pt}
\begin{tabular}{@{}llrrrrrr@{}}
\toprule
 & & & \multicolumn{3}{c}{$L_q$} & \multicolumn{2}{c}{$G_{\log}$} \\
\cmidrule(lr){4-6}\cmidrule(lr){7-8}
Setting & Test set & $L_0$ & Aux. & MLP & LR & MLP & LR \\
\midrule
\multicolumn{8}{@{}l}{\textit{Gender}}\\
No removal & IJB-C & 0.681 & -- & 0.299 & 0.622 & 0.56 & 0.09 \\
 & RFW & 0.554 & -- & 0.291 & 0.587 & 0.48 & -0.06 \\
\addlinespace[1pt]
TEMPER, $k{=}42$, $\lambda{=}3$ & IJB-C & 0.681 & 0.778 & 1.057 & 0.681 & -0.55 & -0.00 \\
 & RFW & 0.554 & 0.650 & 0.887 & 0.556 & -0.60 & -0.00 \\
\addlinespace[1pt]
TEMPER, $k{=}42$, $\lambda{=}30$ & IJB-C & 0.681 & 0.713 & 0.883 & 0.681 & -0.30 & -0.00 \\
 & RFW & 0.554 & 0.577 & 0.731 & 0.554 & -0.32 & -0.00 \\
\addlinespace[1pt]
TEMPER, $k{=}96$, $\lambda{=}30$ & IJB-C & 0.681 & 0.725 & 0.963 & 0.676 & -0.42 & 0.01 \\
 & RFW & 0.554 & 0.588 & 0.789 & 0.551 & -0.42 & 0.00 \\
\addlinespace[1pt]
Sample-side variant, $\lambda{=}1000$ & IJB-C & 0.681 & 0.681 & 0.637 & 0.658 & 0.06 & 0.03 \\
 & RFW & 0.554 & 0.554 & 0.556 & 0.545 & -0.00 & 0.02 \\
\addlinespace[1pt]
PCA truncation, $k{=}12$ & IJB-C & 0.681 & -- & 0.751 & 0.681 & -0.10 & -0.00 \\
 & RFW & 0.554 & -- & 0.628 & 0.558 & -0.13 & -0.01 \\
\addlinespace[1pt]
DVPF (code), $\alpha{=}10$ & IJB-C & 0.681 & -- & 0.681 & 0.680 & -0.00 & 0.00 \\
 & RFW & 0.554 & -- & 0.555 & 0.556 & -0.00 & -0.00 \\
\midrule
\multicolumn{8}{@{}l}{\textit{Ethnicity}}\\
No removal & IJB-C & 1.059 & -- & 0.790 & 0.999 & 0.25 & 0.06 \\
 & RFW & 1.386 & -- & 0.593 & 1.447 & 0.57 & -0.04 \\
\addlinespace[1pt]
TEMPER, $k{=}42$, $\lambda{=}3$ & IJB-C & 1.059 & 1.240 & 1.569 & 1.058 & -0.48 & 0.00 \\
 & RFW & 1.386 & 1.632 & 2.272 & 1.385 & -0.64 & 0.00 \\
\addlinespace[1pt]
Sample-side variant, $\lambda{=}1000$ & IJB-C & 1.059 & 1.064 & 1.104 & 0.962 & -0.04 & 0.09 \\
 & RFW & 1.386 & 1.389 & 1.454 & 1.247 & -0.05 & 0.10 \\
\addlinespace[1pt]
PCA truncation, $k{=}12$ & IJB-C & 1.059 & -- & 1.211 & 1.058 & -0.14 & 0.00 \\
 & RFW & 1.386 & -- & 1.663 & 1.394 & -0.20 & -0.01 \\
\bottomrule
\end{tabular}

\end{table}

\subsection{Target-Domain Attackers}
\label{app:leakage-target}

Every leakage value in the frontiers and tables of the main paper is the leakage of a transfer attack, since the attackers are trained on VGGFace2 and tested on IJB-C or RFW (Table~\ref{tab:s-cross}).
An attacker with labelled templates from the target domain corresponds to a different setting.
We measure the leakage in this setting by splitting the leakage sets of IJB-C and RFW into halves of disjoint subjects, stratified by class.
We train the attackers on one half, test them on the other half and then swap the halves.
The attackers are the MLP, LR and $k$NN attackers of Appendix~\ref{app:protocol-attackers}, on inputs standardised with the training half.
The $k$NN attacker uses the whole training half as reference.
Each target-domain attacker is trained on 17{,}000 to 19{,}000 templates, against 3.1 million for the transfer attackers.
The target-domain leakage is the highest balanced accuracy over these attackers, each averaged over both halves.
The transfer values in Table~\ref{tab:s-target} are the worst case of the same 3 attacker families trained on VGGFace2, without the SVC attacker.

\begin{table}[!t]
\centering
\caption{Transfer and target-domain attackers. Highest balanced accuracy (\%) of the MLP, LR and $k$NN attackers on IJB-C and RFW, for attackers trained on VGGFace2 (transfer) and on identity-disjoint halves of the target set.}
\label{tab:s-target}
\scriptsize
\setlength{\tabcolsep}{3pt}
\begin{tabular}{@{}lrrrr@{}}
\toprule
 & \multicolumn{2}{c}{IJB-C} & \multicolumn{2}{c}{RFW} \\
\cmidrule(lr){2-3}\cmidrule(l){4-5}
Setting & Transfer & Target domain & Transfer & Target domain \\
\midrule
\multicolumn{5}{@{}l}{\textit{Gender} (chance 50\%)}\\
No removal & 96.0 & 95.6 & 96.0 & 95.6 \\
TEMPER, $k{=}22$, $\lambda{=}3$ & 52.0 & 51.7 & 51.3 & 53.5 \\
TEMPER, $k{=}42$, $\lambda{=}3$ & 54.5 & 53.8 & 53.1 & 56.8 \\
TEMPER, $k{=}42$, $\lambda{=}30$ & 51.7 & 53.6 & 51.4 & 53.6 \\
TEMPER, $k{=}62$, $\lambda{=}3$ & 57.0 & 56.2 & 55.7 & 59.5 \\
TEMPER, $k{=}96$, $\lambda{=}3$ & 60.5 & 57.1 & 59.6 & 61.8 \\
Sample-side variant, $\lambda{=}1000$ & 70.4 & 55.5 & 68.4 & 56.0 \\
PE-MIU, $s{=}4$ & 51.4 & 51.0 & 50.8 & 51.1 \\
PE-MIU, $s{=}8$ & 52.6 & 50.6 & 52.7 & 51.3 \\
PE-MIU, $s{=}16$ & 59.1 & 52.0 & 57.7 & 53.0 \\
PCA truncation, $k{=}12$ & 54.1 & 51.2 & 52.5 & 51.7 \\
PCA truncation, $k{=}22$ & 55.8 & 53.4 & 55.5 & 55.4 \\
Random dimensions, $k{=}12$ & 54.7 & 54.9 & 53.7 & 55.9 \\
Random dimensions, $k{=}22$ & 59.4 & 59.4 & 56.6 & 58.1 \\
IVE, 480 removed & 63.3 & 61.9 & 61.7 & 62.5 \\
IVE, 490 removed & 58.2 & 58.7 & 56.3 & 58.8 \\
IVE, 500 removed & 53.3 & 54.8 & 53.4 & 54.4 \\
PCA + LEACE, $k{=}22$ & 55.5 & 53.2 & 55.2 & 54.3 \\
DVPF (code), $\alpha{=}2$ & 68.8 & 61.8 & 67.6 & 60.2 \\
DVPF (code), $\alpha{=}5$ & 56.0 & 52.2 & 55.0 & 52.2 \\
DVPF (code), $\alpha{=}10$ & 52.1 & 51.0 & 51.3 & 50.9 \\
FaRM, 0.3 epochs & 61.0 & 54.9 & 59.9 & 56.2 \\
FaRM, 2 epochs & 53.3 & 53.5 & 53.5 & 54.5 \\
FaRM, 5 epochs & 52.2 & 53.5 & 52.0 & 53.9 \\
FaRM, 30 epochs & 51.0 & 51.5 & 50.7 & 52.5 \\
Multi-IVE, 500 removed & 52.1 & 53.1 & 52.2 & 57.0 \\
Multi-IVE, 470 removed, 5 locked & 53.1 & 52.9 & 52.3 & 55.2 \\
Multi-IVE, 450 removed, 5 locked & 55.8 & 55.0 & 55.9 & 55.7 \\
SensitiveNets, 4 layers & 94.7 & 92.5 & 94.0 & 92.9 \\
SensitiveNets (Fig.~1 detector), 5 layers & 93.6 & 86.3 & 91.9 & 87.7 \\
PFRNet, $\lambda_{\mathrm{dis}}{=}10$ & 50.0 & 50.0 & 50.0 & 50.0 \\
\midrule
\multicolumn{5}{@{}l}{\textit{Ethnicity} (chance 25\%)}\\
No removal & 81.4 & 79.9 & 90.0 & 94.8 \\
TEMPER, $k{=}22$, $\lambda{=}3$ & 27.6 & 28.2 & 26.5 & 30.0 \\
TEMPER, $k{=}42$, $\lambda{=}3$ & 29.0 & 30.0 & 28.2 & 34.6 \\
TEMPER, $k{=}42$, $\lambda{=}10$ & 27.0 & 28.0 & 26.4 & 28.7 \\
TEMPER, $k{=}62$, $\lambda{=}3$ & 29.5 & 30.6 & 27.3 & 35.2 \\
TEMPER, $k{=}128$, $\lambda{=}3$ & 37.0 & 33.7 & 36.2 & 53.4 \\
Sample-side variant, $\lambda{=}1000$ & 44.6 & 38.8 & 46.8 & 43.7 \\
PE-MIU, $s{=}4$ & 25.7 & 25.2 & 27.4 & 28.0 \\
PE-MIU, $s{=}16$ & 30.9 & 26.6 & 31.6 & 31.0 \\
PCA truncation, $k{=}12$ & 28.2 & 27.9 & 28.0 & 29.9 \\
PCA truncation, $k{=}22$ & 32.5 & 30.3 & 32.8 & 35.4 \\
PCA truncation, $k{=}42$ & 43.0 & 36.7 & 44.4 & 52.5 \\
Random dimensions, $k{=}22$ & 32.8 & 34.7 & 33.7 & 39.7 \\
IVE, 470 removed & 45.7 & 39.5 & 46.8 & 55.8 \\
IVE, 490 removed & 33.8 & 32.4 & 33.5 & 39.0 \\
IVE, 500 removed & 28.9 & 29.7 & 28.8 & 32.9 \\
PCA + LEACE, $k{=}22$ & 30.3 & 28.6 & 31.0 & 33.4 \\
PCA + LEACE, $k{=}42$ & 39.7 & 34.6 & 41.1 & 47.8 \\
DVPF (paper), $\alpha{=}10$ & 36.9 & 36.9 & 36.3 & 37.2 \\
DVPF (code), $\alpha{=}2$ & 41.8 & 33.1 & 43.5 & 44.3 \\
DVPF (code), $\alpha{=}3$ & 34.9 & 31.2 & 35.7 & 34.9 \\
DVPF (code), $\alpha{=}5$ & 29.3 & 28.3 & 30.0 & 29.7 \\
DVPF (code), $\alpha{=}10$ & 26.3 & 26.2 & 26.0 & 26.9 \\
FaRM, 0.3 epochs & 34.7 & 32.1 & 34.5 & 37.6 \\
FaRM, 1 epoch & 32.2 & 29.7 & 32.7 & 36.1 \\
FaRM, 20 epochs & 29.8 & 27.8 & 29.7 & 31.8 \\
Multi-IVE, 506 removed, 5 locked & 26.9 & 26.9 & 26.5 & 27.5 \\
Multi-IVE, 500 removed, 5 locked & 28.5 & 28.1 & 28.1 & 30.8 \\
Multi-IVE, 490 removed, 5 locked & 32.1 & 29.8 & 31.3 & 34.4 \\
Multi-IVE, 470 removed & 41.3 & 34.8 & 41.7 & 49.8 \\
SensitiveNets, 5 layers & 78.7 & 71.8 & 85.7 & 89.9 \\
SensitiveNets (Fig.~1 detector), 5 layers & 47.3 & 42.1 & 49.3 & 50.7 \\
PFRNet, $\lambda_{\mathrm{dis}}{=}1$ & 26.3 & 29.4 & 26.4 & 32.7 \\
PFRNet, $\lambda_{\mathrm{dis}}{=}3$ & 26.5 & 30.8 & 26.4 & 36.1 \\
\bottomrule
\end{tabular}

\end{table}

Table~\ref{tab:s-target} lists the settings of Table~\ref{tab:matched} for TEMPER, PE-MIU, FaRM, IVE, Multi-IVE, PCA + LEACE, DVPF and PFRNet, and some settings of the label-free controls.
The table lists both SensitiveNets variants at the lowest leakage of each variant (Appendix~\ref{app:baselines-snets}).
INLP, PASS and LEACE have no setting in Table~\ref{tab:matched} for either attribute and are not listed.

The leakage without removal changes from 96.0\% to 95.6\% for gender on IJB-C under the target-domain attackers. This leakage increases from 90.0\% to 94.8\% for ethnicity on RFW.
The target-domain attackers attain less than the transfer attackers on most gender settings on IJB-C.
The target-domain attackers attain less than the transfer attackers on the sample-side variant, with 55.5\% against 70.4\% for gender on IJB-C.
The largest increase for gender on IJB-C is 1.9\% for TEMPER at $k=42$, $\lambda=30$, which then leaks 53.6\% and exceeds the near-chance threshold.
TEMPER at $k=22$, $\lambda=3$ stays near chance under these attackers, at 51.7\%.
The leakage of FaRM after 5 epochs, the near-chance setting of FaRM in Table~\ref{tab:matched}, increases from 52.2\% to 53.5\%, above the near-chance threshold.
FaRM after 30 epochs stays near chance, at 51.5\%.

The target-domain attackers attain more than the transfer attackers for most ethnicity settings on RFW.
The increase at the settings nearest chance is at most 3.5\% for TEMPER, against 4.1\% for IVE with 500 removed coordinates.
The increase near chance is at most 1.9\% for PE-MIU at $s=4$, PCA truncation at $k=12$ and DVPF (code) at $\alpha=10$.
The tested ethnicity settings of TEMPER are not near chance under these attackers, and the 2 settings nearest chance leak 30.0\% and 28.7\%.
The leakage of PFRNet, which does not retain a usable TMR at FMR~$10^{-3}$, increases by up to 9.7\% near chance.

The largest increase for ethnicity on RFW is 17.2\% for TEMPER at $k=128$, $\lambda=3$, which is about twice the largest increase for IVE or PCA truncation.
This setting of TEMPER is selected in the ethnicity columns at 40\% and 50\% of Table~\ref{tab:matched}.
PE-MIU at $s=16$ leaks less than this TEMPER setting under these attackers, at 31.0\% against 53.4\%, although these 3 attacker families do not include the attackers that exploit the permutation (Appendix~\ref{app:baselines-pemiu}).
The settings of FaRM and Multi-IVE selected in these columns also leak less than this TEMPER setting, at most 49.8\%.
PE-MIU would then attain the highest TMR at FMR~$10^{-3}$ in the ethnicity columns at 40\% and 50\% leakage.

\subsection{Pooled Templates}
\label{app:leakage-pooled}

The leakage of the main paper is measured per face crop.
Each IJB-C template used in verification is pooled from the transformed crops and the transformed mirror images of the template (Appendix~\ref{app:protocol-verification}).
We measure the leakage of the stored templates by transforming every crop and the mirrored crop and pooling the results as in verification.
The MLP, LR and $k$NN attackers of Appendix~\ref{app:protocol-attackers}, trained on VGGFace2, are then applied to the 23{,}096 pooled templates of the 1:1 protocol.
Each pooled template is labelled with the attribute of the subject, and we drop the templates of excluded subjects.
Table~\ref{tab:s-pooled} lists the worst case over these attackers on the pooled templates and the per-crop worst case of the same 3 families.
The table lists the gender settings of Table~\ref{tab:matched} for TEMPER, PE-MIU, FaRM, IVE, Multi-IVE, PCA + LEACE, DVPF (code) and PFRNet.
The label-free controls are listed at $k=12$ and 22.
The table lists both SensitiveNets variants at the lowest leakage of each variant.
DVPF (paper), INLP, PASS and LEACE have no gender setting in Table~\ref{tab:matched} and are not listed.

Pooling raises the leakage by at most 2.4\% and reduces the leakage by at most 3.5\%.
The largest decrease is for DVPF (code) at $\alpha=5$.
The MLP attacker predicts one class on the pooled templates of several methods.
The worst case of these methods therefore comes from the LR and $k$NN attackers.
Pooling raises the highest balanced accuracy of the MLP, LR and $k$NN attackers on TEMPER at $k=22$, $\lambda=3$ from 52.0\% to 52.9\%.
This setting thereby exceeds the near-chance threshold of 52.5\%.
TEMPER at $k=42$, $\lambda=30$ stays near chance, at 52.4\%.
The leakage of FaRM after 5 epochs increases from 52.2\% to 52.8\%, also above this threshold.
FaRM after 30 epochs stays near chance, at 51.3\%.

\begin{table}[!t]
\centering
\caption{Leakage of pooled IJB-C templates, gender. Worst case (\%) over the MLP, LR and $k$NN attackers per face crop and on the pooled templates of the 1:1 protocol, with the MLP and LR attackers on the pooled templates.}
\label{tab:s-pooled}
\footnotesize
\setlength{\tabcolsep}{4pt}
\begin{tabular}{@{}lrrrr@{}}
\toprule
 & Per crop & \multicolumn{3}{c}{Pooled templates} \\
\cmidrule(l){3-5}
Setting & Worst & Worst & MLP & LR \\
\midrule
No removal & 96.0 & 97.5 & 96.7 & 67.2 \\
TEMPER, $k{=}22$, $\lambda{=}3$ & 52.0 & 52.9 & 52.8 & 51.8 \\
TEMPER, $k{=}42$, $\lambda{=}3$ & 54.5 & 55.6 & 55.6 & 50.7 \\
TEMPER, $k{=}42$, $\lambda{=}30$ & 51.7 & 52.4 & 52.4 & 48.9 \\
TEMPER, $k{=}62$, $\lambda{=}3$ & 57.0 & 59.3 & 59.3 & 50.9 \\
TEMPER, $k{=}96$, $\lambda{=}3$ & 60.5 & 62.6 & 62.6 & 48.9 \\
PE-MIU, $s{=}4$ & 51.4 & 51.9 & 51.4 & 51.9 \\
PE-MIU, $s{=}8$ & 52.6 & 53.6 & 53.6 & 51.8 \\
PE-MIU, $s{=}16$ & 59.1 & 61.4 & 61.4 & 53.5 \\
PCA truncation, $k{=}12$ & 54.1 & 52.9 & 50.0 & 52.1 \\
PCA truncation, $k{=}22$ & 55.8 & 56.8 & 50.0 & 52.4 \\
Random dimensions, $k{=}12$ & 54.7 & 55.6 & 50.4 & 52.0 \\
Random dimensions, $k{=}22$ & 59.4 & 60.0 & 49.8 & 55.2 \\
IVE, 480 removed & 63.3 & 63.8 & 50.0 & 51.0 \\
IVE, 490 removed & 58.2 & 58.2 & 50.1 & 50.9 \\
IVE, 500 removed & 53.3 & 53.6 & 50.4 & 50.2 \\
PCA + LEACE, $k{=}22$ & 55.5 & 55.3 & 50.0 & 48.3 \\
DVPF (code), $\alpha{=}2$ & 68.8 & 66.4 & 66.4 & 50.0 \\
DVPF (code), $\alpha{=}5$ & 56.0 & 52.5 & 52.1 & 50.0 \\
DVPF (code), $\alpha{=}10$ & 52.1 & 50.2 & 50.1 & 50.0 \\
FaRM, 0.3 epochs & 61.0 & 63.4 & 63.4 & 54.7 \\
FaRM, 2 epochs & 53.3 & 53.6 & 50.7 & 52.0 \\
FaRM, 5 epochs & 52.2 & 52.8 & 50.0 & 51.0 \\
FaRM, 30 epochs & 51.0 & 51.3 & 50.0 & 50.0 \\
Multi-IVE, 500 removed & 52.1 & 52.4 & 52.4 & 51.0 \\
Multi-IVE, 470 removed, 5 locked & 53.1 & 52.8 & 50.0 & 52.8 \\
Multi-IVE, 450 removed, 5 locked & 55.8 & 55.0 & 50.1 & 54.0 \\
SensitiveNets, 4 layers & 94.7 & 96.2 & 96.2 & 61.0 \\
SensitiveNets (Fig.~1 detector), 5 layers & 93.6 & 95.9 & 95.9 & 60.3 \\
PFRNet, $\lambda_{\mathrm{dis}}{=}10$ & 50.0 & 50.0 & 50.0 & 50.0 \\
\bottomrule
\end{tabular}

\end{table}

\subsection{MLP Attacker Selected on Held-Out Identities}
\label{app:leakage-heldout}

The MLP attacker is fitted to the training identities, with a large gap between the training accuracy and the balanced accuracy on unseen subjects (Appendix~\ref{app:protocol-attackers}).
The attribute is constant within an identity, and identity stays predictable from the transformed template.
We test whether an MLP attacker selected on held-out identities generalises better to new identities than an attacker that memorises the training identities.
We hold out 10\% of the VGGFace2 training identities, stratified by class, and train the MLP attacker on the rest with 3 regularisation settings.
The weight decay and dropout of these settings are $(10^{-4},0.1)$, $(10^{-2},0.3)$ and $(10^{-1},0.5)$.
Each setting is trained for up to 15 epochs, and we keep the setting and epoch with the highest balanced accuracy on the held-out identities.
We test the selected attacker on the leakage sets of the main paper.
This attacker is not part of the worst case, since the attackers of the main paper do not use model selection (Section~\ref{sec:setup-leakage}).

\begin{table}[!t]
\centering
\caption{MLP attacker selected on held-out identities, against the MLP attacker of the main paper. Selected weight decay $w$, dropout $p$ and epoch $e$, balanced accuracy (\%) of both attackers and log loss of the selected attacker (nats, reweighted to the test-set prior), on IJB-C for gender and on RFW for ethnicity.}
\label{tab:s-heldout}
\footnotesize
\setlength{\tabcolsep}{4pt}
\begin{tabular}{@{}lrrrr@{}}
\toprule
 & & \multicolumn{2}{c}{Balanced acc.} & Log loss \\
\cmidrule(lr){3-4}
Setting & Selected $(w, p, e)$ & MLP & Selected & Selected \\
\midrule
\multicolumn{5}{@{}l}{\textit{Gender} (chance 50\%, IJB-C)}\\
No removal & $(10^{-1}, 0.5, 10)$ & 95.8 & 95.8 & 0.126 \\
TEMPER, $k{=}22$, $\lambda{=}3$ & $(10^{-1}, 0.5, 1)$ & 52.0 & 52.8 & 0.687 \\
TEMPER, $k{=}42$, $\lambda{=}3$ & $(10^{-4}, 0.1, 7)$ & 54.5 & 54.0 & 0.958 \\
TEMPER, $k{=}42$, $\lambda{=}30$ & $(10^{-4}, 0.1, 11)$ & 51.7 & 52.3 & 0.903 \\
TEMPER, $k{=}62$, $\lambda{=}3$ & $(10^{-4}, 0.1, 12)$ & 57.0 & 57.0 & 1.167 \\
TEMPER, $k{=}96$, $\lambda{=}30$ & $(10^{-4}, 0.1, 15)$ & 54.1 & 54.3 & 1.073 \\
Sample-side variant, $\lambda{=}1000$ & $(10^{-2}, 0.3, 14)$ & 70.4 & 70.1 & 0.587 \\
PE-MIU, $s{=}4$ & $(10^{-4}, 0.1, 3)$ & 50.6 & 51.8 & 0.680 \\
PCA truncation, $k{=}12$ & $(10^{-1}, 0.5, 1)$ & 54.1 & 54.6 & 0.679 \\
IVE, 500 removed & $(10^{-1}, 0.5, 1)$ & 53.0 & 54.6 & 0.675 \\
DVPF (code), $\alpha{=}10$ & $(10^{-1}, 0.5, 14)$ & 52.1 & 51.9 & 0.680 \\
FaRM, 0.3 epochs & $(10^{-2}, 0.3, 10)$ & 61.0 & 60.7 & 0.711 \\
FaRM, 2 epochs & $(10^{-4}, 0.1, 12)$ & 53.3 & 53.2 & 1.004 \\
FaRM, 5 epochs & $(10^{-4}, 0.1, 11)$ & 52.2 & 52.3 & 0.935 \\
FaRM, 30 epochs & $(10^{-4}, 0.1, 15)$ & 51.0 & 51.2 & 0.928 \\
Multi-IVE, 500 removed & $(10^{-1}, 0.5, 2)$ & 52.1 & 52.4 & 0.680 \\
Multi-IVE, 470 removed, 5 locked & $(10^{-1}, 0.5, 1)$ & 53.1 & 54.0 & 0.676 \\
Multi-IVE, 450 removed, 5 locked & $(10^{-1}, 0.5, 3)$ & 55.8 & 56.3 & 0.670 \\
SensitiveNets, 4 layers & $(10^{-2}, 0.3, 12)$ & 94.7 & 95.1 & 0.178 \\
SensitiveNets (Fig.~1 detector), 5 layers & $(10^{-2}, 0.3, 14)$ & 93.6 & 93.7 & 0.196 \\
\midrule
\multicolumn{5}{@{}l}{\textit{Ethnicity} (chance 25\%, RFW)}\\
No removal & $(10^{-4}, 0.1, 3)$ & 90.0 & 91.4 & 0.295 \\
TEMPER, $k{=}22$, $\lambda{=}3$ & $(10^{-1}, 0.5, 1)$ & 26.5 & 26.7 & 1.407 \\
TEMPER, $k{=}42$, $\lambda{=}3$ & $(10^{-1}, 0.5, 13)$ & 28.2 & 25.8 & 1.585 \\
TEMPER, $k{=}42$, $\lambda{=}10$ & $(10^{-1}, 0.5, 8)$ & 26.4 & 25.7 & 1.429 \\
PE-MIU, $s{=}4$ & $(10^{-4}, 0.1, 2)$ & 26.1 & 28.2 & 1.382 \\
PCA truncation, $k{=}12$ & $(10^{-1}, 0.5, 1)$ & 28.0 & 29.0 & 1.387 \\
IVE, 500 removed & $(10^{-2}, 0.3, 1)$ & 28.8 & 31.0 & 1.395 \\
DVPF (code), $\alpha{=}10$ & $(10^{-1}, 0.5, 10)$ & 26.0 & 25.7 & 1.387 \\
FaRM, 0.3 epochs & $(10^{-4}, 0.1, 2)$ & 34.5 & 34.4 & 1.549 \\
FaRM, 1 epoch & $(10^{-4}, 0.1, 3)$ & 32.7 & 32.9 & 1.588 \\
FaRM, 20 epochs & $(10^{-1}, 0.5, 2)$ & 29.7 & 29.6 & 1.414 \\
Multi-IVE, 506 removed, 5 locked & $(10^{-1}, 0.5, 3)$ & 26.5 & 26.2 & 1.388 \\
Multi-IVE, 500 removed, 5 locked & $(10^{-1}, 0.5, 8)$ & 28.1 & 28.9 & 1.404 \\
Multi-IVE, 490 removed, 5 locked & $(10^{-1}, 0.5, 3)$ & 31.3 & 33.1 & 1.368 \\
Multi-IVE, 470 removed & $(10^{-1}, 0.5, 1)$ & 41.7 & 44.6 & 1.258 \\
SensitiveNets, 5 layers & $(10^{-2}, 0.3, 9)$ & 85.7 & 85.3 & 0.473 \\
SensitiveNets (Fig.~1 detector), 5 layers & $(10^{-1}, 0.5, 13)$ & 49.3 & 50.5 & 1.157 \\
\bottomrule
\end{tabular}

\end{table}

Table~\ref{tab:s-heldout} lists no removal, the sample-side variant and low-leakage settings of TEMPER, PE-MIU, PCA truncation, IVE and DVPF (code).
The table also lists the settings of FaRM, Multi-IVE and SensitiveNets in Table~\ref{tab:s-target}.
We trained this attacker for the 37 settings of the table only.
PFRNet, DVPF (paper), INLP, PASS, LEACE, PCA + LEACE and random dimensions are therefore not listed.
The selected MLP attacker differs from the MLP attacker of the main paper by $-$2.4\% to $+$2.9\% over the 37 settings of the table.
The selected attacker exceeds the worst case of the main paper by at most 1.6\%, for IVE with 500 removed coordinates for ethnicity.
The excess over the worst case is at most 0.6\% on TEMPER.

The selected attacker attains 52.8\% for gender on TEMPER at $k=22$, $\lambda=3$, above the near-chance threshold of 52.5\%.
FaRM after 5 epochs stays near chance under this attacker, at 52.3\%.
The gender setting of TEMPER at $k=42$, $\lambda=30$ and the ethnicity setting at $k=22$, $\lambda=3$ stay near chance.
The log loss of the selected attacker on the leakage sets stays above the prior entropy on every setting of TEMPER.
The held-out identities belong to the training split of TEMPER.
The selected attackers achieve 48.4\% to 53.0\% on these identities for gender on the transformed templates of TEMPER.
The balanced accuracy used for the selection is therefore close to chance for TEMPER, and a stronger attacker is not found by this selection.

\section{Sample-Side Variant}
\label{app:ssv}

We describe the sample-side variant of TEMPER in this appendix and report the measurements behind Sections~\ref{sec:res-attacker} and~\ref{sec:res-ablation} of the main paper.
The entropy term of the sample-side variant acts on the sampled latent, and the auxiliary classifier is trained on the unstandardised posterior mean.
The rows of Table~\ref{tab:ablation} with the privacy loss on the sampled latent are the sample-side variant with a wider auxiliary classifier.
Table~\ref{tab:s-ablation} lists the sample-side variant with both auxiliary classifiers.

\subsection{Configuration}
\label{app:ssv-config}

We give the configuration of the sample-side variant in full.
The sample-side variant has the variational autoencoder of TEMPER, with a template branch, a class branch and a decoder.
The template branch of the sample-side variant is nonlinear and randomly initialised, and the variant has neither the relational term nor the memory.
The encoder of the template branch is an MLP with 4 linear layers and PReLU activations, of which the first 3 have width 512.
The last layer of this MLP outputs the mean $\bm{\mu}_r$ and the log-variance of a $d_r=480$-dimensional latent, and the mean and the log-variance share the first 3 layers.
The transformed template therefore has 480 dimensions, which is the template dimension of the nonlinear encoder in Table~\ref{tab:ablation}.
The log-variance of the template branch is clipped to $[-4,2]$.
Therefore, $\bm{\sigma}_r\in[e^{-2},e]$. The lower bound $\sigma_0=e^{-2}\approx0.135$ is the noise level of Proposition~\ref{prop:s-mask} once every log-variance is clipped to $-4$.
The log-variance of the class branch has no bound, as in TEMPER.
The class encoder is a separate MLP with 4 linear layers, of which the first 3 have width 512, and maps the embedding to the mean and the log-variance of the $d_c=32$-dimensional class latent.
The class latent has the class-conditional prior $\mathcal{N}(\bm{e}_c,\bm{I})$ of TEMPER, and the decoder has one hidden layer of 512 PReLU units and an $\ell_2$-normalised output, as in TEMPER.
The auxiliary classifier is an MLP with 3 hidden layers of 256 LeakyReLU units with slope 0.2 and dropout 0.2, and the classifier has no linear head.
The auxiliary classifier takes one Adam step at learning rate $10^{-3}$ in each minibatch, on the cross-entropy of the predictions from $\bm{\mu}_r$ without gradients to the encoder.
Both encoders, the decoder and the prior means then take one Adam step at learning rate $10^{-4}$ on the objective~\eqref{eq:total} with $\gamma=0$.
The entropy term of this step is the predictive entropy of the frozen classifier on the sample $\bm{z}_r=\bm{\mu}_r+\bm{\sigma}_r\odot\bm{\epsilon}$, without standardisation.
The transformed template is $\bm{r}=\bm{\mu}_r/\|\bm{\mu}_r\|_2$, which never enters the auxiliary classifier during training.
The other settings are those of TEMPER, with $\beta_r=0.1$, $\beta_c=1$, the KL terms divided by $d_r$ and $d_c$ and a reconstruction weight of 1.
Each minibatch has 16 identities of 16 images each, balanced over the classes.
We train for 10 epochs and evaluate the model of the last epoch.
The privacy weights of the reported models are listed in Tables~\ref{tab:s-diag} and~\ref{tab:s-ablation}.

\subsection{Representation Statistics}
\label{app:ssv-diag}

To explain why the auxiliary classifier of the sample-side variant is at chance, we measure the scale of the mean and of the noise of the template latent (Table~\ref{tab:s-diag}).
Part~(a) of Table~\ref{tab:s-diag} lists the median per-dimension standard deviation of $\bm{\mu}_r$ over images for the gender models of the sample-side variant.
Part~(a) also lists the balanced accuracy of an MLP attacker on the transformed template, on $\bm{\mu}_r$ and on one posterior sample $\bm{z}_r$.
The last 2 columns list the balanced accuracy of the auxiliary classifier of the model on $\bm{\mu}_r$ and on $\bm{z}_r$.
Part~(a) is computed on the IJB-C leakage set with an MLP attacker trained for 5 epochs on $10^{6}$ training templates.
The values of part~(a) therefore differ from the values of the leakage protocol.
The MLP attacker attains 67.9\% on the transformed template and 53.0\% on $\bm{z}_r$ at $\lambda=1000$, whereas the auxiliary classifier attains chance on $\bm{\mu}_r$ and on $\bm{z}_r$.
Part~(b) lists the sample-side variant and TEMPER without the relational term and the memory.
TEMPER without the relational term and the memory has a linear template branch, and the statistics of part~(b) are computed on 20{,}000 VGGFace2 training embeddings.
The accuracy and the entropy of the auxiliary classifier in part~(b) are averages over the training batches of the last epoch.

\begin{table}[!t]
\centering
\caption{Representation statistics of the sample-side variant and of TEMPER without the relational term and the memory. (a) Gender models of the sample-side variant, balanced accuracy (\%) on the IJB-C leakage set. (b) Template branch, auxiliary classifier (last epoch, training batches), IJB-C TMR and leakage (\%) on IJB-C (gender) or RFW (ethnicity).}
\label{tab:s-diag}
\footnotesize
\setlength{\tabcolsep}{4pt}
\textit{(a) Sample-side variant, gender}\\[2pt]
\begin{tabular}{@{}rrrrrrr@{}}
\toprule
 & Median std & \multicolumn{3}{c}{MLP attacker on} & \multicolumn{2}{c}{Auxiliary classifier on} \\
\cmidrule(lr){3-5}\cmidrule(lr){6-7}
$\lambda$ & of $\bm{\mu}_r$ & transformed template & $\bm{\mu}_r$ & $\bm{z}_r$ & $\bm{\mu}_r$ & $\bm{z}_r$ \\
\midrule
10 & 0.273 & 84.5 & 84.8 & 75.6 & 51.8 & 51.5 \\
100 & 0.107 & 71.1 & 70.7 & 59.9 & 50.1 & 50.0 \\
1000 & 0.029 & 67.9 & 67.3 & 53.0 & 50.0 & 50.0 \\
10000 & 0.001 & 62.2 & 62.4 & 50.9 & 50.0 & 50.4 \\
\bottomrule
\end{tabular}

\vspace{6pt}
\textit{(b) Template branch, auxiliary classifier and operating point}\\[2pt]
\begin{tabular}{@{}lrrrrrrrr@{}}
\toprule
 & Median & At clip & Median & Offset / & Aux.\ acc.\ (\%) & TMR & Worst & MLP / LR \\
Model & $\bm{\sigma}$ & (\%) & std of $\bm{\mu}$ & spread & / entropy & $10^{-3}$ & case & \\
\midrule
\multicolumn{9}{@{}l}{\textit{Sample-side variant}}\\
\quad gender, $\lambda{=}0$ & 0.98 & 0 & 0.15 & 0.03 & 72.8 / 0.116 & 97.3 & 94.1 & 94.1 / 56.8 \\
\quad gender, $\lambda{=}100$ & 0.135 & 92.2 & 0.108 & 4.6 & 51.4 / 0.692 & 71.3 & 73.9 & 73.9 / 60.8 \\
\quad gender, $\lambda{=}1000$ & 0.135 & 99.99 & 0.030 & 12.4 & 50.4 / 0.6927 & 46.1 & 70.4 & 70.4 / 58.9 \\
\quad ethnicity, $\lambda{=}1000$ & 0.135 & 100 & 0.024 & 30.0 & 25.5 / 1.3857 & 41.9 & 46.8 & 46.8 / 43.5 \\
\multicolumn{9}{@{}l}{\textit{TEMPER without the relational term and the memory}}\\
\quad gender, $k{=}42$, $\lambda{=}0$ & 0.38 & 0 & 0.93 & 0.02 & 84.0 / 0.461 & 93.2 & 66.6 & 65.1 / 53.9 \\
\quad gender, $k{=}42$, $\lambda{=}3$ & 0.39 & 0 & 1.00 & 0.05 & 58.2 / 0.680 & 77.1 & 56.1 & 56.1 / 51.2 \\
\quad gender, $k{=}42$, $\lambda{=}30$ & 0.67 & 0 & 1.56 & 0.03 & 55.4 / 0.686 & 77.4 & 51.2 & 51.2 / 50.2 \\
\quad gender, $k{=}62$, $\lambda{=}30$ & 0.71 & 0 & 1.70 & 0.03 & 56.1 / 0.684 & 78.3 & 52.9 & 52.1 / 52.9 \\
\quad ethnicity, $k{=}42$, $\lambda{=}0.1$ & 0.36 & 0 & 0.88 & 0.03 & 84.8 / 0.426 & 91.8 & 34.6 & 33.8 / 25.1 \\
\bottomrule
\end{tabular}
\end{table}

The share of log-variances of the template branch clipped to $-4$ increases from 92.2\% at $\lambda=100$ to 99.99\% at $\lambda=1000$ for gender.
The median standard deviation of $\bm{\mu}_r$ is below $\sigma_0$ at these privacy weights.
The noise of the sampled latent therefore exceeds the spread of $\bm{\mu}_r$ at these privacy weights, as in Proposition~\ref{prop:s-mask}.
The auxiliary classifier attains chance on the sampled latents of the training batches at $\lambda=1000$.
The log loss of the auxiliary classifier on IJB-C is within 0.005~nats of the log loss of the class prior (Table~\ref{tab:s-logloss}).
The MLP attacker attains 70.4\% balanced accuracy on IJB-C at 46.1\% TMR at FMR~$10^{-3}$ (Table~\ref{tab:s-ablation}), with a log loss below the log loss of the class prior.
The common offset of the trained $\bm{\mu}_r$ is 12.4 times the spread for gender and 30.0 times for ethnicity at $\lambda=1000$, as assumed in Remark~\ref{rem:s-offset}.
The auxiliary classifier of the sample-side variant therefore attains chance through the noise of the sampled latent, while the attribute stays predictable from the transformed template.

These properties are not found in TEMPER without the relational term and the memory.
The log-variances of this model are not clipped to $-4$, and the median standard deviation of $\bm{\mu}_\theta$ is more than twice the median $\bm{\sigma}_\theta$.
The offset of $\bm{\mu}_\theta$ is at most 0.05 times the spread.
The auxiliary classifier of this model achieves a higher accuracy on the training batches than the MLP attacker on IJB-C, at $k=42$ and $\lambda=30$.
The LR attacker achieves a higher balanced accuracy than the MLP attacker at $k=62$ and the same $\lambda$.
The near-chance leakage of this model is therefore not obtained through the noise of the latent.

\subsection{Ethnicity}
\label{app:ssv-ethnicity}

We repeat the measurements for the ethnicity model of the sample-side variant at $\lambda=1000$.
All log-variances of the template branch of this model are clipped to $-4$, and the spread of $\bm{\mu}_r$ is 0.18 times $\sigma_0$ (Table~\ref{tab:s-diag}).
The auxiliary classifier attains chance on the training batches, with an entropy of 1.3857~nats against $\log4\approx1.3863$~nats.
The MLP attacker attains 46.8\% balanced accuracy on RFW, against a chance level of 25\%.
This model is therefore a second case in which the auxiliary classifier attains chance while the attribute stays predictable from the transformed template.

\section{Ablation}
\label{app:ablation}

We report in this appendix the full ablation of TEMPER, summarised in Table~\ref{tab:ablation} of the main paper, and the measurements of the class latent.

\subsection{Full Ablation}
\label{app:ablation-ladder}

To support the conclusions of Section~\ref{sec:res-ablation} at more values of $\lambda$ and for ethnicity, we list the models of Table~\ref{tab:ablation} as a chain from the sample-side variant (Appendix~\ref{app:ssv}) to TEMPER (Table~\ref{tab:s-ablation}).
The first 3 blocks marked + differ from the sample-side variant by the stated elements, and each later block marked + adds the stated element to the block before.
The mark pc in the attacker column denotes the $k$NN attacker with prior-corrected votes.

To test whether a privacy loss on the sampled latent removes the attribute from the transformed template, we train the sample-side variant and the same model with a wider auxiliary classifier with 3 updates per minibatch.
Every usable gender setting leaks at least 73.7\% with the nonlinear encoder of $d_r=480$ dimensions and the normalised mean as the transformed template.
The gender leakage stays above 60\% at every $\lambda$ in the block with the wider auxiliary classifier, and the TMR at FMR~$10^{-3}$ decreases to 11.0\% at $\lambda=100$.
A privacy loss on the sampled latent therefore does not remove the attribute from the transformed template.

To test the nonlinear encoder with the privacy loss on the transformed template, we move the auxiliary classifier and the entropy term of the sample-side variant to the standardised transformed template.
The model with the narrow auxiliary classifier reduces the leakage only together with the TMR at FMR~$10^{-3}$.
This model retains 37.8\% TMR at FMR~$10^{-3}$ with 69.2\% gender leakage at $\lambda=30$, whereas the sample-side variant at $\lambda=1000$ retains 46.1\% TMR at this FMR with 70.4\% leakage (Appendix~\ref{app:ssv-diag}).
The wider auxiliary classifier acts on the transformed template in the block with the nonlinear encoder, and the MLP attacker attains 50.0\% at $\lambda=30$.
The $k$NN attacker with prior-corrected votes attains 64.3\% on the same model at 0.4\% TMR at FMR~$10^{-3}$.
The attribute therefore stays predictable from the transformed template of this model while the MLP attacker is at chance.
The gender leakage stays above 92\% at every $\lambda$ with a linear encoder of $d_r=480$ dimensions.
We observe that a linear encoder retains a usable TMR at FMR~$10^{-3}$ with the privacy loss on the transformed template, and that the leakage of the linear encoder decreases only with the reduction to $k$ dimensions.

To test whether the encoder needs the PCA initialisation, we initialise the linear encoder at random, in TEMPER without the relational term and the memory and in TEMPER.
The TMR at FMR~$10^{-3}$ of the linear model with $k=42$ decreases by about half at $\lambda=3$ without these elements and from a random initialisation.
This model retains 1.7\% TMR at FMR~$10^{-3}$ with 71.8\% leakage at $\lambda=30$.
The same model from the PCA solution retains 77.4\% TMR at FMR~$10^{-3}$ with 51.2\% leakage.
TEMPER from a random initialisation retains a gender TMR at FMR~$10^{-3}$ within 1\% of the TMR at this FMR of TEMPER from the PCA initialisation.
The gender leakage of the randomly initialised TEMPER increases with $\lambda$, from 54.6\% at $\lambda=3$ to 65.8\% at $\lambda=30$ (Table~\ref{tab:ablation}).
The PCA initialisation is therefore necessary with and without the relational term and the memory.

To measure the gain of the relational term, we add the relational term, the memory and the linear head of the auxiliary classifier to the PCA-initialised model.
We also add the relational term alone to the model without the variational autoencoder and the memory.
The relational term, the memory and the linear head raise the gender TMR of the PCA-initialised model at FMR~$10^{-3}$ and $\lambda=30$ from 77.4\% to 81.0\%. The leakage of the 2 models is 51.2\% and 51.7\%.
The linear head of the auxiliary classifier was ablated only together with the relational term and the memory.
The same addition raises the ethnicity TMR at FMR~$10^{-3}$ and equal leakage from 50.8\% at $\lambda=30$ to 72.5\% at $\lambda=10$.
The relational term raises the gender TMR at FMR~$10^{-3}$ and $\lambda=30$ from 70.9\% to 84.0\% in the model without the variational autoencoder and the memory.
The relational term gives the largest gain in TMR at FMR~$10^{-3}$ of the terms added to the PCA solution, for both attributes.
The PCA initialisation itself gives a larger gain in the gender TMR at FMR~$10^{-3}$.
We therefore attribute the improvement of TEMPER over prior methods near chance to the PCA initialisation and the relational term.

To measure the effect of the memory, we add the memory to the models without the variational autoencoder, with and without the relational term.
The memory reduces the gender TMR at FMR~$10^{-3}$ and $\lambda=30$ from 70.9\% to 49.6\% in the model without the variational autoencoder and the relational term. The leakage of the 2 models differs by 0.4\%.
Adding the memory to the relational term changes the gender operating points by less than 3\% without the variational autoencoder.
The same addition raises both the TMR at FMR~$10^{-3}$ and the leakage for ethnicity.
The memory therefore does not raise the TMR at equal leakage.

To measure the effect of the variational autoencoder, we remove the variational autoencoder from TEMPER and from TEMPER without the relational term and the memory.
We train 3 seeds of TEMPER with and without the variational autoencoder.
The variational autoencoder is ablated only as a whole.
The effect of one of its terms is therefore not isolated from the others.
Removing the variational autoencoder raises the gender leakage at $\lambda=30$ from 51.7\% to 52.6\%.
The same removal reduces the ethnicity TMR at FMR~$10^{-3}$ and $\lambda=3$ from 88.4\% to 87.6\%.
The ranges over the 3 seeds of TEMPER and over the 3 seeds without the variational autoencoder are disjoint for both quantities.
All 3 seeds of TEMPER at $\lambda=30$ are near chance for gender, whereas one of the 3 seeds without the variational autoencoder is near chance.
The variational autoencoder changes the operating point of TEMPER by about 1\% at the seeded settings.
The variational autoencoder has larger effects of opposite sign for gender and ethnicity in the model without the relational term and the memory.
Removing the variational autoencoder from this model reduces the gender TMR at FMR~$10^{-3}$ and $\lambda=30$ from 77.4\% to 70.9\%.
The same removal raises the ethnicity TMR at FMR~$10^{-3}$ and $\lambda=3$ from 30.8\% to 41.0\%.
The variational autoencoder therefore adds a smaller gain than the relational term.

\newcommand{\SAblationCaption}{Full ablation of TEMPER. TMR (\%) on IJB-C at FMR $10^{-4}$ and $10^{-3}$ and on RFW at FMR~$10^{-3}$, and leakage (\%) on IJB-C for gender and on RFW for ethnicity.}

\begingroup\footnotesize\setlength{\tabcolsep}{4pt}
\begin{longtable}{@{}lrrrrlrr@{}}
\caption{\SAblationCaption}\label{tab:s-ablation}\\
\toprule
 & \multicolumn{3}{c}{TMR (\%)} & \multicolumn{4}{c}{Leakage (\%)} \\
\cmidrule(lr){2-4}\cmidrule(lr){5-8}
Setting & IJB-C $10^{-4}$ & IJB-C $10^{-3}$ & RFW $10^{-3}$ & Worst & Attacker & MLP & LR \\
\midrule
\endfirsthead
\multicolumn{8}{@{}l}{\footnotesize Table~\ref{tab:s-ablation} (continued).}\\
\toprule
 & \multicolumn{3}{c}{TMR (\%)} & \multicolumn{4}{c}{Leakage (\%)} \\
\cmidrule(lr){2-4}\cmidrule(lr){5-8}
Setting & IJB-C $10^{-4}$ & IJB-C $10^{-3}$ & RFW $10^{-3}$ & Worst & Attacker & MLP & LR \\
\midrule
\endhead
\bottomrule
\endlastfoot
\multicolumn{8}{@{}l}{\textbf{Gender} (chance 50\%)}\\
\multicolumn{8}{@{}p{0.95\textwidth}}{\textit{Sample-side variant: MLP encoder ($d_r{=}480$), auxiliary classifier with 3 hidden layers of 256 units on $\bm{\mu}_r$ (one update), entropy on $\bm{z}_r$, minibatches of 16 identities with 16 images}}\\
\quad $\lambda{=}1$ & 81.7 & 88.2 & 77.1 & 94.5 & MLP & 94.5 & 58.3 \\
\quad $\lambda{=}3$ & 76.3 & 84.9 & 62.3 & 92.5 & MLP & 92.5 & 59.2 \\
\quad $\lambda{=}10$ & 69.3 & 79.3 & 46.9 & 86.4 & MLP & 86.4 & 58.0 \\
\quad $\lambda{=}30$ & 64.8 & 76.7 & 39.2 & 76.9 & MLP & 76.9 & 62.8 \\
\quad $\lambda{=}100$ & 57.3 & 71.3 & 28.0 & 73.9 & MLP & 73.9 & 60.8 \\
\quad $\lambda{=}300$ & 44.4 & 60.8 & 14.5 & 73.7 & MLP & 73.7 & 62.0 \\
\quad $\lambda{=}1000$ & 26.2 & 46.1 & 7.6 & 70.4 & MLP & 70.4 & 58.9 \\
\multicolumn{8}{@{}p{0.95\textwidth}}{\textit{+ wider auxiliary classifier (3 hidden layers of 1024 units, 3 updates), still on $\bm{\mu}_r$ and $\bm{z}_r$}}\\
\quad $\lambda{=}1$ & 75.8 & 84.1 & 69.6 & 91.9 & MLP & 91.9 & 60.5 \\
\quad $\lambda{=}3$ & 75.1 & 84.1 & 67.1 & 87.4 & MLP & 87.4 & 62.1 \\
\quad $\lambda{=}10$ & 64.8 & 76.9 & 38.8 & 81.2 & MLP & 81.2 & 59.0 \\
\quad $\lambda{=}30$ & 50.4 & 66.4 & 19.4 & 82.3 & MLP & 82.3 & 61.2 \\
\quad $\lambda{=}100$ & 2.9 & 11.0 & 1.2 & 71.3 & MLP & 71.3 & 60.7 \\
\quad $\lambda{=}1000$ & 0.0 & 0.0 & 0.0 & 63.4 & MLP & 63.4 & 55.6 \\
\quad $\lambda{=}3000$ & 0.0 & 0.0 & 0.0 & 60.9 & MLP & 60.9 & 55.0 \\
\quad $\lambda{=}10^{4}$ & 0.0 & 0.0 & 0.2 & 62.4 & MLP & 62.4 & 59.5 \\
\multicolumn{8}{@{}p{0.95\textwidth}}{\textit{+ auxiliary classifier and entropy on the standardised transformed template (3 hidden layers of 256 units, one update)}}\\
\quad $\lambda{=}1$ & 90.0 & 93.3 & 65.6 & 91.7 & MLP & 91.7 & 61.2 \\
\quad $\lambda{=}3$ & 87.3 & 91.5 & 56.2 & 89.1 & MLP & 89.1 & 68.8 \\
\quad $\lambda{=}10$ & 43.0 & 58.1 & 11.4 & 83.2 & MLP & 83.2 & 75.2 \\
\quad $\lambda{=}30$ & 21.9 & 37.8 & 4.4 & 69.2 & MLP & 69.2 & 60.4 \\
\quad $\lambda{=}100$ & 10.1 & 21.5 & 2.5 & 59.4 & MLP & 59.4 & 55.1 \\
\quad $\lambda{=}300$ & 4.7 & 12.0 & 2.6 & 58.4 & MLP & 58.4 & 51.9 \\
\quad $\lambda{=}1000$ & 6.4 & 14.2 & 2.1 & 68.8 & MLP & 68.8 & 51.7 \\
\multicolumn{8}{@{}p{0.95\textwidth}}{\textit{+ both (transformed template, 3 hidden layers of 1024 units, 3 updates)}}\\
\quad $\lambda{=}3$ & 9.8 & 20.1 & 6.4 & 61.8 & MLP & 61.8 & 51.2 \\
\quad $\lambda{=}10$ & 0.6 & 7.1 & 3.0 & 61.8 & $k$NN-51\,pc & 59.4 & 51.0 \\
\quad $\lambda{=}30$ & 0.0 & 0.4 & 0.0 & 64.3 & $k$NN-51\,pc & 50.0 & 50.8 \\
\quad $\lambda{=}100$ & 0.0 & 0.4 & 0.0 & 62.5 & $k$NN-51\,pc & 51.0 & 51.9 \\
\multicolumn{8}{@{}p{0.95\textwidth}}{\textit{+ linear encoder ($d_r{=}480$)}}\\
\quad $\lambda{=}3$ & 83.8 & 89.4 & 62.1 & 93.4 & MLP & 93.4 & 64.5 \\
\quad $\lambda{=}10$ & 73.9 & 83.0 & 64.6 & 93.0 & MLP & 93.0 & 61.9 \\
\quad $\lambda{=}30$ & 67.2 & 77.5 & 52.7 & 92.3 & MLP & 92.3 & 60.2 \\
\quad $\lambda{=}100$ & 64.6 & 75.6 & 46.4 & 92.4 & MLP & 92.4 & 60.8 \\
\quad $\lambda{=}300$ & 64.0 & 74.7 & 44.5 & 92.7 & MLP & 92.7 & 60.3 \\
\multicolumn{8}{@{}p{0.95\textwidth}}{\textit{+ linear $k$-dimensional encoder, minibatches of 64 identities with 4 images, random initialisation}}\\
\quad $k{=}22$, $\lambda{=}0$ & 74.4 & 84.6 & 25.1 & 56.9 & SVC & 54.7 & 53.6 \\
\quad $k{=}22$, $\lambda{=}3$ & 1.0 & 5.7 & 1.0 & 63.3 & MLP & 63.3 & 50.9 \\
\quad $k{=}22$, $\lambda{=}30$ & 0.4 & 1.2 & 0.5 & 61.1 & MLP & 61.1 & 52.7 \\
\quad $k{=}22$, $\lambda{=}300$ & 0.5 & 1.2 & 0.4 & 60.1 & MLP & 60.1 & 53.8 \\
\quad $k{=}42$, $\lambda{=}0$ & 89.6 & 93.2 & 66.9 & 66.3 & SVC & 64.5 & 53.3 \\
\quad $k{=}42$, $\lambda{=}3$ & 28.9 & 50.2 & 6.0 & 74.0 & MLP & 74.0 & 50.3 \\
\quad $k{=}42$, $\lambda{=}30$ & 0.7 & 1.7 & 0.5 & 71.8 & MLP & 71.8 & 52.8 \\
\quad $k{=}42$, $\lambda{=}300$ & 0.6 & 2.2 & 0.4 & 72.0 & MLP & 72.0 & 51.3 \\
\quad $k{=}62$, $\lambda{=}0$ & 93.1 & 95.4 & 79.9 & 74.9 & MLP & 74.9 & 54.1 \\
\quad $k{=}62$, $\lambda{=}3$ & 49.7 & 67.2 & 13.9 & 82.4 & MLP & 82.4 & 54.8 \\
\quad $k{=}62$, $\lambda{=}30$ & 0.5 & 2.1 & 0.4 & 79.1 & MLP & 79.1 & 50.7 \\
\quad $k{=}62$, $\lambda{=}300$ & 0.6 & 2.4 & 0.3 & 79.6 & MLP & 79.6 & 55.0 \\
\multicolumn{8}{@{}p{0.95\textwidth}}{\textit{+ PCA initialisation, $k{=}42$ (TEMPER without the relational term and the memory)}}\\
\quad $k{=}42$, $\lambda{=}0$ & 89.4 & 93.2 & 67.7 & 66.6 & SVC & 65.1 & 53.9 \\
\quad $k{=}42$, $\lambda{=}0.3$ & 84.7 & 90.2 & 51.1 & 56.2 & MLP & 56.2 & 51.2 \\
\quad $k{=}42$, $\lambda{=}1$ & 67.7 & 78.5 & 23.8 & 54.8 & MLP & 54.8 & 50.6 \\
\quad $k{=}42$, $\lambda{=}3$ & 65.4 & 77.1 & 20.2 & 56.1 & MLP & 56.1 & 51.2 \\
\quad $k{=}42$, $\lambda{=}30$ & 65.6 & 77.4 & 24.7 & 51.2 & MLP & 51.2 & 50.2 \\
\quad $k{=}42$, $\lambda{=}300$ & 57.9 & 72.0 & 18.5 & 51.8 & LR & 51.7 & 51.8 \\
\multicolumn{8}{@{}p{0.95\textwidth}}{\textit{+ relational term, memory and linear head of the auxiliary classifier (TEMPER), $k{=}42$}}\\
\quad $k{=}42$, $\lambda{=}3$ & 87.4 & 92.1 & 61.2 & 54.5 & MLP & 54.5 & 50.6 \\
\quad $k{=}42$, $\lambda{=}10$ & 81.3 & 88.2 & 43.8 & 52.5 & MLP & 52.5 & 50.2 \\
\quad $k{=}42$, $\lambda{=}30$ & 69.1 & 81.0 & 26.8 & 51.7 & MLP & 51.7 & 50.1 \\
\quad $k{=}42$, $\lambda{=}100$ & 54.5 & 70.8 & 13.8 & 52.0 & LR & 51.5 & 52.0 \\
\quad $k{=}42$, $\lambda{=}300$ & 45.2 & 63.4 & 9.1 & 53.5 & LR & 51.4 & 53.5 \\
\quad $k{=}42$, $\lambda{=}30$, seed 1 & 69.5 & 80.9 & 23.0 & 51.8 & MLP & 51.8 & 51.8 \\
\quad $k{=}42$, $\lambda{=}30$, seed 2 & 69.0 & 80.8 & 24.8 & 52.0 & MLP & 52.0 & 51.8 \\
\multicolumn{8}{@{}p{0.95\textwidth}}{\textit{TEMPER from a random initialisation (relational term and memory kept), $k{=}42$}}\\
\quad $\lambda{=}0$ & 88.2 & 92.4 & 63.1 & 58.9 & SVC & 58.3 & 51.2 \\
\quad $\lambda{=}3$ & 86.5 & 91.3 & 56.9 & 54.6 & MLP & 54.6 & 49.8 \\
\quad $\lambda{=}10$ & 82.1 & 88.7 & 47.1 & 58.8 & MLP & 58.8 & 50.8 \\
\quad $\lambda{=}30$ & 71.4 & 81.9 & 26.5 & 65.8 & MLP & 65.8 & 49.5 \\
\multicolumn{8}{@{}p{0.95\textwidth}}{\textit{TEMPER without the variational autoencoder (no class encoder, class prior, KL terms or decoder), $k{=}42$}}\\
\quad $k{=}42$, $\lambda{=}0$ & 87.7 & 92.3 & 61.2 & 62.3 & SVC & 60.1 & 53.9 \\
\quad $k{=}42$, $\lambda{=}1$ & 87.6 & 92.2 & 60.2 & 56.7 & SVC & 56.3 & 52.5 \\
\quad $k{=}42$, $\lambda{=}3$ & 87.3 & 92.0 & 58.1 & 54.5 & MLP & 54.5 & 50.8 \\
\quad $k{=}42$, $\lambda{=}30$ & 69.2 & 81.1 & 26.8 & 52.6 & LR & 51.3 & 52.6 \\
\quad $k{=}42$, $\lambda{=}300$ & 42.4 & 61.3 & 10.0 & 53.9 & LR & 51.8 & 53.9 \\
\quad $k{=}42$, $\lambda{=}30$, seed 1 & 68.2 & 80.3 & 27.2 & 52.3 & MLP & 52.3 & 51.4 \\
\quad $k{=}42$, $\lambda{=}30$, seed 2 & 69.7 & 81.3 & 25.7 & 52.7 & MLP & 52.7 & 52.3 \\
\multicolumn{8}{@{}p{0.95\textwidth}}{\textit{TEMPER without the variational autoencoder and the relational term (memory only), $k{=}42$}}\\
\quad $k{=}42$, $\lambda{=}0$ & 87.9 & 92.0 & 61.2 & 70.3 & SVC & 67.7 & 54.0 \\
\quad $k{=}42$, $\lambda{=}1$ & 31.8 & 50.5 & 7.8 & 53.2 & LR & 51.3 & 53.2 \\
\quad $k{=}42$, $\lambda{=}3$ & 30.7 & 49.9 & 7.6 & 53.5 & LR & 51.4 & 53.5 \\
\quad $k{=}42$, $\lambda{=}30$ & 31.1 & 49.6 & 7.3 & 53.2 & LR & 51.1 & 53.2 \\
\quad $k{=}42$, $\lambda{=}300$ & 29.2 & 48.3 & 7.0 & 53.7 & LR & 52.2 & 53.7 \\
\multicolumn{8}{@{}p{0.95\textwidth}}{\textit{TEMPER without the variational autoencoder and the memory (relational term only), $k{=}42$}}\\
\quad $k{=}42$, $\lambda{=}3$ & 86.9 & 91.7 & 57.9 & 54.1 & MLP & 54.1 & 50.9 \\
\quad $k{=}42$, $\lambda{=}30$ & 74.1 & 84.0 & 34.5 & 52.4 & MLP & 52.4 & 51.7 \\
\multicolumn{8}{@{}p{0.95\textwidth}}{\textit{TEMPER without the variational autoencoder, the relational term and the memory, $k{=}42$}}\\
\quad $k{=}42$, $\lambda{=}3$ & 57.4 & 71.6 & 18.1 & 53.0 & LR & 52.2 & 53.0 \\
\quad $k{=}42$, $\lambda{=}30$ & 56.7 & 70.9 & 18.7 & 53.6 & LR & 52.0 & 53.6 \\
\midrule
\multicolumn{8}{@{}l}{\textbf{Ethnicity} (chance 25\%)}\\
\multicolumn{8}{@{}p{0.95\textwidth}}{\textit{Sample-side variant: MLP encoder ($d_r{=}480$), auxiliary classifier with 3 hidden layers of 256 units on $\bm{\mu}_r$ (one update), entropy on $\bm{z}_r$, minibatches of 16 identities with 16 images}}\\
\quad $\lambda{=}1$ & 80.8 & 87.7 & 91.1 & 87.4 & MLP & 87.4 & 37.3 \\
\quad $\lambda{=}3$ & 69.2 & 79.3 & 64.8 & 78.2 & MLP & 78.2 & 34.6 \\
\quad $\lambda{=}10$ & 63.0 & 74.6 & 44.5 & 73.7 & MLP & 73.7 & 32.1 \\
\quad $\lambda{=}30$ & 59.3 & 72.5 & 34.1 & 61.4 & MLP & 61.4 & 35.1 \\
\quad $\lambda{=}100$ & 57.3 & 70.5 & 28.3 & 53.4 & MLP & 53.4 & 39.2 \\
\quad $\lambda{=}300$ & 40.0 & 56.8 & 12.0 & 47.8 & MLP & 47.8 & 41.1 \\
\quad $\lambda{=}1000$ & 21.3 & 41.9 & 7.1 & 46.8 & MLP & 46.8 & 43.5 \\
\multicolumn{8}{@{}p{0.95\textwidth}}{\textit{+ auxiliary classifier and entropy on the standardised transformed template (3 hidden layers of 256 units, one update)}}\\
\quad $\lambda{=}1$ & 63.5 & 74.8 & 20.4 & 85.3 & MLP & 85.3 & 37.3 \\
\quad $\lambda{=}3$ & 51.9 & 65.0 & 12.7 & 85.7 & MLP & 85.7 & 36.6 \\
\quad $\lambda{=}10$ & 50.6 & 64.3 & 13.8 & 74.5 & MLP & 74.5 & 42.5 \\
\quad $\lambda{=}30$ & 13.8 & 23.3 & 4.7 & 44.9 & MLP & 44.9 & 33.7 \\
\quad $\lambda{=}100$ & 9.6 & 18.9 & 3.9 & 36.0 & MLP & 36.0 & 29.5 \\
\quad $\lambda{=}300$ & 2.3 & 5.2 & 1.5 & 33.0 & MLP & 33.0 & 29.8 \\
\quad $\lambda{=}1000$ & 0.6 & 2.1 & 0.7 & 29.0 & MLP & 29.0 & 28.5 \\
\multicolumn{8}{@{}p{0.95\textwidth}}{\textit{+ PCA initialisation, $k{=}42$ (TEMPER without the relational term and the memory)}}\\
\quad $k{=}42$, $\lambda{=}0$ & 89.9 & 93.3 & 69.8 & 50.8 & SVC & 50.3 & 28.3 \\
\quad $k{=}42$, $\lambda{=}0.1$ & 87.2 & 91.8 & 59.6 & 34.6 & SVC & 33.8 & 25.1 \\
\quad $k{=}42$, $\lambda{=}0.3$ & 19.9 & 35.1 & 3.8 & 29.3 & MLP & 29.3 & 26.7 \\
\quad $k{=}42$, $\lambda{=}1$ & 10.0 & 23.2 & 2.4 & 33.0 & MLP & 33.0 & 27.5 \\
\quad $k{=}42$, $\lambda{=}3$ & 14.2 & 30.8 & 4.3 & 30.4 & MLP & 30.4 & 26.8 \\
\quad $k{=}42$, $\lambda{=}30$ & 31.2 & 50.8 & 5.9 & 26.4 & MLP & 26.4 & 26.0 \\
\quad $k{=}42$, $\lambda{=}300$ & 21.1 & 40.3 & 3.7 & 26.0 & SVC & 25.7 & 23.4 \\
\multicolumn{8}{@{}p{0.95\textwidth}}{\textit{+ relational term, memory and linear head of the auxiliary classifier (TEMPER), $k{=}42$}}\\
\quad $k{=}42$, $\lambda{=}1$ & 88.1 & 92.6 & 67.3 & 31.5 & SVC & 30.0 & 25.9 \\
\quad $k{=}42$, $\lambda{=}3$ & 81.5 & 88.4 & 43.3 & 28.6 & SVC & 28.2 & 26.7 \\
\quad $k{=}42$, $\lambda{=}10$ & 57.1 & 72.5 & 17.6 & 26.4 & MLP & 26.4 & 25.3 \\
\quad $k{=}42$, $\lambda{=}30$ & 39.3 & 59.6 & 8.9 & 26.1 & LR & 25.6 & 26.1 \\
\quad $k{=}42$, $\lambda{=}3$, seed 1 & 80.9 & 88.1 & 42.9 & 28.2 & SVC & 27.6 & 26.8 \\
\quad $k{=}42$, $\lambda{=}3$, seed 2 & 81.6 & 88.5 & 49.5 & 29.1 & SVC & 28.3 & 26.0 \\
\multicolumn{8}{@{}p{0.95\textwidth}}{\textit{TEMPER from a random initialisation (relational term and memory kept), $k{=}42$}}\\
\quad $\lambda{=}0$ & 88.9 & 92.9 & 68.4 & 35.5 & SVC & 33.2 & 27.2 \\
\quad $\lambda{=}1$ & 85.9 & 90.9 & 60.5 & 31.0 & SVC & 29.6 & 25.2 \\
\quad $\lambda{=}3$ & 76.7 & 85.2 & 39.5 & 28.8 & SVC & 28.8 & 25.2 \\
\quad $\lambda{=}10$ & 60.6 & 74.9 & 18.6 & 30.6 & MLP & 30.6 & 26.0 \\
\multicolumn{8}{@{}p{0.95\textwidth}}{\textit{TEMPER without the variational autoencoder (no class encoder, class prior, KL terms or decoder), $k{=}42$}}\\
\quad $k{=}42$, $\lambda{=}0$ & 87.8 & 92.5 & 64.6 & 35.7 & SVC & 32.8 & 27.4 \\
\quad $k{=}42$, $\lambda{=}1$ & 87.5 & 92.0 & 65.2 & 31.7 & SVC & 29.5 & 25.8 \\
\quad $k{=}42$, $\lambda{=}3$ & 80.2 & 87.6 & 46.2 & 28.8 & SVC & 27.7 & 26.7 \\
\quad $k{=}42$, $\lambda{=}30$ & 39.4 & 59.6 & 7.9 & 25.9 & LR & 25.8 & 25.9 \\
\quad $k{=}42$, $\lambda{=}300$ & 14.4 & 33.3 & 2.7 & 25.6 & LR & 25.4 & 25.6 \\
\quad $k{=}42$, $\lambda{=}3$, seed 1 & 80.3 & 87.5 & 47.3 & 28.3 & SVC & 27.4 & 25.2 \\
\quad $k{=}42$, $\lambda{=}3$, seed 2 & 80.3 & 87.6 & 47.4 & 28.1 & SVC & 27.7 & 25.6 \\
\multicolumn{8}{@{}p{0.95\textwidth}}{\textit{TEMPER without the variational autoencoder and the memory (relational term only), $k{=}42$}}\\
\quad $k{=}42$, $\lambda{=}1$ & 86.4 & 91.5 & 62.6 & 29.2 & SVC & 28.5 & 24.9 \\
\quad $k{=}42$, $\lambda{=}3$ & 74.0 & 83.9 & 37.1 & 27.2 & SVC & 27.0 & 26.2 \\
\multicolumn{8}{@{}p{0.95\textwidth}}{\textit{TEMPER without the variational autoencoder, the relational term and the memory, $k{=}42$}}\\
\quad $k{=}42$, $\lambda{=}1$ & 22.9 & 42.5 & 4.2 & 26.0 & LR & 26.0 & 26.0 \\
\quad $k{=}42$, $\lambda{=}3$ & 21.3 & 41.0 & 3.4 & 25.9 & SVC & 25.4 & 25.3 \\
\end{longtable}
\endgroup

\subsection{The Class Latent}
\label{app:ablation-zc}

To test whether the attribute is encoded in the class branch, we measure the prior means and the class mean of the models with a class branch (Table~\ref{tab:s-zc}).
The objective has no penalty on coinciding prior means $\bm{e}_c$.
The minimum of the single term $\mathcal{L}_{\mathrm{KL}}^{c}$ is attained both by coinciding prior means and by distinct prior means matched by the posterior means of the class encoder.
Table~\ref{tab:s-zc} lists, for models with a class branch, the distances between the learnt prior means and the KL term of the last epoch.
Table~\ref{tab:s-zc} also lists the worst-case balanced accuracy of the MLP and LR attackers on the class mean $\bm{\mu}_\vartheta$, tested on IJB-C for gender and on RFW for ethnicity.
The last 2 columns list the IJB-C TMR at FMR~$10^{-3}$ when the template is $\bm{\mu}_\vartheta$ or the normalised concatenation of the transformed template and $\bm{\mu}_\vartheta$.

The prior means are initialised from $\mathcal{N}(\bm{0},\bm{I})$ in 32 dimensions, where 2 independent draws have an expected squared distance of 64.
The learnt distances of 6.1 to 15.4 are of the order of the distances at initialisation.
The prior means therefore stay distinct in every model of TEMPER, with or without the relational term and the memory.

The attackers predict the attribute from the class mean of these models with 86 to 96\% balanced accuracy.
The TMR at FMR~$10^{-3}$ with the class mean as the template is 2 to 4\% for these models.
Concatenating the class mean with the transformed template reduces the TMR at FMR~$10^{-3}$ in 11 of the 14 models (Table~\ref{tab:s-zc}).
The concatenation raises the TMR at FMR~$10^{-3}$ only for TEMPER without the relational term and the memory at $\lambda\ge30$, by at most 6.1\%.
The attribute is therefore encoded in the class branch of TEMPER, whereas the class mean does not give a usable TMR at FMR~$10^{-3}$.

The ethnicity model of the sample-side variant at $\lambda=0$ is the only model in Table~\ref{tab:s-zc} with coinciding prior means and a KL term near zero.
The prior means of this model are within 0.05 of each other.
The attackers predict the attribute from the class mean of this model with 86.9\% balanced accuracy.
Identity is also encoded in the class branch, since the TMR at FMR~$10^{-3}$ with the class mean as the template is 34.0\% for this model.
Coinciding prior means therefore do not imply an uninformative class mean.
The accuracy of the attackers on $\bm{\mu}_\vartheta$ is a more direct measure of the class latent than the distances between the prior means.

\begin{table}[!t]
\centering
\caption{The class latent. Euclidean distances between the learnt prior means (range over class pairs), $\mathcal{L}_{\mathrm{KL}}^{c}$ in the last epoch (nats), worst-case balanced accuracy (\%) of the MLP and LR attackers on $\bm{\mu}_\vartheta$, and IJB-C TMR (\%) at FMR~$10^{-3}$ from $\bm{\mu}_\vartheta$ and from the concatenation $[\bm{r};\bm{\mu}_\vartheta]$.}
\label{tab:s-zc}
\footnotesize
\setlength{\tabcolsep}{4pt}
\begin{tabular}{@{}lrrrrr@{}}
\toprule
 & & & Attribute & \multicolumn{2}{c}{Verification from} \\
\cmidrule(lr){5-6}
Model & Distances & KL & from $\bm{\mu}_\vartheta$ & $\bm{\mu}_\vartheta$ & $[\bm{r};\bm{\mu}_\vartheta]$ \\
\midrule
\multicolumn{6}{@{}l}{\textit{TEMPER}}\\
\quad gender, $k{=}42$, $\lambda{=}3$ & 6.3 & 0.029 & 95.4 & 3.0 & 89.7 \\
\quad gender, $k{=}42$, $\lambda{=}30$ & 6.1 & 0.029 & 95.6 & 2.2 & 80.6 \\
\quad ethnicity, $k{=}42$, $\lambda{=}3$ & 8.7--13.3 & 0.063 & 86.3 & 1.6 & 81.1 \\
\quad ethnicity, $k{=}42$, $\lambda{=}10$ & 8.4--13.5 & 0.106 & 86.7 & 1.9 & 71.1 \\
\multicolumn{6}{@{}l}{\textit{TEMPER without the relational term and the memory}}\\
\quad gender, $\lambda{=}0$ & 6.8 & 0.033 & 95.3 & 3.7 & 91.3 \\
\quad gender, $\lambda{=}30$ & 6.1 & 0.029 & 95.4 & 3.2 & 78.7 \\
\quad gender, $\lambda{=}300$ & 6.1 & 0.029 & 95.2 & 3.1 & 73.8 \\
\quad ethnicity, $\lambda{=}0$ & 8.2--15.4 & 0.060 & 87.1 & 1.8 & 90.3 \\
\quad ethnicity, $\lambda{=}0.1$ & 8.7--14.5 & 0.063 & 87.1 & 1.6 & 88.0 \\
\quad ethnicity, $\lambda{=}30$ & 7.1--12.5 & 0.299 & 86.9 & 3.9 & 56.9 \\
\multicolumn{6}{@{}l}{\textit{Sample-side variant}}\\
\quad gender, $\lambda{=}0$ & 6.7 & 0.030 & 95.3 & 2.1 & 92.9 \\
\quad gender, $\lambda{=}1000$ & 5.4 & 0.144 & 95.6 & 6.5 & 41.3 \\
\quad ethnicity, $\lambda{=}0$ & 0.0--0.0 & 0.000 & 86.9 & 34.0 & 94.6 \\
\quad ethnicity, $\lambda{=}1000$ & 4.7--9.8 & 0.463 & 88.4 & 3.5 & 15.5 \\
\bottomrule
\end{tabular}

\end{table}

\section{Bias and Fairness Analysis}
\label{app:fairness}

We report the absolute error rates of the gender groups behind Table~\ref{tab:fairness} of the main paper, and the demographic differentials on RFW at the equal error rate and on IJB-C and RFW at FMR~$10^{-2}$ and $10^{-1}$.
We describe error rates only at these operating points and do not make claims about recognition bias in general.

The FMR and FNMR of the 2 gender groups of IJB-C at one global threshold are listed in Table~\ref{tab:s-fairness}.
The threshold is the evaluated threshold whose FMR over all non-mated comparisons is nearest to $10^{-3}$ on a logarithmic scale.
The global FMR at this threshold is between 0.87 and 1.08 per thousand for the 3 settings of the table.
The FMR of a group is computed over the non-mated comparisons within the group, and the global FMR is computed over every non-mated comparison.
Table~\ref{tab:fairness} lists the FMR of each group relative to the geometric mean of the group FMRs.
The same table also lists the Gini index of ISO/IEC 19795-10, $\frac{n}{n-1}\sum_{i,j}|x_i-x_j|/(2n^2\bar{x})$ over the rates $x_1,\dots,x_n$ of the $n$ groups with mean $\bar{x}$.
The settings are those of Table~\ref{tab:matched} at a gender leakage of at most 60\%, with TEMPER at $k=62$, $\lambda=3$ and PCA truncation at $k=22$.

The female FMR without removal is almost twice the male FMR, and the female FNMR is also almost twice the male FNMR.
The FMRs of the 2 groups are close at the selected TEMPER setting, with 0.97 and 0.89 per thousand for the female and male groups.
The female FNMR stays above the male FNMR, at a larger overall FNMR than without removal.
The FMRs of the 2 groups are also close for PCA truncation at $k=22$, at a larger overall FNMR than for TEMPER.
A smaller FMR gap at a larger overall FNMR has to be read together with this loss of verification accuracy.

Table~\ref{tab:fairness} lists the ethnicity differentials on IJB-C, with our annotation of the IJB-C subjects.
We do not report the RFW ethnicity groups at FMR~$10^{-3}$, where each group FMR is based on fewer than 10 false matches among 3{,}000 non-mated pairs.
We report the RFW groups at the equal error rate, the threshold at which the FMR and the FNMR over all 24{,}000 RFW pairs are equal, at the settings of Table~\ref{tab:fairness} (Table~\ref{tab:s-fairness-rfw}).
The smallest group count at this threshold is 3 false matches, for the Caucasian pairs without removal.
The FMRs of the female and male pairs without removal are closer than the FMRs of the ethnicity groups, with a Gini index of 0.05 against 0.41, and the Caucasian pairs have the lowest FMR, as on IJB-C.
Every method that reduces the Gini index of the ethnicity FMRs raises the EER.
TEMPER reduces the index from 0.41 to 0.12.
The EER of this setting is 3.3\%, against 1.1\% without removal.
PE-MIU reduces the index to 0.19 at a lower EER, and FaRM to 0.04 at an EER above 16\%.
The index of the gender FMRs stays within 0.06 for every method except PFRNet and LEACE.
The removal methods differ in the EER for gender, which is lowest for TEMPER at 4.9\% among the methods not marked $^{\dagger}$.
The Gini index of the group FNMRs of TEMPER is higher than without removal for ethnicity, 0.18 against 0.15, and lower for gender.

\begin{table}[!t]
\centering
\caption{Demographic differentials on RFW at the equal error rate, at the settings of Table~\ref{tab:fairness}. EER (\%) over all RFW pairs, the FMR of each group relative to the geometric mean of the group FMRs, and the Gini indices of the group FMRs and FNMRs. $^{\dagger}$ marks a method shown at its lowest leakage with at least 1\% TMR at FMR~$10^{-3}$, as in Table~\ref{tab:fairness}. Bold marks the lowest EER among the removal methods without $^{\dagger}$ and, among those below 50\% EER, the lowest Gini index and the relative FMRs of the method whose largest deviation from 1 is smallest.}
\label{tab:s-fairness-rfw}
\fontsize{6}{7}\selectfont
\setlength{\tabcolsep}{1.2pt}
\begin{tabular}{@{}cl*{15}{c}@{}}
\toprule
 & & No removal & PFRNet & DVPF (paper) & DVPF (code) & SensitiveNets & SensitiveNets (Fig.~1 detector) & INLP & IVE & Multi-IVE & PE-MIU & PASS & FaRM & LEACE & PCA + LEACE & TEMPER (ours) \\
\midrule
\multirow{6}{*}{\rotatebox{90}{Gender}}
 & Leakage (\%) & 96.0 & 92.9$^{\dagger}$ & 80.1$^{\dagger}$ & 56.0 & 94.7$^{\dagger}$ & 93.6$^{\dagger}$ & 95.8$^{\dagger}$ & 59.2 & 56.6 & 54.9 & 91.0$^{\dagger}$ & 53.3 & 96.0$^{\dagger}$ & 57.9 & 57.0 \\
 & EER (\%) & 1.1 & 2.2 & 36.1 & 36.0 & 4.5 & 7.0 & 1.2 & 13.7 & 17.0 & 9.1 & 9.1 & 17.1 & 1.1 & 13.5 & \textbf{4.9} \\
 & Rel.\ FMR, female & 0.95 & 0.77 & 0.99 & 0.97 & 1.06 & 0.95 & 1.01 & \textbf{1.01} & 0.98 & 1.05 & 1.00 & 0.94 & 0.90 & 0.94 & 0.94 \\
 & Rel.\ FMR, male & 1.05 & 1.30 & 1.01 & 1.04 & 0.94 & 1.06 & 0.99 & \textbf{0.99} & 1.02 & 0.95 & 1.00 & 1.07 & 1.11 & 1.06 & 1.06 \\
 & Gini, FMR & 0.05 & 0.25 & 0.01 & 0.04 & 0.06 & 0.06 & 0.01 & \textbf{0.01} & 0.02 & 0.05 & 0.00 & 0.06 & 0.11 & 0.06 & 0.06 \\
 & Gini, FNMR & 0.36 & 0.33 & 0.00 & \textbf{0.03} & 0.24 & 0.22 & 0.31 & 0.16 & 0.17 & 0.22 & 0.22 & 0.12 & 0.35 & 0.16 & 0.25 \\
\midrule\midrule
\multirow{8}{*}{\rotatebox{90}{Ethnicity}}
 & Leakage (\%) & 90.0 & 88.5$^{\dagger}$ & 36.3 & 35.7 & 85.7$^{\dagger}$ & 49.3$^{\dagger}$ & 88.4$^{\dagger}$ & 35.4 & 33.0 & 38.7 & 71.9$^{\dagger}$ & 34.5 & 89.5$^{\dagger}$ & 33.4 & 36.2 \\
 & EER (\%) & 1.1 & 2.8 & 36.1 & 27.8 & 5.1 & 20.0 & 1.5 & 14.0 & 13.3 & \textbf{2.7} & 7.7 & 16.7 & 1.1 & 14.4 & 3.3 \\
 & Rel.\ FMR, African & 2.60 & 1.73 & \textbf{0.96} & 1.03 & 1.53 & 1.43 & 2.07 & 1.09 & 1.15 & 1.33 & 1.22 & 1.07 & 2.82 & 1.03 & 1.20 \\
 & Rel.\ FMR, Asian & 1.63 & 1.37 & \textbf{0.97} & 1.07 & 1.27 & 0.96 & 1.52 & 1.10 & 0.89 & 1.12 & 1.12 & 0.94 & 1.56 & 0.94 & 1.07 \\
 & Rel.\ FMR, Caucasian & 0.14 & 0.32 & \textbf{1.04} & 1.00 & 0.25 & 0.52 & 0.21 & 0.81 & 0.97 & 0.56 & 0.52 & 0.98 & 0.14 & 1.06 & 0.72 \\
 & Rel.\ FMR, Indian & 1.72 & 1.31 & \textbf{1.03} & 0.91 & 2.10 & 1.41 & 1.52 & 1.02 & 1.01 & 1.20 & 1.40 & 1.01 & 1.65 & 0.97 & 1.07 \\
 & Gini, FMR & 0.41 & 0.30 & \textbf{0.02} & 0.04 & 0.38 & 0.25 & 0.35 & 0.08 & 0.07 & 0.19 & 0.21 & 0.04 & 0.44 & 0.03 & 0.12 \\
 & Gini, FNMR & 0.15 & 0.11 & 0.09 & \textbf{0.03} & 0.15 & 0.11 & 0.18 & 0.06 & 0.12 & 0.12 & 0.09 & 0.08 & 0.16 & 0.12 & 0.18 \\
\bottomrule
\end{tabular}

\end{table}

Tables~\ref{tab:s-fairness-rfw-fmr2} and~\ref{tab:s-fairness-rfw-fmr1} repeat the RFW comparison at global FMR~$10^{-2}$ and $10^{-1}$, at the threshold at which the FMR over all RFW pairs equals the target.
Tables~\ref{tab:s-fairness-ijbc-fmr2} and~\ref{tab:s-fairness-ijbc-fmr1} repeat the IJB-C comparison of Table~\ref{tab:fairness} at the same two operating points, at the evaluated threshold nearest to the target.
On IJB-C, TEMPER has the lowest FNMR among the methods not marked $^{\dagger}$ at both operating points and for both attributes, and reduces the Gini index of the group FMRs at every operating point, from 0.18 to 0.02 for gender and from 0.20 to 0.08 for ethnicity at FMR~$10^{-2}$.
The FMRs of the female and male pairs on RFW differ little without removal at these operating points, with a Gini index of at most 0.05.
Every method not marked $^{\dagger}$ raises the FNMR at these operating points, and the index of the gender FMRs increases to 0.10 for TEMPER at FMR~$10^{-2}$.
The index of the ethnicity FMRs on RFW decreases with every method not marked $^{\dagger}$, as at the equal error rate.

\begin{table}[!t]
\centering
\caption{Demographic differentials on RFW at global FMR~$10^{-2}$, at the settings of Table~\ref{tab:fairness}. FMR and FNMR (\%) over all RFW pairs at the threshold at which the FMR equals the target, the FMR of each group relative to the geometric mean of the group FMRs, and the Gini indices of the group FMRs and FNMRs; $^{\dagger}$ and bold as in Table~\ref{tab:s-fairness-rfw}, with the lowest FNMR in place of the lowest EER.}
\label{tab:s-fairness-rfw-fmr2}
\fontsize{6}{7}\selectfont
\setlength{\tabcolsep}{1.2pt}
\begin{tabular}{@{}cl*{15}{c}@{}}
\toprule
 & & No removal & PFRNet & DVPF (paper) & DVPF (code) & SensitiveNets & SensitiveNets (Fig.~1 detector) & INLP & IVE & Multi-IVE & PE-MIU & PASS & FaRM & LEACE & PCA + LEACE & TEMPER (ours) \\
\midrule
\multirow{7}{*}{\rotatebox{90}{Gender}}
 & Leakage (\%) & 96.0 & 92.9$^{\dagger}$ & 80.1$^{\dagger}$ & 56.0 & 94.7$^{\dagger}$ & 93.6$^{\dagger}$ & 95.8$^{\dagger}$ & 59.2 & 56.6 & 54.9 & 91.0$^{\dagger}$ & 53.3 & 96.0$^{\dagger}$ & 57.9 & 57.0 \\
 & FMR (\%) & 1.0 & 1.0 & 1.0 & 1.0 & 1.0 & 1.0 & 1.0 & 1.0 & 1.0 & 1.0 & 1.0 & 1.0 & 1.0 & 1.0 & 1.0 \\
 & FNMR (\%) & 1.2 & 3.4 & 94.0 & 94.4 & 11.7 & 20.9 & 1.3 & 51.3 & 60.2 & 21.9 & 28.6 & 56.7 & 1.2 & 47.9 & \textbf{12.3} \\
 & Rel.\ FMR, female & 0.95 & 0.60 & 0.68 & 0.80 & 1.03 & 0.96 & 0.98 & 1.01 & 1.07 & \textbf{1.09} & 1.15 & 0.82 & 0.91 & 0.84 & 0.90 \\
 & Rel.\ FMR, male & 1.05 & 1.66 & 1.47 & 1.25 & 0.97 & 1.04 & 1.02 & 0.99 & 0.93 & \textbf{0.92} & 0.87 & 1.22 & 1.09 & 1.19 & 1.11 \\
 & Gini, FMR & 0.05 & 0.47 & 0.36 & 0.22 & 0.03 & 0.04 & 0.02 & 0.01 & 0.07 & \textbf{0.09} & 0.14 & 0.19 & 0.09 & 0.17 & 0.10 \\
 & Gini, FNMR & 0.38 & 0.34 & 0.00 & 0.00 & 0.20 & 0.16 & 0.31 & 0.08 & 0.07 & 0.17 & 0.15 & 0.07 & 0.36 & \textbf{0.09} & 0.19 \\
\midrule\midrule
\multirow{9}{*}{\rotatebox{90}{Ethnicity}}
 & Leakage (\%) & 90.0 & 88.5$^{\dagger}$ & 36.3 & 35.7 & 85.7$^{\dagger}$ & 49.3$^{\dagger}$ & 88.4$^{\dagger}$ & 35.4 & 33.0 & 38.7 & 71.9$^{\dagger}$ & 34.5 & 89.5$^{\dagger}$ & 33.4 & 36.2 \\
 & FMR (\%) & 1.0 & 1.0 & 1.0 & 1.0 & 1.0 & 1.0 & 1.0 & 1.0 & 1.0 & 1.0 & 1.0 & 1.0 & 1.0 & 1.0 & 1.0 \\
 & FNMR (\%) & 1.2 & 4.9 & 94.2 & 87.1 & 13.8 & 72.7 & 1.9 & 50.7 & 46.4 & \textbf{3.8} & 20.1 & 57.8 & 1.2 & 53.2 & 6.3 \\
 & Rel.\ FMR, African & 2.71 & 2.00 & 1.01 & 0.97 & 1.53 & 2.78 & 2.07 & 1.16 & \textbf{1.16} & 1.28 & 1.43 & 1.03 & 2.46 & 0.75 & \textbf{1.15} \\
 & Rel.\ FMR, Asian & 1.83 & 1.27 & 0.83 & 1.13 & 0.93 & 0.84 & 1.38 & 1.29 & \textbf{0.88} & 1.27 & 1.01 & 0.89 & 1.61 & 0.94 & \textbf{1.01} \\
 & Rel.\ FMR, Caucasian & 0.11 & 0.32 & 1.00 & 1.13 & 0.30 & 0.23 & 0.25 & 0.68 & \textbf{1.32} & 0.50 & 0.41 & 1.13 & 0.15 & 1.08 & \textbf{0.74} \\
 & Rel.\ FMR, Indian & 1.88 & 1.23 & 1.20 & 0.80 & 2.37 & 1.83 & 1.38 & 0.98 & \textbf{0.74} & 1.24 & 1.68 & 0.96 & 1.71 & 1.31 & \textbf{1.17} \\
 & Gini, FMR & 0.40 & 0.35 & 0.09 & 0.10 & 0.44 & 0.51 & 0.36 & 0.16 & 0.16 & 0.19 & 0.31 & 0.06 & 0.40 & 0.15 & \textbf{0.12} \\
 & Gini, FNMR & 0.17 & 0.13 & 0.01 & 0.01 & 0.10 & 0.03 & 0.20 & 0.02 & \textbf{0.05} & 0.12 & 0.09 & 0.04 & 0.16 & 0.06 & 0.18 \\
\bottomrule
\end{tabular}

\end{table}

\begin{table}[!t]
\centering
\caption{Demographic differentials on RFW at global FMR~$10^{-1}$, as in Table~\ref{tab:s-fairness-rfw-fmr2}.}
\label{tab:s-fairness-rfw-fmr1}
\fontsize{6}{7}\selectfont
\setlength{\tabcolsep}{1.2pt}
\begin{tabular}{@{}cl*{15}{c}@{}}
\toprule
 & & No removal & PFRNet & DVPF (paper) & DVPF (code) & SensitiveNets & SensitiveNets (Fig.~1 detector) & INLP & IVE & Multi-IVE & PE-MIU & PASS & FaRM & LEACE & PCA + LEACE & TEMPER (ours) \\
\midrule
\multirow{7}{*}{\rotatebox{90}{Gender}}
 & Leakage (\%) & 96.0 & 92.9$^{\dagger}$ & 80.1$^{\dagger}$ & 56.0 & 94.7$^{\dagger}$ & 93.6$^{\dagger}$ & 95.8$^{\dagger}$ & 59.2 & 56.6 & 54.9 & 91.0$^{\dagger}$ & 53.3 & 96.0$^{\dagger}$ & 57.9 & 57.0 \\
 & FMR (\%) & 10.0 & 10.0 & 10.0 & 10.0 & 10.0 & 10.0 & 10.0 & 10.0 & 10.0 & 10.0 & 10.0 & 10.0 & 10.0 & 10.0 & 10.0 \\
 & FNMR (\%) & 0.3 & 0.7 & 70.5 & 71.6 & 2.3 & 5.0 & 0.3 & 17.7 & 24.3 & 8.7 & 8.4 & 24.6 & 0.3 & 16.7 & \textbf{2.8} \\
 & Rel.\ FMR, female & 0.99 & 0.89 & 0.98 & 0.90 & 1.04 & 1.01 & 1.00 & 1.01 & 0.96 & 1.05 & 1.01 & 0.92 & 1.00 & 0.96 & \textbf{1.00} \\
 & Rel.\ FMR, male & 1.01 & 1.12 & 1.02 & 1.11 & 0.96 & 0.99 & 1.00 & 0.99 & 1.04 & 0.95 & 0.99 & 1.09 & 1.00 & 1.05 & \textbf{1.00} \\
 & Gini, FMR & 0.01 & 0.11 & 0.02 & 0.10 & 0.04 & 0.01 & 0.00 & 0.01 & 0.04 & 0.05 & 0.01 & 0.08 & 0.00 & 0.05 & \textbf{0.00} \\
 & Gini, FNMR & 0.54 & 0.42 & 0.01 & 0.01 & 0.28 & 0.23 & 0.46 & 0.15 & 0.16 & 0.22 & 0.22 & \textbf{0.11} & 0.62 & 0.15 & 0.26 \\
\midrule\midrule
\multirow{9}{*}{\rotatebox{90}{Ethnicity}}
 & Leakage (\%) & 90.0 & 88.5$^{\dagger}$ & 36.3 & 35.7 & 85.7$^{\dagger}$ & 49.3$^{\dagger}$ & 88.4$^{\dagger}$ & 35.4 & 33.0 & 38.7 & 71.9$^{\dagger}$ & 34.5 & 89.5$^{\dagger}$ & 33.4 & 36.2 \\
 & FMR (\%) & 10.0 & 10.0 & 10.0 & 10.0 & 10.0 & 10.0 & 10.0 & 10.0 & 10.0 & 10.0 & 10.0 & 10.0 & 10.0 & 10.0 & 10.0 \\
 & FNMR (\%) & 0.3 & 1.3 & 71.3 & 55.1 & 2.6 & 33.8 & 0.5 & 18.4 & 16.4 & \textbf{1.3} & 6.0 & 23.3 & 0.3 & 19.1 & 1.5 \\
 & Rel.\ FMR, African & 1.66 & 1.56 & 0.99 & 1.01 & 1.43 & 1.66 & 1.55 & 1.08 & 1.17 & 1.18 & 1.19 & 1.14 & 1.69 & \textbf{1.06} & 1.15 \\
 & Rel.\ FMR, Asian & 1.41 & 1.33 & 0.87 & 1.14 & 1.19 & 0.97 & 1.30 & 1.06 & 0.90 & 1.19 & 1.12 & 0.92 & 1.47 & \textbf{0.93} & 1.05 \\
 & Rel.\ FMR, Caucasian & 0.29 & 0.37 & 1.01 & 1.04 & 0.33 & 0.45 & 0.39 & 0.80 & 0.98 & 0.63 & 0.57 & 0.98 & 0.29 & \textbf{1.01} & 0.78 \\
 & Rel.\ FMR, Indian & 1.47 & 1.30 & 1.15 & 0.84 & 1.78 & 1.39 & 1.28 & 1.09 & 0.97 & 1.13 & 1.32 & 0.97 & 1.38 & \textbf{1.00} & 1.07 \\
 & Gini, FMR & 0.29 & 0.26 & 0.07 & 0.08 & 0.32 & 0.30 & 0.26 & 0.07 & 0.07 & 0.14 & 0.18 & 0.05 & 0.30 & \textbf{0.03} & 0.09 \\
 & Gini, FNMR & 0.24 & 0.13 & 0.04 & 0.02 & 0.24 & 0.09 & 0.26 & \textbf{0.06} & 0.10 & 0.14 & 0.08 & 0.07 & 0.23 & 0.12 & 0.16 \\
\bottomrule
\end{tabular}

\end{table}

\begin{table}[!t]
\centering
\caption{Demographic differentials on IJB-C at global FMR~$10^{-2}$, at the settings and with the rules of Table~\ref{tab:fairness}. Leakage (\%), FNMR (\%), the FMR of each group relative to the geometric mean of the group FMRs, and the Gini indices of the group FMRs and FNMRs.}
\label{tab:s-fairness-ijbc-fmr2}
\scriptsize
\setlength{\tabcolsep}{3pt}
\begin{tabular}{@{}lcccccc@{}}
\toprule
 & Leak. & FNMR & \multicolumn{2}{c}{Relative FMR} & \multicolumn{2}{c}{Gini} \\
\cmidrule(lr){4-5}\cmidrule(l){6-7}
Gender & (\%) & (\%) & Female & Male & FMR & FNMR \\
\midrule
No removal & 96.0 & 1.5 & 1.21 & 0.83 & 0.18 & 0.32 \\
PFRNet$^{\dagger}$ & 92.9 & 6.0 & 1.32 & 0.76 & 0.27 & 0.27 \\
DVPF (paper)$^{\dagger}$ & 80.1 & 28.8 & 0.72 & 1.39 & 0.32 & 0.11 \\
DVPF (code) & 56.0 & 50.1 & 1.05 & 0.95 & 0.05 & 0.05 \\
SensitiveNets$^{\dagger}$ & 94.7 & 2.7 & 1.39 & 0.72 & 0.32 & 0.35 \\
SensitiveNets (Fig.~1 detector)$^{\dagger}$ & 93.6 & 3.0 & 1.39 & 0.72 & 0.32 & 0.36 \\
INLP$^{\dagger}$ & 95.8 & 1.6 & 1.20 & 0.84 & 0.18 & 0.31 \\
IVE & 59.2 & 11.4 & 1.04 & 0.96 & 0.04 & 0.22 \\
Multi-IVE & 56.6 & 16.7 & 1.10 & 0.91 & 0.09 & \textbf{0.14} \\
PE-MIU & 54.9 & 9.1 & 0.98 & 1.02 & 0.02 & 0.23 \\
PASS$^{\dagger}$ & 91.0 & 14.1 & 1.38 & 0.72 & 0.31 & 0.15 \\
FaRM & 53.3 & 11.2 & 0.98 & 1.03 & 0.03 & 0.16 \\
LEACE$^{\dagger}$ & 96.0 & 1.5 & 1.20 & 0.83 & 0.18 & 0.33 \\
PCA + LEACE & 57.9 & 9.5 & \textbf{1.01} & \textbf{0.99} & \textbf{0.01} & 0.18 \\
TEMPER (ours) & 57.0 & \textbf{3.6} & 1.02 & 0.98 & 0.02 & 0.27 \\
\bottomrule
\end{tabular}

\vspace{6pt}
\begin{tabular}{@{}lcccccccc@{}}
\toprule
 & Leak. & FNMR & \multicolumn{4}{c}{Relative FMR} & \multicolumn{2}{c}{Gini} \\
\cmidrule(lr){4-7}\cmidrule(l){8-9}
Ethnicity & (\%) & (\%) & Afr. & Asi. & Cau. & Ind. & FMR & FNMR \\
\midrule
No removal & 90.0 & 1.5 & 1.27 & 1.34 & 0.48 & 1.21 & 0.20 & 0.33 \\
PFRNet$^{\dagger}$ & 88.5 & 5.5 & 1.50 & 1.37 & 0.54 & 0.90 & 0.26 & 0.10 \\
DVPF (paper) & 36.3 & 29.4 & 1.18 & 0.96 & 1.12 & 0.80 & 0.11 & 0.11 \\
DVPF (code) & 35.7 & 31.8 & 1.69 & 1.18 & 0.71 & 0.71 & 0.27 & \textbf{0.06} \\
SensitiveNets$^{\dagger}$ & 85.7 & 3.6 & 1.48 & 1.22 & 0.47 & 1.18 & 0.23 & 0.16 \\
SensitiveNets (Fig.~1 detector)$^{\dagger}$ & 49.3 & 21.2 & 1.48 & 1.37 & 0.42 & 1.17 & 0.25 & 0.07 \\
INLP$^{\dagger}$ & 88.4 & 1.8 & 1.30 & 1.30 & 0.54 & 1.09 & 0.20 & 0.32 \\
IVE & 35.4 & 11.7 & 1.01 & 1.04 & 0.90 & 1.05 & 0.04 & 0.07 \\
Multi-IVE & 33.0 & 9.5 & \textbf{1.05} & \textbf{1.00} & \textbf{0.96} & \textbf{0.99} & \textbf{0.02} & \textbf{0.06} \\
PE-MIU & 38.7 & 4.2 & 1.05 & 1.07 & 0.89 & 0.99 & 0.05 & 0.16 \\
PASS$^{\dagger}$ & 71.9 & 14.1 & 1.46 & 1.28 & 0.49 & 1.10 & 0.24 & 0.06 \\
FaRM & 34.5 & 10.9 & 1.06 & 1.03 & 0.89 & 1.03 & 0.04 & 0.07 \\
LEACE$^{\dagger}$ & 89.5 & 1.5 & 1.32 & 1.36 & 0.50 & 1.11 & 0.21 & 0.33 \\
PCA + LEACE & 33.4 & 11.5 & 1.00 & 1.04 & 0.93 & 1.03 & 0.03 & 0.08 \\
TEMPER (ours) & 36.2 & \textbf{3.2} & 1.12 & 1.10 & 0.84 & 0.97 & 0.08 & 0.15 \\
\bottomrule
\end{tabular}

\end{table}

\begin{table}[!t]
\centering
\caption{Demographic differentials on IJB-C at global FMR~$10^{-1}$, as in Table~\ref{tab:s-fairness-ijbc-fmr2}.}
\label{tab:s-fairness-ijbc-fmr1}
\scriptsize
\setlength{\tabcolsep}{3pt}
\begin{tabular}{@{}lcccccc@{}}
\toprule
 & Leak. & FNMR & \multicolumn{2}{c}{Relative FMR} & \multicolumn{2}{c}{Gini} \\
\cmidrule(lr){4-5}\cmidrule(l){6-7}
Gender & (\%) & (\%) & Female & Male & FMR & FNMR \\
\midrule
No removal & 96.0 & 0.9 & 1.08 & 0.93 & 0.08 & 0.33 \\
PFRNet$^{\dagger}$ & 92.9 & 2.5 & 1.16 & 0.86 & 0.15 & 0.20 \\
DVPF (paper)$^{\dagger}$ & 80.1 & 10.3 & 0.85 & 1.18 & 0.17 & 0.14 \\
DVPF (code) & 56.0 & 28.3 & 0.99 & 1.01 & 0.01 & \textbf{0.09} \\
SensitiveNets$^{\dagger}$ & 94.7 & 1.0 & 1.19 & 0.84 & 0.17 & 0.31 \\
SensitiveNets (Fig.~1 detector)$^{\dagger}$ & 93.6 & 1.1 & 1.25 & 0.80 & 0.22 & 0.26 \\
INLP$^{\dagger}$ & 95.8 & 0.9 & 1.07 & 0.93 & 0.07 & 0.34 \\
IVE & 59.2 & 4.4 & 1.01 & 0.99 & 0.01 & 0.27 \\
Multi-IVE & 56.6 & 5.6 & 1.04 & 0.96 & 0.04 & 0.21 \\
PE-MIU & 54.9 & 5.1 & 0.99 & 1.01 & 0.01 & 0.24 \\
PASS$^{\dagger}$ & 91.0 & 6.0 & 1.18 & 0.85 & 0.17 & 0.17 \\
FaRM & 53.3 & 4.9 & \textbf{1.00} & \textbf{1.00} & \textbf{0.00} & 0.22 \\
LEACE$^{\dagger}$ & 96.0 & 0.9 & 1.07 & 0.93 & 0.07 & 0.35 \\
PCA + LEACE & 57.9 & 3.7 & \textbf{1.00} & \textbf{1.00} & \textbf{0.00} & 0.24 \\
TEMPER (ours) & 57.0 & \textbf{1.8} & 1.01 & 0.99 & 0.01 & 0.29 \\
\bottomrule
\end{tabular}

\vspace{6pt}
\begin{tabular}{@{}lcccccccc@{}}
\toprule
 & Leak. & FNMR & \multicolumn{4}{c}{Relative FMR} & \multicolumn{2}{c}{Gini} \\
\cmidrule(lr){4-7}\cmidrule(l){8-9}
Ethnicity & (\%) & (\%) & Afr. & Asi. & Cau. & Ind. & FMR & FNMR \\
\midrule
No removal & 90.0 & 0.9 & 1.09 & 1.10 & 0.75 & 1.11 & 0.09 & 0.37 \\
PFRNet$^{\dagger}$ & 88.5 & 2.3 & 1.28 & 1.22 & 0.72 & 0.89 & 0.16 & 0.18 \\
DVPF (paper) & 36.3 & 10.7 & 1.10 & 0.94 & 1.05 & 0.91 & 0.06 & 0.13 \\
DVPF (code) & 35.7 & 15.3 & 1.34 & 1.12 & 0.85 & 0.78 & 0.16 & \textbf{0.06} \\
SensitiveNets$^{\dagger}$ & 85.7 & 1.4 & 1.24 & 1.08 & 0.66 & 1.13 & 0.14 & 0.28 \\
SensitiveNets (Fig.~1 detector)$^{\dagger}$ & 49.3 & 6.5 & 1.32 & 1.22 & 0.55 & 1.14 & 0.19 & 0.14 \\
INLP$^{\dagger}$ & 88.4 & 1.0 & 1.11 & 1.11 & 0.80 & 1.01 & 0.09 & 0.36 \\
IVE & 35.4 & 4.5 & 1.01 & 1.01 & 0.95 & 1.04 & 0.02 & 0.14 \\
Multi-IVE & 33.0 & 3.7 & \textbf{1.00} & \textbf{1.02} & \textbf{0.98} & \textbf{1.01} & \textbf{0.01} & 0.19 \\
PE-MIU & 38.7 & 2.6 & 1.01 & 1.02 & 0.96 & 1.01 & 0.02 & 0.21 \\
PASS$^{\dagger}$ & 71.9 & 6.0 & 1.23 & 1.11 & 0.70 & 1.05 & 0.14 & 0.06 \\
FaRM & 34.5 & 4.2 & 1.09 & 1.06 & 0.88 & 0.98 & 0.06 & 0.11 \\
LEACE$^{\dagger}$ & 89.5 & 0.9 & 1.13 & 1.13 & 0.78 & 1.01 & 0.10 & 0.38 \\
PCA + LEACE & 33.4 & 4.2 & \textbf{1.00} & \textbf{1.01} & \textbf{0.98} & \textbf{1.01} & \textbf{0.01} & 0.14 \\
TEMPER (ours) & 36.2 & \textbf{1.6} & 1.08 & 1.05 & 0.93 & 0.95 & 0.05 & 0.22 \\
\bottomrule
\end{tabular}

\end{table}

\begin{table}[!t]
\centering
\caption{Per-group error rates for gender on IJB-C at the global threshold closest to FMR~$10^{-3}$. FMR (in units of $10^{-3}$) and FNMR of the female (F) and male (M) groups, and the global FNMR.}
\label{tab:s-fairness}
\footnotesize
\setlength{\tabcolsep}{4pt}
\begin{tabular}{@{}lrrrrr@{}}
\toprule
 & Global & \multicolumn{2}{c}{FMR ($10^{-3}$)} & \multicolumn{2}{c}{FNMR (\%)} \\
\cmidrule(lr){3-4}\cmidrule(lr){5-6}
Setting & FNMR (\%) & F & M & F & M \\
\midrule
No removal & 2.3 & 2.7 & 1.5 & 3.2 & 1.7 \\
TEMPER, $k{=}62$, $\lambda{=}3$ & 5.9 & 0.97 & 0.89 & 7.8 & 4.5 \\
PCA truncation, $k{=}22$ & 18.4 & 0.94 & 0.90 & 21.2 & 16.3 \\
\bottomrule
\end{tabular}
\end{table}

\section{IResNet100 Trained on Glint360k}
\label{app:backbone}

Every experiment of the main paper, apart from the reduced grid of this appendix, uses the embeddings of an IResNet50 trained with ArcFace on MS1MV3.
We repeat a reduced grid on the embeddings of an IResNet100 trained on Glint360k to test whether the results hold beyond the IResNet50.
The IResNet100 is the recognition model of the InsightFace antelopev2 package and takes the same aligned crops.
The configurations, hyperparameters, protocols and attackers are those of the main paper.
The grid comprises TEMPER at $k\in\{22,42,62,96\}$ with $\lambda\in\{3,10,30\}$ for gender and at $k\in\{22,42,96\}$ with $\lambda\in\{3,10\}$ for ethnicity.
The grid also comprises PCA truncation at 5 values of $k$ and the embeddings without removal.
We did not rerun the prior methods on the IResNet100.

TEMPER leaks less than PCA truncation on both backbones at every $k$ that both grids contain (Table~\ref{tab:s-backbone}).
The TMR at FMR~$10^{-3}$ without removal is 0.5\% higher on the IResNet100 than on the IResNet50, on IJB-C and on RFW.
The IJB-C TMR at FMR~$10^{-3}$ of TEMPER differs between the backbones by at most 2\% for gender and by at most 4\% for ethnicity, at the same $k$ and $\lambda$.
The largest of these differences is found at $k=96$, $\lambda=3$ for ethnicity, where the IResNet100 attains a 3.8\% higher TMR at FMR~$10^{-3}$ with a lower leakage.
The RFW TMR at FMR~$10^{-3}$ of TEMPER differs more between the backbones, by up to 21.9\% at the same setting.
The threshold at FMR~$10^{-3}$ on RFW is set by 12 non-mated pairs (Appendix~\ref{app:protocol-verification}).

The leakage of TEMPER differs between the backbones by at most 1\% near chance, except at $k=96$, $\lambda=10$ for ethnicity.
The leakage of this setting on the IResNet100 is 27.6\%, just above the near-chance limit of 27.5\%.
The gender leakage of TEMPER is higher on the IResNet100 by 1.5\% to 6.0\% at 3 settings with $k\ge62$.
The LR attacker attains the worst case on the IResNet100 at these 3 settings.
PCA truncation leaks less on the IResNet100 at every $k$ except $k=12$, by 5.1\% for gender and by 7.3\% for ethnicity at $k=42$.
The LR attacker attains 88.3\% balanced accuracy for gender on the IResNet100 without removal, against 65.1\% on the IResNet50.
The MLP attacker attains 96\% on both backbones.
Gender is therefore encoded more linearly in the IResNet100 embeddings than in the IResNet50 embeddings.

\begin{table}[!t]
\centering
\caption{TEMPER and PCA truncation on the IResNet50 trained on MS1MV3 (R50) and the IResNet100 trained on Glint360k (R100). TMR (\%) at FMR~$10^{-3}$ on IJB-C and RFW, and worst-case leakage (\%) on IJB-C for gender and on RFW for ethnicity.}
\label{tab:s-backbone}
\footnotesize
\setlength{\tabcolsep}{4pt}
\begin{tabular}{@{}lrrrrrr@{}}
\toprule
 & \multicolumn{2}{c}{IJB-C TMR} & \multicolumn{2}{c}{RFW TMR} & \multicolumn{2}{c}{Leakage} \\
\cmidrule(lr){2-3}\cmidrule(lr){4-5}\cmidrule(l){6-7}
Setting & R50 & R100 & R50 & R100 & R50 & R100 \\
\midrule
\multicolumn{7}{@{}l}{\textit{Gender} (chance 50\%)} \\
No removal & 97.6 & 98.1 & 96.8 & 97.3 & 96.0 & 95.8 \\
PCA truncation $k{=}12$ & 60.8 & 62.7 & 10.7 & 11.6 & 54.6 & 55.0 \\
PCA truncation $k{=}22$ & 82.3 & 84.0 & 24.3 & 30.6 & 58.4 & 57.1 \\
PCA truncation $k{=}42$ & 92.0 & 92.6 & 61.2 & 57.9 & 70.7 & 65.6 \\
PCA truncation $k{=}62$ & 94.9 & 95.2 & 77.6 & 74.7 & 80.5 & 74.9 \\
PCA truncation $k{=}128$ & 97.0 & 97.4 & 94.5 & 93.0 & 93.4 & 90.9 \\
TEMPER $k{=}22$, $\lambda{=}3$ & 82.4 & 83.6 & 27.8 & 30.8 & 52.2 & 51.6 \\
TEMPER $k{=}22$, $\lambda{=}10$ & 82.2 & 83.5 & 26.6 & 30.0 & 51.1 & 50.9 \\
TEMPER $k{=}22$, $\lambda{=}30$ & 79.9 & 80.6 & 27.6 & 27.7 & 51.1 & 51.0 \\
TEMPER $k{=}42$, $\lambda{=}3$ & 92.1 & 92.6 & 61.2 & 56.6 & 54.5 & 53.7 \\
TEMPER $k{=}42$, $\lambda{=}10$ & 88.2 & 87.7 & 43.8 & 38.8 & 52.5 & 52.3 \\
TEMPER $k{=}42$, $\lambda{=}30$ & 81.0 & 80.3 & 26.8 & 20.9 & 51.7 & 52.3 \\
TEMPER $k{=}62$, $\lambda{=}3$ & 94.2 & 94.6 & 75.4 & 69.4 & 57.0 & 55.2 \\
TEMPER $k{=}62$, $\lambda{=}10$ & 88.7 & 88.8 & 41.3 & 40.5 & 53.0 & 56.2 \\
TEMPER $k{=}62$, $\lambda{=}30$ & 82.5 & 82.4 & 22.3 & 21.2 & 52.6 & 58.6 \\
TEMPER $k{=}96$, $\lambda{=}3$ & 95.2 & 95.8 & 81.9 & 78.6 & 60.5 & 57.0 \\
TEMPER $k{=}96$, $\lambda{=}10$ & 91.4 & 91.5 & 56.0 & 54.7 & 54.5 & 53.6 \\
TEMPER $k{=}96$, $\lambda{=}30$ & 87.9 & 88.2 & 45.8 & 42.5 & 54.1 & 55.6 \\
\midrule
\multicolumn{7}{@{}l}{\textit{Ethnicity} (chance 25\%)} \\
No removal & 97.6 & 98.1 & 96.8 & 97.3 & 90.0 & 89.1 \\
PCA truncation $k{=}12$ & 60.9 & 62.8 & 10.5 & 11.3 & 28.8 & 28.9 \\
PCA truncation $k{=}22$ & 82.2 & 83.8 & 26.3 & 32.2 & 35.1 & 31.5 \\
PCA truncation $k{=}42$ & 91.9 & 92.6 & 60.0 & 58.8 & 48.5 & 41.2 \\
PCA truncation $k{=}62$ & 94.9 & 95.1 & 77.4 & 74.1 & 57.6 & 48.4 \\
PCA truncation $k{=}128$ & 97.0 & 97.4 & 94.2 & 92.8 & 86.5 & 79.4 \\
TEMPER $k{=}22$, $\lambda{=}3$ & 79.6 & 79.4 & 25.9 & 21.5 & 26.5 & 26.3 \\
TEMPER $k{=}22$, $\lambda{=}10$ & 73.9 & 70.6 & 19.9 & 17.3 & 26.3 & 25.9 \\
TEMPER $k{=}42$, $\lambda{=}3$ & 88.4 & 88.1 & 43.3 & 35.0 & 28.6 & 28.2 \\
TEMPER $k{=}42$, $\lambda{=}10$ & 72.5 & 72.2 & 17.6 & 19.0 & 26.4 & 26.4 \\
TEMPER $k{=}96$, $\lambda{=}3$ & 90.8 & 94.6 & 56.0 & 77.9 & 30.3 & 29.7 \\
TEMPER $k{=}96$, $\lambda{=}10$ & 73.3 & 70.5 & 18.1 & 13.6 & 26.2 & 27.6 \\
\bottomrule
\end{tabular}

\end{table}

\end{document}